\documentclass[10pt,twocolumn,letterpaper]{article}

\pdfoutput=1

\usepackage{cvpr}
\usepackage{times}
\usepackage{epsfig}
\usepackage{graphicx}
\usepackage{amsmath}
\usepackage{amssymb}
\usepackage{color}
\usepackage{cite}

\usepackage{multirow}
\newcommand\blfootnote[1]{%
  \begingroup
  \renewcommand\thefootnote{}\footnote{#1}%
  \addtocounter{footnote}{-1}%
  \endgroup
}
\renewcommand\baselinestretch{0.97} 

\usepackage[pagebackref=true,breaklinks=true,letterpaper=true,colorlinks,bookmarks=false]{hyperref}

\cvprfinalcopy 

\def\cvprPaperID{720} 

\begin{document}

\title{Learning Temporal Regularity in Video Sequences}


\author{
Mahmudul Hasan\hspace{1em}Jonghyun Choi$^\dagger$\hspace{1em}Jan Neumann$^\dagger$\hspace{1em}Amit K. Roy-Chowdhury\hspace{1em}Larry S. Davis$^\ddagger$
\vspace{0.5em}
\\
UC, Riverside\hspace{2em}
Comcast Labs, DC$^\dagger$\hspace{2em}
University of Maryland, College Park$^\ddagger$
\\
{\tt\footnotesize \{mhasa004@,amitrc@ee.\}ucr.edu}~
{\tt\footnotesize \{jonghyun\_choi,jan\_neumann\}@cable.comcast.com}~
{\tt\footnotesize lsd@umiacs.umd.edu}
}

\maketitle

\blfootnote{This work is partially done during M. Hasan's internship at Comcast Labs DC.}

\begin{abstract}

Perceiving meaningful activities in a long video sequence is a challenging problem due to ambiguous definition of `meaningfulness' as well as clutters in the scene.
We approach this problem by learning a generative model for regular motion patterns (termed as regularity) using multiple sources with very limited supervision.
Specifically, we propose two methods that are built upon the autoencoders for their ability to work with little to no supervision.
We first leverage the conventional handcrafted spatio-temporal local features and learn a fully connected autoencoder on them.
Second, we build a fully convolutional feed-forward autoencoder to learn both the local features and the classifiers as an end-to-end learning framework.
Our model can capture the regularities from multiple datasets.
We evaluate our methods in both qualitative and quantitative ways - showing the learned regularity of videos in various aspects and demonstrating competitive performance on anomaly detection datasets as an application.

\end{abstract}


\section{Introduction}
\label{sec:intro}

The availability of large numbers of uncontrolled videos gives rise to the problem of watching long hours of meaningless scenes~\cite{sunFS14}. 
Automatic segmentation of `meaningful' moments in such videos without supervision or with very limited supervision is a fundamental problem for various computer vision applications such as video annotation~\cite{vondrick2013efficiently}, summarization~\cite{songVSJ15,chuSJ15}, indexing or temporal segmentation~\cite{laptevMSR08}, anomaly detection~\cite{popoola2012video}, and activity recognition~\cite{karpathyTSLSF14}.
We address this problem by modeling temporal regularity of videos with limited supervision, rather than modeling the sparse irregular or meaningful moments in a supervised manner. 

Learning temporal visual characteristics of meaningful or salient moments is very challenging as the definition of such moments is ill-defined \ie, visually unbounded. 
On the other hand, learning temporal visual characteristics of ordinary moments is relatively easier as they often exhibit temporally regular dynamics such as periodic crowd motions.
We focus on learning the characteristics of regular temporal patterns with a very limited form of labeling - we assume that all events in the training videos are part of the regular patterns. 
Especially, we use multiple video sources, e.g., different datasets, to learn the regular temporal appearance changing pattern of videos in a single model that can then be used for multiple videos.


\begin{figure}[t]
	\centering
	\includegraphics[scale=0.48]{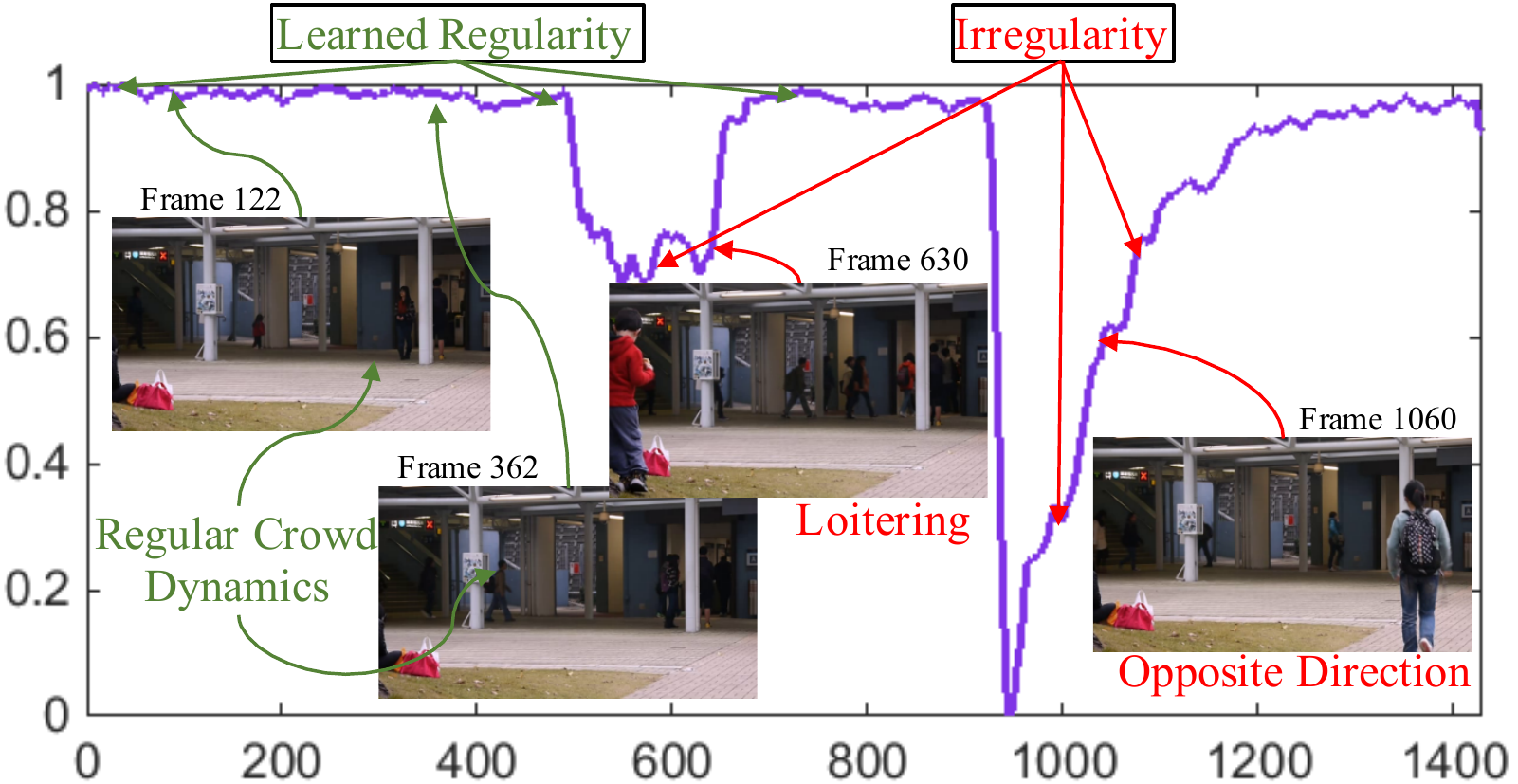}
	\caption{Learned regularity of a video sequence. Y-axis refers to regularity score and X-axis refers to frame number. When there are irregular motions, the regularity score drops significantly (from CUHK-Avenue dataset \cite{lu2013abnormal}).}
	\vspace{-6mm}
	\label{fig:teaser_new}
\end{figure}



Given the training data of regular videos only, learning the temporal dynamics of regular scenes is an unsupervised learning problem. 
A state-of-the-art approach for such unsupervised modeling involves a combination of sparse coding and bag-of-words~\cite{zhao2011online,congYL11,lu2013abnormal}.
However, bag-of-words does not preserve spatio-temporal structure of the words and requires prior information about the number of words. 
Additionally, optimization involved in sparse coding for both training and testing is computationally expensive, especially with large data such as videos.

We present an approach based on autoencoders. 
Its objective function is computationally more efficient than sparse coding and it preserves spatio-temporal information while encoding dynamics.
The learned autoencoder reconstructs regular motion with low error but incurs higher reconstruction error for irregular motions.
Reconstruction error has been widely used for abnormal event detection \cite{popoola2012video}, since it is a function of frame visual statistics and abnormalities manifest themselves as deviations from normal visual patterns.
Figure~\ref{fig:teaser_new} shows an example of learned regularity, which is computed from the reconstruction error by a learned model (Eq.\ref{eq:test} and Eq.\ref{eq:test1}).

We propose to learn an autoencoder for temporal regularity based on two types of features as follows.
First, we use state-of-the-art handcrafted motion features and learn a neural network based deep autoencoder consisting of seven fully connected layers.
The state-of-the-art motion features, however, may be suboptimal for learning temporal regularity as they are not designed or optimized for this problem.
Subsequently, we directly learn both the motion features and the discriminative regular patterns using a fully convolutional neural network based autoencoder. 

We train our models using multiple datasets including CUHK Avenue \cite{lu2013abnormal}, Subway (Enter and Exit) \cite{adam2008robust}, and UCSD Pedestrian datasets (Ped1 and Ped2) \cite{mahadevan2010anomaly}, without compensating the dataset bias~\cite{torralbaEfros11}. 
Therefore, the learned model is generalizable across the datasets. 
We show that our methods discover temporally regular appearance-changing patterns of videos with various applications - synthesizing the most regular frame from a video, delineating objects involved in irregular motions, and predicting the past and the future regular motions from a single frame. 
Our model also performs comparably to the state-of-the-art methods on anomaly detection task evaluated on multiple datasets including recently released public ones. 

Our contributions are summarized as follows:
\vspace{-2mm}
\begin{itemize}
	\setlength{\itemsep}{-3pt}
	\item Showing that an autoencoder effectively learns the regular dynamics in long-duration videos and can be applied to identify irregularity in the videos.
	\item Learning the low level motion features for our proposed method using a fully convolutional autoencoder.
	\item Applying the model to various applications including learning temporal regularity, detecting objects associated with irregular motions, past and future frame prediction, and abnormal event detection.
\end{itemize}
\vspace{-2mm}



\section{Related Work}
\label{sec:related}

\noindent
{\bf Learning Motion Patterns Without Supervision.}
Learning motion patterns without supervision has received much attention in recent years~\cite{leZYN11,jhuangSWP07,taylorFLB10}.
Goroshin\etal~\cite{goroshinBTEC15} trained a regularized high capacity (\emph{i.e.}, deep) neural network based autoencoder using a temporal coherency prior on adjacent frames. 
Ramanathan\etal~\cite{ramanathanTMF15} trained a network to learn the motion signature of the same temporal coherency prior and used it for event retrieval.

To analyze temporal information, recurrent neural networks (RNN) have been widely used for analyzing speech and audio data~\cite{graves2013speech}. 
For video analysis, Donahue\etal~take advantage of long short term memory (LSTM) based RNN for visual recognition with the large scale labeled data~\cite{donahue2014long}. 
Du\etal~built an RNN in a hierarchical way to recognize actions~\cite{du2015hierarchical}. 
The supervised action recognition setup requires human supervision to train the models. 
Ranzato\etal~used the RNN for motion prediction \cite{ranzatoSBMCC14}, while we model the temporal regularity in a video sequence.

\vspace{.5em} \noindent
{\bf Anomaly Detection.}
One of the applications of our model is abnormal or anomalous event detection.
The survey paper \cite{popoola2012video} contains a comprehensive review of this topic. 
Most video based anomaly detection approaches involve a local feature extraction step followed by learning a model on training video. 
Any event that is an outlier with respect to the learned model is regarded as the anomaly. 
These models include mixtures of probabilistic principal components on optical flow \cite{kim2009observe}, sparse dictionary \cite{zhao2011online,lu2013abnormal}, Gaussian regression based probabilistic framework \cite{cheng2015video}, spatio-temporal context \cite{xiao2015learning, zhu2013contextj}, sparse autoencoder \cite{sabokrou2015real}, codebook based spatio-temporal volumes analysis \cite{roshtkhari2013online}, and shape \cite{vaswani2005shape}.
Xu\etal~\cite{xu2015learning} proposed a deep model for anomalous event detection that uses a stacked autoencoder for feature learning and a linear classifier for event classification. 
In contrast, our model is an end-to-end trainable generative one that is generalizable across multiple datasets.

\vspace{.5em} \noindent
{\bf Convolutional Neural Network (CNN).}
Since Krizhevsky\etal's work on image classification~\cite{krizhevsky2012imagenet}, CNN has been widely applied to various computer vision tasks such as feature extraction~\cite{razavianASC14}, image classification~\cite{simonyanZ14}, object detection~\cite{deanRSSVY13,renHGS13}, face verification~\cite{taigmanYRW14}, semantic embedding~\cite{fromeCSBDRM13,karpathyF14}, video analysis~\cite{ranzatoSBMCC14,taylorFLB10}, and \emph{etc}. 
Particularly in video, Karpathy\etal~and Ng\etal~recently proposed a supervised CNN to classify actions in videos~\cite{karpathyTSLSF14,ngHVVMT15}. 
Xu\etal~trained a CNN to detect events in videos~\cite{xu2014discriminative}. 
Wang\etal~learned a CNN to pool the trajectory information for recognizing actions~\cite{wang2015action}. 
These methods, however, require human supervision as they are supervised classification tasks. 

\begin{figure*}[ht]
	\centering
	\includegraphics[scale=0.59]{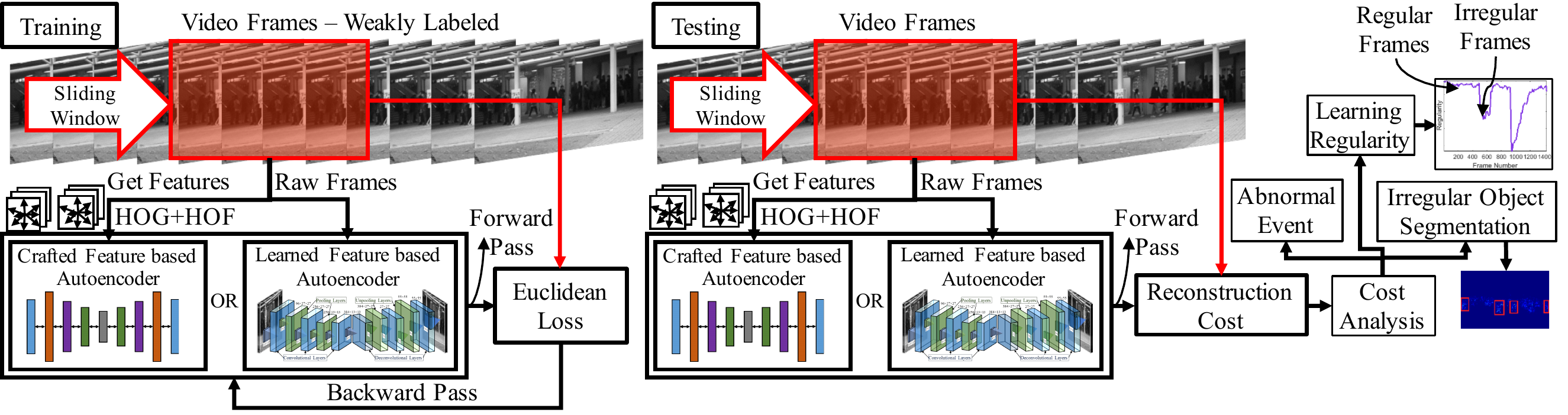}
	\caption{Overview of our approach. It utilizes either state-of-the-art motion features or learned features combined with autoencoder to reconstruct the scene. The reconstruction error is used to measure the regularity score that can be further analyzed for different applications.}
	\vspace{-5mm}
	\label{fig:framework}
\end{figure*}

%

\vspace{.5em} \noindent
{\bf Convolutional Autoencoder.}
For an end-to-end learning system for regularity in videos, we employ the convolutional autoencoder.
Zhao\etal~proposed a unified loss function to train a convolutional autoencoder for classification purposes~\cite{zhaoMGL15}.
Noh\etal~\cite{nohHH15} used convolutional autoencoders for semantic segmentation.


\section{Approach}
\label{sec:approach}

We use an autoencoder to learn regularity in video sequences. 
The intuition is that the learned autoencoder will reconstruct the motion signatures present in regular videos with low error but will not accurately reconstruct motions in irregular videos. 
In other words, the autoencoder can model the complex distribution of the regular dynamics of appearance changes. 

As an input to the autoencoder, initially, we use state-of-the-art handcrafted motion features that consist of HOG and HOF with improved trajectory features \cite{wang2013action}. Then we learn the regular motion signatures by a (fully-connected) neural network based autoencoder. 
However, even the state-of-the-art motion features may not be optimal for learning regularity as they are not specifically designed for this purpose.
Thus, we use the video as an input and learn both local motion features and the autoencoder by an end-to-end learning model based on a fully convolutional neural network. 
We illustrate the overview of our the approach in Fig.~\ref{fig:framework}.

\subsection{Learning Motions on Handcrafted Features}
\label{sec:local_approach}

We first extract handcrafted appearance and motion features from the video frames. 
We then use the extracted features as input to a fully connected neural network based autoencoder to learn the temporal regularity in the videos, similar to \cite{hinton2006reducing,hasanR14}.

\vspace{.5em}
\noindent \textbf{Low-Level Motion Information in a Small Temporal Cuboid.}
We use Histograms of Oriented Gradients (HOG)~\cite{DT05,laptevMSR08} and Histograms of Optical Flows (HOF)~\cite{dalal2006human} in a temporal cuboid as a spatio-temporal appearance feature descriptor for their efficiency in encoding appearance and motion information respectively.

\vspace{.5em}
\noindent \textbf{Trajectory Encoding.}
In order to extract HOG and HOF features along with the trajectory information, we use the improved trajectory (IT) features from Wang\etal~ \cite{wang2013action}.
It is based on the trajectory of local features, which has shown impressive performance in many human activity recognition benchmarks~\cite{wang2013action,gaurZSR11}. 

As a first step of feature extraction, interest points are densely sampled at dense grid locations of every five pixels.
Eight spatial scales are used for scale invariance.
Interest points located in the homogeneous texture areas are excluded based on the eigenvalues of the auto-correlation matrix. 
Then, the interest points in the current frame are tracked to the next frame by median filtering a dense optical flow field~\cite{wangKSC11}.
This tracking is normally carried out up to a fixed number of frames ($L$) in order to avoid drifting. 
Finally, trajectories with sudden displacement are removed from the set \cite{wang2013action}.

\vspace{.5em}
\noindent \textbf{Final Motion Feature.}
Local appearance and motion features around the trajectories are encoded with the HOG and HOF descriptors. 
We finally concatenate them to form a 204 dimensional feature as an input to the autoencoder.

%

\subsubsection{Model Architecture}

\begin{figure}[h]
	\centering
	\includegraphics[scale=0.6]{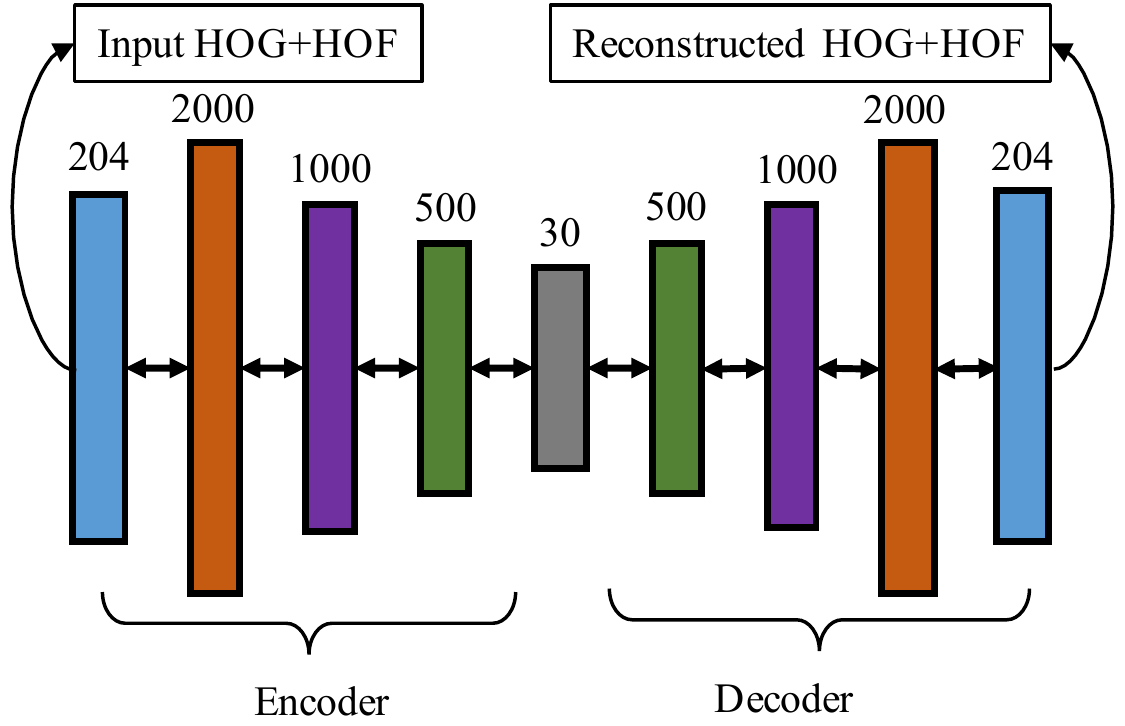}
	\caption{Structure of our autoencoder taking the HOG+HOF feature as input.}
	\vspace{-2mm}
	\label{fig:it_auto}
\end{figure}

Next, we learn a model for regular motion patterns on the motion features in an unsupervised manner. 
We propose to use a deep autoencoder with an architecture similar to Hinton\etal~ \cite{hinton2006reducing} as shown in Figure \ref{fig:it_auto}. 

Our autoencoder takes the 204 dimensional HOG+HOF feature as the input to an encoder and a decoder sequentially. 
The encoder has four hidden layers with 2,000, 1,000, 500, and 30 neurons respectively, whereas the decoder has three hidden layers with 500, 1,000 and 2,000 neurons respectively.
The small-sized middle layers are for learning compact semantics as well as reducing noisy information.

Since both the input and the reconstructed signals of the autoencoder are HOG+HOF histograms, their magnitude of them should be bounded in the range from 0 to 1. 
Thus, we use either sigmoid or hyperbolic tangent ($\tanh$) as the activation function instead of the rectified linear unit (ReLU). ReLU is not suitable for a network that has large receptive fields for each neuron as the sum of the inputs to a neuron can become very large.

In addition, we use the sparse weight initialization technique described in \cite{sutskever2013importance} for the large receptive field. 
In the initialization step, each neuron is connected to $k$ randomly chosen units in the previous layer, whose weights are drawn from a unit Gaussian with zero bias.
As a result, the total number of inputs to each neuron is a constant, which prevents the large input problem.

We define the objective function of the autoencoder by an Euclidean loss of input feature ($\textbf{x}_i$) and the reconstructed feature ($f_W(\textbf{x}_i)$) with an $L_2$ regularization term as shown in Eq.\ref{eqn:it_auto_loss}.
Intuitively, we want to learn a non-linear classifier so that the overall reconstruction cost for the $i^\text{th}$ training features $\textbf{x}_i$ is minimized.
\begin{equation}
	\hat{f}_W = \arg \min_{W} \frac{1}{2N} \sum_i \| \textbf{x}_i - f_W(\textbf{x}_i) \|_2^2 + \gamma \|W\|_2^2,
	\label{eqn:it_auto_loss}
\end{equation}
where $N$ is the size of mini batch, $\gamma$ is a hyper-parameter to balance the loss and the regularization and $f_W(\cdot)$ is a non-linear classifier such as a neural network associated with its weights $W$.

\subsection{Learning Features and Motions}

Even though we use the state-of-the-art motion feature descriptors, they may not be optimal for learning regular patterns in videos. 
To learn highly tuned low level features that best learn temporal regularity, we propose to learn a fully convolutional autoencoder that takes short video clips in a temporal sliding window as the input. 
We use fully convolutional network because it does not contain fully connected layers. 
Fully connected layers loses spatial information \cite{long2015fully}.  
For our model, the spatial information needs to be preserved for reconstructing the input frames.
We present the details of training data, network architecture, and the loss function of our fully convolutional autoencoder model, the training procedure, and parameters in the following subsections.

\subsubsection{Model Architecture}
\begin{figure}[h]
	\centering
	\includegraphics[scale=0.44]{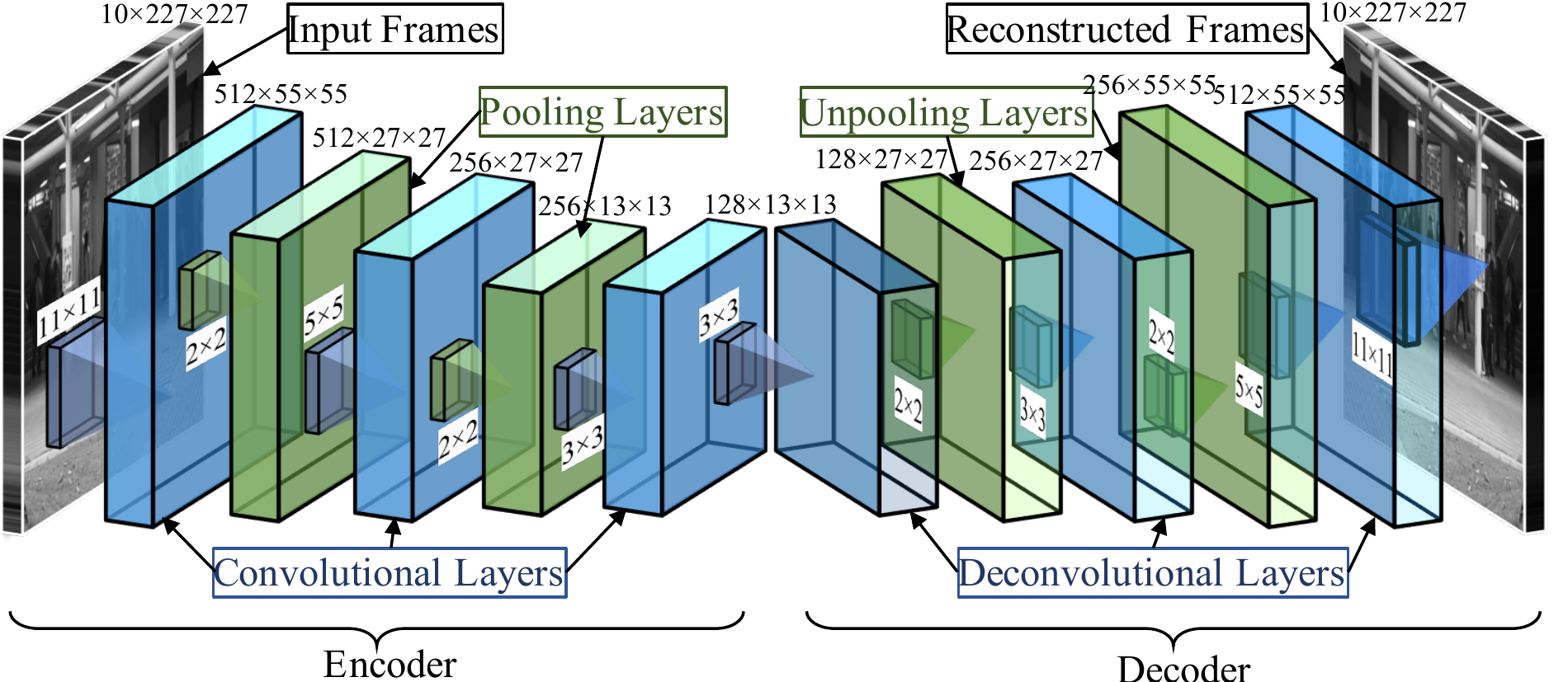}
	\caption{Structure of our fully convolutional autoencoder.}
	\vspace{-2mm}
	\label{fig:conv_auto}
\end{figure}

Figure \ref{fig:conv_auto} illustrates the architecture of our fully convolutional autoencoder.
The encoder consists of convolutional layers \cite{krizhevsky2012imagenet} and the decoder consists of deconvolutional layers that are the reverse of the encoder with padding removal at the boundary of images. 

We use three convolutional layers and two pooling layers on the encoder side and three deconvolutional layers and two unpooling layers on the decoder side by considering the size of input cuboid and training data.

The first convolutional layer has $512$ filters with a stride of $4$. 
It produces $512$ feature maps with a resolution of $55\times55$ pixels. 
Both of the pooling layers have kernel of size $2\times2$ pixels and perform max poling. 
The first pooling layer produces $512$ feature maps of size $27\times27$ pixels. The second and third convolutional layers have $256$ and $128$ filters respectively. 
Finally, the encoder produces $128$ feature maps of size $13 \times 13$ pixels. 
Then, the decoder reconstructs the input by deconvolving and unpooling the input in reverse order of size. 
The output of final deconvolutional layer is the reconstructed version of the input. 

\vspace{.5em}
\noindent \textbf{Input Data Layer.}
Most of convolutional neural networks are for classifying images and take an input of three channels (for R,G, and B color channel). 
Our input, however, is a video, which consists of an arbitrary number of channels.
Recent works \cite{karpathyTSLSF14,donahue2014long} extract features for each video frame, then use several feature fusion schemes to construct the input features to the network, similar to our first approach described in Sec.~\ref{sec:local_approach}.

\begin{figure}[h]
	\centering
	\includegraphics[scale=0.48]{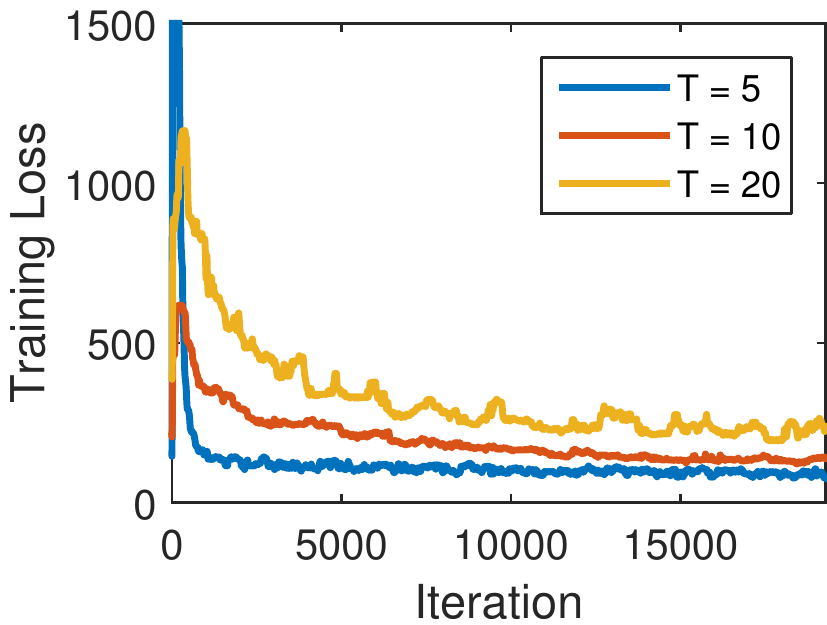}
	\includegraphics[scale=0.48]{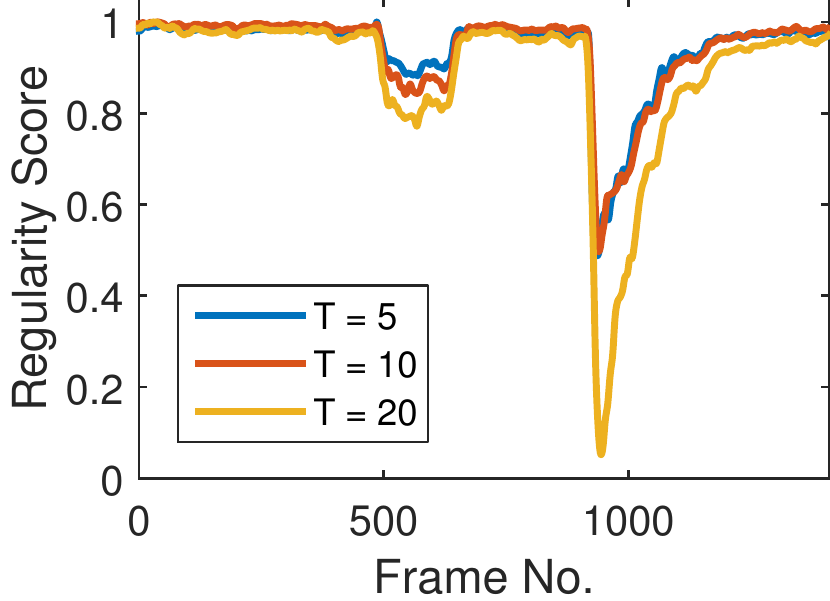}
	\caption{Effect of temporal length ($T$) of input video cuboid. (Left) X-axis is the increasing number of iterations, Y-axis is the training loss, and three plots correspond to three different values of $T$. (Right) X-axis is the increasing number of video frames and Y-axis is the regularity score. As $T$ increases, the training loss takes more iterations to converge as it is more likely that the inputs with more channels have more irregularity to hamper learning regularity. On the other hand, once the model is learned, the regularity score is more distinguishable for higher values of $T$ between regular and irregular regions (note that there are irregular motions in the frame from 480 to 680, and 950 to 1250).}
	\vspace{-4mm}
	\label{fig:trloss_n}
\end{figure}

We, however, construct the input by a temporal cuboid using a sliding window technique without any feature transform. 
Specifically, we stack $T$ frames together and use them as the input to the autoencoder, where $T$ is the length of the sliding window. 
Our experiment shows that increasing $T$ results in a more discriminative regularity score as it incorporates longer motions or temporal information as shown in Fig.~\ref{fig:trloss_n}.

\vspace{.5em}
\noindent \textbf{Data Augmentation In the Temporal Dimension.}
As the number of parameters in the autoencoder is large, we need large amounts of training data.
The size of a given training datasets, however, may not be large enough to train the network.
Thus, we increase the size of the input data by generating more input cuboids with possible transformations to the given data.
To this end, we concatenate frames with various skipping strides to construct $T$-sized input cuboid.
We sample three types of cuboids from the video sequences - stride-1, stride-2, and stride-3. 
In stride-1 cuboids, all $T$ frames are consecutive, whereas in stride-2 and stride-3 cuboids, we skip one and two frames, respectively.
The stride used for sampling cuboids is two frames.
%
%

We also performed experiments with precomputed optical flows. 
Given the gradients and the magnitudes of optical flows between two frames, we compute a single gray scale frame by linearly combining the gradients and magnitudes.
It increases the temporal dimension of the input cuboid from $T$ to $2T$. 
Channels $1, \ldots, T$ contain gray scale video frames, whereas channels $T+1, \ldots, 2T$ contain gray scale flow information. 
This information fusion scheme was used in \cite{ngHVVMT15}.
Our experiments reveal that the overall improvement is insignificant.

\vspace{.5em}
\noindent \textbf{Convolutional and Deconvolutional Layer.}
A convolutional layers connects multiple input activations within the fixed receptive field of a filter to a single activation output.
It abstracts the information of a filter cuboid into a scalar value.

On the other hand, deconvolution layers densify the sparse signal by convolution-like operations with multiple learned filters; thus they associate a single input activation with patch outputs by an inverse operation of convolution.
Thus, the output of deconvolution is larger than the original input due to the superposition of the filters multiplied by the input activation at the boundaries. 
To keep the size of the output mapping identical to the preceding layer, we crop out the boundary of the output that is larger than the input.

The learned filters in the deconvolutional layers serve as bases to reconstruct the shape of an input motion cuboid.
As we stack the convolutional layers at the beginning of the network, we stack the deconvolutional layers to capture different levels of shape details for building an autoencoder.
The filters in early layers of convolutional and the later layers of deconvolutional layers tend to capture specific motion signature of input video frames while high level motion abstractions are encoded in the filters in later layers.

\vspace{.5em}
\noindent \textbf{Pooling and Unpooling Layer.}
Combined with a convolutional layer, the pooling layer further abstracts the activations for various purposes such as translation invariance after the convolutional layer.
Types of pooling operations include `max' and `average.'
We use `max' for translation invariance.
It is known to help classifying images by making convolutional filter output to be spatially invariant~\cite{krizhevsky2012imagenet}.

By using `max' pooling, however, spatial information is lost, which is important for location specific regularity.
Thus, we employ the unpooling layers in the deconvolution network, which perform the reverse operation of pooling and reconstruct the original size of activations~\cite{nohHH15,zeiler2014visualizing,zeiler2011adaptive}. 

We implement the unpooling layer in the same way as \cite{zeiler2014visualizing,zeiler2011adaptive} which records the locations of maximum activations selected during a pooling operation in switch variables and use them to place each activation back to the originally pooled location.

\vspace{.5em}
\noindent \textbf{Optimization Objective.}
Similar to Eq.\ref{eqn:it_auto_loss}, we use Euclidean loss with $L_2$ regularization as an objective function on the temporal cuboids:
\begin{equation}
	\hat{f}_W = \arg \min_{W} \frac{1}{2N} \sum_i \| \textbf{X}_i - f_W(\textbf{X}_i) \|_2^2 + \gamma \|W\|_2^2,
	\label{eqn:conv_auto_loss}
\end{equation}
where $\textbf{X}_i$ is $i^\text{th}$ cuboid, $N$ is the size of mini batch, $\gamma$ is a hyper-parameter to balance the loss and the regularization and $f_W(\cdot)$ is a non-linear classifier - a fully convolutional-deconvolutional neural network with its weights $W$.

\subsection{Optimization and Initialization}
To optimize the autoencoders of Eq.\ref{eqn:it_auto_loss} and Eq.\ref{eqn:conv_auto_loss}, we use a stochastic gradient descent with an adaptive sub-gradient method called AdaGrad~\cite{duchi2011adaptive}.
AdaGrad computes a dimension-wise learning rate that adapts the rate of gradients by a function of all previous updates on each dimension.
It is widely used for its strong theoretical guarantee of convergence and empirical successes.
We also tested Adam~\cite{kingmaB15} and RMSProp~\cite{tielemanH12} but empirically chose to use AdaGrad.

\begin{figure}[h]
	\centering
		\includegraphics[scale=0.55]{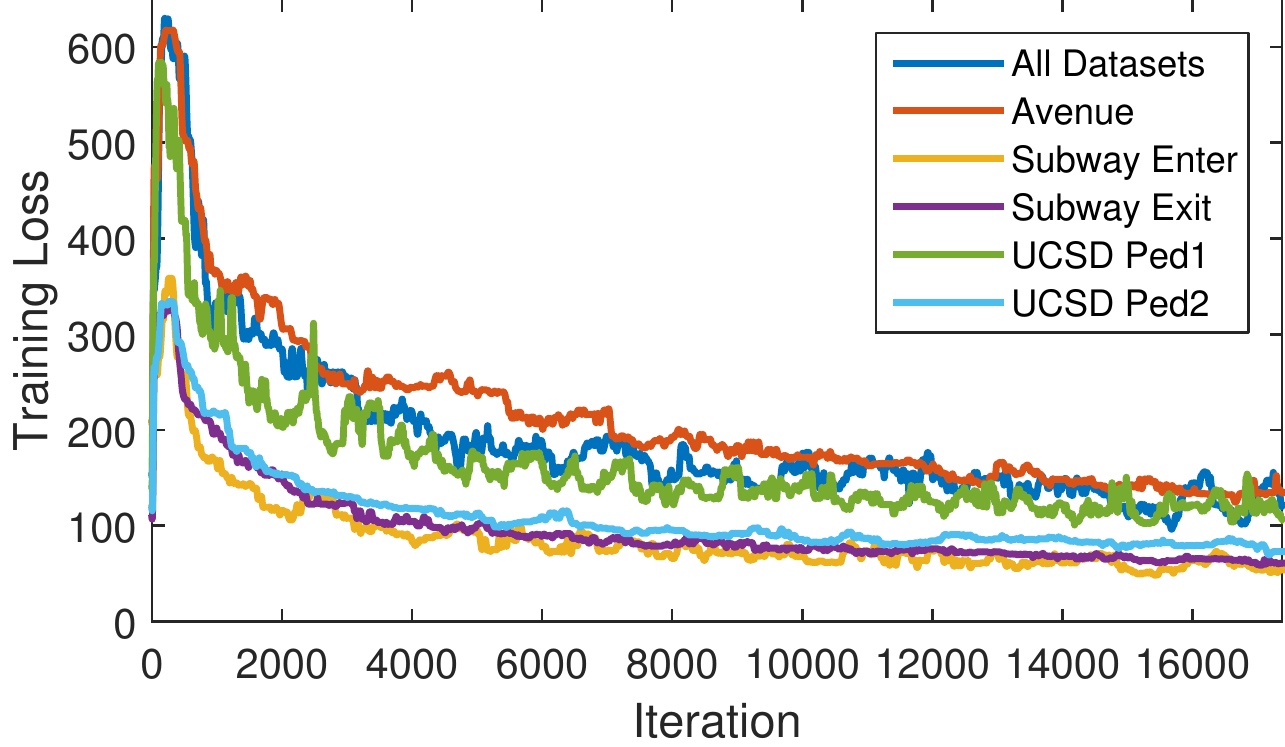}
	\caption{Loss value of models trained on each dataset and all datasets as a function of optimization iterations.}
	\label{fig:loss_datasets}
	\vspace{-3mm}
\end{figure}

We train the network using multiple datasets.
Fig.~\ref{fig:loss_datasets} shows the learning curves trained with different datasets as a function of iterations.
We start with a learning rate of $0.001$ and reduce it when the training loss stops decreasing.
For the autoencoder on the improved trajectory features, we use mini-batches of size $1,024$ and weight decay of $0.0005$. 
For the fully convolutional autoencoder, we use mini batch size of $32$ and start training the network with learning rate $0.01$. 

We initialized the weights using the Xavier algorithm \cite{glorot2010understanding} since Gaussian initialization for various network structure has the following problems.
First, if the weights in a network are initialized with too small values, then the signal shrinks as it passes through each layer until it becomes too small in value to be useful. 
Second, if the weights in a network are initialized with too large values, then the signal grows as it passes through each layer until it becomes too large to be useful.
The Xavier initialization automatically determines the scale of initialization based on the number of input and output neurons, keeping the signal in a reasonable range of values through many layers. 
We empirically observed that the Xavier initialization is noticeably more stable than Gaussian.

\subsection{Regularity Score}
Once we trained the model, we compute the reconstruction error of a pixel's intensity value $I$ at location $(x,y)$ in frame $t$ of the video sequence as following:
\begin{equation}
e(x,y,t) = \| I(x,y,t) - f_{W}(I(x,y,t)) \|_2,
\label{eq:test}
\end{equation}
where $f_W$ is the learned model by the fully convolutional autoencoder. Given the reconstruction errors of the pixels of a frame $t$, we compute the reconstruction error of a frame by summing up all the pixel-wise errors: $e(t) = \sum_{(x,y)} e(x,y,t)$. We compute the regularity score $s(t)$ of a frame $t$ as follows:
\begin{equation}
s(t) = 1 - \frac{e(t) - \min_t e(t)}{\max_t e(t)}.
\label{eq:test1}
\end{equation}
For the autoencoder on the improved trajectory feature, we can simply replace $I(x,y)$ with $p(x,y)$ where $p(\cdot)$ is an improved trajectory feature descriptor of a patch that covers the location of $(x,y)$.


\section{Experiments}
\label{sec:exp}

We learn the model using multiple video datasets, totaling 1 hour 50 minutes, and evaluate our method both qualitatively and quantitatively.
We modify\footnote{\url{https://github.com/mhasa004/caffe}} and use Caffe~\cite{caffe14} for all of our experiments on NVIDIA Tesla K80 GPUs.

For qualitative analysis, we generate the most regular image from a video and visualize the pixel-level irregularity.
In addition, we show that the learned model based on the convolutional autoencoder can be used to forecast future frames and estimate past frames.

For quantitative analysis, we temporally segment the anomalous events in video and compare performance against the state of the arts.
Note that our model is not fine-tuned to one dataset. It is general enough to capture regularities across multiple datasets.

\subsection{Datasets}

We use three datasets to train and demonstrate our models.
They are curated for anomaly or abnormal event detection and are referred to as Avenue \cite{lu2013abnormal}, UCSD pedestrian \cite{mahadevan2010anomaly}, and Subway \cite{adam2008robust} datasets.
We describe the details of datasets in the supplementary material.

\subsection{Learning a General Model Across Datasets}

\begin{figure}[h]
	\centering
	\hspace{-1.5em}
	\resizebox{8.8cm}{!}{%
	\begin{tabular}{ccc}
		\includegraphics[scale=0.20]{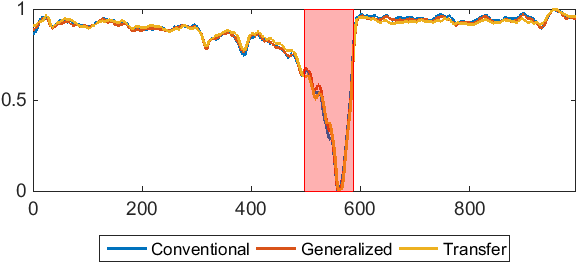}
		&\includegraphics[scale=0.20]{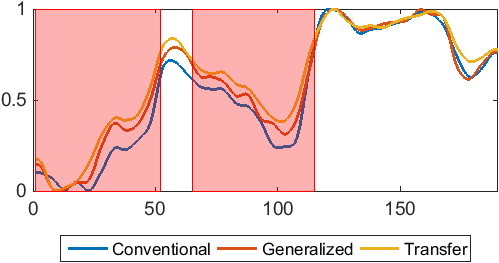}
		&\includegraphics[scale=0.20]{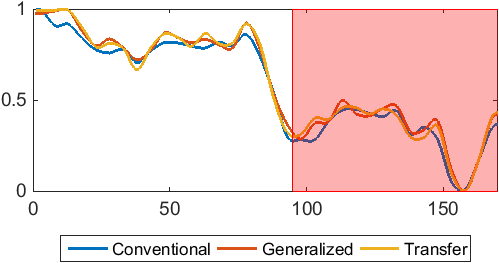}\\
		{\footnotesize CUHK Avenue-\# 15} 
		&{\footnotesize UCSD Ped1-\# 32}
		&{\footnotesize UCSD Ped2-\# 02}\\
		\multicolumn{3}{c}{\includegraphics[scale=0.18]{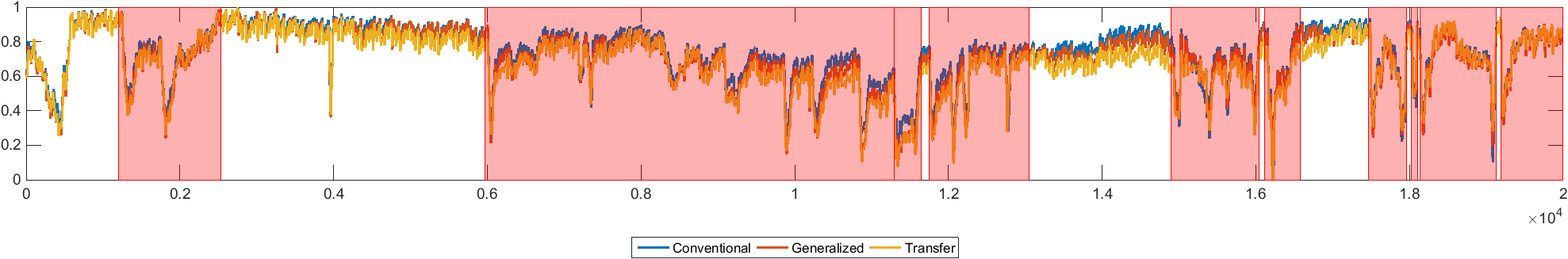}}\\
		\multicolumn{3}{c}{\footnotesize Subway Enter-\#1}\\
		\multicolumn{3}{c}{\includegraphics[scale=0.22]{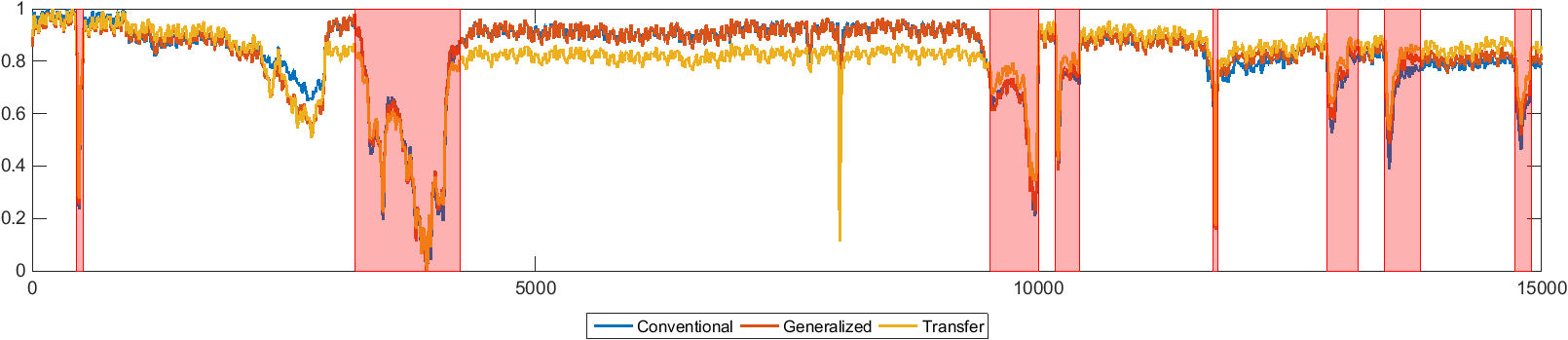}}\\
		\multicolumn{3}{c}{\footnotesize Subway-Exit-\#1}\\
	\end{tabular}
	}
	\caption{Generalizability of Models by Obtained Regularity Scores. `Conventional' is by a model trained on the \emph{specific target} dataset. `Generalized' is by a model trained on \emph{all} datasets. `Transfer' is by a model trained on \emph{all datasets except that specific target} datasets. Best viewed in zoom.}
	\label{fig:gen}
	\vspace{-1em}
\end{figure}

We compare the generalizability of the trained model using various training setups in terms of regularity scores obtained by each model in Fig. \ref{fig:gen}.
Blue (conventional) represents the score obtained by a model trained on the \emph{specific target} dataset.
Red (generalized) represents the score obtained by a model trained on \emph{all} datasets, which is the model we use for all other experiments.
Yellow (transfer) represents the score obtained by a model trained on \emph{all datasets except that specific target} dataset.
 Red shaded regions represent ground truth temporal segments of the abnormal events.

By comparing `conventional' and `generalized', we observe that the model is not degraded by other datasets.
At the same time, by comparing `transfer' and either `generalized' or `conventional', we observe that the model is not too much overfitted to the given dataset as it can generalize to \emph{unseen} videos in spite of potential dataset biases.
Consequently, we believe that the proposed network structure is well balanced between overfitting and underfitting.

\subsection{Visualizing Temporal Regularity}

The learned model measures the intensity of regularity up to pixel precision. 
We synthesize the most regular frame from the test video by collecting the pixels that have the highest regularity score by our convolutional autoencoder (conv-autoencoder) and autoencoder on improved trajectories (IT-autoencoder).

\begin{figure}[h]
	\centering
	\begin{tabular}{c}
		\includegraphics[scale=0.115]{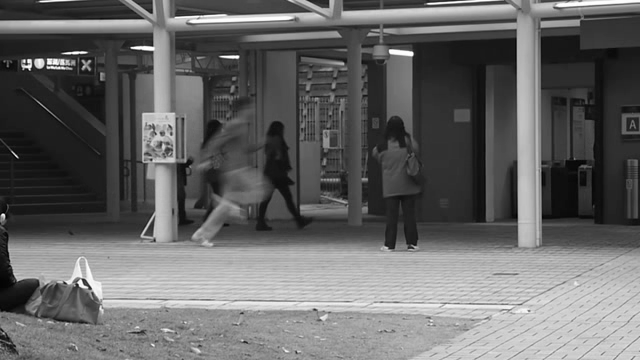}
		\includegraphics[scale=0.115]{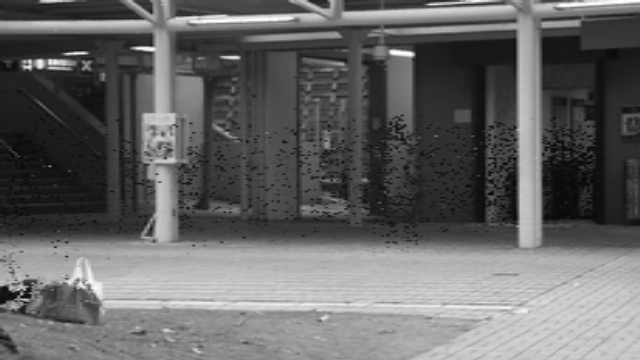}
		\includegraphics[scale=0.158]{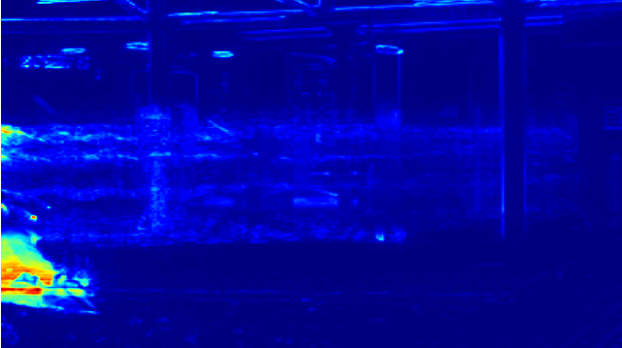}\\
		{\footnotesize Avenue Dataset}\\
		\includegraphics[scale=0.4]{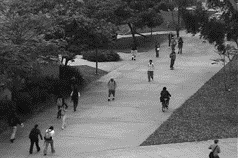}
		\includegraphics[scale=0.3]{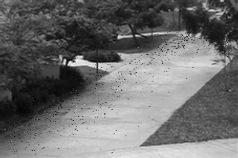} 
		\includegraphics[scale=0.155]{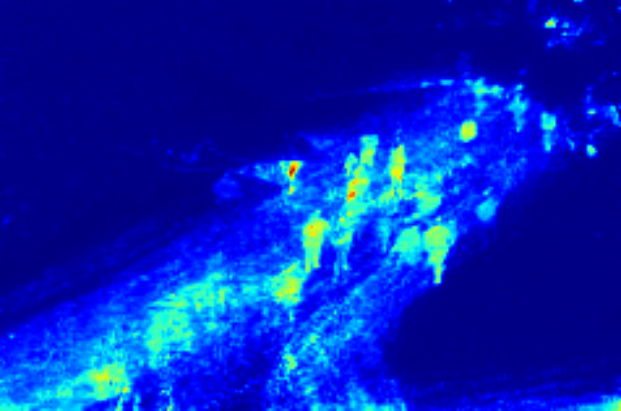} \\
		{\footnotesize UCSD Ped1 Dataset}\\
		\includegraphics[scale=0.265]{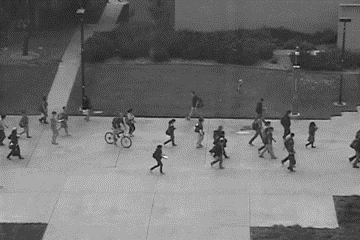}
		\includegraphics[scale=0.2]{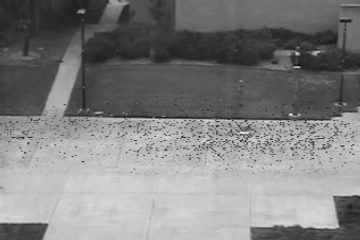}
		\includegraphics[scale=0.155]{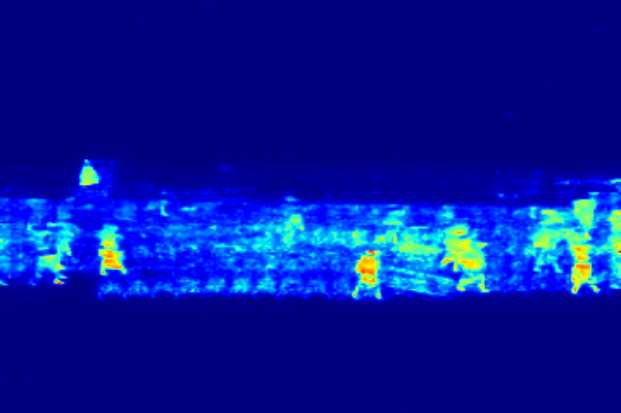}\\
		{\footnotesize UCSD Ped2 Dataset}\\
		\includegraphics[scale=0.186]{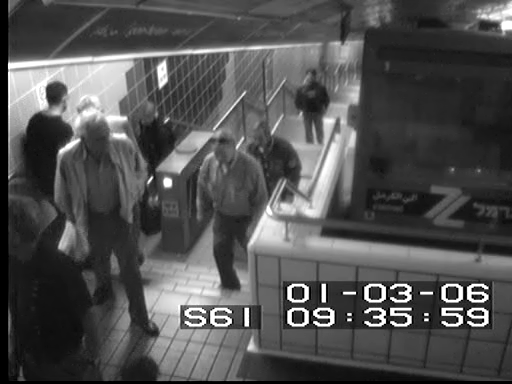}
		\includegraphics[scale=0.14]{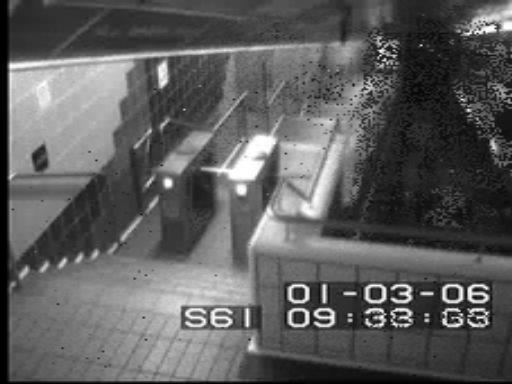}
		\includegraphics[scale=0.155]{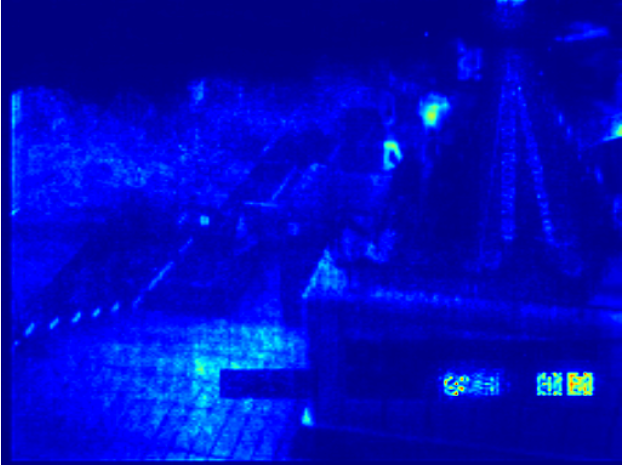}\\
		{\footnotesize Subway Exit Dataset}\\		
	\end{tabular}
	\caption{(Left) A sample irregular frame. (Middle) Synthesized regular frame. (Right) Regularity Scores of the frame. Blue represents regular pixel. Red represents irregular pixel.}
	\label{fig:super_unsuper}
\end{figure}

The first column of Fig.~\ref{fig:super_unsuper} shows sample images that contain irregular motions.
The second column shows the synthesized regular frame.
Each pixel of the synthesized image corresponds to the pixel for which reconstruction cost is minimum along the temporal dimension.
The right most column shows the corresponding regularity score.
Blue represents high score, red represents low.

\begin{figure}[h]
	\centering
	\begin{tabular}{cccc}
		\hspace{-1.5em}
		\includegraphics[scale=0.10]{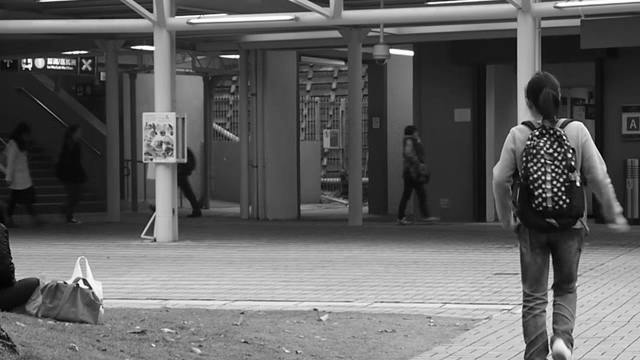}&\hspace{-1em}
		\includegraphics[scale=0.14]{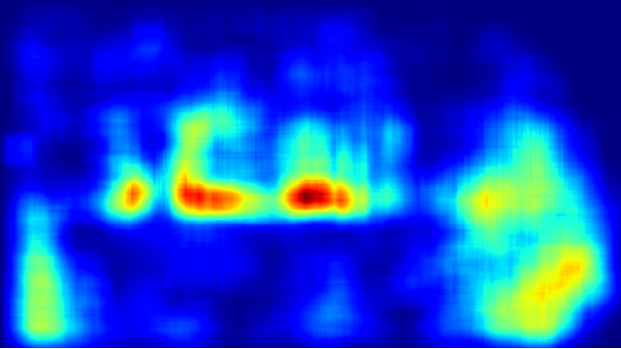}&\hspace{-1em}
		\includegraphics[scale=0.20]{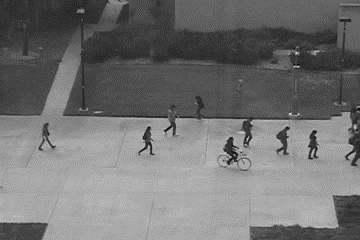}&\hspace{-1em}
		\includegraphics[scale=0.125]{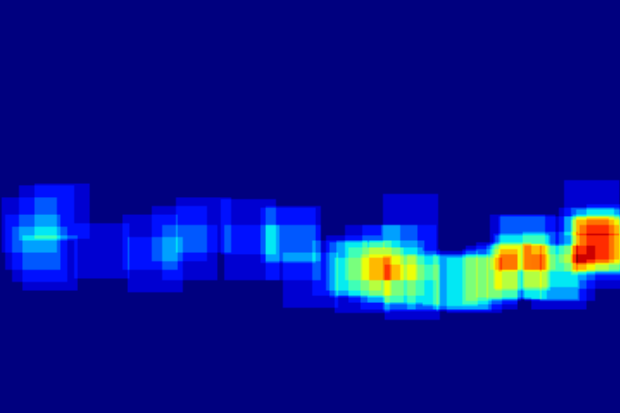}\\
		\multicolumn{2}{c}{\footnotesize Avenue Dataset}
		&\multicolumn{2}{c}{\footnotesize UCSD Ped2 Dataset}\\

	\end{tabular}
	\caption{Learned regularity by improved trajectory features. (Left) Frames with irregular motion. (Right) Learned regularity on the entire video sequence. Blue represents regular region. Red represents irregular region.}
	\vspace{-1em}
	\label{fig:super_it}
\end{figure}

Fig. \ref{fig:super_it} shows the results using IT-autoencoder. 
The left column shows the sample irregular frame of a video sequences, and the right column is the pixel-wise regularity score for that video sequence.
It captures irregularity to patch precision; thus the spatial location is not pixel-precise as obtained by conv-autoencoder.


%
%
%

\subsection{Predicting the Regular Past and the Future}

Our convolutional autoencoder captures temporal appearance changes since it takes a short video clip as input.
Using a clip that is blank except for the center frame, we can predict both near past and future frames of a regular video clip for the given center frame.

\begin{figure}[h]
	\centering
	\includegraphics[scale=0.43]{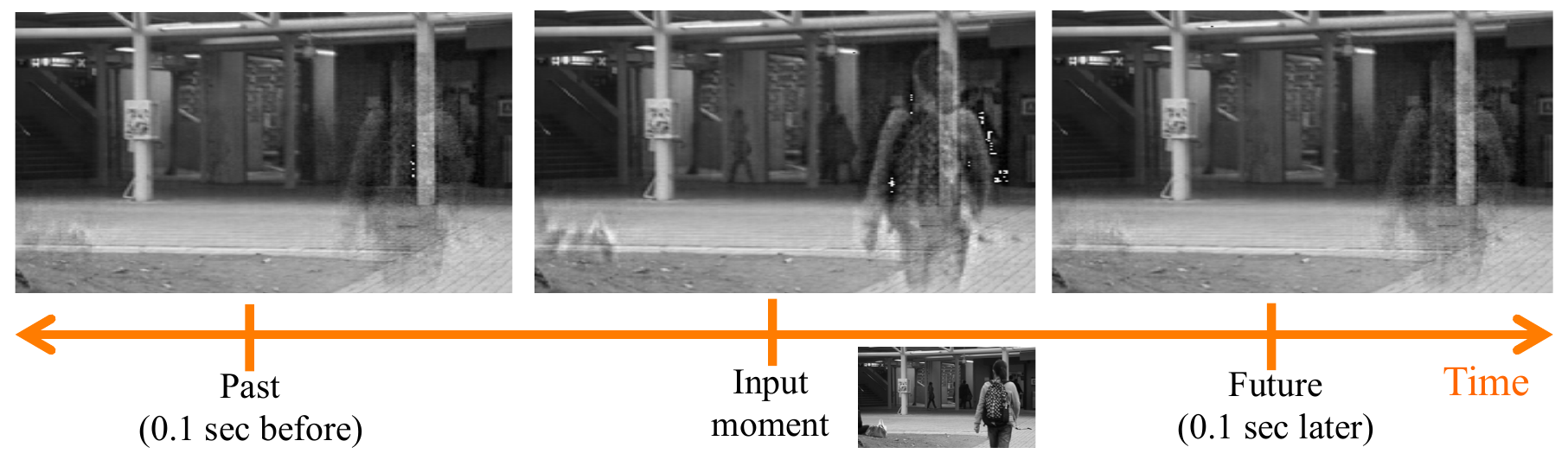}\\
	\includegraphics[scale=0.57]{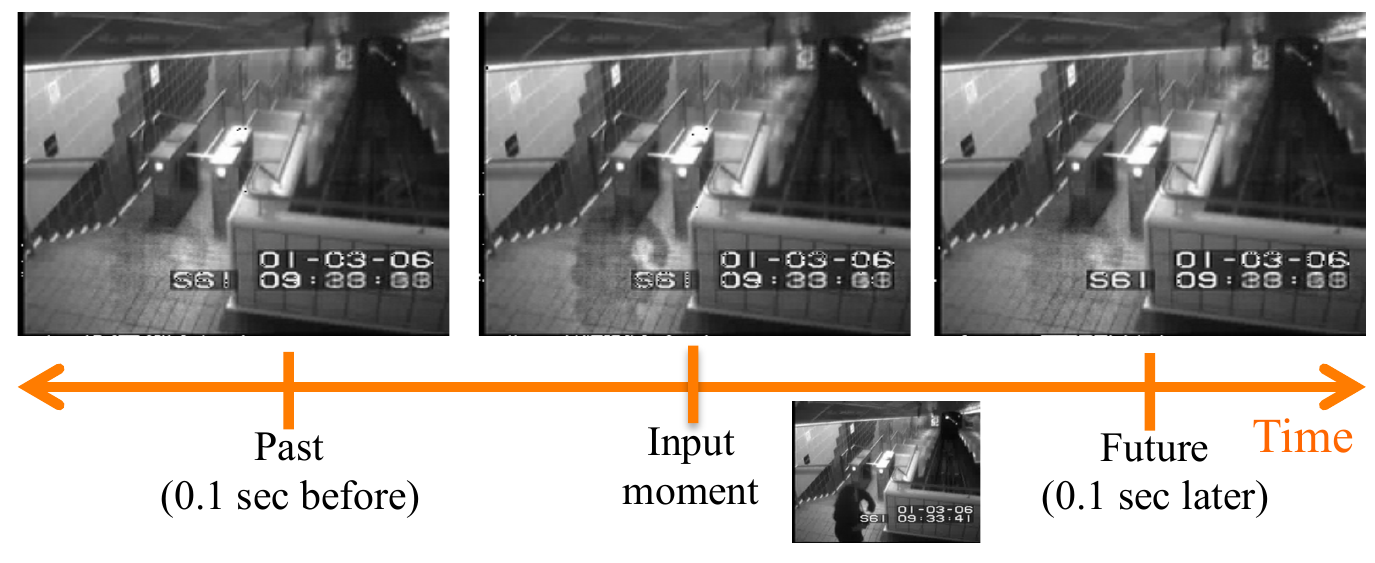}\\

	\caption{Synthesizing a video of regular motion from a single seed image (at the center). Upper: CUHK-Avenue. Bottom: Subway-Exit.}
	\vspace{-.5em}
	\label{fig:pred_fore}
\end{figure}

Given a single irregular image, we construct a temporal cube as the input to our network by padding other frames with all zero values. 
Then we pass the cube through our learned model to extrapolate the past and the future of that center frame.
Fig.~\ref{fig:pred_fore} shows some examples of generated videos. 
The objects in an irregular motion start appearing from the past and gradually disappearing in the future.

Since the network is trained with regular videos, it learns the regular motion patterns. With this experiment, we showed that the network can predict the regular motion of the objects in a given frame up to a few number of past and future frames.

\begin{figure*}[t]
	\centering
	\includegraphics[scale=0.45]{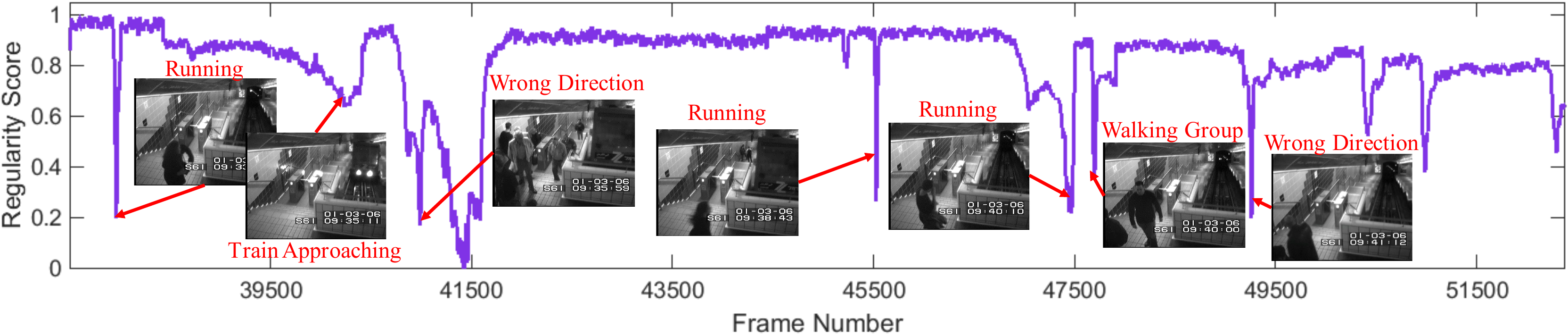}\\
	\vspace{2mm}
	\includegraphics[scale=0.55]{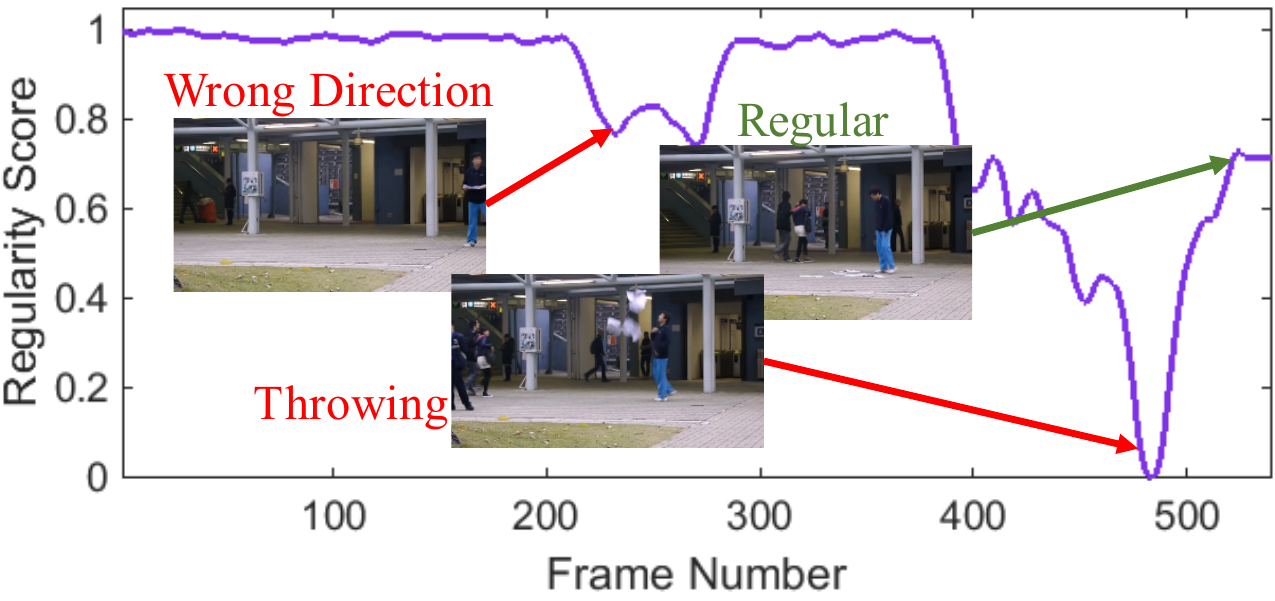}
	\includegraphics[scale=0.45]{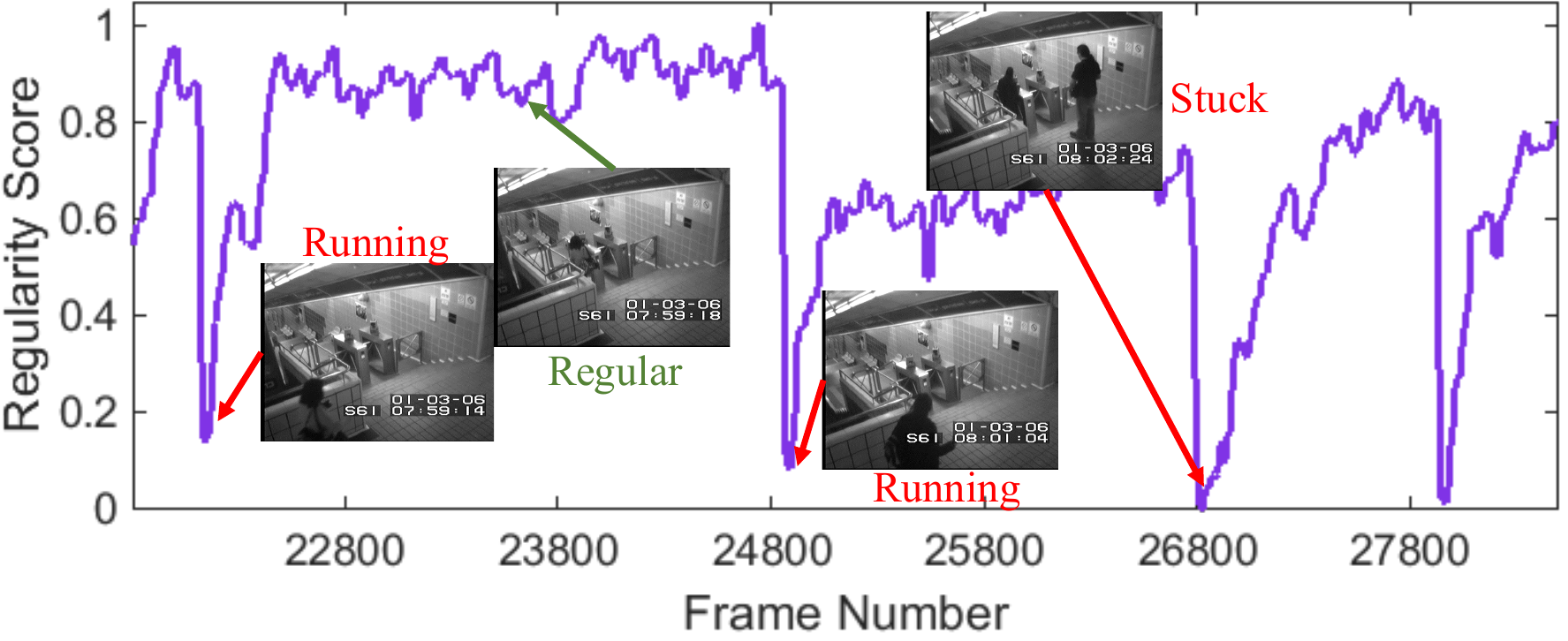}
	\caption{Regularity score (Eq.\ref{eq:test}) of each frame of three video sequences. (Top) Subway Exit, (Bottom-Left) Avenue, and (Bottom-Right) Subway Enter datasets. Green and red colors represent regular and irregular frames respectively.}
	\label{fig:plot_recon_cost}
\end{figure*}


\begin{table*}[t]
	\centering
	\resizebox{17.5cm}{!}{%
	\begin{tabular}{|c|c|c||c||c|c|c|c|c|c|}
		\hline
		\multicolumn{3}{|c||}{Dataset} & Regularity & \multicolumn{6}{c|}{Anomaly Detection}\\
		\hline
		 & \ & \# Regular & Conv-AE &  \# Anomalous & \multicolumn{3}{c|}{Correct Detection / False Alarm} & \multicolumn{2}{c|}{AUC/EER} \\
		\cline{6-10} 
		Name & \# Frames & Frames & Correct Detect / FA &  Event & Conv-AE & IT-AE & State of the art & Conv-AE & State of the art\\
		\hline
		\hline 
		CUHK Avenue & $15,324$ & \centering $11,504$ & \centering $11,419/355$ & $47$ & $45/4$ & $43/8$ & $12/1$ (Old Dataset) \cite{lu2013abnormal} & $70.2 / 25.1$ & N/A\\
		\hline
		UCSD Ped1 & $7,200$ & \centering  $3,195$ & \centering $3,135/310$ & $40$ & $38/6$ & $36/11$ & N/A & $81.0 / 27.9$ & $92.7 / 16.0$~\cite{saligramaC12}\\
		\hline
		UCSD Ped2 & $2,010$ & \centering $374$ & \centering $374/50$ & $12$ & $12/1$ & $12/3$ & N/A & $90.0 / 21.7$ & $90.8 /16.0$~\cite{xu2015learning}\\
		\hline
		Subway Entrance & $121,749$ & \centering  $119,349$ & \centering $112,188/4,154$ & $66$ & $61/15$ &  $55/17$ & $57/4$ \cite{lu2013abnormal} & $94.3 / 26.0$ & N/A\\
		\hline
		Subway Exit & $64,901$ & \centering $64,181$ & \centering $62,871/1,125$ & $19$ & $17/5$ & $17/9$ & $19/2$ \cite{lu2013abnormal} & $80.7 / 9.9$ & N/A\\
		\hline
	\end{tabular}%
	}
	\vspace{2mm}
	\caption{Comparing abnormal event detection performance. AE refers to auto-encoder. IT refers to improved trajectory.}
	\vspace{-1em}
	\label{table:anomaly_regularity}
\end{table*}

\subsection{Anomalous Event Detection}

As our model learns the temporal regularity, it can be used for detecting anomalous events in a weakly supervised manner.
Fig.~\ref{fig:plot_recon_cost} shows the regularity scores as a function of frame number.
Table~\ref{table:anomaly_regularity} compares the anomaly detection accuracies of our autoencoders against state-of-the-art methods.
To the best of our knowledge, there are no correct detection or false alarm results reported for UCSD Ped1 and Ped2 datasets in the literature. 
We provide the EER and AUC measures from \cite{saligramaC12} for reference.
Additionally, the state-of-the-art results for the avenue dataset from \cite{lu2013abnormal} are not directly comparable as it is reported on the old version of the Avenue dataset that is smaller than the current version.

We find the local minimas in the time series of regularity scores to detect abnormal events. 
However, these local minima are very noisy and not all of them are meaningful local minima.
We use the persistence1D \cite{persistence1D} algorithm to identify meaningful local minima and span the region with a fixed temporal window (50 frames) and group nearby expanded local minimal regions when they overlap to obtain the final abnormal temporal regions. 
Specifically, if two local minima are within fifty frames of one another, they are considered to be a part of same abnormal event. 
We consider a detected abnormal region as a correct detection if it has at least fifty percent overlap with the ground truth.

Our model outperforms or performs comparably to the state-of-the-art abnormal event detection methods but with a few more false alarms.
It is because our method identifies any deviations from regularity, many of which have not been annotated as abnormal events in those datasets while competing approaches focused on the identification of abnormal events.
For example, in the top figure of Fig. \ref{fig:plot_recon_cost}, the `running' event is detected as an irregularity due to its unusual motion pattern by our model, but in the ground truth it is a normal event and considered as a false alarm during evaluation.

\subsection{Filter Responses}
\begin{figure}[h]
	\centering
		\includegraphics[scale=0.21]{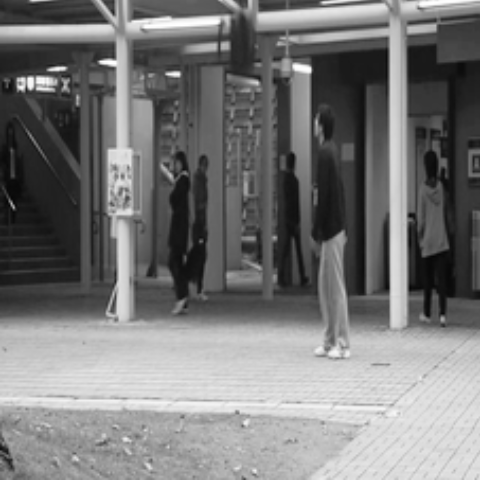}
		\includegraphics[scale=0.2]{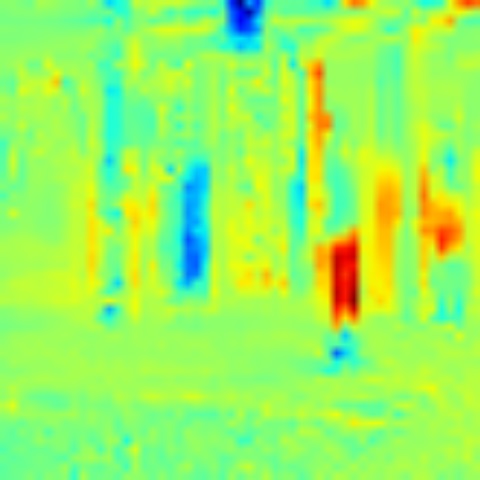}
		\includegraphics[scale=0.2]{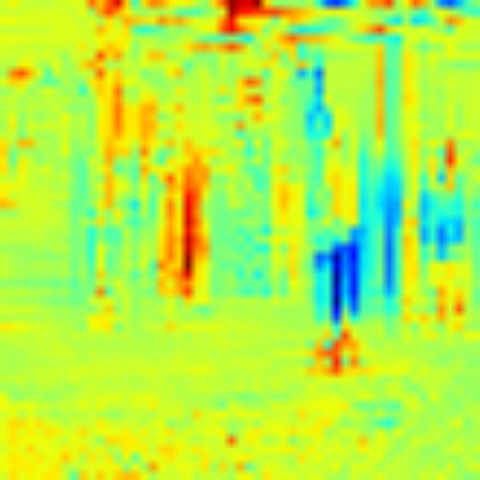}\\
		{\footnotesize \hspace{-3.5em}(a) Input data frame \hspace{5.5em} (b) Conv1 filter responses.} \\ 
		\vspace{2mm}
		\includegraphics[scale=0.15]{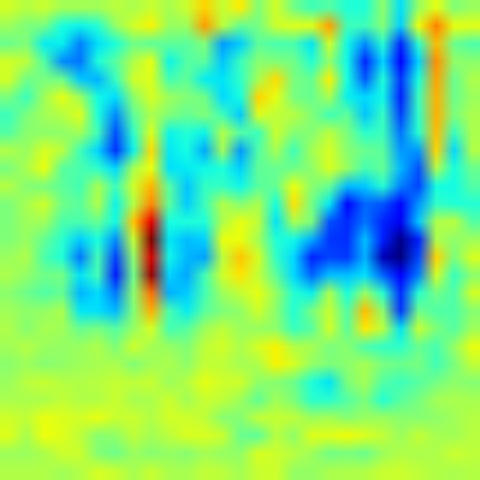}
		\includegraphics[scale=0.15]{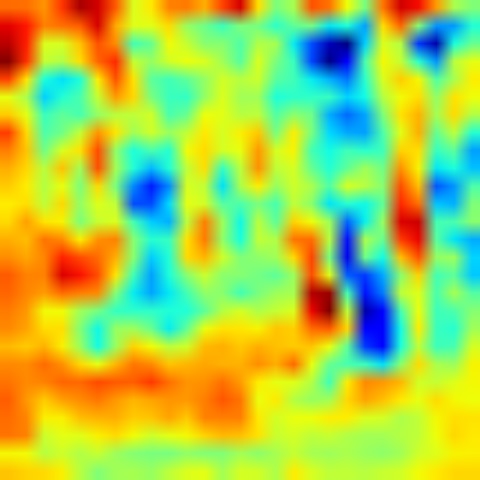}\hspace{.5em}
		\includegraphics[scale=0.15]{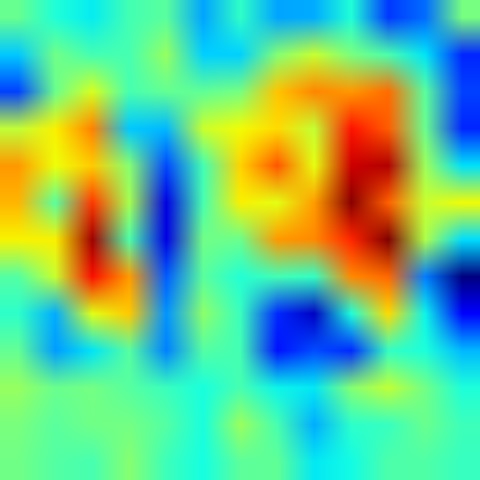}
		\includegraphics[scale=0.15]{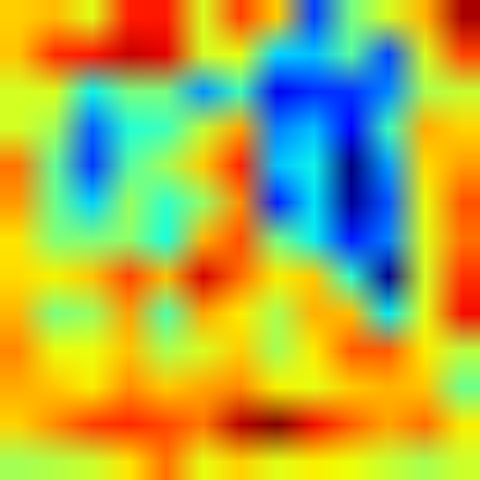}\\
		{\footnotesize (c) Conv2 filter responses. \hspace{4em} (d) Conv3 filter responses.} \\
	\vspace{1mm}
	\caption{Filter responses of the convolutional autoencoder trained on the Avenue dataset. Early layers (conv1) captures fine grained regular motion pattern whereas the deeper layers (conv3) captures higher level information.}
	\vspace{-1em}
	\label{fig:filter_vis}
\end{figure}

We visualize some of the learned filter responses of our model on Avenue datasets in Fig.~\ref{fig:filter_vis}.
The first row visualizes one channel of the input data and two filter responses of the conv1 layer. 
These two filters show completely opposite responses to the irregular object - the bag in the top of the frame. 
The first filter provides very low response (blue color) to it, whereas the second filter provides very high response (red color).
The first filter can be described as the filter that detects regularity, whereas the second filter detects irregularity.
All other filters show similar characteristics.
The second row of Fig.~\ref{fig:filter_vis} shows the responses of the filters from conv2 and conv3 layers respectively.

Additional results can be found in the supplementary material. Data, codes, and videos are available online\footnote{\url{http://www.ee.ucr.edu/~mhasan/regularity.html}}.

\section{Conclusion}
\label{sec:conclusion}

We present a method to learn regular patterns using autoencoders with limited supervision.
We first take advantage of the conventional spatio-temporal local features and learn a fully connected autoencoder.
Then, we build a fully convolutional autoencoder to learn both the local features and the classifiers in a single learning framework.
Our model is generalizable across multiple datasets even with potential dataset biases.
We analyze our learned models in a number of ways such as visualizing the regularity in frames and pixels and predicting a regular video of past and future given only a single image. 
For quantitative analysis, we show that our method performs competitively to the state-of-the-art anomaly detection methods.

{
\small
\bibliographystyle{IEEEtran}
\bibliography{bibs/activity,bibs/anomaly,bibs/cnn,bibs/ppolon_out}
}

\clearpage
\onecolumn

\begin{center} 
{\bf \Large Supplementary Materials}
\end{center}
\vspace{1em}

\begin{center}
	{\bf \large Table of Contents}
\end{center}
\begin{table*}[h]
	\vspace{-5mm}
	\centering
	\begin{tabular}{|r|p{12cm}|}
		\hline
		Section & Contents\\
		\hline \hline \hline
		\ref{sec:datasets} & {\bf Dataset Details} \\
		\hline \hline
		\ref{sec:temp_regularity} & {\bf Learned Temporal Regularity} \\
		\hline
		\ref{sec:temp_regularity_avenue} & CUHK Avenue \\
		\hline
		\ref{sec:temp_regularity_ped1} & UCSD Ped1 \\
		\hline
		\ref{sec:temp_regularity_ped2} & UCSD Ped2\\
		\hline
		\ref{sec:temp_regularity_enter} & Subway Enter \\
		\hline
		\ref{sec:temp_regularity_exit} & Subway Exit \\
		\hline \hline
		 \ref{sec:obj_det} & {\bf Object Detection in Irregular Motion} \\
		\hline
		\ref{sec:obj_det_avenue} & CUHK Avenue\\
		\hline
		\ref{sec:obj_det_ped1} & UCSD Ped1 \\
		\hline
		\ref{sec:obj_det_ped2} & UCSD Ped2 \\
		\hline
		\ref{sec:obj_det_enter} & Subway Enter \\
		\hline
		\ref{sec:obj_det_exit} & Subway Exit \\
		\hline \hline
		\ref{sec:frame_pred} & {\bf Predicting Past and Future Regular Frames}  \\
		\hline
		\ref{sec:frame_pred_avenue} & CUHK Avenue\\
		\hline
		\ref{sec:frame_pred_ped1} & UCSD Ped1 \\
		\hline
		\ref{sec:frame_pred_ped2} & UCSD Ped2 \\
		\hline
		\ref{sec:frame_pred_enter} & Subway Enter \\
		\hline
		\ref{sec:frame_pred_exit} & Subway Exit \\
		\hline \hline
		\ref{sec:anomaly} & {\bf Anomalous Event Detection and Generalization Analysis on Multiple Datasets} \\
		\hline
		\ref{sec:anomaly_avenue} & CUHK Avenue\\
		\hline
		\ref{sec:anomaly_ped1} & UCSD Ped1 \\
		\hline
		\ref{sec:anomaly_ped2} & UCSD Ped2 \\
		\hline
		\ref{sec:anomaly_enter} & Subway Enter \\
		\hline
		\ref{sec:anomaly_exit} & Subway Exit \\
		\hline \hline
		\ref{sec:filter_res_vis} & {\bf Filter Response Visualization} \\
		\hline
		\ref{sec:filter_res_vis_avenue} & CUHK Avenue\\
		\hline
		\ref{sec:filter_res_vis_ped1} & UCSD Ped1 \\
		\hline
		\ref{sec:filter_res_vis_ped2} & UCSD Ped2 \\
		\hline
		\ref{sec:filter_res_vis_enter} & Subway Enter \\
		\hline
		\ref{sec:filter_res_vis_exit} & Subway Exit \\
		\hline \hline
		\ref{sec:filter_w_vis}  & {\bf Filter Weights Visualization} \\
		\hline
	\end{tabular}
\end{table*}

\clearpage


\section{Dataset Details}
\label{sec:datasets}

We use three challenging datasets to demonstrate our methods. 
They are curated for anomaly or abnormal event detection and are referred to as Avenue \cite{lu2013abnormal}, UCSD pedestrian \cite{mahadevan2010anomaly}, and Subway \cite{adam2008robust} datasets.
We describe the details of datasets in the supplementary material.

\paragraph{\bf Avenue.}
There are total 16 training and 21 testing video sequences. 
Each of the sequences is short; about 1 to 2 minutes long.
The total number of training frames is $15,328$ and testing frame is $15,324$. 
Resolution of each frame is $640\times360$ pixels.

\paragraph{\bf UCSD Pedestrian.}
This dataset has two different scenes - Ped1 and Ped2.

\vspace{1em}
\noindent {\bf UCSD-Ped1.} It has 34 short clips for training, and another 36 clips for testing. 
All testing video clips have frame-level ground truth labels. 
Each clip has $200$ frames, with a resolution of $238\times 158$ pixels.

\vspace{1em}
\noindent {\bf UCSD-Ped2.} It has 16 short clips for training, and another 12 clips for testing. 
Each clip has 150 to 200 frames, with a resolution of $360\times240$ pixels.

\paragraph{\bf Subway.}
The videos are taken from two surveillance cameras in a subway station. 
One monitors the exit and the other monitors the entrance. 
In both videos, there are roughly 10 people walking around in a frame. The resolution is $512\times384$ pixels.

\vspace{1em}
\noindent {\bf Subway-Entrance.}
It is 1 hour 36 minutes long with $144,249$ frames in total. 
There are $66$ unusual events of five different types: (a) walking in the wrong direction (WD); (b) no payment (NP); (c) loitering (LT); (d) irregular interactions between people (II) and (e) misc, including sudden stop, running fast.

\vspace{1em}
\noindent {\bf Subway-Exit.}
It is 43 minutes long with $64,901$ frames. Three types of unusual events are defined in the subway exit video: (a) walking in the wrong direction (WD), (b) loitering near the exit (LT), and (c) miscellaneous, including suddenly stop and look around, janitor cleaning the wall, someone gets off the train and gets on again very soon. 
In total, 19 unusual events are defined as ground truth.

\begin{center}
	\hyperlink{page.11}{Go to Table of Contents}
\end{center}
\clearpage


\section{Learned Temporal Regularity}
\label{sec:temp_regularity}

In Section 4.2 in the main paper, we visualize the temporal regularity by 1) synthesizing the regular frame and 2) visualizing accumulated regularity score within a video as a heat-map obtained by convolutional autoencoder (conv-autoencoder).
Here, we present more examples per each dataset with a heat-map obtained by the improved trajectory based autoencoder (IT-autoencoder) for comparison.
Compared to conv-autoencoder's regular score, the regular score by IT-autoencoder is up to patch precision and cannot capture the regularity well.

\subsection{CUHK Avenue Dataset}
\label{sec:temp_regularity_avenue}

\begin{figure}[h]
	\centering
	\begin{tabular}{c}
		\includegraphics[scale=0.17]{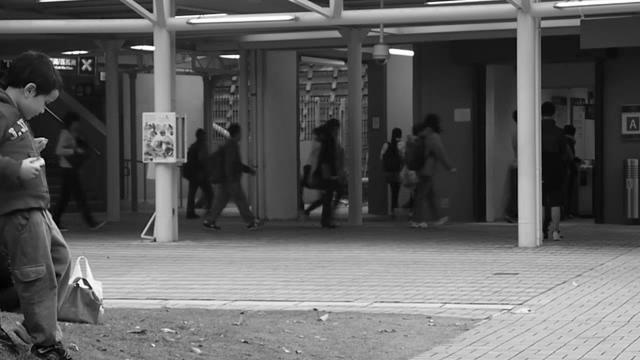}
		\includegraphics[scale=0.17]{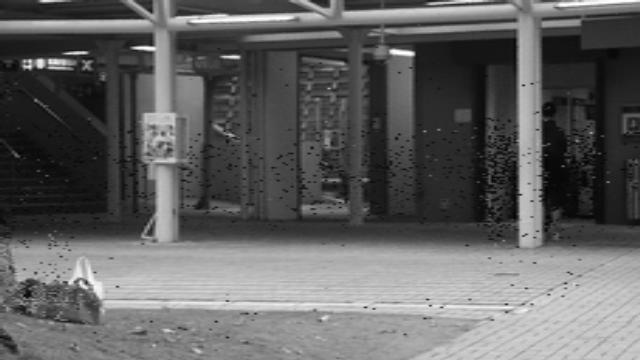}
		\includegraphics[scale=0.175]{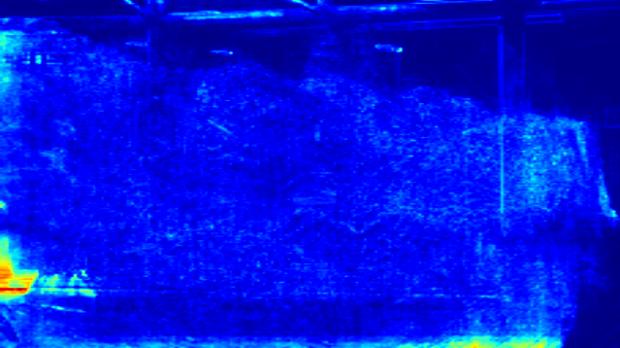}
		\includegraphics[scale=0.175]{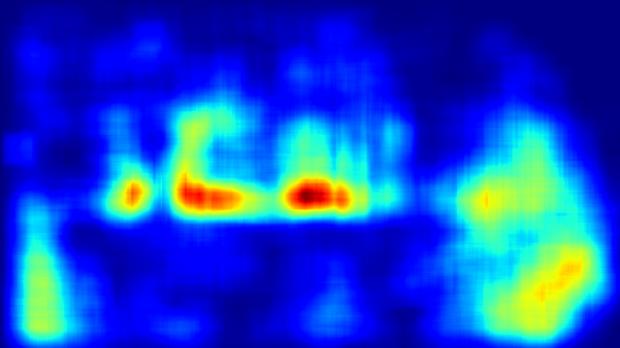}\\
		{\footnotesize Video \# 1} \\
		\includegraphics[scale=0.17]{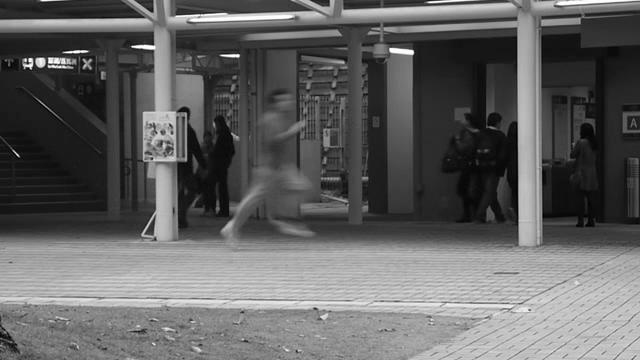}
		\includegraphics[scale=0.17]{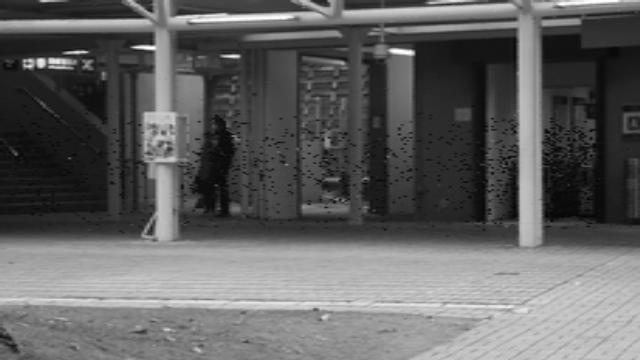}
		\includegraphics[scale=0.175]{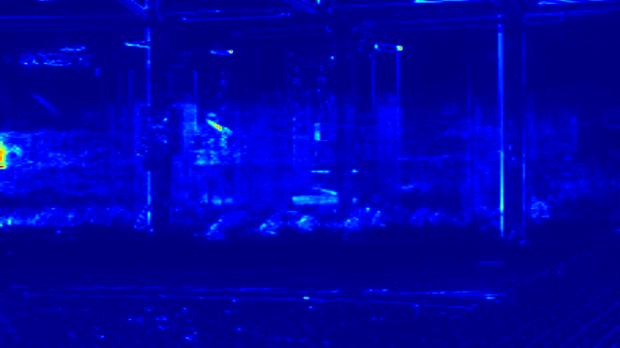}
		\includegraphics[scale=0.175]{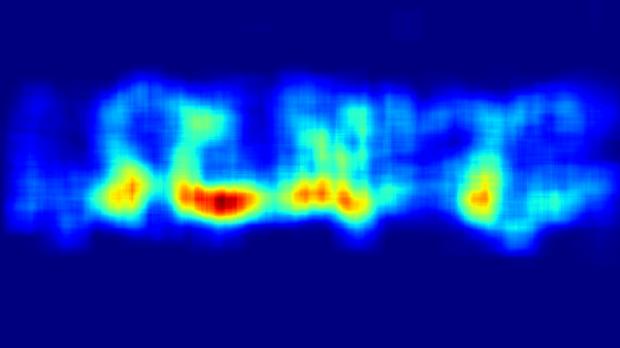}\\
		{\footnotesize Video \# 3} \\
		\includegraphics[scale=0.17]{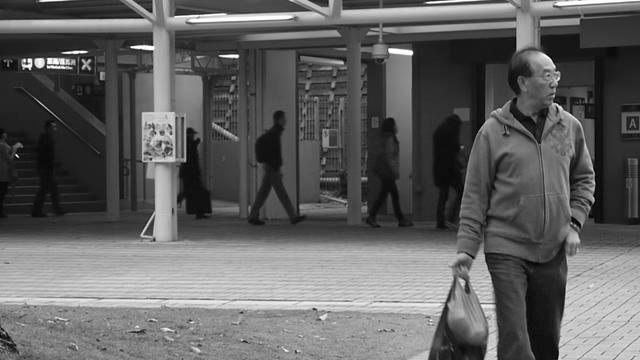}
		\includegraphics[scale=0.17]{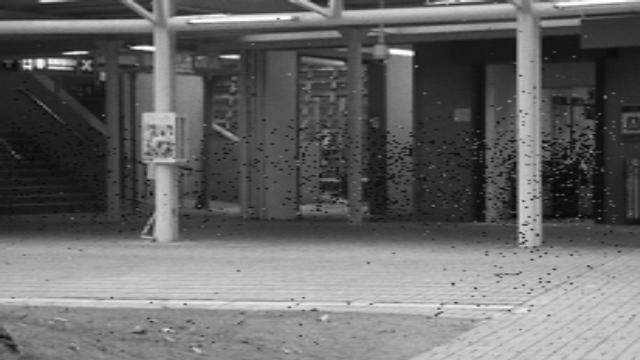}
		\includegraphics[scale=0.175]{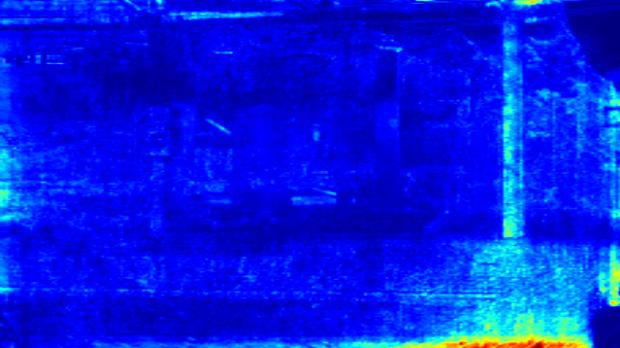}
		\includegraphics[scale=0.175]{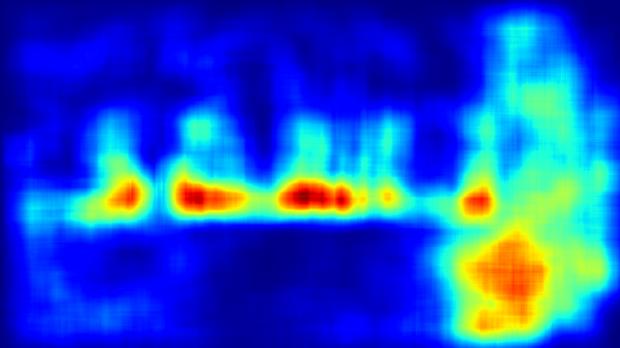}\\
		{\footnotesize Video \# 6} \\
		\includegraphics[scale=0.17]{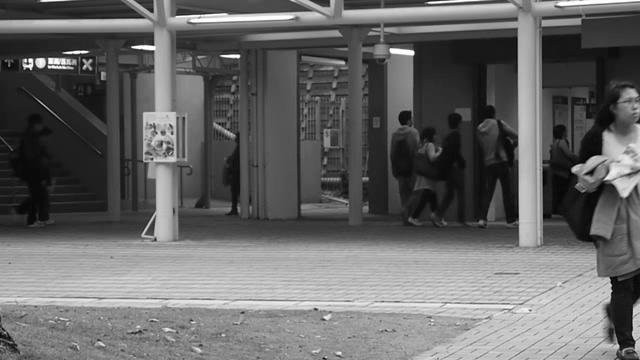}
		\includegraphics[scale=0.17]{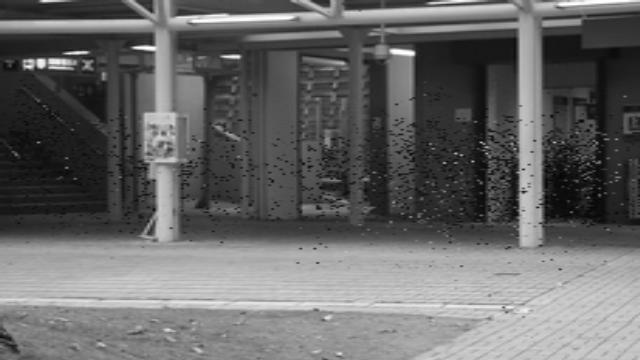}
		\includegraphics[scale=0.175]{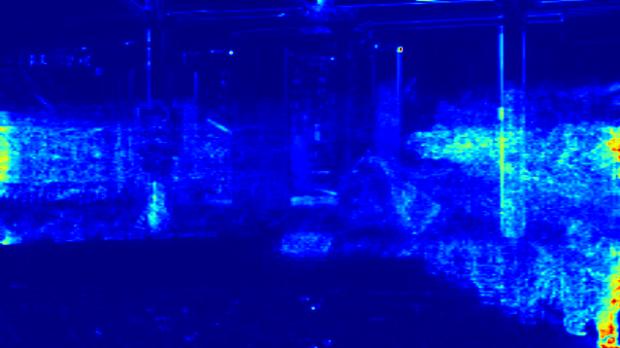}
		\includegraphics[scale=0.175]{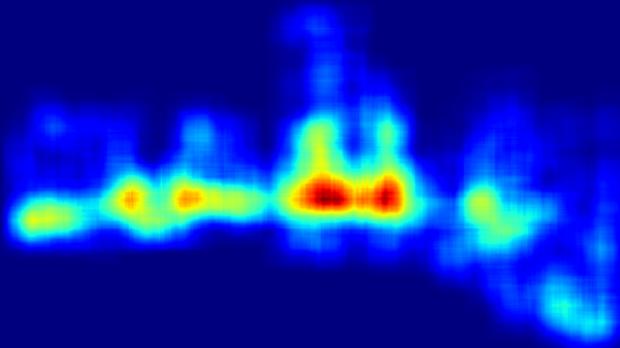}\\
		{\footnotesize Video \# 11} \\
		\includegraphics[scale=0.17]{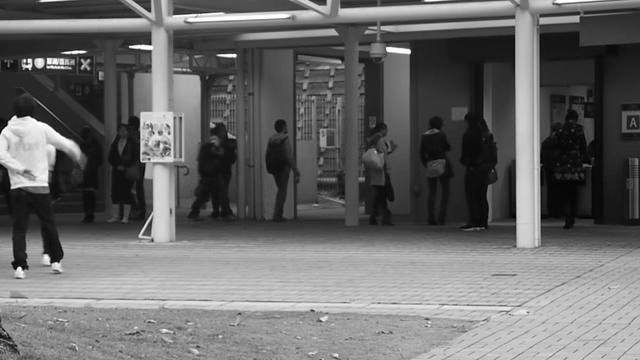}
		\includegraphics[scale=0.17]{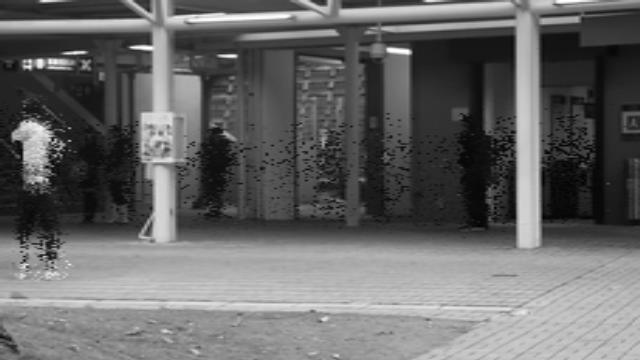}
		\includegraphics[scale=0.175]{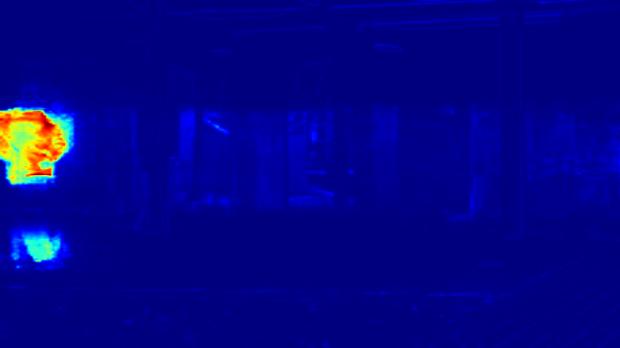}
		\includegraphics[scale=0.175]{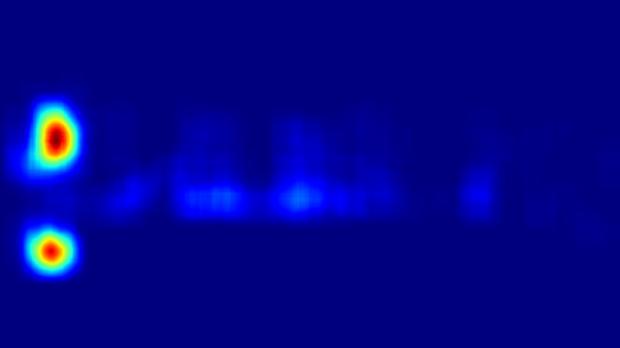}\\
		{\footnotesize Video \# 17} \\
		\hline \vspace{-.5em}\\
		\includegraphics[scale=0.17]{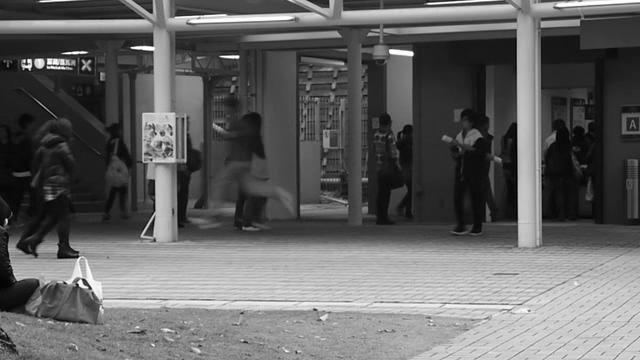}
		\includegraphics[scale=0.17]{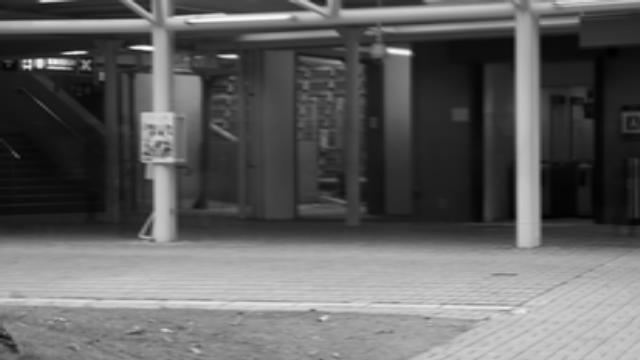}
		\includegraphics[scale=0.175]{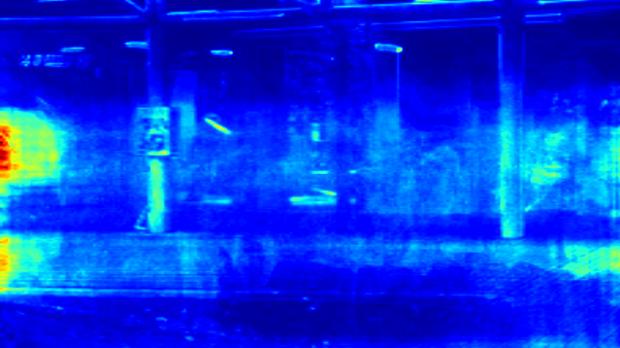}
		\includegraphics[scale=0.175]{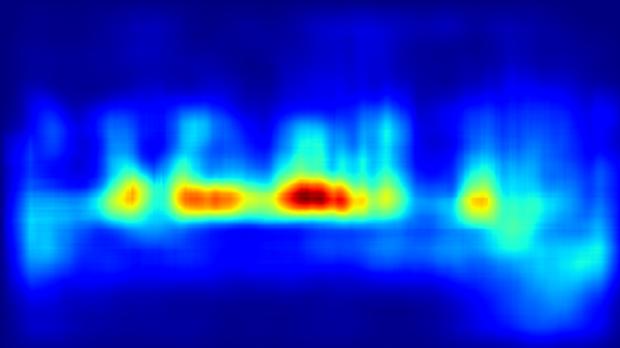}\\
		{\footnotesize All Videos}
	\end{tabular}
	\label{fig:temp_regularity_avenue}	
	\caption{(Left) A sample irregular frame. (Second) A synthesized regular frame obtained by the pixel value of lowest reconstruction score across all frames of a video. (Third) Accumulated regularity score obtained by convolutional-autoencoder. (Fourth) Accumulated regularity score obtained by IT-autoencoder.}
\end{figure}

\vspace{-1.5em}
\begin{center}
	\hyperlink{page.11}{Go to Table of Contents}
\end{center}
\clearpage

\subsection{UCSD Ped1}
\label{sec:temp_regularity_ped1}

\begin{figure}[h]
	\centering
	\begin{tabular}{c}
		\includegraphics[scale=0.47]{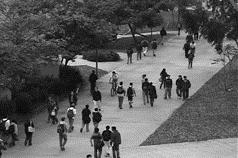}
		\includegraphics[scale=0.47]{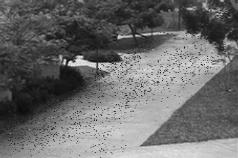}
		\includegraphics[scale=0.18]{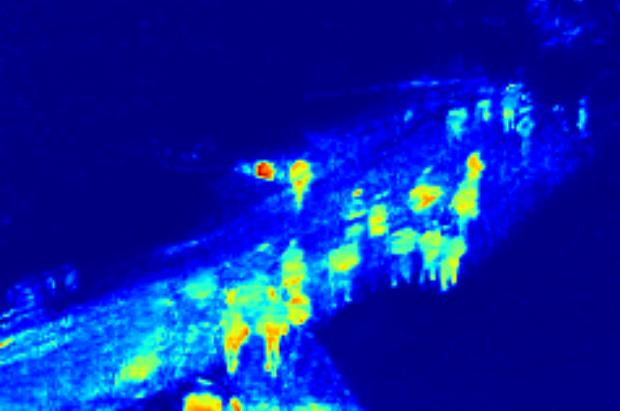}
		\includegraphics[scale=0.18]{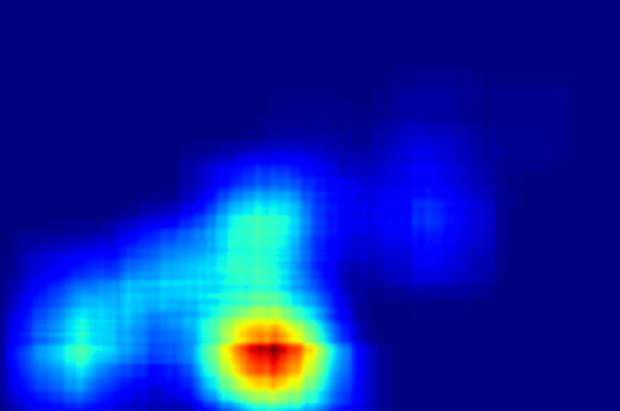}\\
		{\footnotesize Video \# 3} \\
		\includegraphics[scale=0.47]{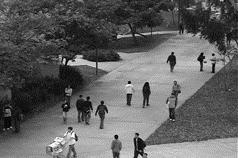}
		\includegraphics[scale=0.47]{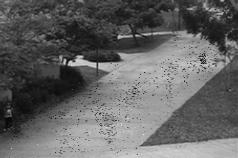}
		\includegraphics[scale=0.18]{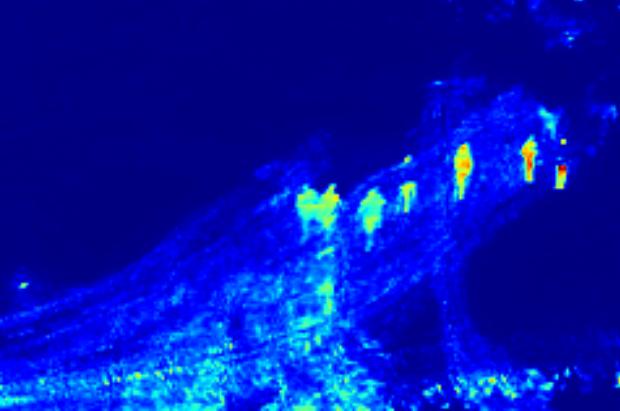}
		\includegraphics[scale=0.18]{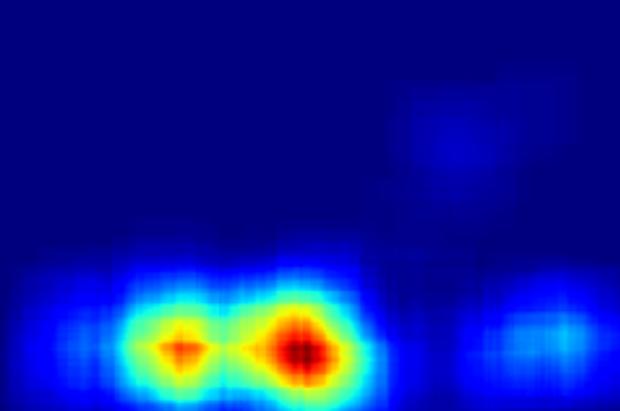}\\
		{\footnotesize Video \# 13} \\
		\includegraphics[scale=0.47]{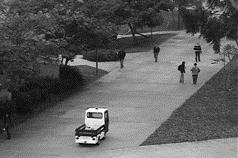}
		\includegraphics[scale=0.47]{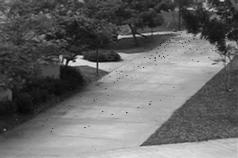}
		\includegraphics[scale=0.18]{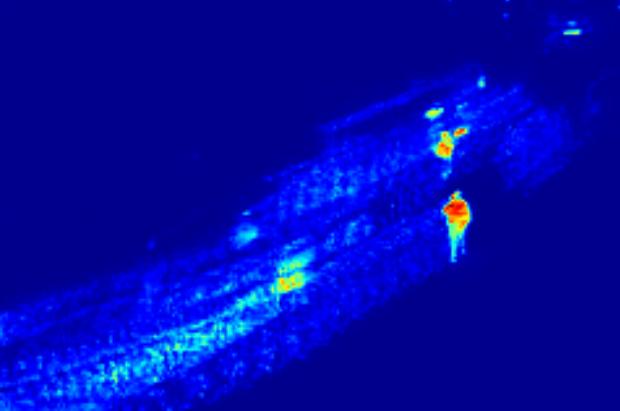}
		\includegraphics[scale=0.18]{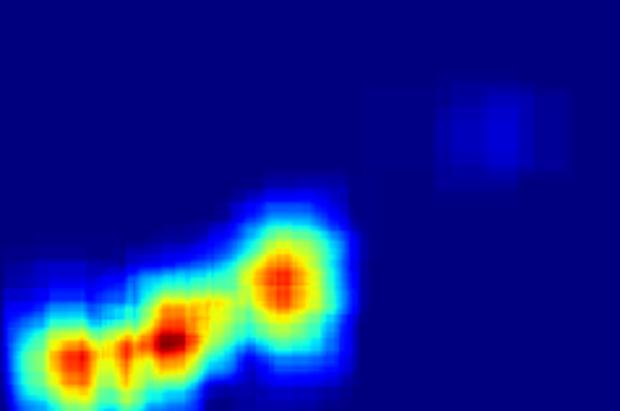}\\
		{\footnotesize Video \# 25} \\
		\includegraphics[scale=0.47]{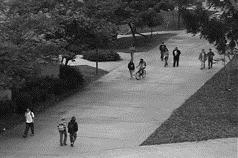}
		\includegraphics[scale=0.47]{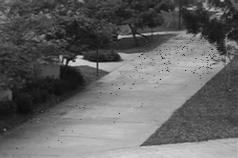}
		\includegraphics[scale=0.18]{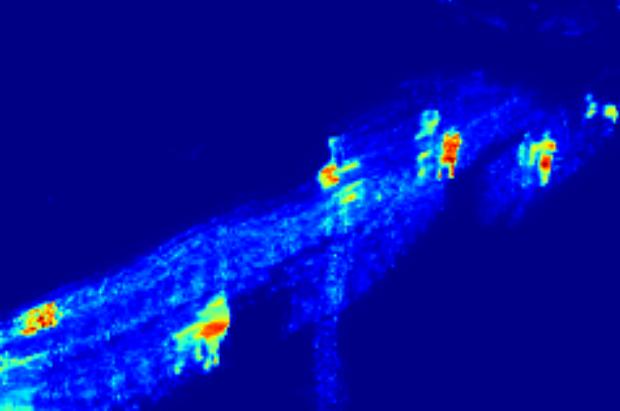}
		\includegraphics[scale=0.18]{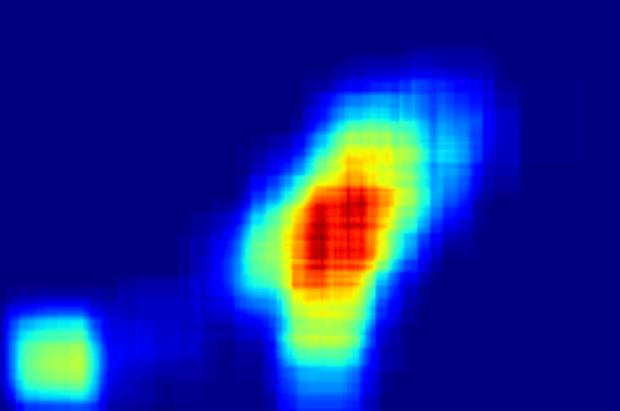}\\
		{\footnotesize Video \# 26} \\
		\includegraphics[scale=0.47]{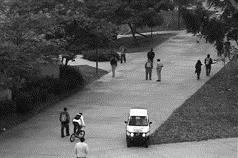}
		\includegraphics[scale=0.47]{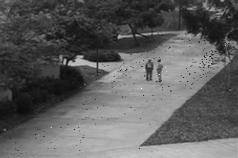}
		\includegraphics[scale=0.18]{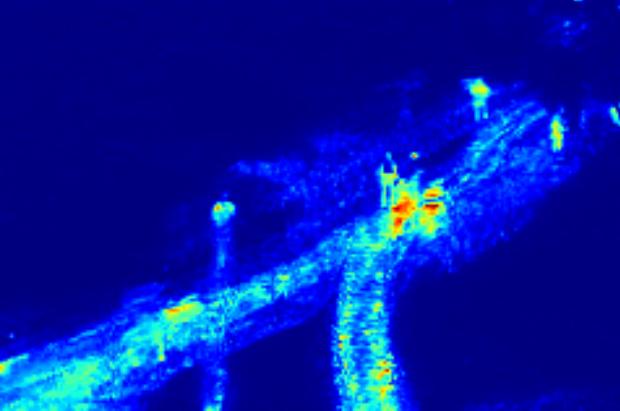}
		\includegraphics[scale=0.18]{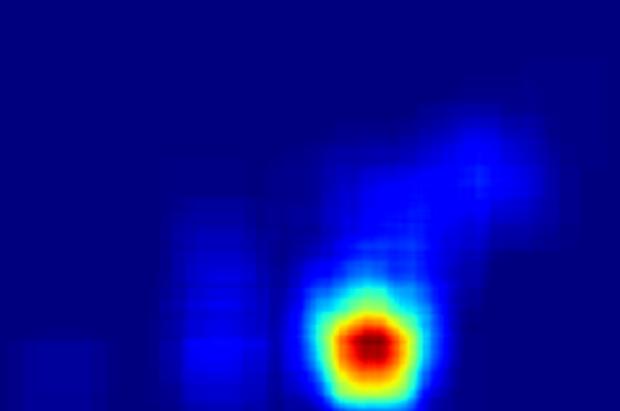}\\
		{\footnotesize Video \# 36} \\
		\hline \vspace{-.5em}\\
		\includegraphics[scale=0.47]{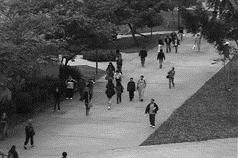}
		\includegraphics[scale=0.47]{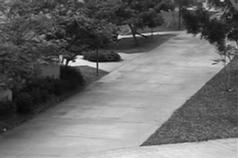}
		\includegraphics[scale=0.18]{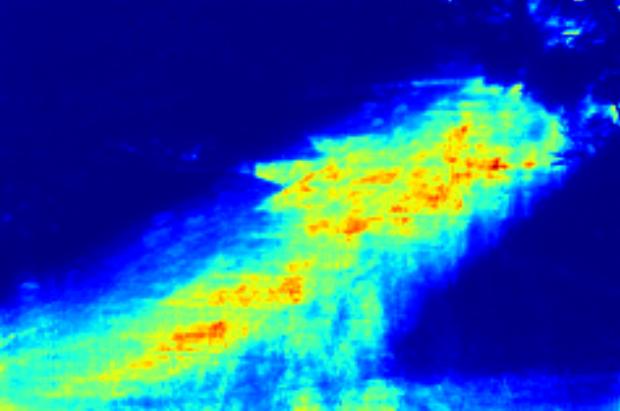}
		\includegraphics[scale=0.18]{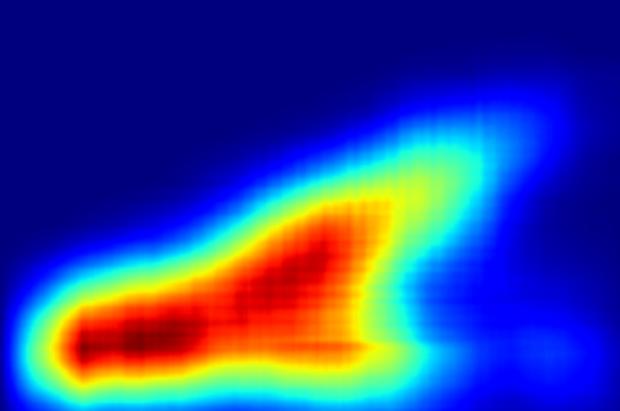}\\
		{\footnotesize All Videos} \\
	\end{tabular}
		\caption{Same layout in all figures in Section \ref{fig:temp_regularity_avenue}. Especially, in video 36, we can observe the trajectory of a SUV in the heatmap of accumulated regular score (third).}
		\label{fig:temp_regularity_ped1}
\end{figure}
\vspace{-1em}
\begin{center}
	\hyperlink{page.11}{Go to Table of Contents}
\end{center}
\clearpage

\subsection{UCSD Ped2}
\label{sec:temp_regularity_ped2}

\begin{figure}[h]
	\centering
	\begin{tabular}{c}
		\includegraphics[scale=0.31]{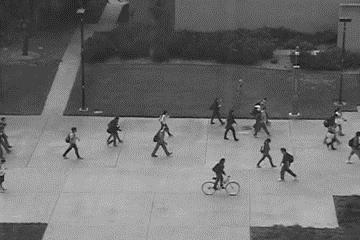}
		\includegraphics[scale=0.31]{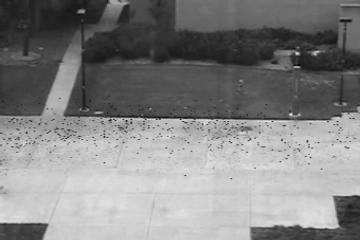}
		\includegraphics[scale=0.18]{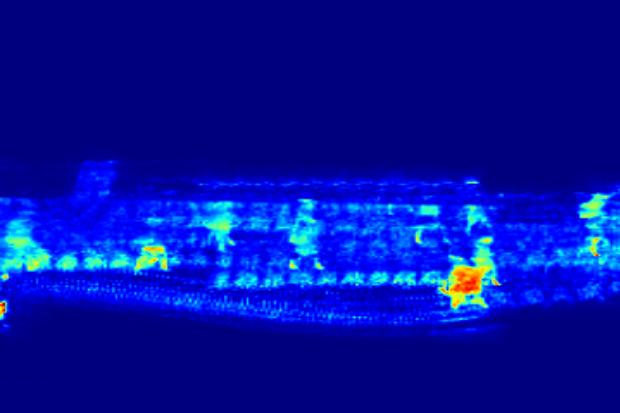}
		\includegraphics[scale=0.18]{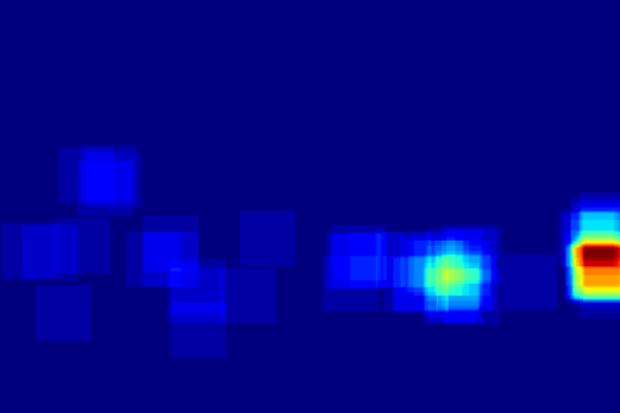}\\
		{\footnotesize Video \# 2} \\
		\includegraphics[scale=0.31]{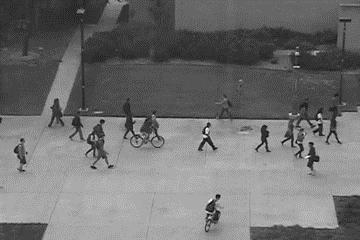}
		\includegraphics[scale=0.31]{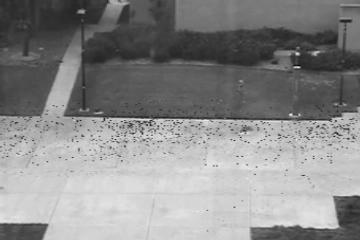}
		\includegraphics[scale=0.18]{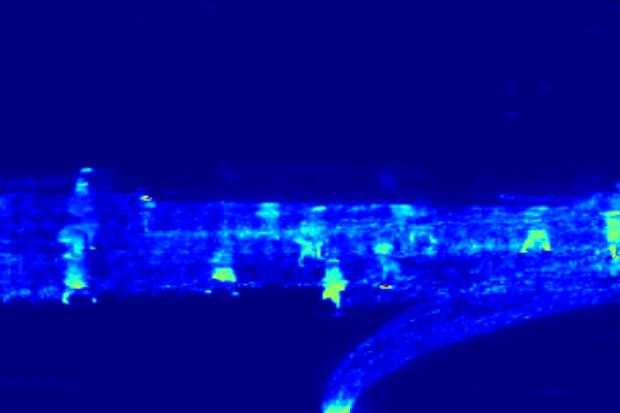}
		\includegraphics[scale=0.18]{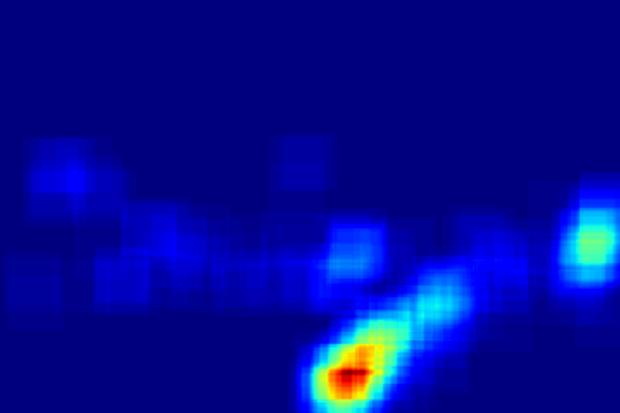}\\
		{\footnotesize Video \# 3} \\
		\includegraphics[scale=0.31]{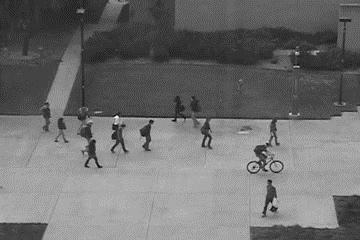}
		\includegraphics[scale=0.31]{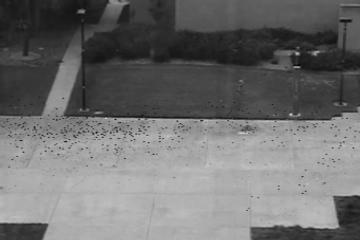}
		\includegraphics[scale=0.18]{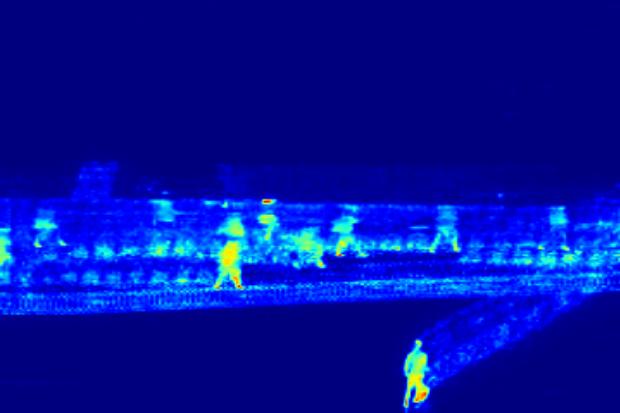}
		\includegraphics[scale=0.18]{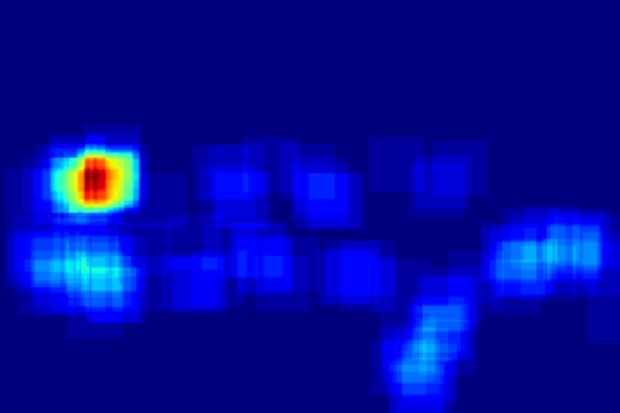}\\
		{\footnotesize Video \# 6} \\
		\includegraphics[scale=0.31]{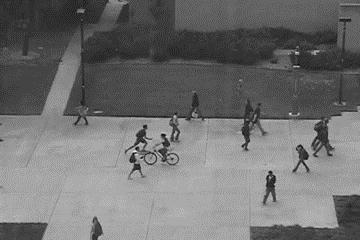}
		\includegraphics[scale=0.31]{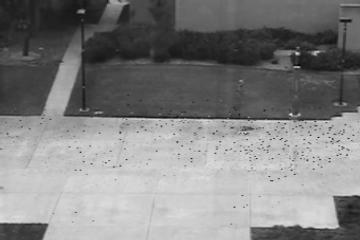}
		\includegraphics[scale=0.18]{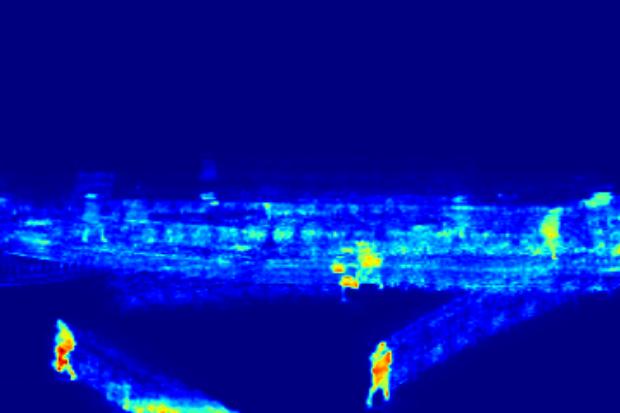}
		\includegraphics[scale=0.18]{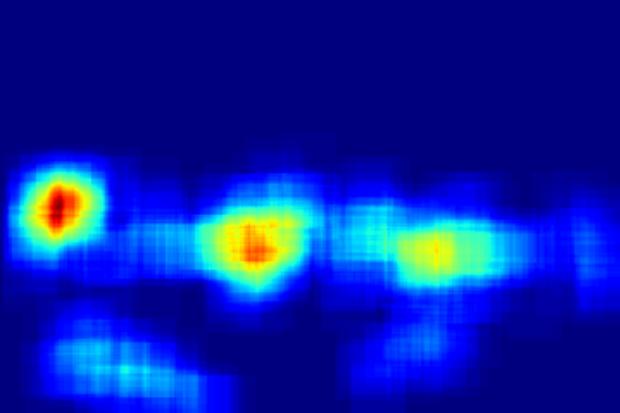}\\
		{\footnotesize Video \# 8} \\
		\includegraphics[scale=0.31]{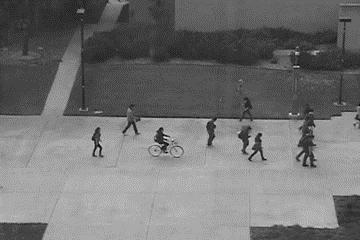}
		\includegraphics[scale=0.31]{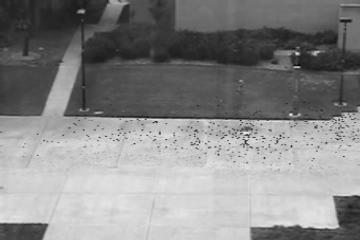}
		\includegraphics[scale=0.18]{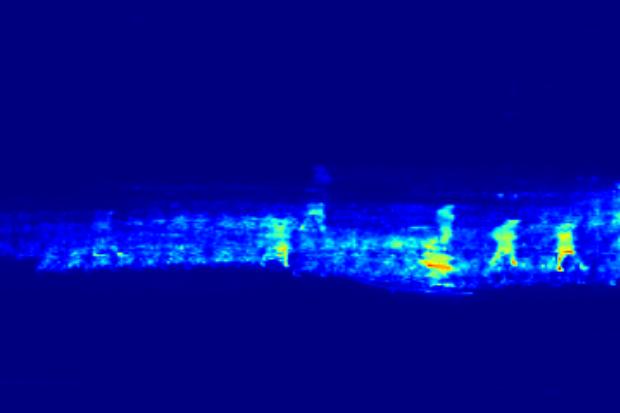}
		\includegraphics[scale=0.18]{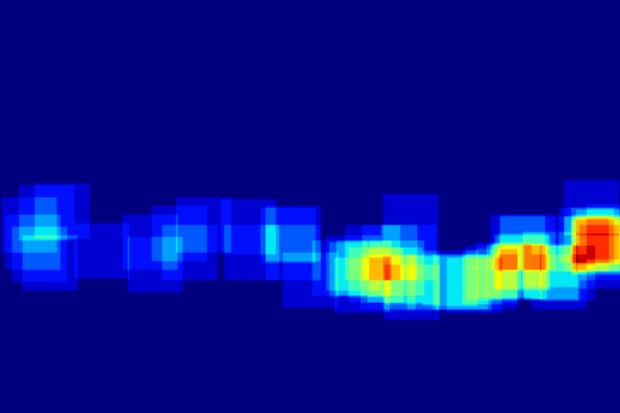}\\
		{\footnotesize Video \# 10} \\
		\hline \vspace{-.5em}\\
		\includegraphics[scale=0.31]{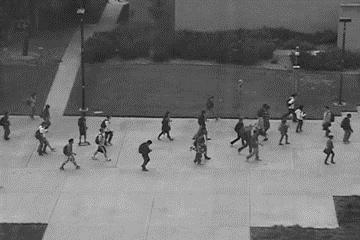}
		\includegraphics[scale=0.31]{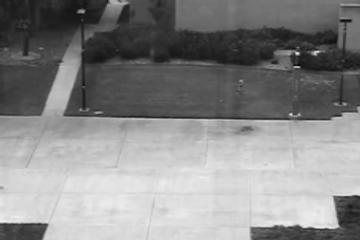}
		\includegraphics[scale=0.18]{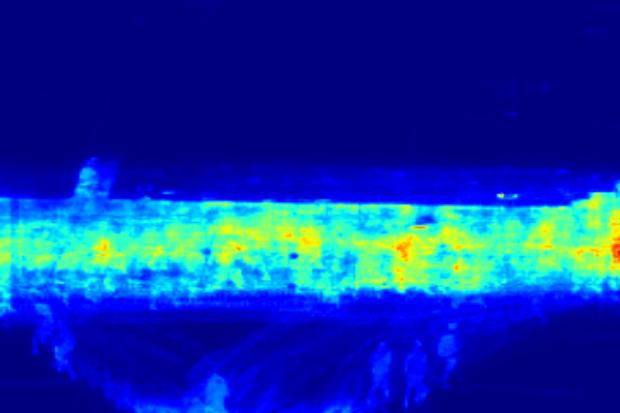}
		\includegraphics[scale=0.18]{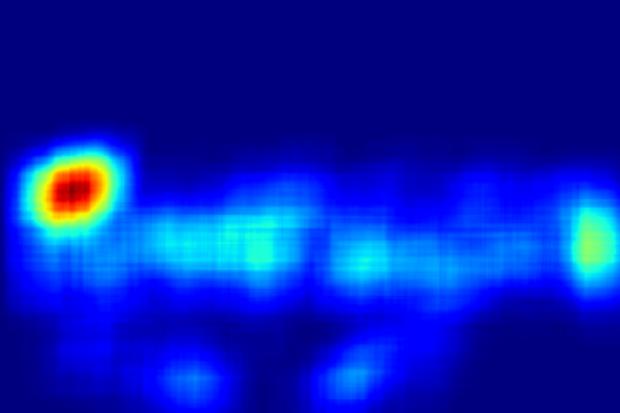}\\
		{\footnotesize All Videos} \\
	\end{tabular}
		\caption{Same layout in all figures in Section \ref{fig:temp_regularity_avenue}. Especially, in video 8, we can clearly observe a trajectory of two people in the heatmap of accumulated regular score (third column). In the synthesized regular frame by conv-autoencoder (second), there are dots. Those dots are outliers in regularity score due to lack of data as there is no dots in the regular frame by all videos thanks to statistically significant amount of data.}
		\label{fig:temp_regularity_ped2}
\end{figure}
\vspace{-1em}
\begin{center}
	\hyperlink{page.11}{Go to Table of Contents}
\end{center}

\subsection{Subway Enter}
\label{sec:temp_regularity_enter}

\begin{figure}[h]
	\centering
	\begin{tabular}{c}
		\includegraphics[scale=0.22]{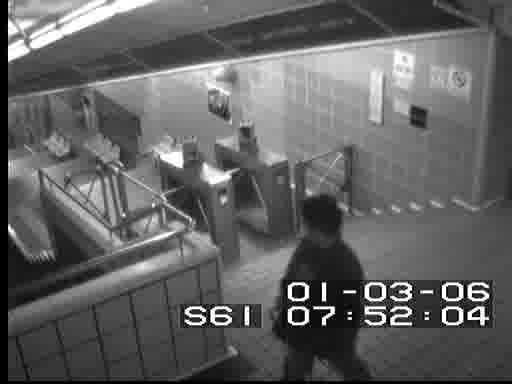}
		\includegraphics[scale=0.22]{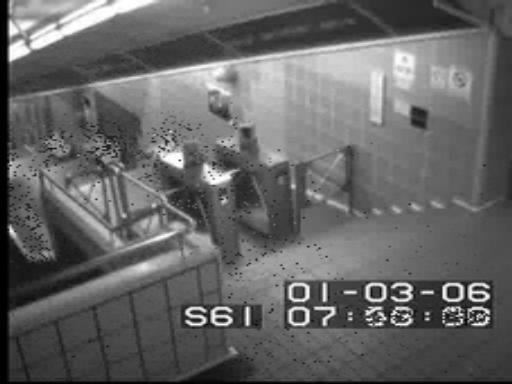}
		\includegraphics[scale=0.182]{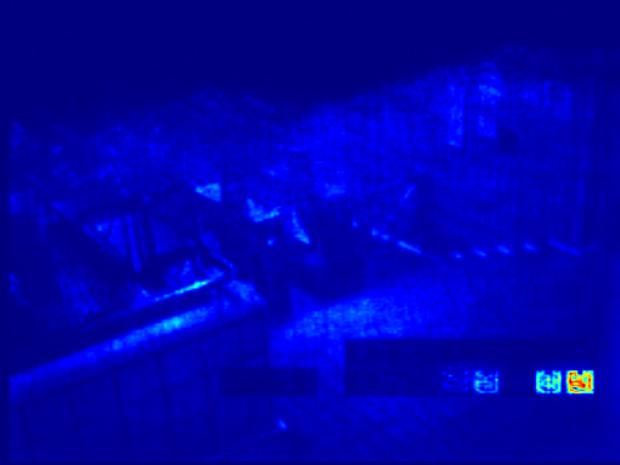}
		\includegraphics[scale=0.182]{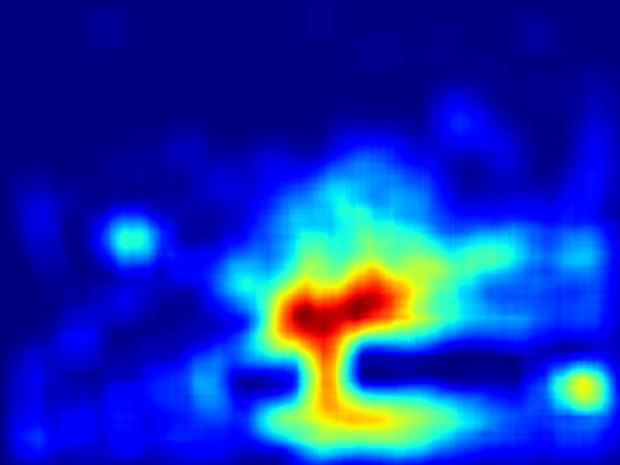}\\
		{\footnotesize Video \# 1} \\
		\includegraphics[scale=0.22]{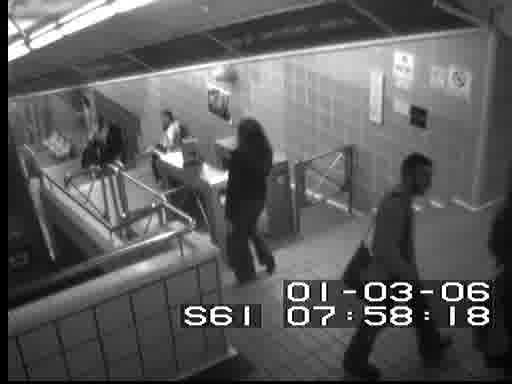}
		\includegraphics[scale=0.22]{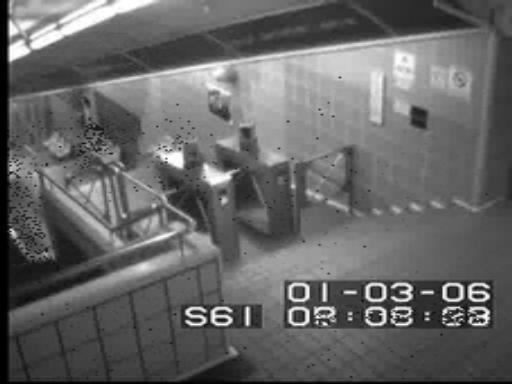}
		\includegraphics[scale=0.182]{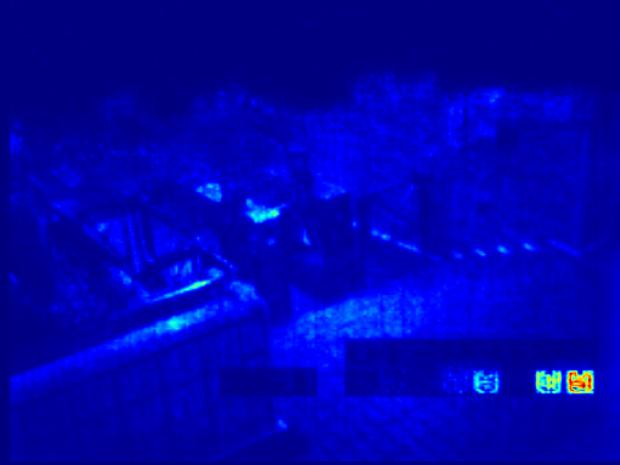}
		\includegraphics[scale=0.182]{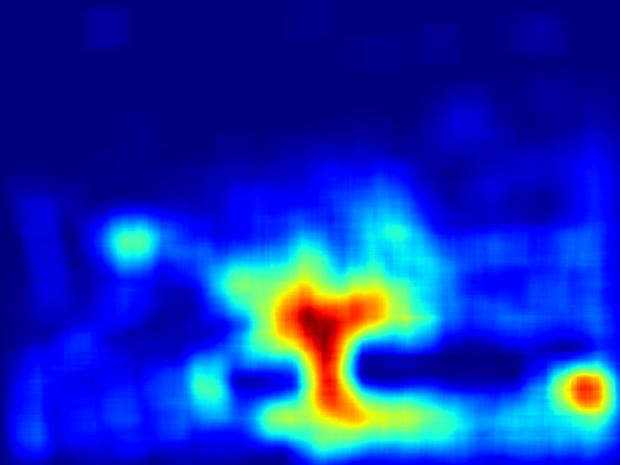}\\
		{\footnotesize Video \# 2} \\
		\includegraphics[scale=0.22]{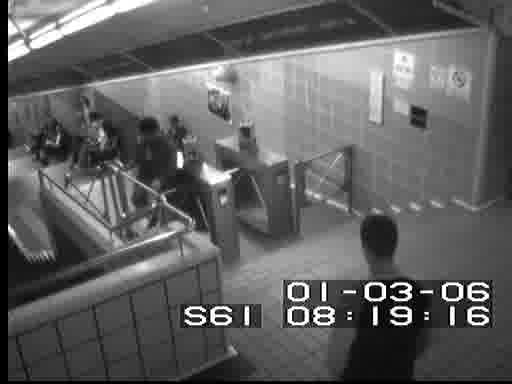}
		\includegraphics[scale=0.22]{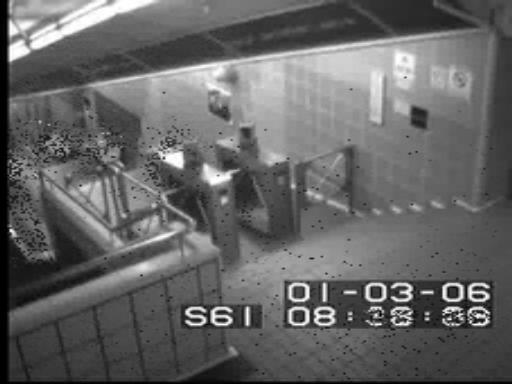}
		\includegraphics[scale=0.182]{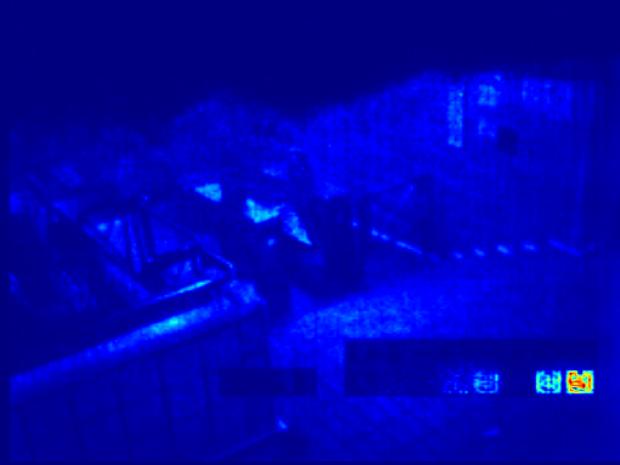}
		\includegraphics[scale=0.182]{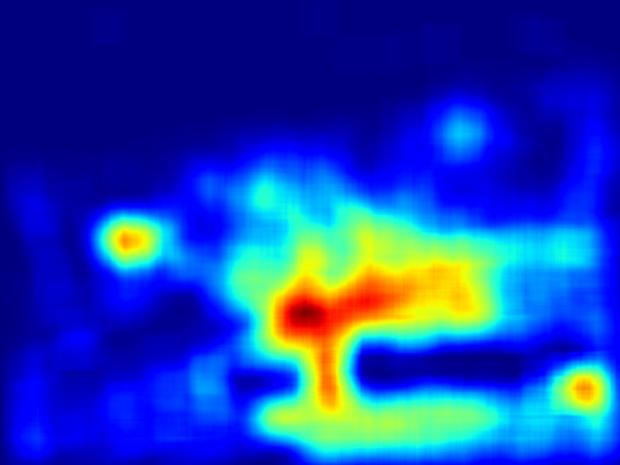}\\
		{\footnotesize Video \# 3} \\
		\includegraphics[scale=0.22]{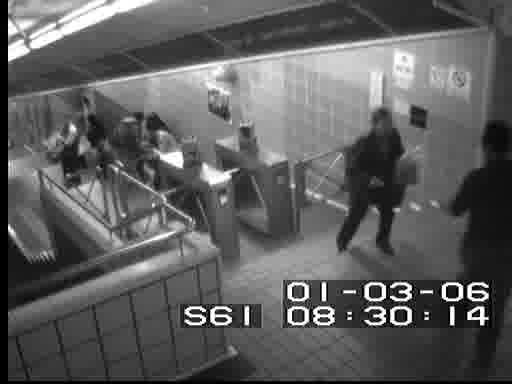}
		\includegraphics[scale=0.22]{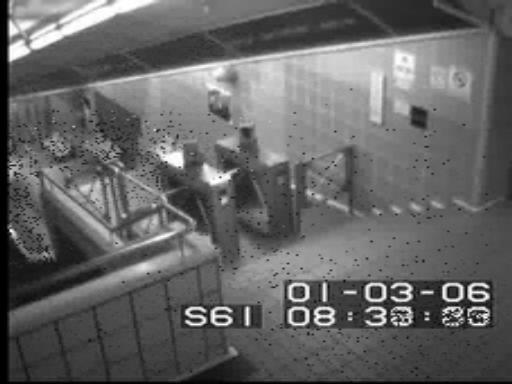}
		\includegraphics[scale=0.182]{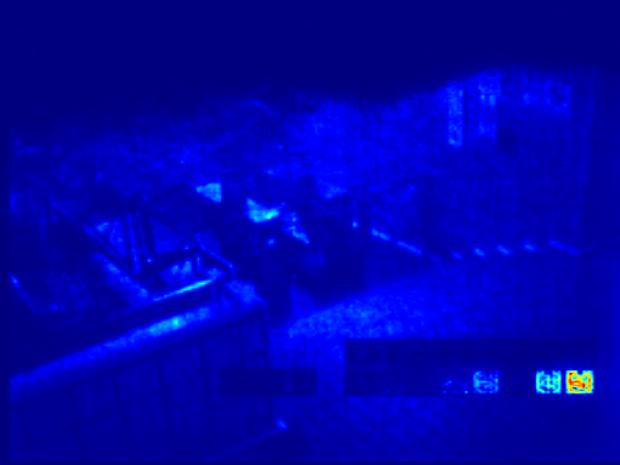}
		\includegraphics[scale=0.182]{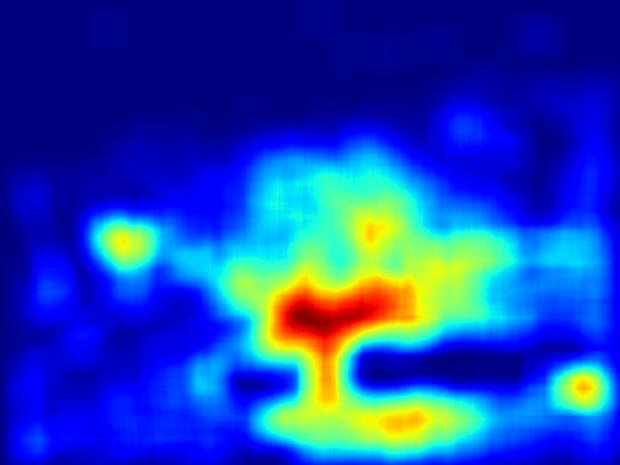}\\
		{\footnotesize Video \# 4} \\
		\hline \vspace{-.5em}\\
		\includegraphics[scale=0.22]{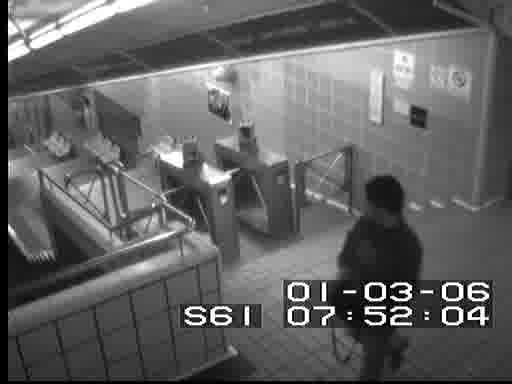}
		\includegraphics[scale=0.22]{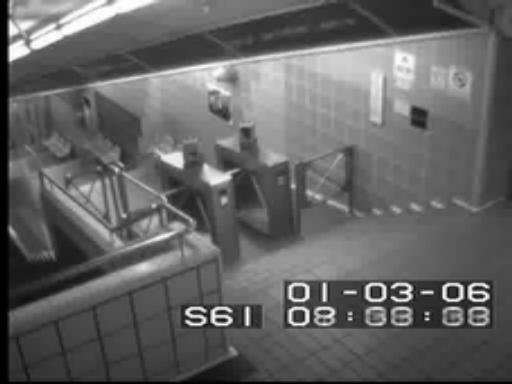}
		\includegraphics[scale=0.182]{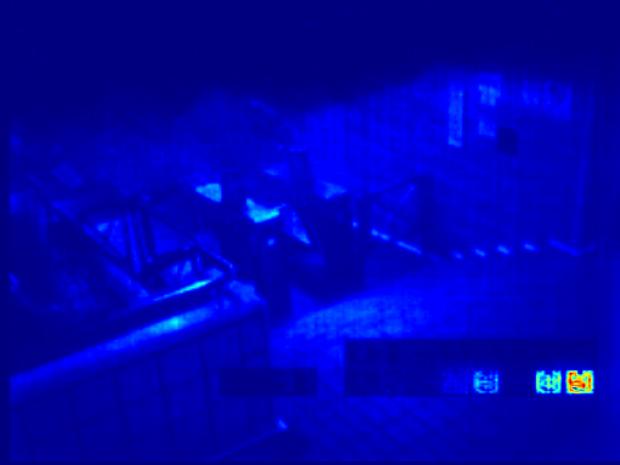}
		\includegraphics[scale=0.182]{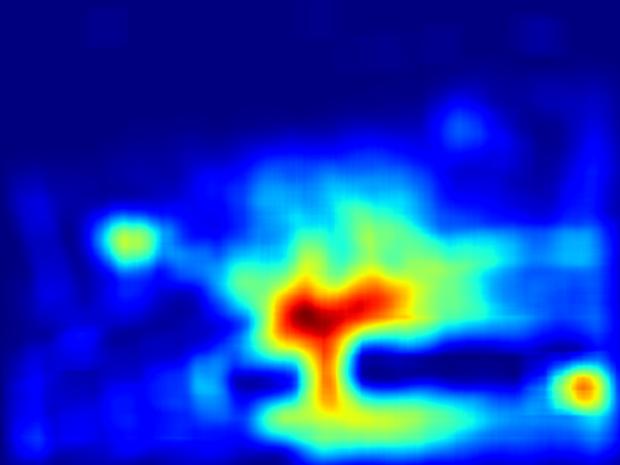}\\
		{\footnotesize All Videos} \\
	\end{tabular}
	\caption{Same layout in all figures in Section \ref{fig:temp_regularity_avenue}. Compared to other datasets, we have relatively high regularity score (more blue). It is because length of the videos is long so that the irregular motion is averaged out in long minutes. Obviously, the clock ticking is not part of regular motions.}
	\label{fig:temp_regularity_enter}
\end{figure}
\vspace{-1em}
\begin{center}
	\hyperlink{page.11}{Go to Table of Contents}
\end{center}
\clearpage

\subsection{Subway Exit}
\label{sec:temp_regularity_exit}

\begin{figure}[h]
	\centering
	\begin{tabular}{c}
		\includegraphics[scale=0.22]{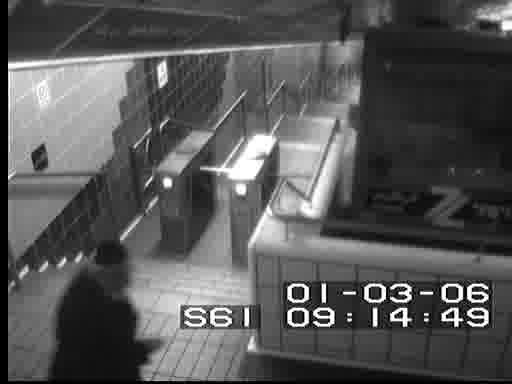}
		\includegraphics[scale=0.22]{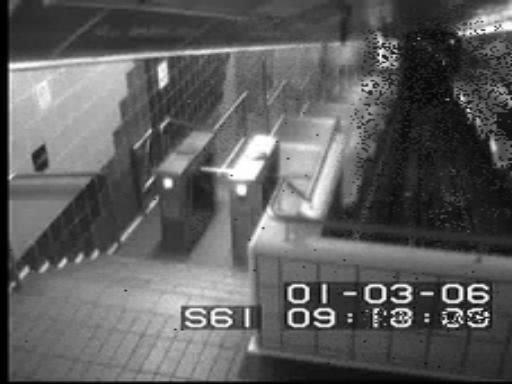}
		\includegraphics[scale=0.182]{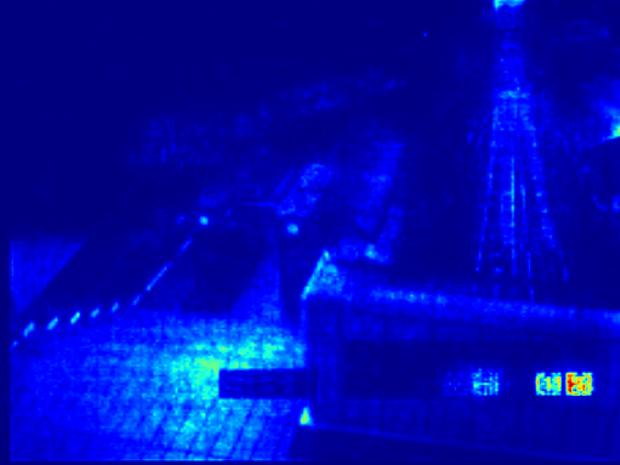}
		\includegraphics[scale=0.182]{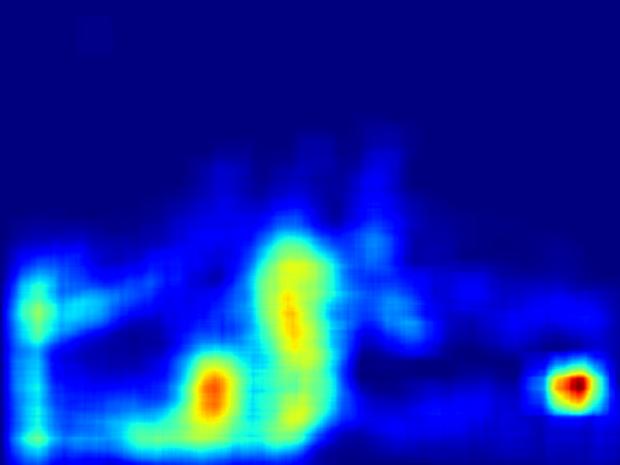}\\
		{\footnotesize Video \# 1} \\
		\includegraphics[scale=0.22]{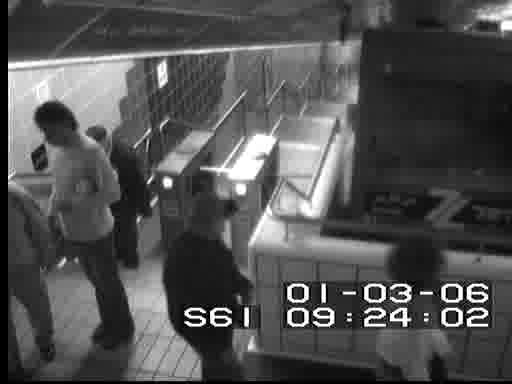}
		\includegraphics[scale=0.22]{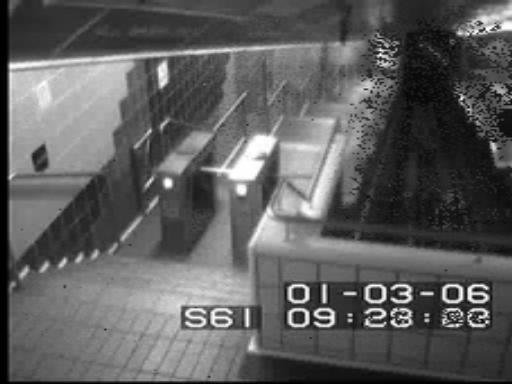}
		\includegraphics[scale=0.182]{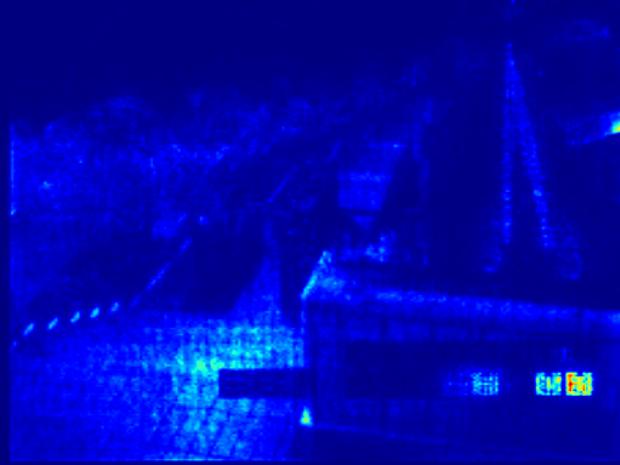}
		\includegraphics[scale=0.182]{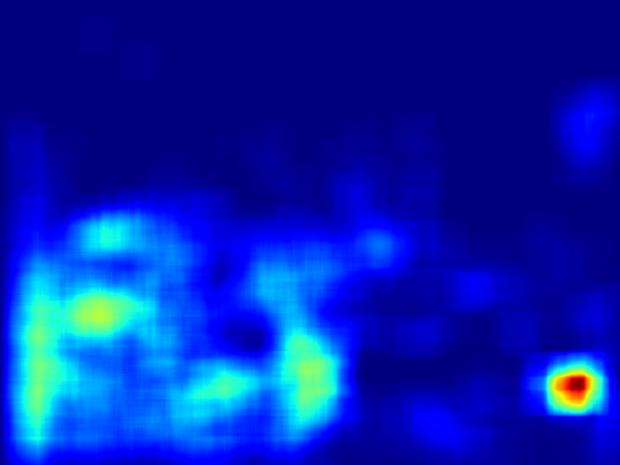}\\
		{\footnotesize Video \# 2} \\
		\includegraphics[scale=0.22]{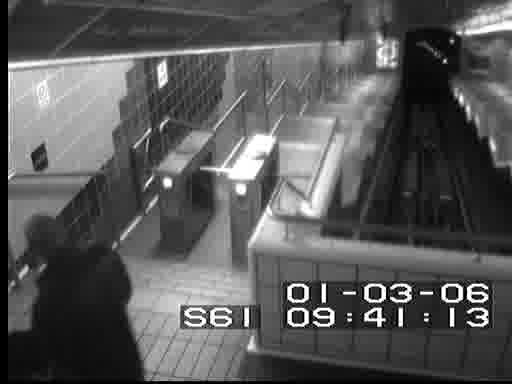}
		\includegraphics[scale=0.22]{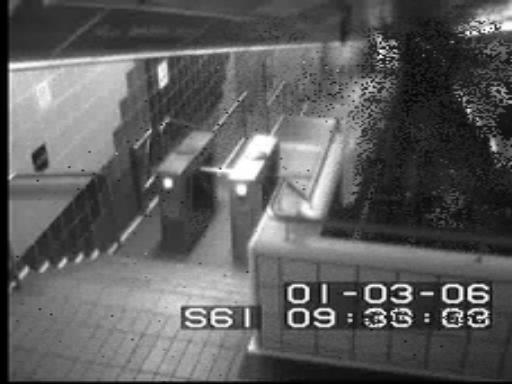}
		\includegraphics[scale=0.182]{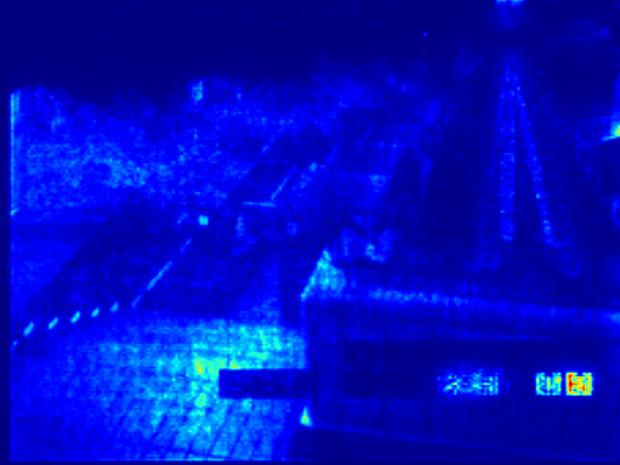}
		\includegraphics[scale=0.182]{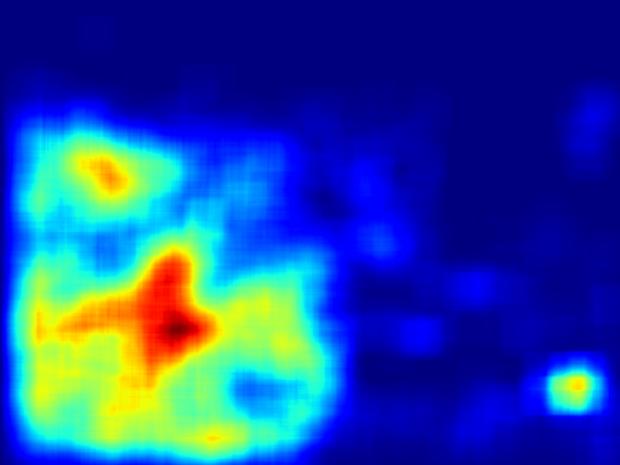}\\
		{\footnotesize Video \# 3} \\
		\includegraphics[scale=0.22]{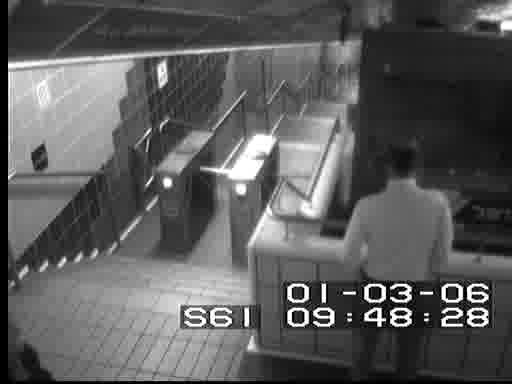}
		\includegraphics[scale=0.22]{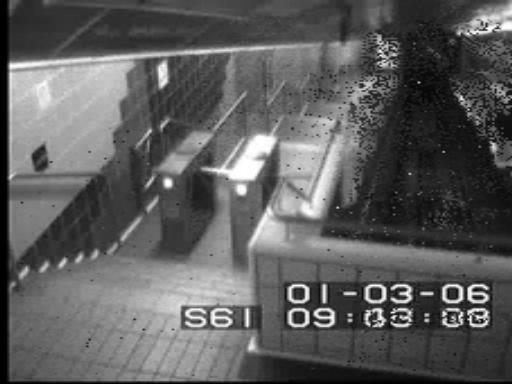}
		\includegraphics[scale=0.182]{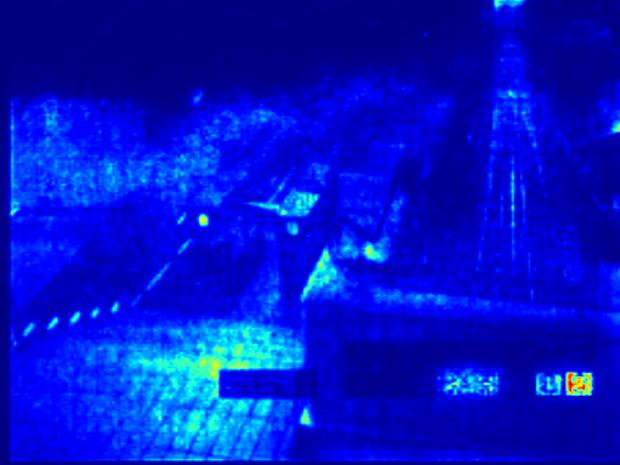}
		\includegraphics[scale=0.182]{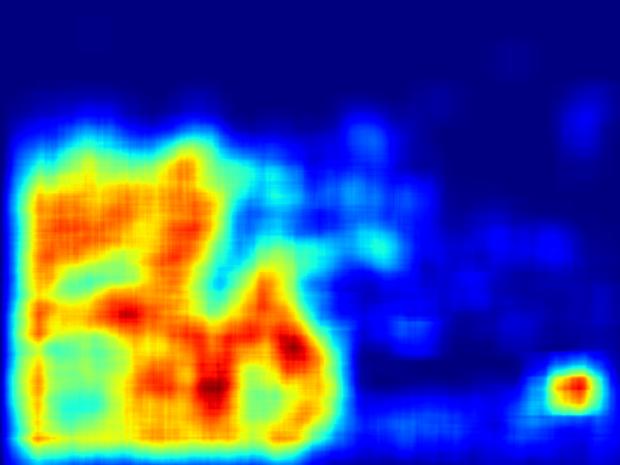}\\
		{\footnotesize Video \# 4} \\
		\includegraphics[scale=0.22]{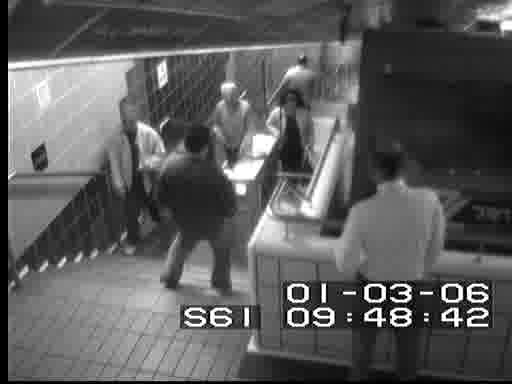}
		\includegraphics[scale=0.22]{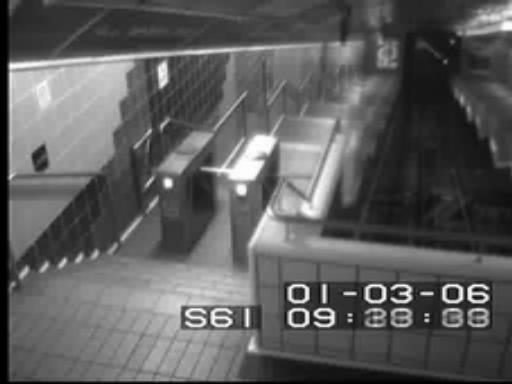}
		\includegraphics[scale=0.182]{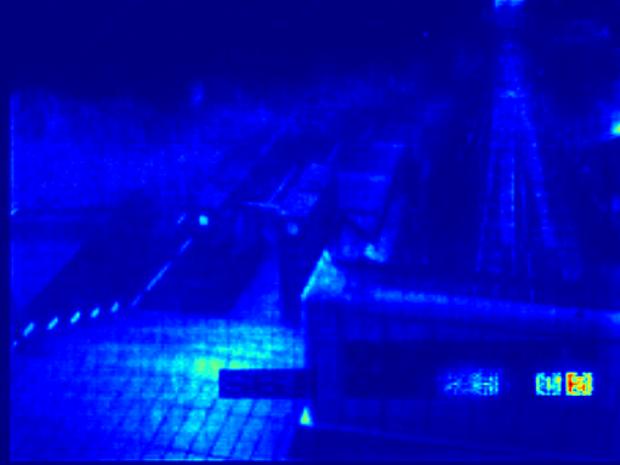}
		\includegraphics[scale=0.182]{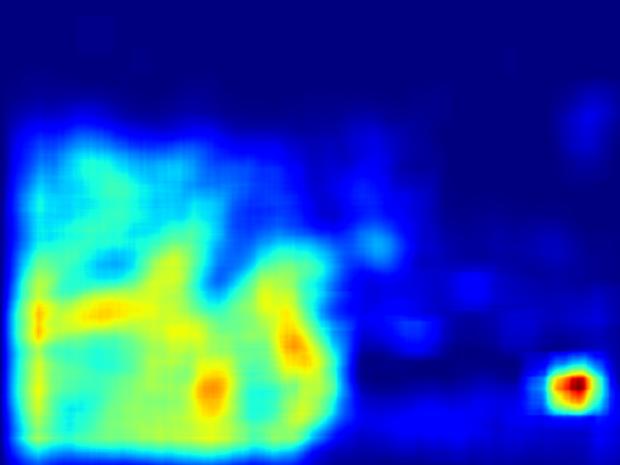}\\
		{\footnotesize All Videos} \\
	\end{tabular}
	\caption{Same layout in all figures in Section \ref{fig:temp_regularity_avenue}. Similar to Subway Enter. But interestingly, IT-autoencoder has a very high accumulated irregular score in the stair regions.}
	\label{fig:temp_regularity_exit}
\end{figure}
\vspace{-1em}
\begin{center}
	\hyperlink{page.11}{Go to Table of Contents}
\end{center}
\clearpage


\section{Object Detection in Irregular Motion}
\label{sec:obj_det}
Using the regularity score, we can obtain locations of objects involved in irregular motion in each frame, which is a usually the objects of interest.
We present several frames with irregular motions for each dataset and its corresponding objects location in the frame.
Note that we have high irregularity response at the edge of the objects where the motion changes most significantly.
It is better presented in video: \texttt{reg\_score\_video.avi}

\subsection{CUHK Avenue Dataset}
\label{sec:obj_det_avenue}

\begin{figure}[h]
	\centering
	\begin{tabular}{cc}
		\includegraphics[scale=0.18]{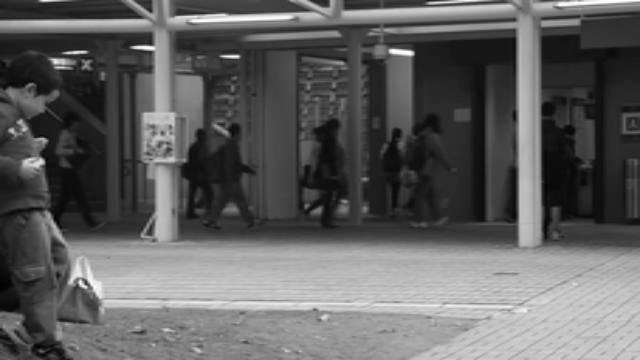}
		\includegraphics[scale=0.26]{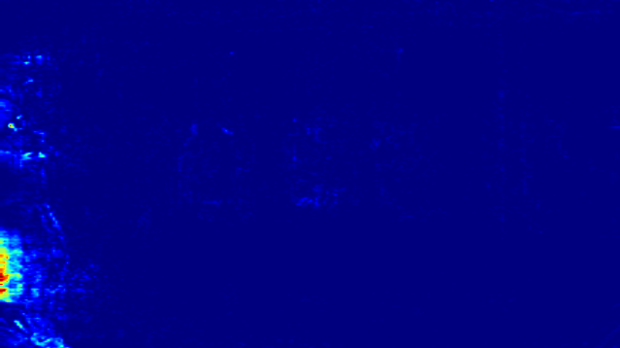}
		&\includegraphics[scale=0.18]{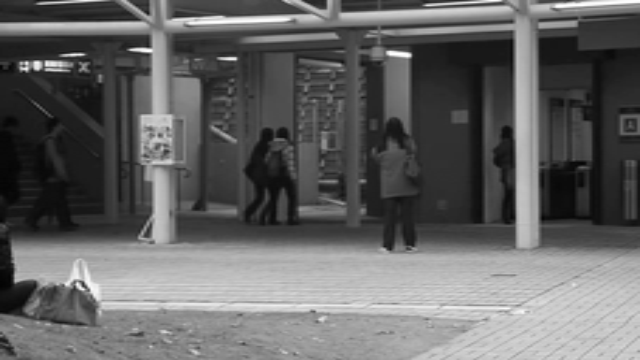}
		\includegraphics[scale=0.26]{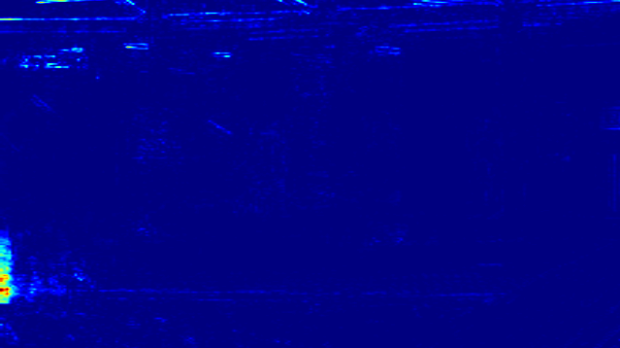}\\
		{\footnotesize Video \# 1, Frame \# 600} 
		& {\footnotesize Video \# 2, Frame \# 1075}\\
		\includegraphics[scale=0.18]{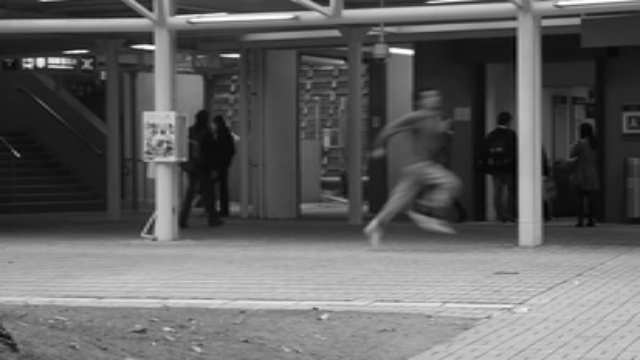}
		\includegraphics[scale=0.26]{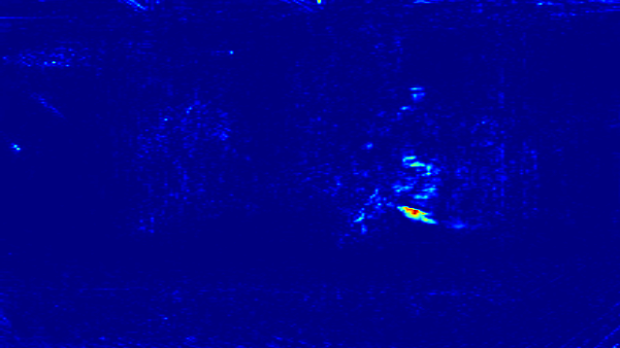}
		&\includegraphics[scale=0.18]{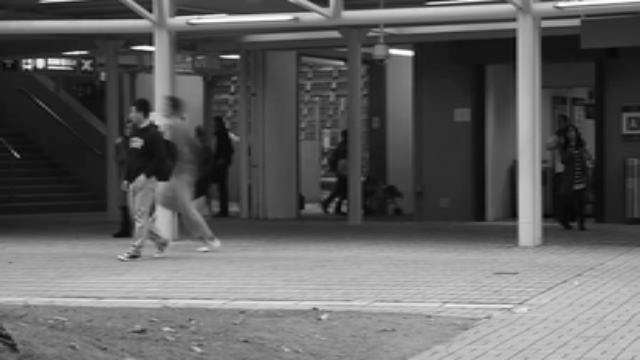}
		\includegraphics[scale=0.26]{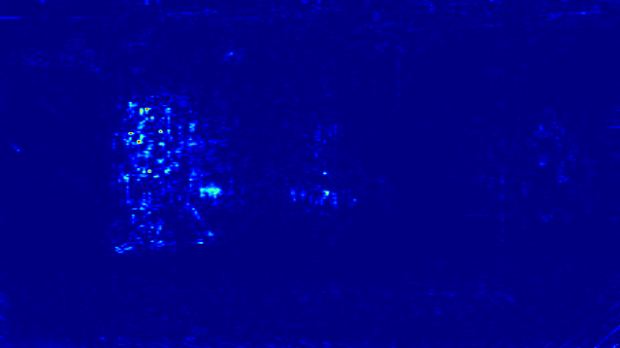}\\
		{\footnotesize Video \# 3, Frame \# 600}
		& {\footnotesize Video \# 4, Frame \# 400}\\
		\includegraphics[scale=0.18]{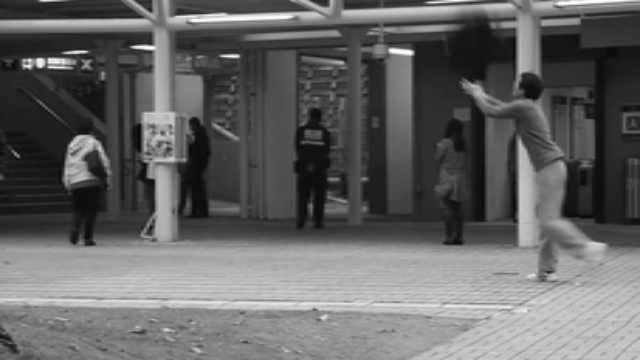}
		\includegraphics[scale=0.26]{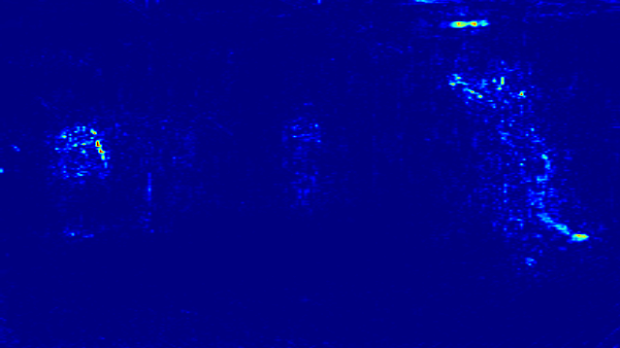}
		&\includegraphics[scale=0.18]{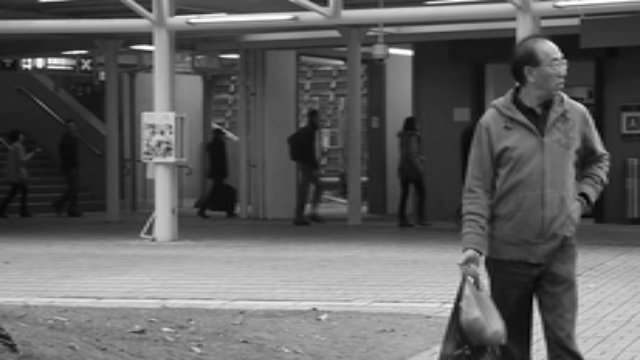}
		\includegraphics[scale=0.26]{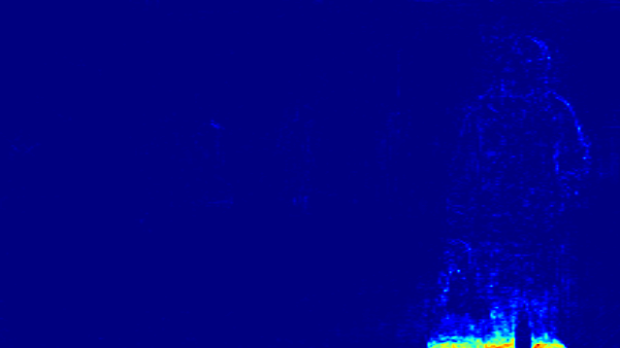}\\
		{\footnotesize Video \# 5, Frame \# 600} 
		& {\footnotesize Video \# 6, Frame \# 500}\\
		\includegraphics[scale=0.18]{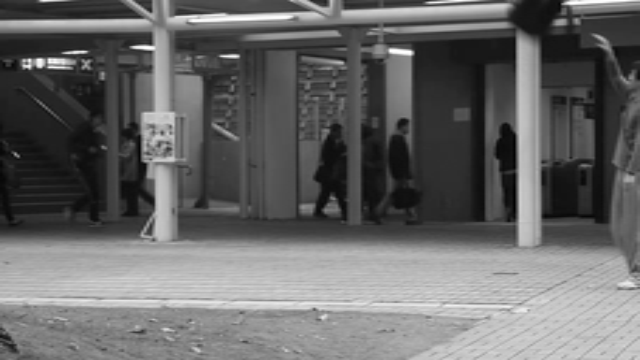}
		\includegraphics[scale=0.26]{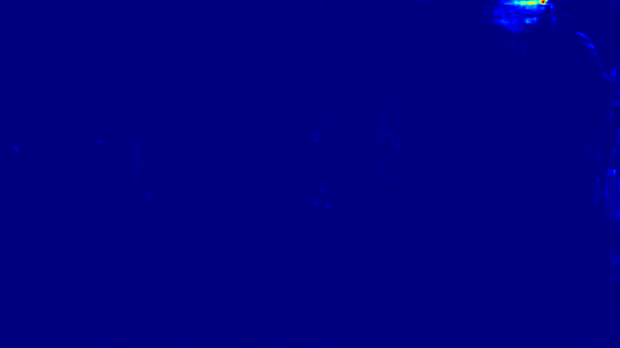}
		&\includegraphics[scale=0.18]{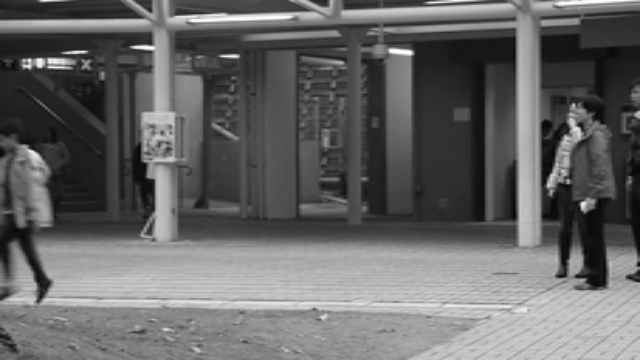}
		\includegraphics[scale=0.26]{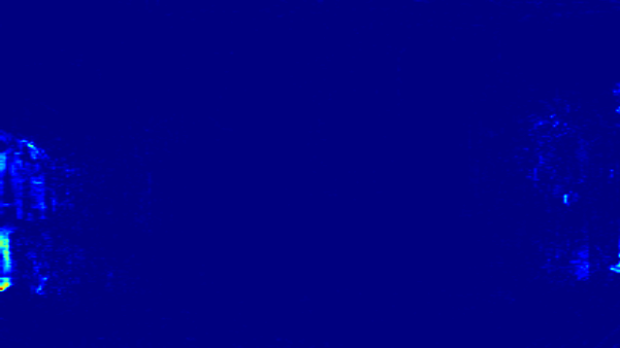}\\
		{\footnotesize Video \# 6, Frame \# 900}
		& {\footnotesize Video \# 7, Frame \# 4800}\\
		\includegraphics[scale=0.18]{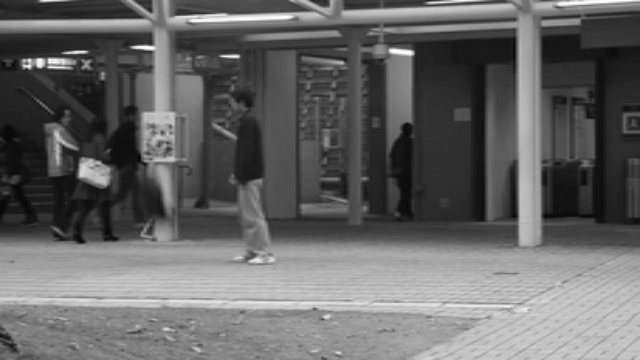}
		\includegraphics[scale=0.26]{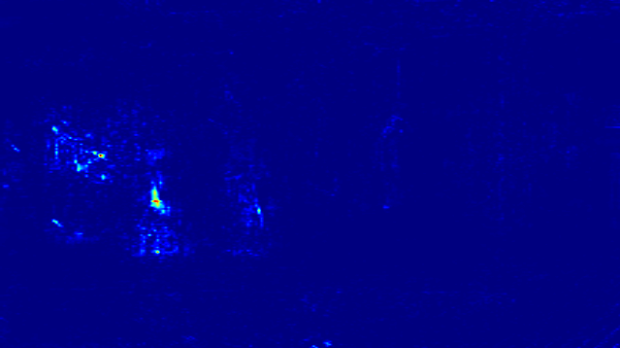}
		&\includegraphics[scale=0.18]{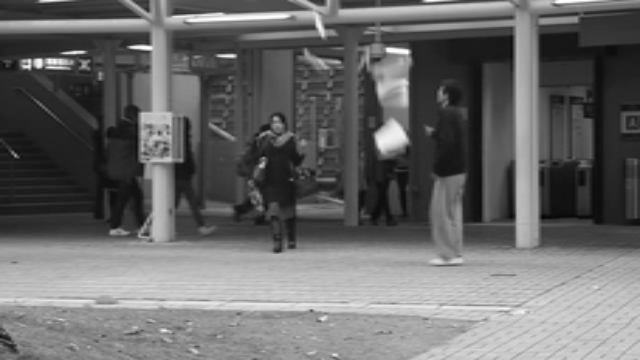}
		\includegraphics[scale=0.26]{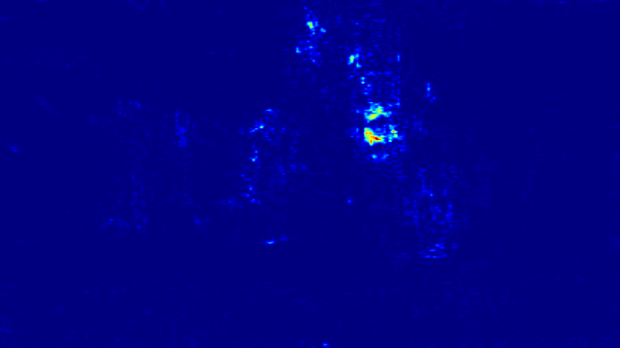}\\
		{\footnotesize Video \# 12, Frame \# 680} 
		& {\footnotesize Video \# 14, Frame \# 420}\\
		\includegraphics[scale=0.18]{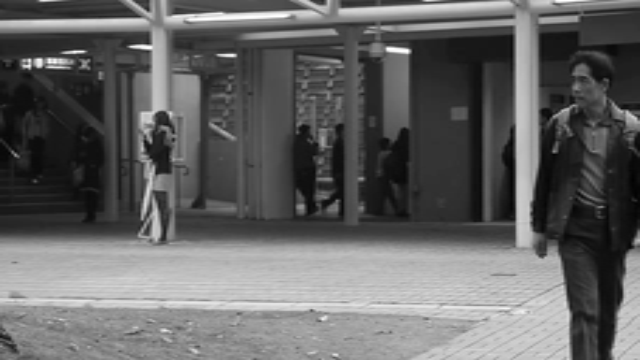}
		\includegraphics[scale=0.26]{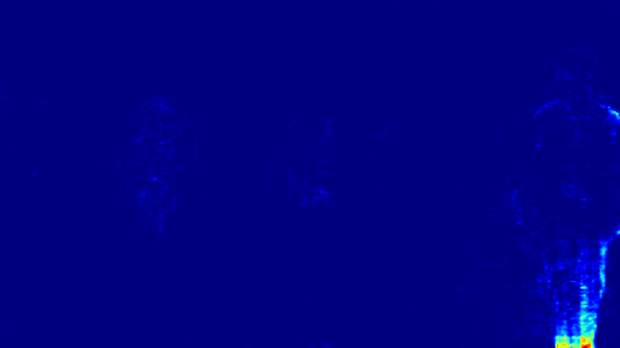}
		&\includegraphics[scale=0.18]{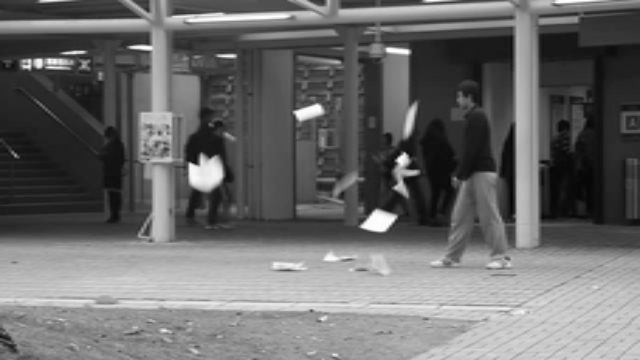}
		\includegraphics[scale=0.26]{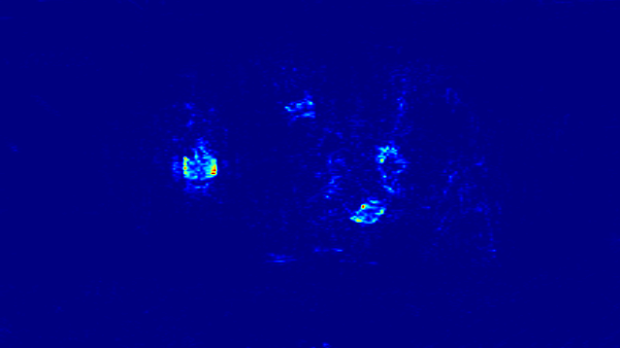}\\
		{\footnotesize Video \# 15, Frame \# 550}
		& {\footnotesize Video \# 20, Frame \# 100}\\
	\end{tabular}
	\caption{(Left) a frame (Right) Object regularity score. In video 6 (frame \# 500), the bottom part of legs, which are the most prominent object involved in a irregular motion, exhibits very high scores to other regions. In video 14 and 20, the flying papers are well captured.}
	\label{fig:obj_det_avenue}
\end{figure}

\begin{center}
	\hyperlink{page.11}{Go to Table of Contents}
\end{center}

\clearpage

\subsection{UCSD Ped1}
\label{sec:obj_det_ped1}

\begin{figure}[h]
	\centering
	\begin{tabular}{cc}
		\includegraphics[scale=0.487]{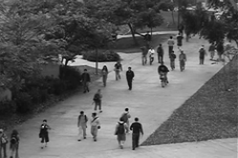}
		\includegraphics[scale=0.26]{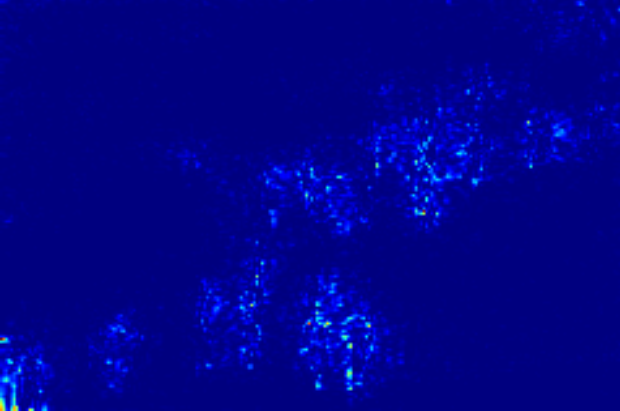}
		&\includegraphics[scale=0.487]{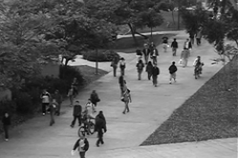}
		\includegraphics[scale=0.26]{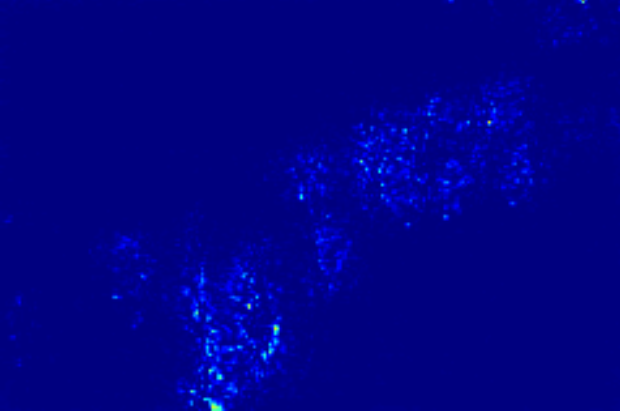}\\
		{\footnotesize Video \# 1, Frame \# 100} 
		& {\footnotesize Video \# 5, Frame \# 150}\\
		\includegraphics[scale=0.487]{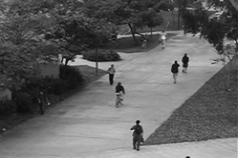}
		\includegraphics[scale=0.26]{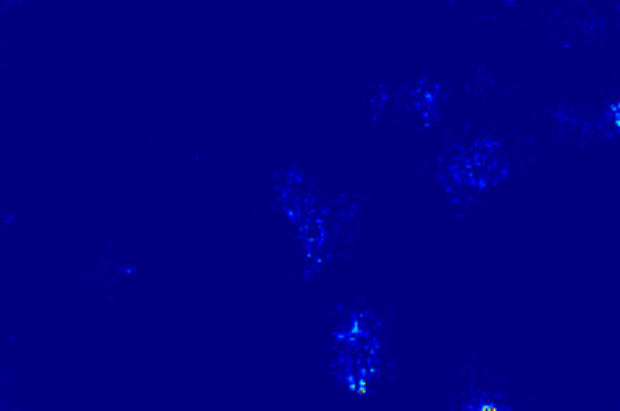}
		&\includegraphics[scale=0.487]{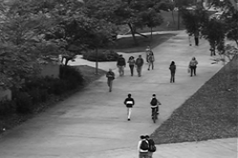}
		\includegraphics[scale=0.26]{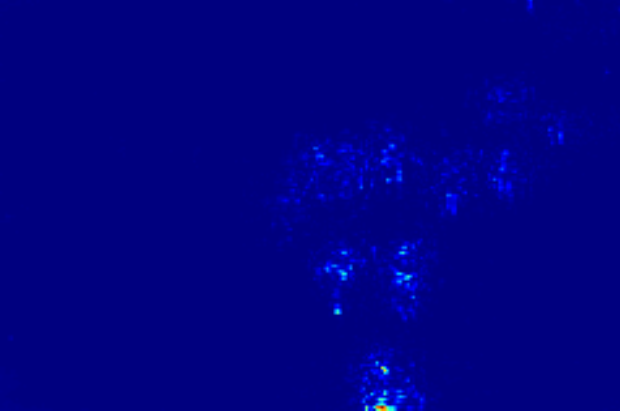}\\
		{\footnotesize Video \# 15, Frame \# 170} 
		& {\footnotesize Video \# 16, Frame \# 150}\\
		\includegraphics[scale=0.487]{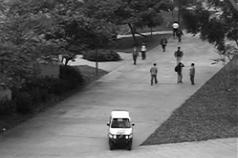}
		\includegraphics[scale=0.26]{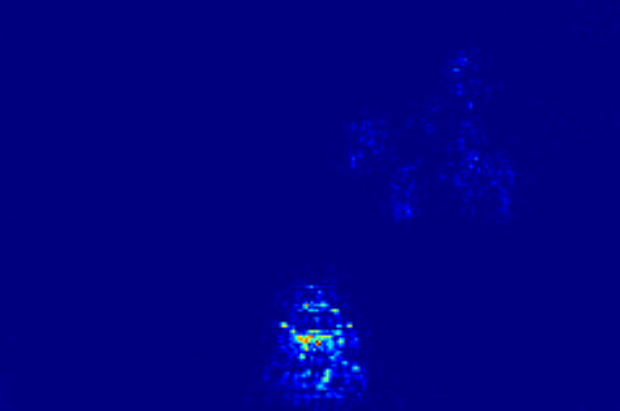}
		&\includegraphics[scale=0.487]{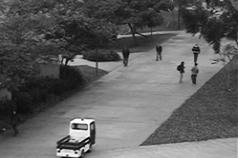}
		\includegraphics[scale=0.26]{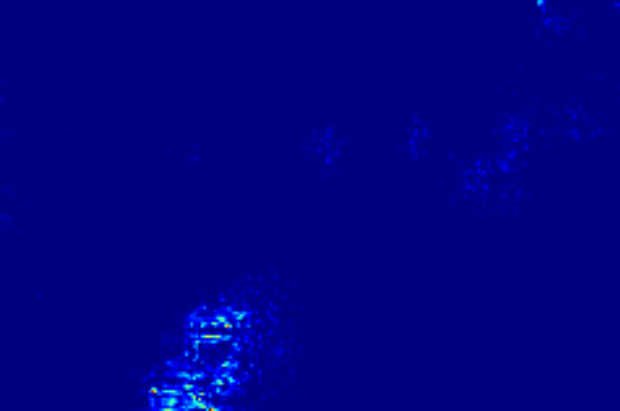}\\
		{\footnotesize Video \# 19, Frame \# 120} 
		& {\footnotesize Video \# 20, Frame \# 60}\\
		\includegraphics[scale=0.487]{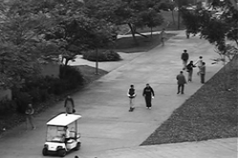}
		\includegraphics[scale=0.26]{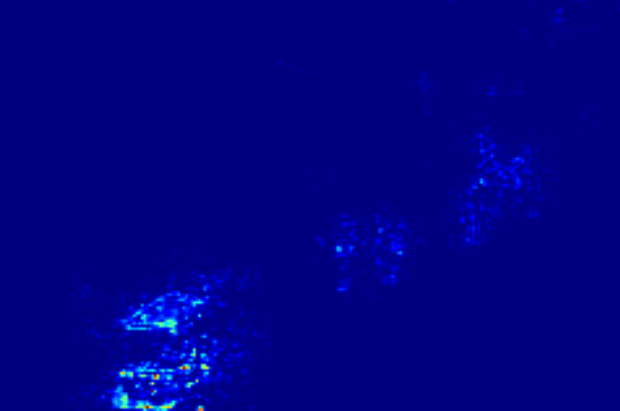}
		&\includegraphics[scale=0.487]{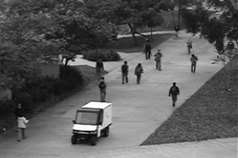}
		\includegraphics[scale=0.26]{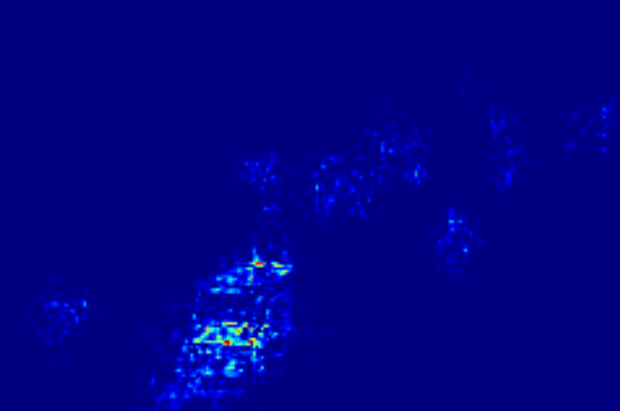}\\
		{\footnotesize Video \# 24, Frame \# 150} 
		& {\footnotesize Video \# 27, Frame \# 90}\\
		\includegraphics[scale=0.487]{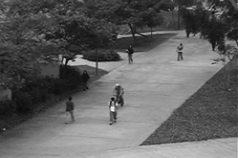}
		\includegraphics[scale=0.26]{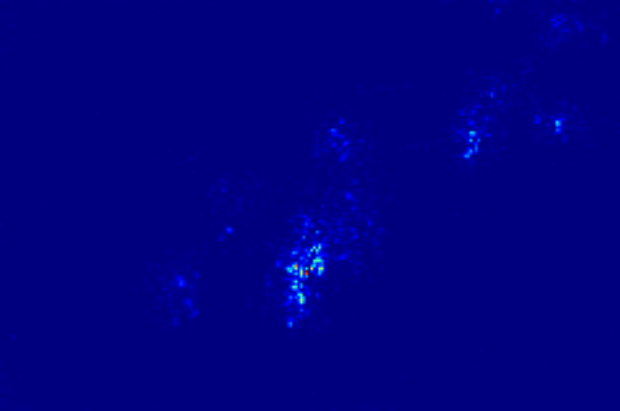}
		&\includegraphics[scale=0.487]{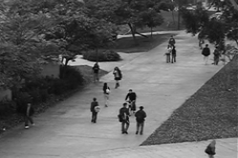}
		\includegraphics[scale=0.26]{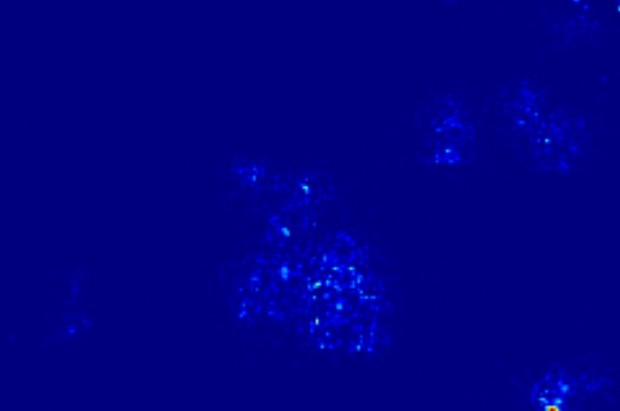}\\
		{\footnotesize Video \# 29, Frame \# 80} 
		& {\footnotesize Video \# 33, Frame \# 50}\\
	\end{tabular}
		\caption{Same layout in all figures as in Section \ref{sec:obj_det_avenue}. The moving cars are easily identified in video \#19, \#20, \#24, and \#27 and fast moving persons in video \#29.}
		\label{fig:obj_det_ped1}
\end{figure}

\begin{center}
	\hyperlink{page.11}{Go to Table of Contents}
\end{center}

\clearpage

\subsection{UCSD Ped2}
\label{sec:obj_det_ped2}

\begin{figure}[h]
	\centering
	\begin{tabular}{cc}
		\includegraphics[scale=0.32]{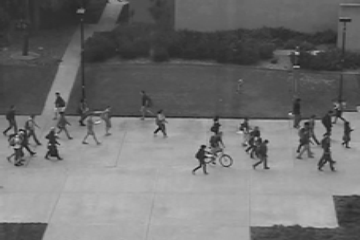}
		\includegraphics[scale=0.26]{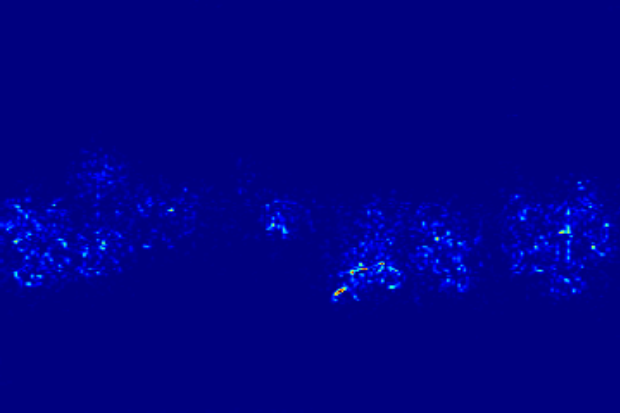}
		&\includegraphics[scale=0.32]{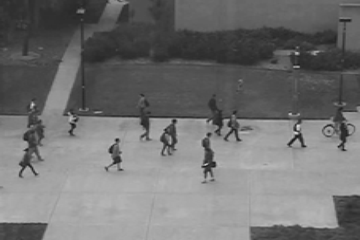}
		\includegraphics[scale=0.26]{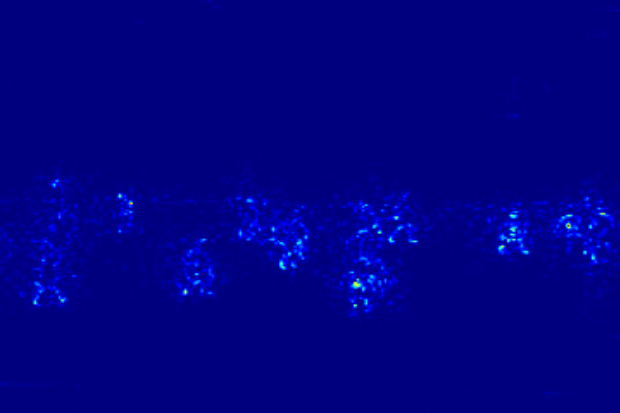}\\
		{\footnotesize Video \# 1, Frame \# 160} 
		& {\footnotesize Video \# 3, Frame \# 120}\\
		\includegraphics[scale=0.32]{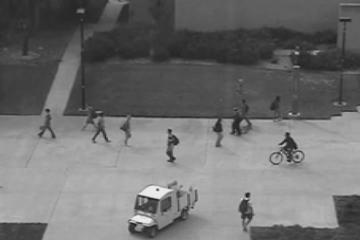}
		\includegraphics[scale=0.26]{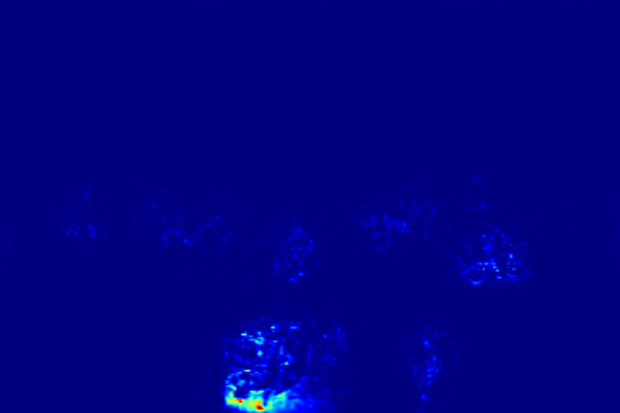}
		&\includegraphics[scale=0.32]{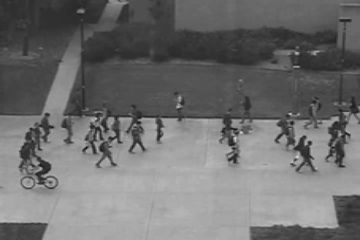}
		\includegraphics[scale=0.26]{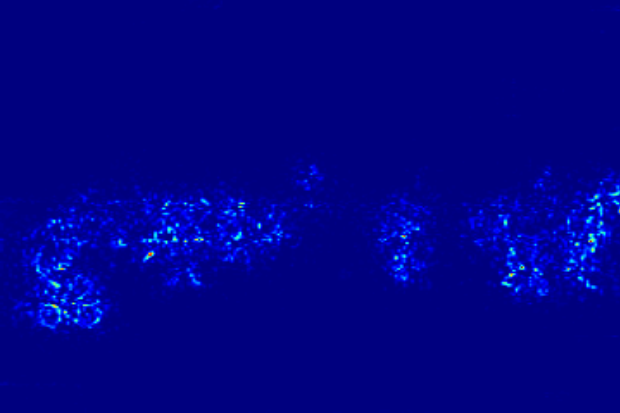}\\
		{\footnotesize Video \# 4, Frame \# 150} 
		& {\footnotesize Video \# 5, Frame \# 100}\\
		\includegraphics[scale=0.32]{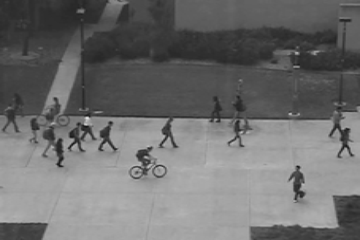}
		\includegraphics[scale=0.26]{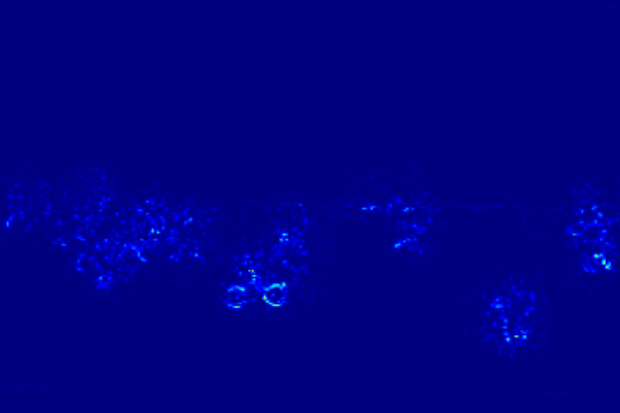}
		&\includegraphics[scale=0.32]{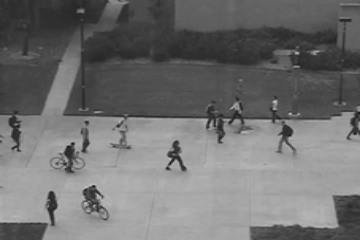}
		\includegraphics[scale=0.26]{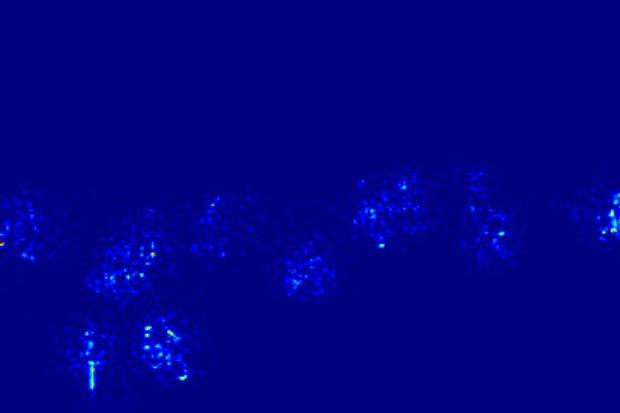}\\
		{\footnotesize Video \# 6, Frame \# 80} 
		& {\footnotesize Video \# 7, Frame \# 120}\\
		\includegraphics[scale=0.32]{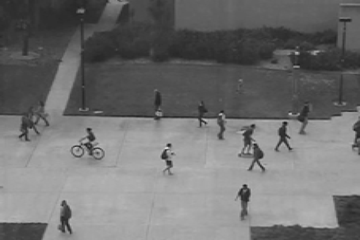}
		\includegraphics[scale=0.26]{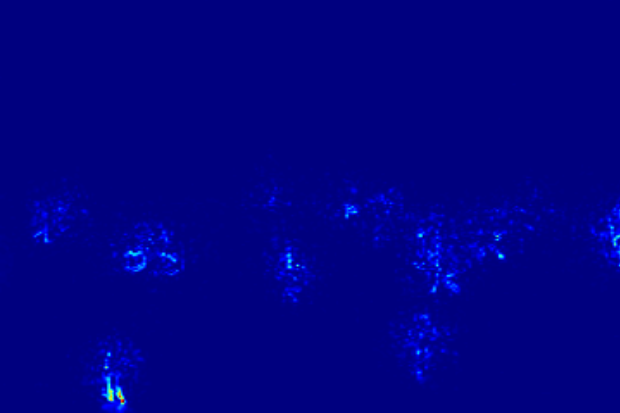}
		&\includegraphics[scale=0.32]{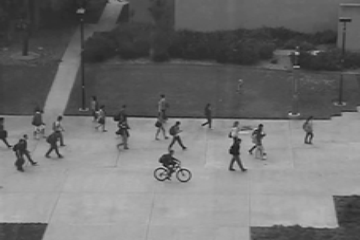}
		\includegraphics[scale=0.26]{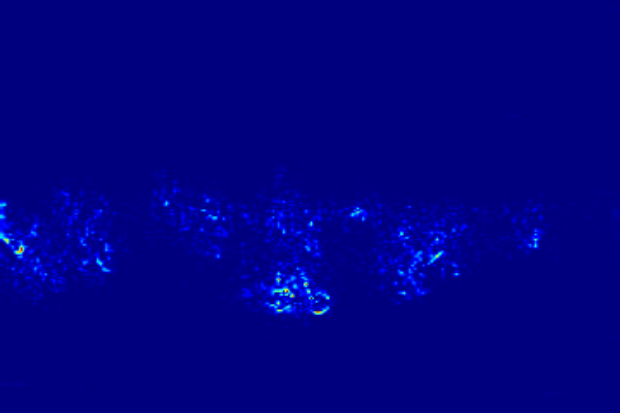}\\
		{\footnotesize Video \# 8, Frame \# 130} 
		& {\footnotesize Video \# 9, Frame \# 100}\\
	\end{tabular}
		\caption{Same layout in all figures as in Section \ref{sec:obj_det_avenue}. The moving cars (frames in video 6) and fast moving persons are easily localized.}
		\label{fig:obj_det_ped2}
\end{figure}

\begin{center}
	\hyperlink{page.11}{Go to Table of Contents}
\end{center}

\clearpage

\subsection{Subway Enter}
\label{sec:obj_det_enter}

\begin{figure}[h]
	\centering
	\begin{tabular}{cc}
		\includegraphics[scale=0.226]{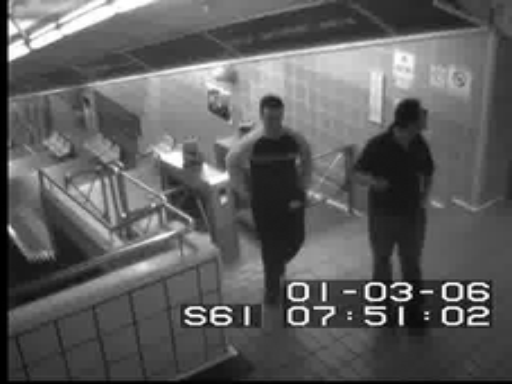}
		\includegraphics[scale=0.26]{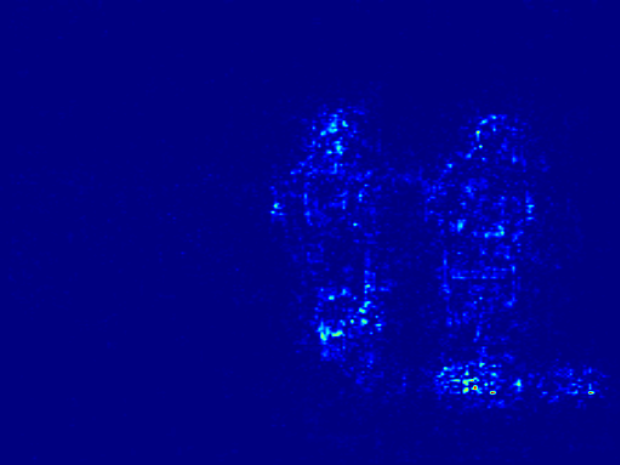}
		&\includegraphics[scale=0.226]{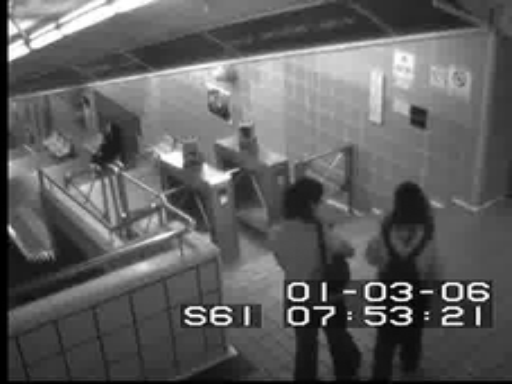}
		\includegraphics[scale=0.26]{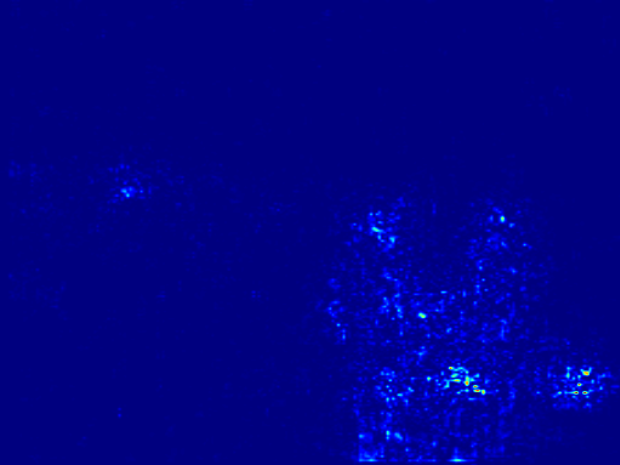}\\
		{\footnotesize Video \# 1, Frame \# 9830} 
		& {\footnotesize Video \# 1, Frame \# 13310}\\
		\includegraphics[scale=0.226]{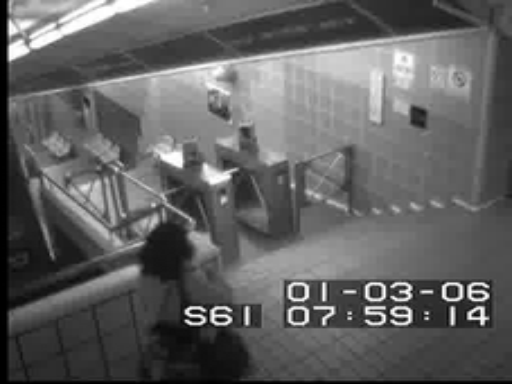}
		\includegraphics[scale=0.26]{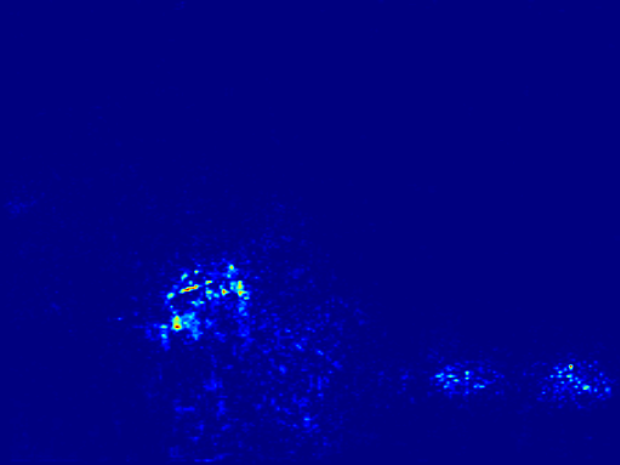}
		&\includegraphics[scale=0.226]{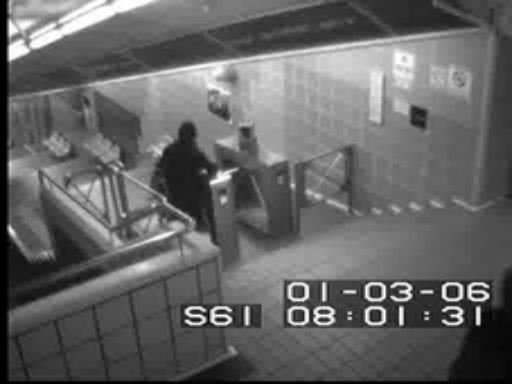}
		\includegraphics[scale=0.26]{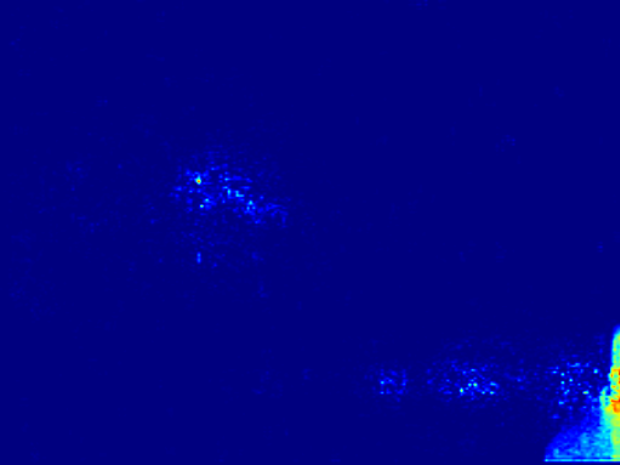}\\
		{\footnotesize Video \# 2, Frame \# 2130} 
		& {\footnotesize Video \# 2, Frame \# 5540}\\
		\includegraphics[scale=0.226]{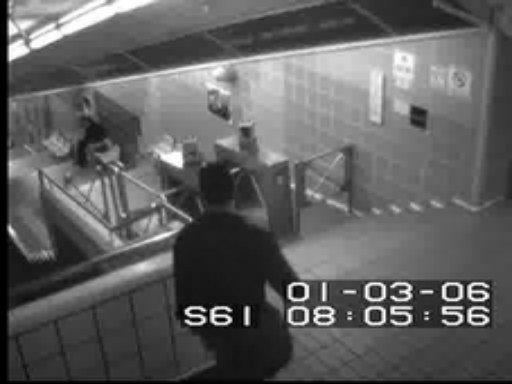}
		\includegraphics[scale=0.26]{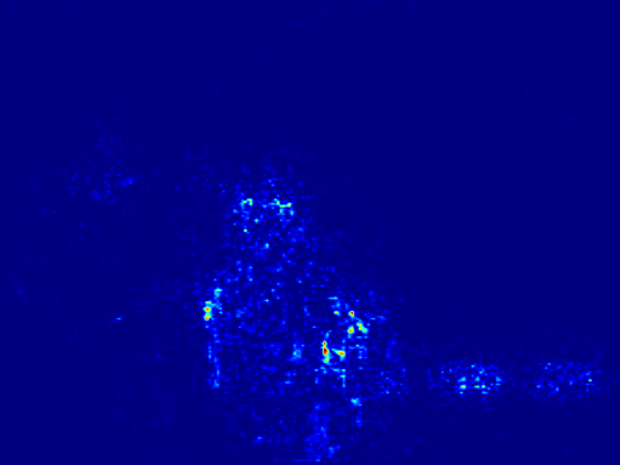}
		&\includegraphics[scale=0.226]{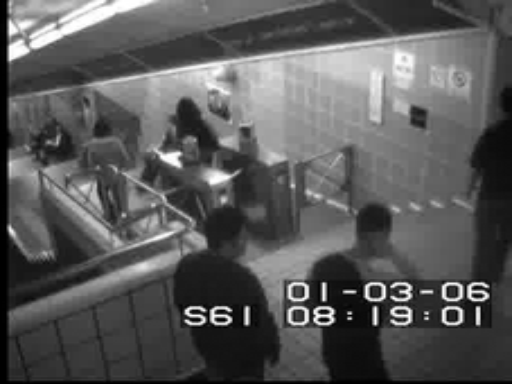}
		\includegraphics[scale=0.26]{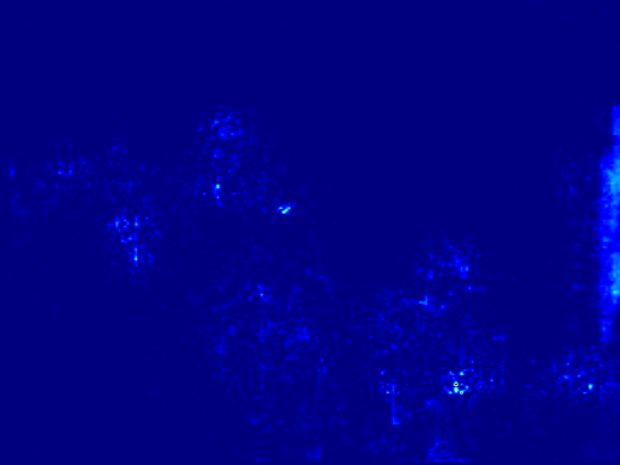}\\
		{\footnotesize Video \# 2, Frame \# 12170} 
		& {\footnotesize Video \# 3, Frame \# 11800}\\
		\includegraphics[scale=0.226]{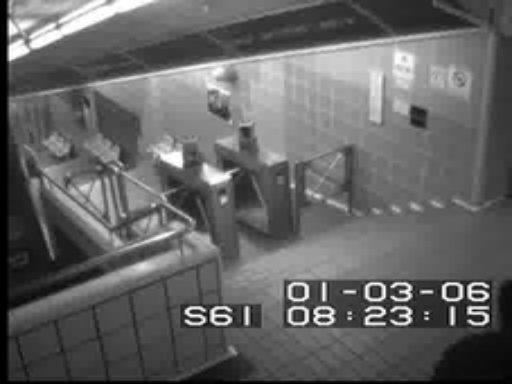}
		\includegraphics[scale=0.26]{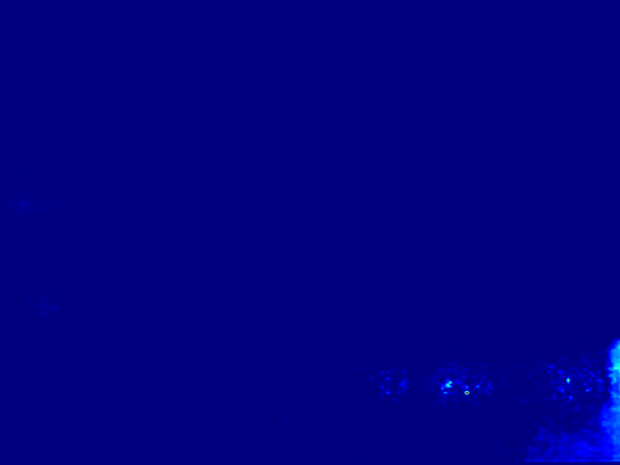}
		&\includegraphics[scale=0.226]{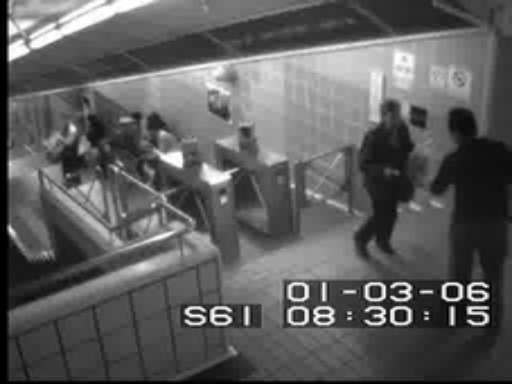}
		\includegraphics[scale=0.26]{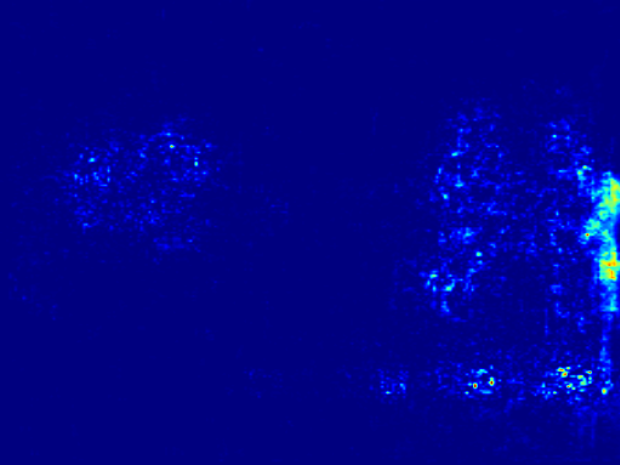}\\
		{\footnotesize Video \# 3, Frame \# 18150} 
		& {\footnotesize Video \# 4, Frame \# 8640}\\
	\end{tabular}
		\caption{Same layout in all figures as in Section \ref{sec:obj_det_avenue}. Moving persons are easily localized.}
		\label{fig:obj_det_enter}
\end{figure}

\begin{center}
	\hyperlink{page.11}{Go to Table of Contents}
\end{center}

\clearpage

\subsection{Subway Exit}
\label{sec:obj_det_exit}

\begin{figure}[h]
	\centering
	\begin{tabular}{cc}
		\includegraphics[scale=0.226]{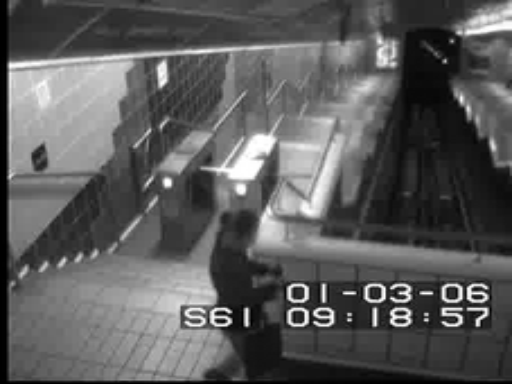}
		\includegraphics[scale=0.26]{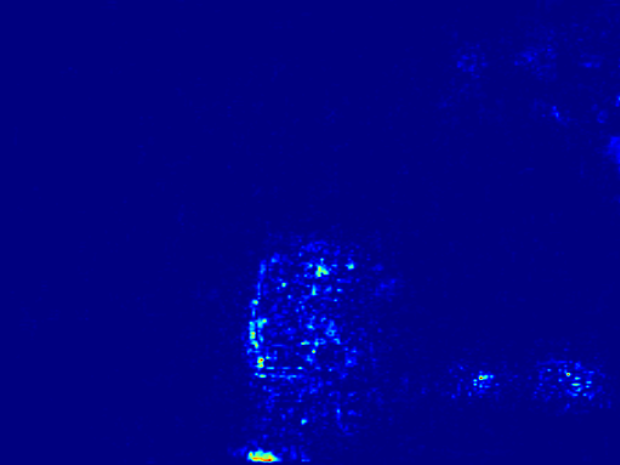}
		&\includegraphics[scale=0.226]{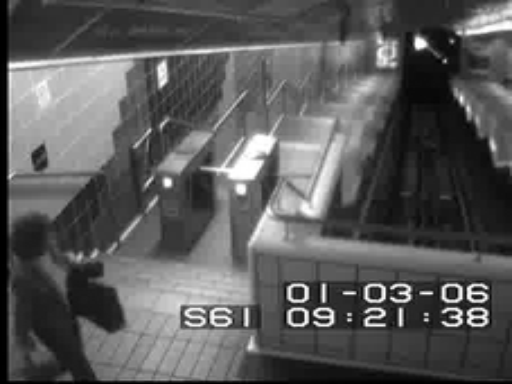}
		\includegraphics[scale=0.26]{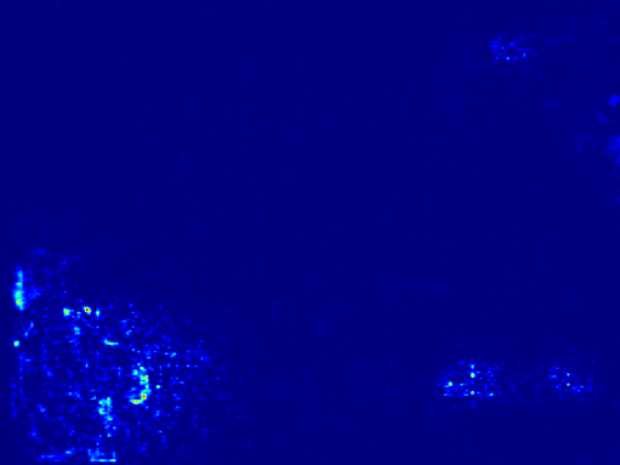}\\
		{\footnotesize Video \# 1, Frame \# 8370}
		& {\footnotesize Video \# 1, Frame \# 12390}\\
		\includegraphics[scale=0.226]{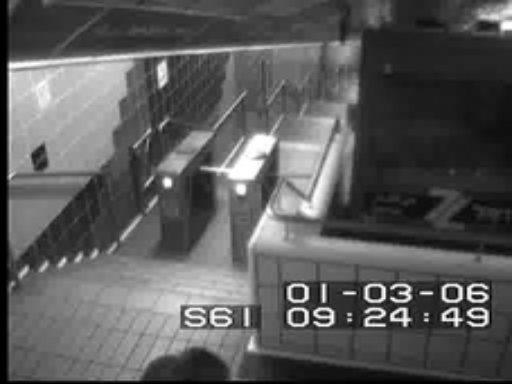}
		\includegraphics[scale=0.26]{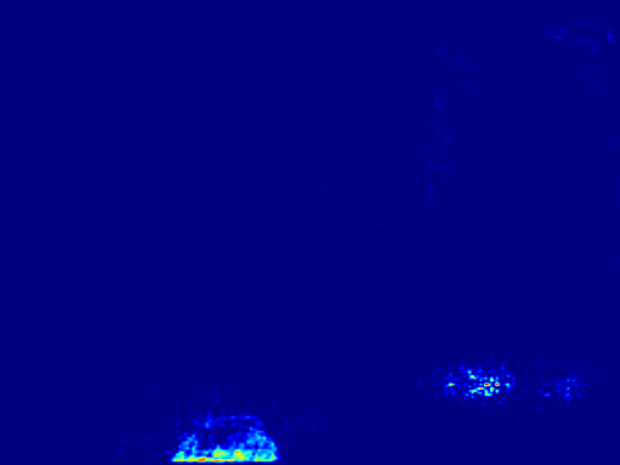}
		&\includegraphics[scale=0.226]{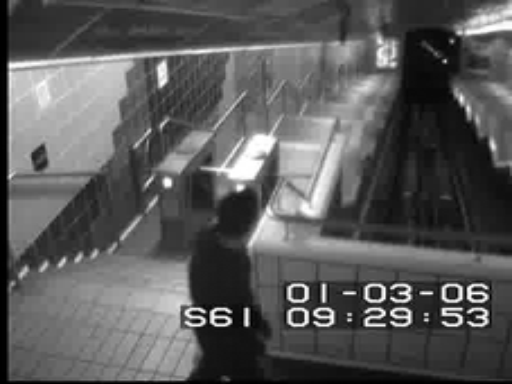}
		\includegraphics[scale=0.26]{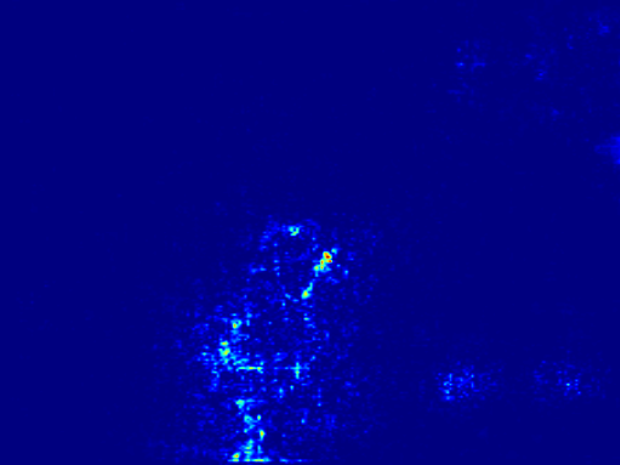}\\
		{\footnotesize Video \# 2, Frame \# 2170} 
		& {\footnotesize Video \# 2, Frame \# 9770}\\
		\includegraphics[scale=0.226]{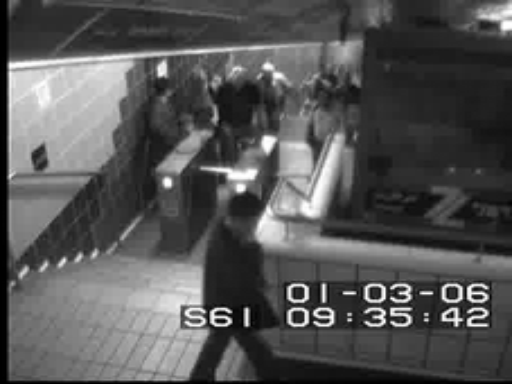}
		\includegraphics[scale=0.26]{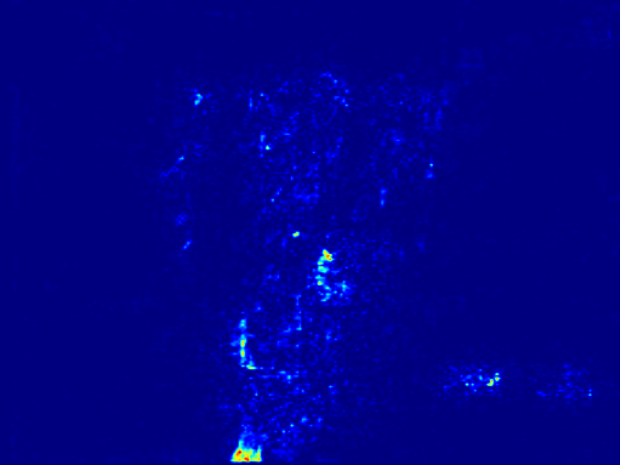}
		&\includegraphics[scale=0.226]{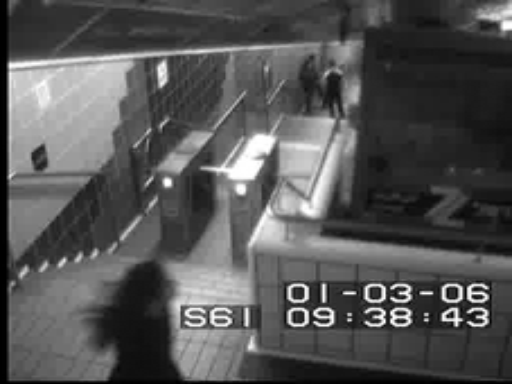}
		\includegraphics[scale=0.26]{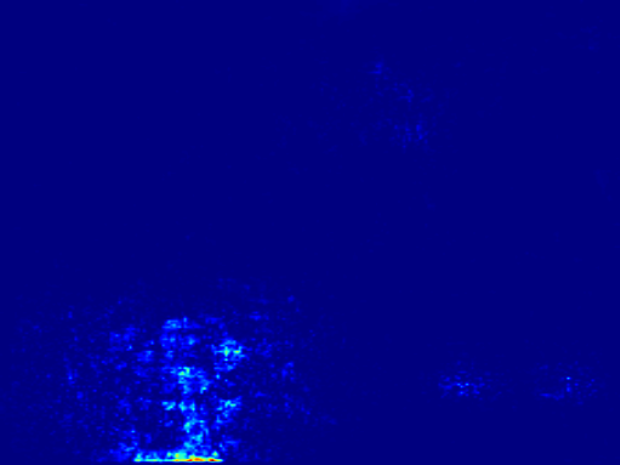}\\
		{\footnotesize Video \# 3, Frame \# 3490} 
		& {\footnotesize Video \# 3, Frame \# 8020}\\
		\includegraphics[scale=0.226]{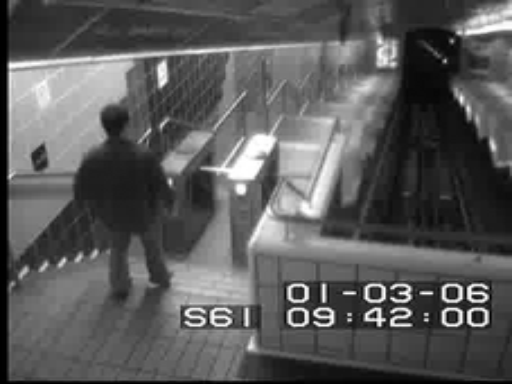}
		\includegraphics[scale=0.26]{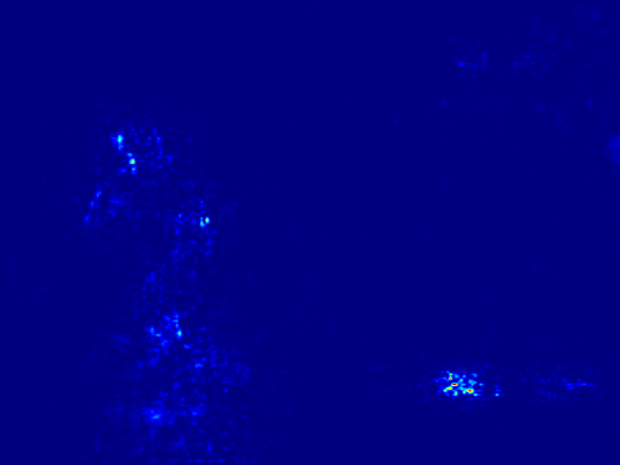}
		&\includegraphics[scale=0.226]{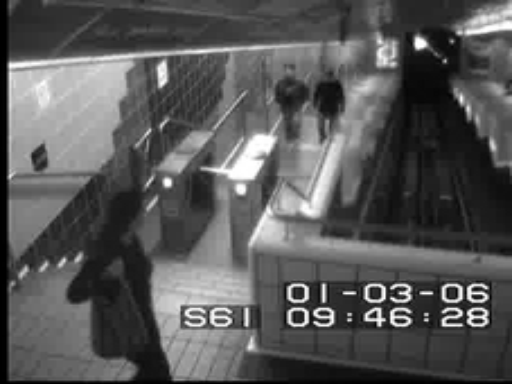}
		\includegraphics[scale=0.26]{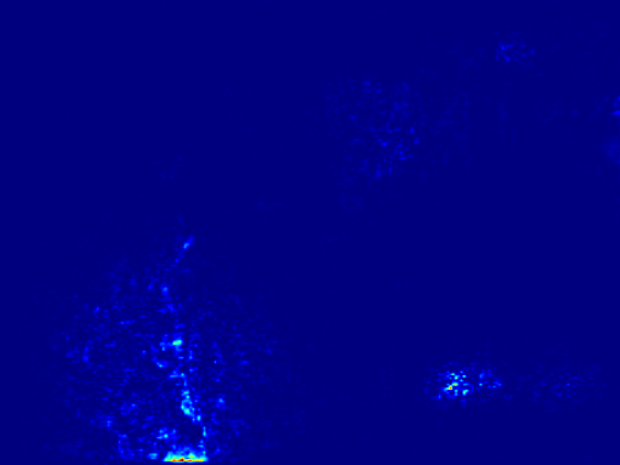}\\
		{\footnotesize Video \# 3, Frame \# 12940} 
		& {\footnotesize Video \# 4, Frame \# 4640}\\
	\end{tabular}
	\caption{Same layout in all figures as in Section \ref{sec:obj_det_avenue}. Similar to Subway Enter dataset; moving persons are easily revealed.}
	\label{fig:obj_det_exit}
\end{figure}

\begin{center}
	\hyperlink{page.11}{Go to Table of Contents}
\end{center}

\clearpage

\section{Predicting Past and Future Regular Frames}
\label{sec:frame_pred}
As in Section 4.4 in the main paper, we present a predicted regular frames of the past and the future of a given single image. 
The left most column in each figure in this section shows the given single image from which we predict the past and the future regular frames. 
Second column presents the images of 0.1 second before the moment of the given image. 
Third column presents the reconstructed `regular' frame of the moment of the given image.
Fourth column presents the images of 0.1 second after the moment of the given image.
Note that the objects involved in the irregular motions are gradually appearing from the past and gradually disappearing in the future.


\subsection{CUHK Avenue Dataset}
\label{sec:frame_pred_avenue}
It is best viewed in a video form: \texttt{frame\_pred\_avenue.mp4}

In the video, we put the all twelve videos into one file for the ease of playing. 
We first show the single seed frame for a second and show the predicted video followed by a blank frames.

\begin{figure}[h]
	\centering
	\begin{tabular}{c|ccc}
		\includegraphics[scale=0.18]{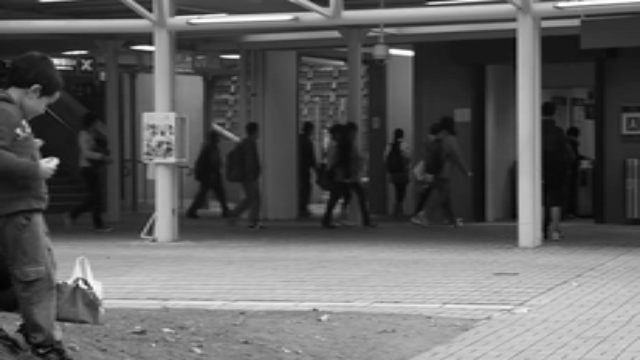}
		&\includegraphics[scale=0.18]{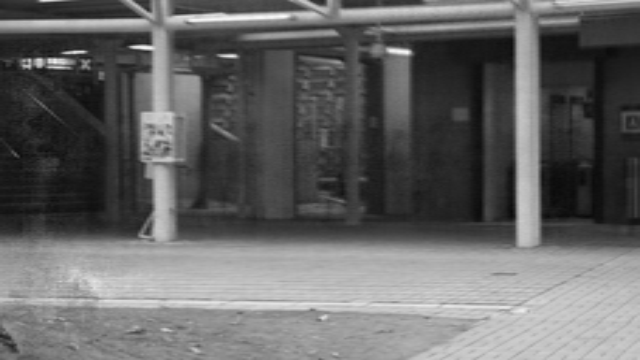}
		&\includegraphics[scale=0.18]{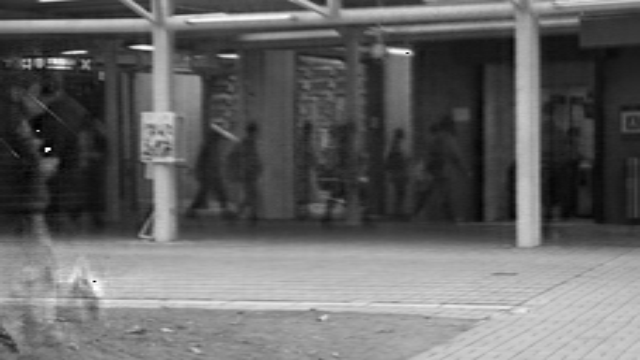}
		&\includegraphics[scale=0.18]{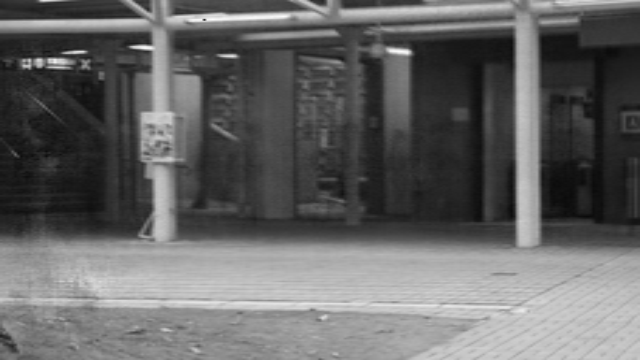}\\
		\multicolumn{4}{c}{{\footnotesize Video \# 1, Frame \# 600}} \\
		\includegraphics[scale=0.18]{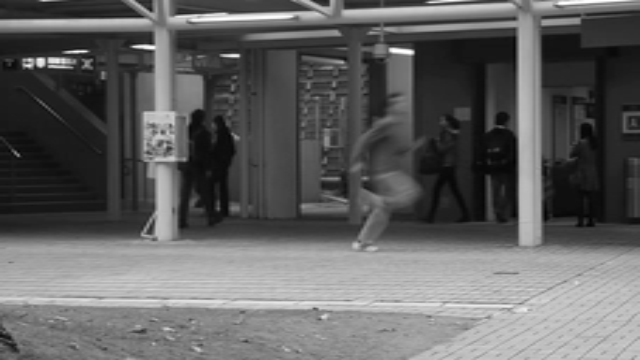}
		&\includegraphics[scale=0.18]{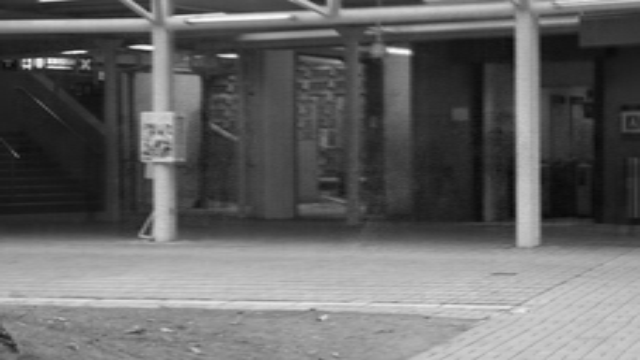}
		&\includegraphics[scale=0.18]{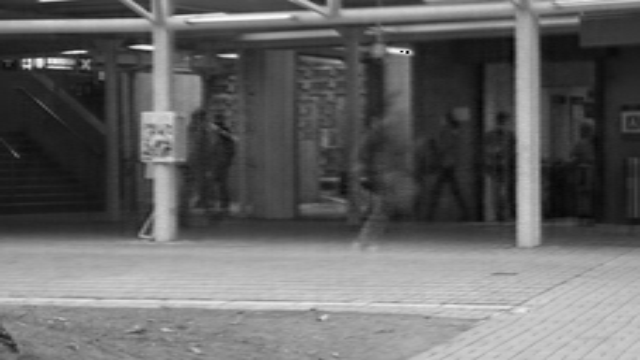}
		&\includegraphics[scale=0.18]{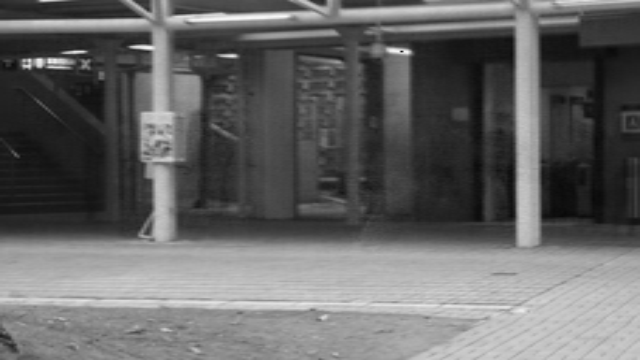}\\
		\multicolumn{4}{c}{{\footnotesize Video \# 3, Frame \# 600}} \\
		\includegraphics[scale=0.18]{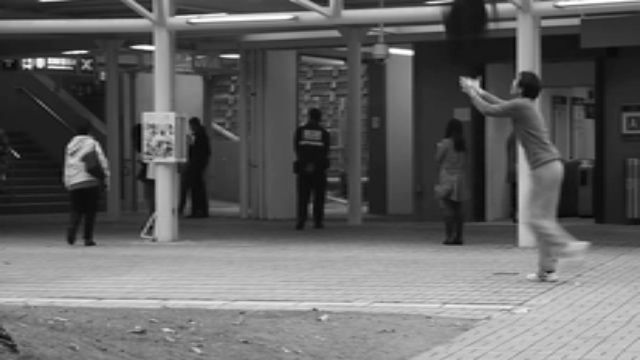}
		&\includegraphics[scale=0.18]{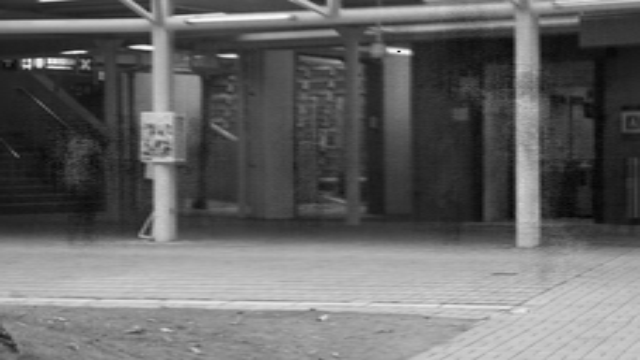}
		&\includegraphics[scale=0.18]{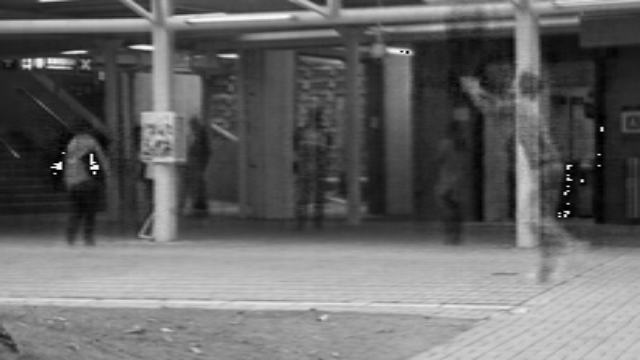}
		&\includegraphics[scale=0.18]{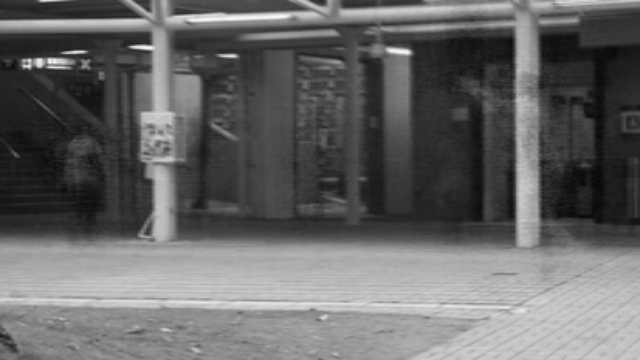}\\
		\multicolumn{4}{c}{{\footnotesize Video \# 5, Frame \# 600}} \\
		\includegraphics[scale=0.18]{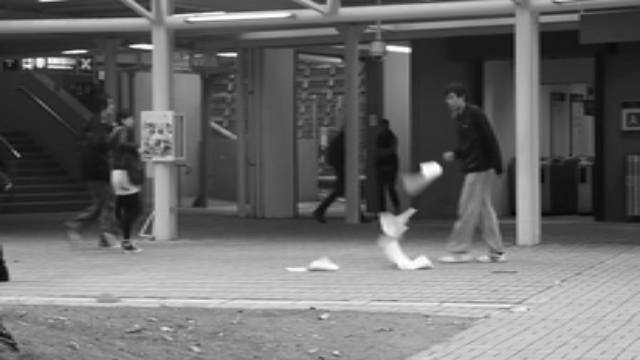}
		&\includegraphics[scale=0.18]{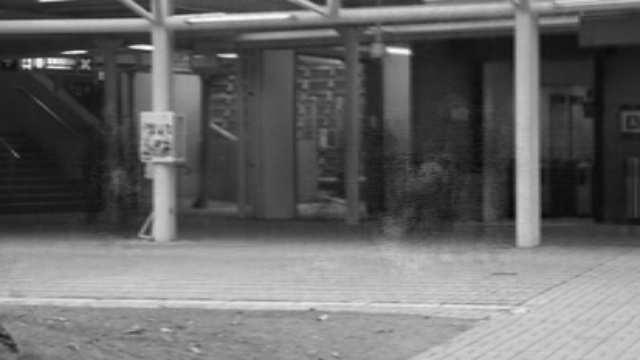}
		&\includegraphics[scale=0.18]{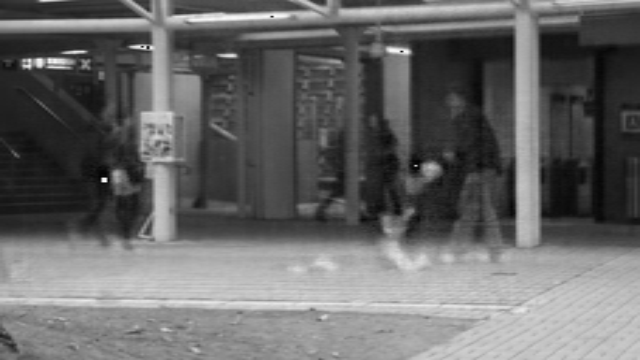}
		&\includegraphics[scale=0.18]{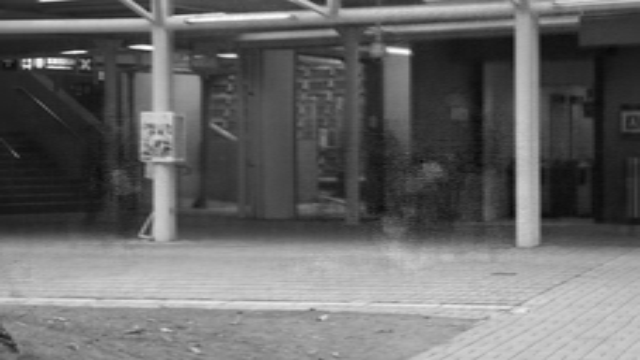}\\
		\multicolumn{4}{c}{{\footnotesize Video \# 13, Frame \# 490}} \\
		\includegraphics[scale=0.18]{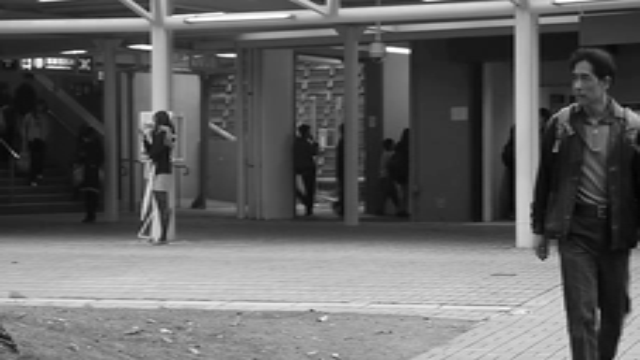}
		&\includegraphics[scale=0.18]{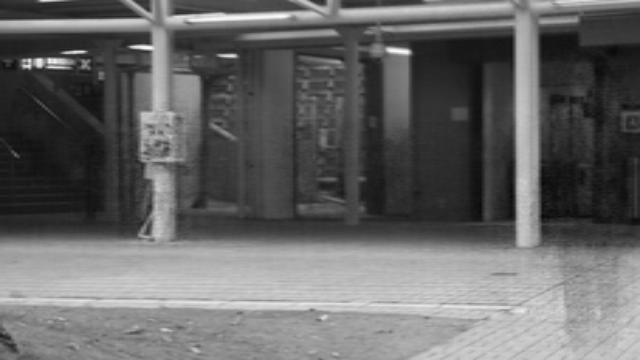}
		&\includegraphics[scale=0.18]{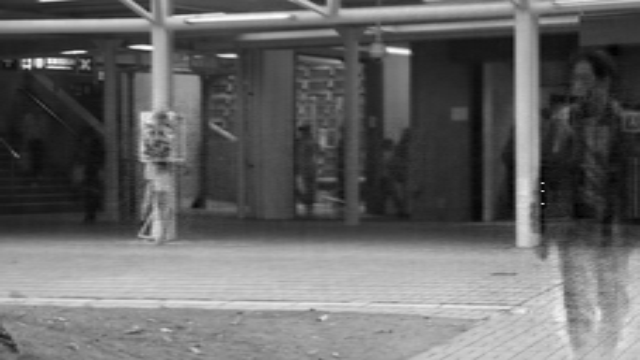}
		&\includegraphics[scale=0.18]{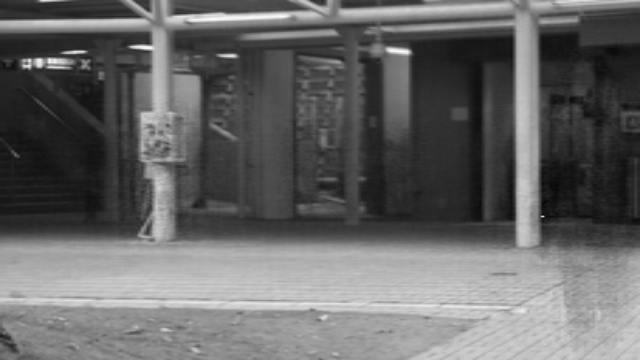}\\
		\multicolumn{4}{c}{{\footnotesize Video \# 15, Frame \# 550}} \\
	\end{tabular}
		\caption{Same layout as discussed in Section \ref{sec:frame_pred}. The regularity enforces that the objects involved in irregular motion gradually appearing and disappearing. In video 1 (first row), the crowd and the person in front are gradually appearing and disappearing. In video 13, in the future frame, the paper is closer to the ground compared to the past.}
		\label{fig:frame_pred_avenue}
\end{figure}

\begin{center}
	\hyperlink{page.11}{Go to Table of Contents}
\end{center}

\clearpage

\subsection{UCSD Ped1}
\label{sec:frame_pred_ped1}
We do not provide a video for this dataset but figure will explain it.

\begin{figure}[h]
	\centering
	\begin{tabular}{c|ccc}
		\includegraphics[scale=0.45]{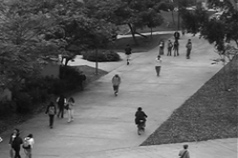}
		&\includegraphics[scale=0.45]{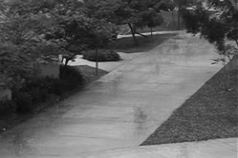}
		&\includegraphics[scale=0.45]{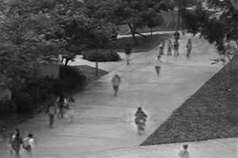}
		&\includegraphics[scale=0.45]{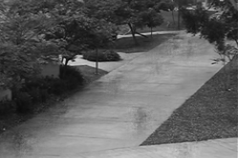}\\
		\multicolumn{4}{c}{{\footnotesize Video \# 2, Frame \# 70}} \\
		\includegraphics[scale=0.45]{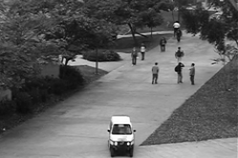}
		&\includegraphics[scale=0.45]{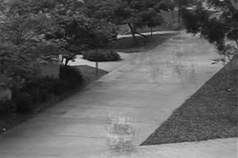}
		&\includegraphics[scale=0.45]{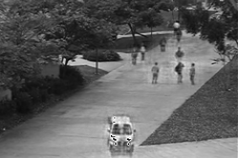}
		&\includegraphics[scale=0.45]{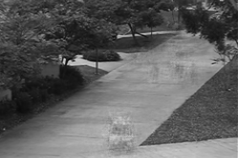}\\
		\multicolumn{4}{c}{{\footnotesize Video \# 19, Frame \# 120}} \\
		\includegraphics[scale=0.45]{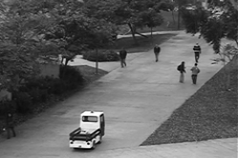}
		&\includegraphics[scale=0.45]{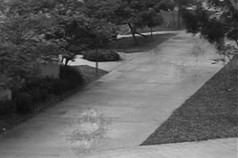}
		&\includegraphics[scale=0.45]{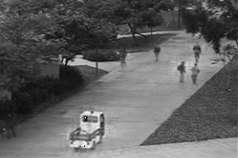}
		&\includegraphics[scale=0.45]{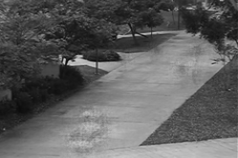}\\
		\multicolumn{4}{c}{{\footnotesize Video \# 20, Frame \# 60}} \\
		\includegraphics[scale=0.45]{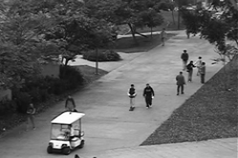}
		&\includegraphics[scale=0.45]{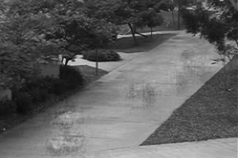}
		&\includegraphics[scale=0.45]{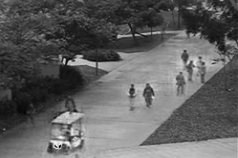}
		&\includegraphics[scale=0.45]{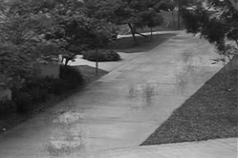}\\
		\multicolumn{4}{c}{{\footnotesize Video \# 24, Frame \# 150}} \\
		\includegraphics[scale=0.45]{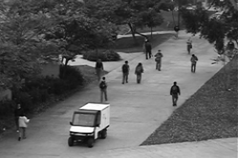}
		&\includegraphics[scale=0.45]{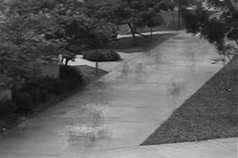}
		&\includegraphics[scale=0.45]{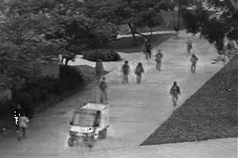}
		&\includegraphics[scale=0.45]{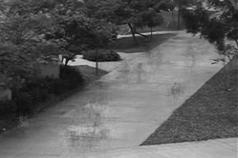}\\
		\multicolumn{4}{c}{{\footnotesize Video \# 27, Frame \# 90}} \\
	\end{tabular}
	\caption{Same layout as discussed in Section \ref{sec:frame_pred}. In video 20, the car moves a little bit upwards in the future frame.}
	\label{fig:frame_pred_ped1}
\end{figure}

\begin{center}
	\hyperlink{page.11}{Go to Table of Contents}
\end{center}

\clearpage

\subsection{UCSD Ped2}
\label{sec:frame_pred_ped2}
We do not provide a video for this dataset but figure will explain it.

\begin{figure}[h]
	\centering
	\begin{tabular}{c|ccc}
		\includegraphics[scale=0.32]{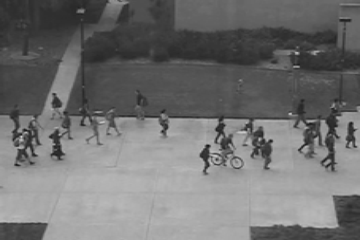}
		&\includegraphics[scale=0.32]{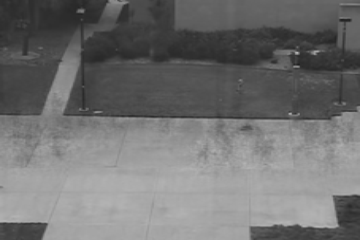}
		&\includegraphics[scale=0.32]{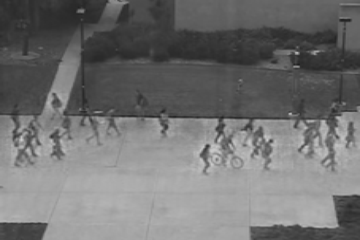}
		&\includegraphics[scale=0.32]{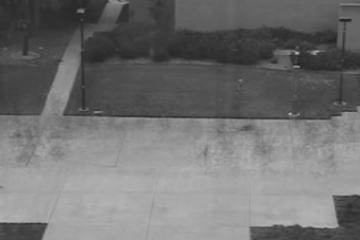}\\
		\multicolumn{4}{c}{{\footnotesize Video \# 1, Frame \# 160}} \\
		\includegraphics[scale=0.32]{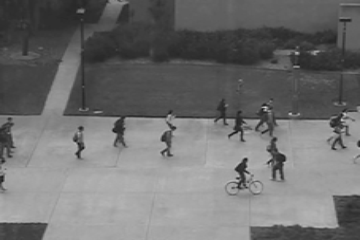}
		&\includegraphics[scale=0.32]{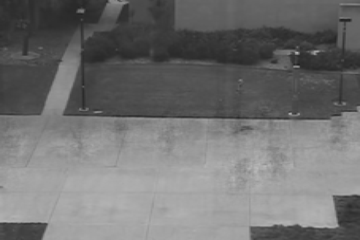}
		&\includegraphics[scale=0.32]{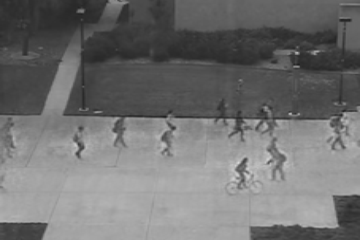}
		&\includegraphics[scale=0.32]{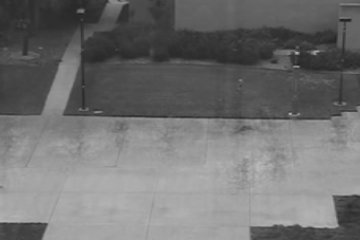}\\
		\multicolumn{4}{c}{{\footnotesize Video \# 2, Frame \# 160}} \\
		\includegraphics[scale=0.32]{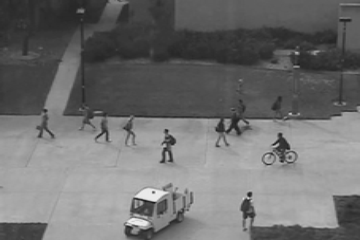}
		&\includegraphics[scale=0.32]{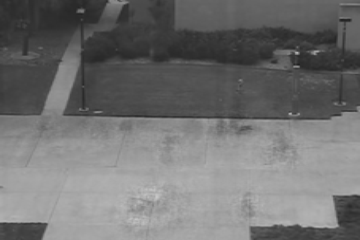}
		&\includegraphics[scale=0.32]{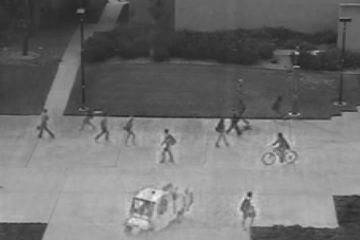}
		&\includegraphics[scale=0.32]{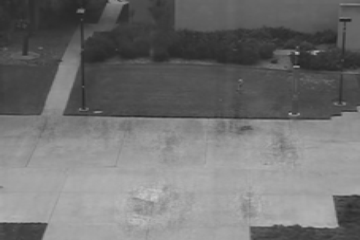}\\
		\multicolumn{4}{c}{{\footnotesize Video \# 4, Frame \# 150}} \\
		\includegraphics[scale=0.32]{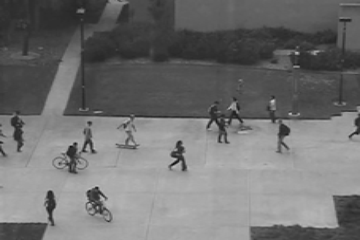}
		&\includegraphics[scale=0.32]{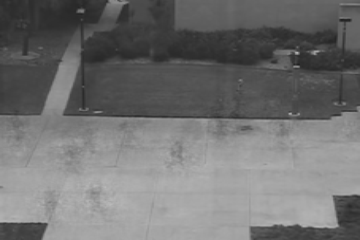}
		&\includegraphics[scale=0.32]{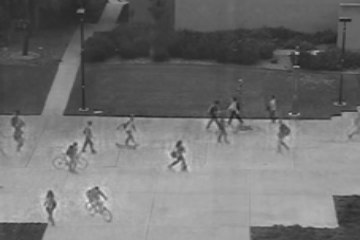}
		&\includegraphics[scale=0.32]{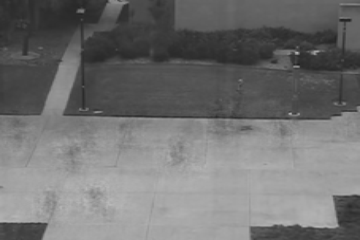}\\
		\multicolumn{4}{c}{{\footnotesize Video \# 7, Frame \# 120}} \\
		\includegraphics[scale=0.32]{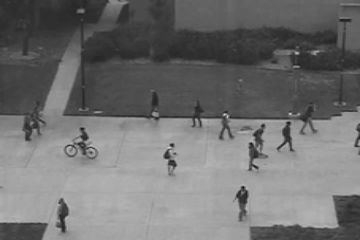}
		&\includegraphics[scale=0.32]{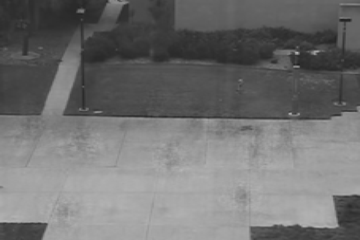}
		&\includegraphics[scale=0.32]{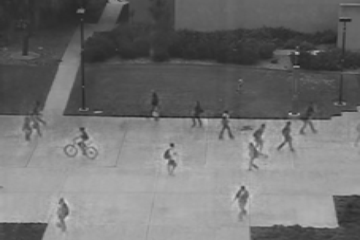}
		&\includegraphics[scale=0.32]{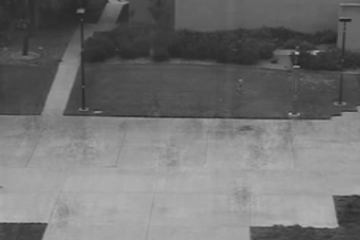}\\
		\multicolumn{4}{c}{{\footnotesize Video \# 8, Frame \# 130}} \\
	\end{tabular}
	\caption{Same layout as discussed in Section \ref{sec:frame_pred}. In video 4, the car appears clearer than the past frame predicted (second) and moves a bit more south than the past frames.}
	\label{fig:frame_pred_ped2}
\end{figure}

\begin{center}
	\hyperlink{page.11}{Go to Table of Contents}
\end{center}

\clearpage

\subsection{Subway Enter}
\label{sec:frame_pred_enter}
We do not provide a video for this dataset but figure will explain it.

\begin{figure}[h]
	\centering
	\begin{tabular}{c|ccc}
		\includegraphics[scale=0.22]{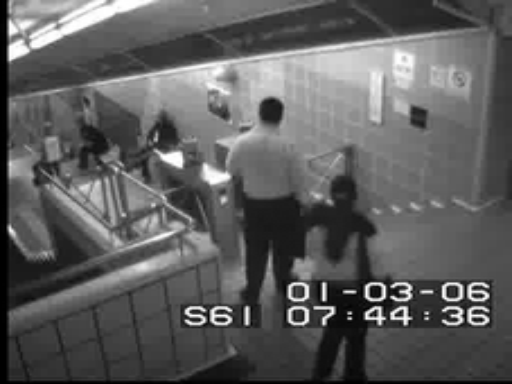}
		&\includegraphics[scale=0.22]{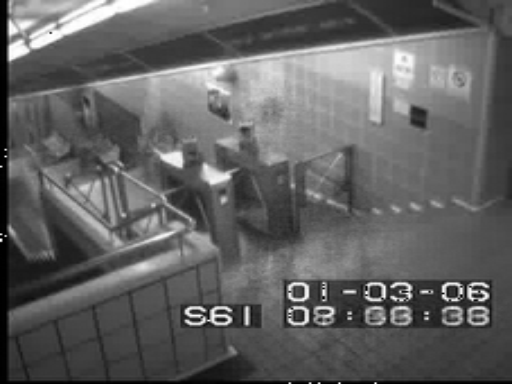}
		&\includegraphics[scale=0.22]{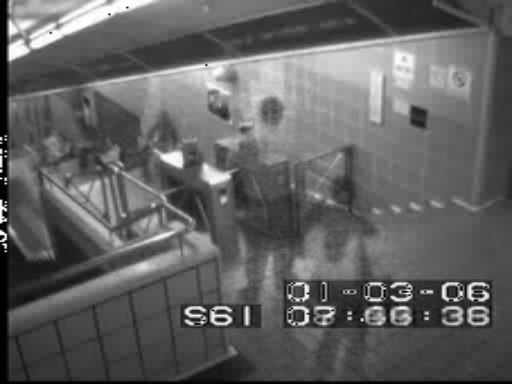}
		&\includegraphics[scale=0.22]{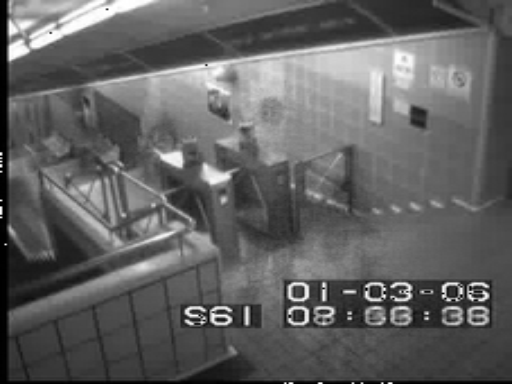}\\
		\multicolumn{4}{c}{{\footnotesize Video \# 1, Frame \# 180}} \\
		\includegraphics[scale=0.22]{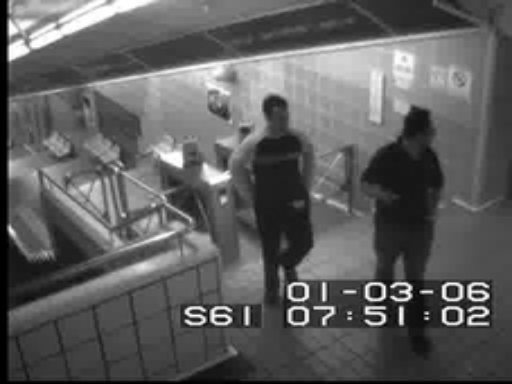}
		&\includegraphics[scale=0.22]{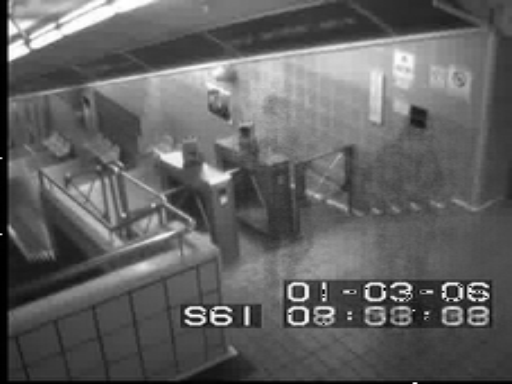}
		&\includegraphics[scale=0.22]{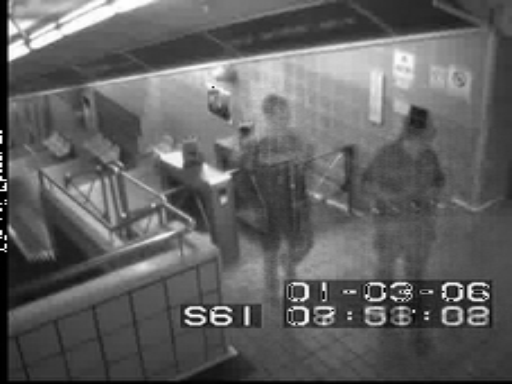}
		&\includegraphics[scale=0.22]{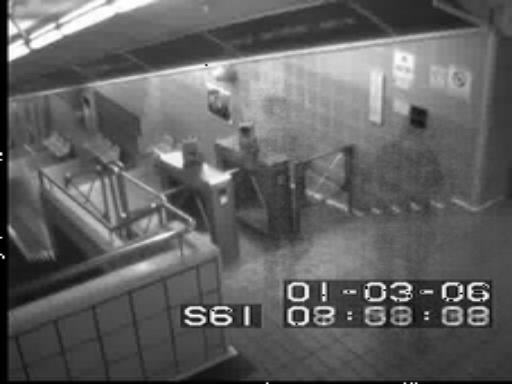}\\
		\multicolumn{4}{c}{{\footnotesize Video \# 1, Frame \# 9830}} \\
		\includegraphics[scale=0.22]{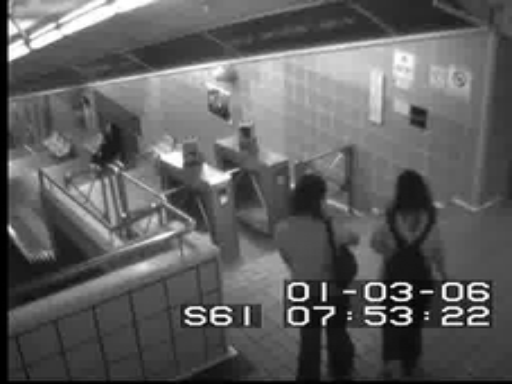}
		&\includegraphics[scale=0.22]{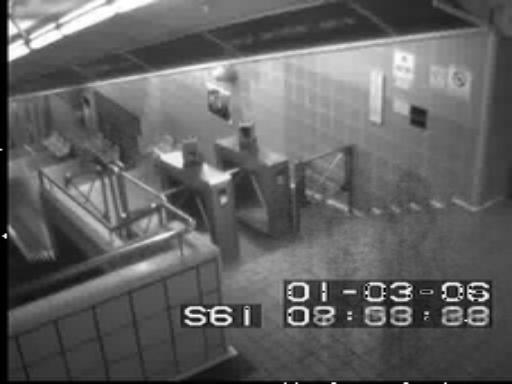}
		&\includegraphics[scale=0.22]{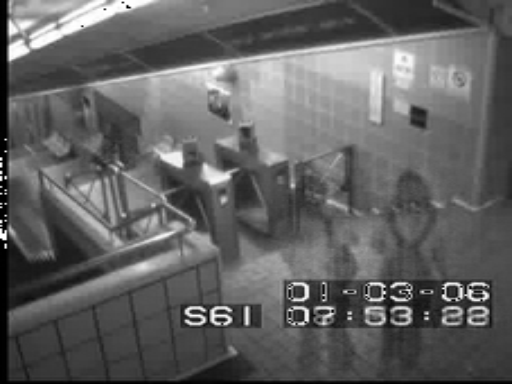}
		&\includegraphics[scale=0.22]{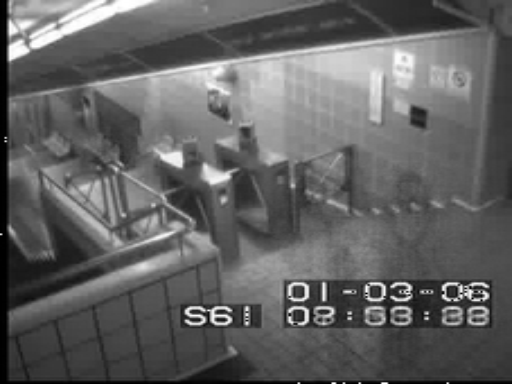}\\
		\multicolumn{4}{c}{{\footnotesize Video \# 1, Frame \# 13310}} \\
		\includegraphics[scale=0.22]{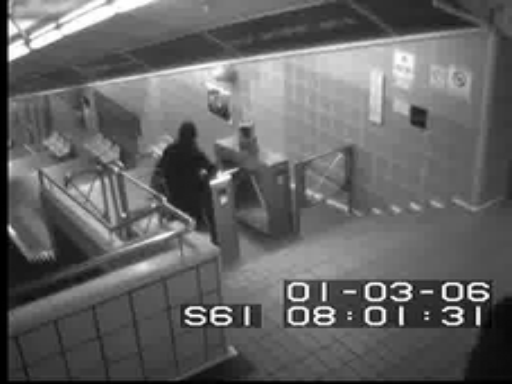}
		&\includegraphics[scale=0.22]{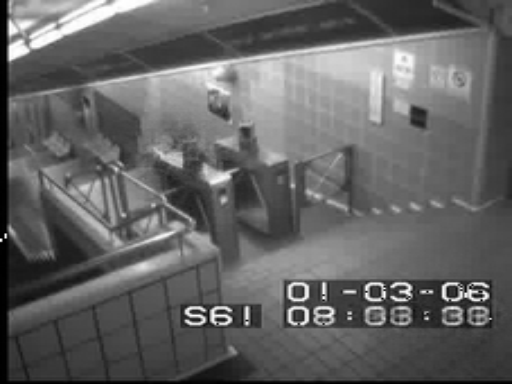}
		&\includegraphics[scale=0.22]{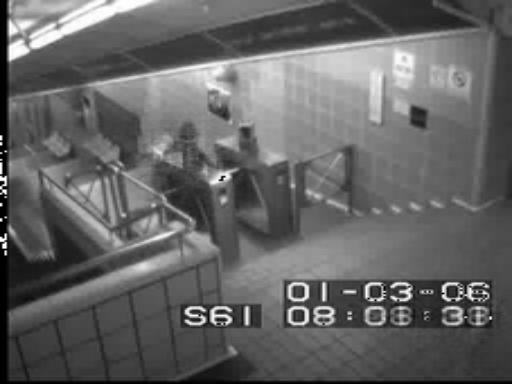}
		&\includegraphics[scale=0.22]{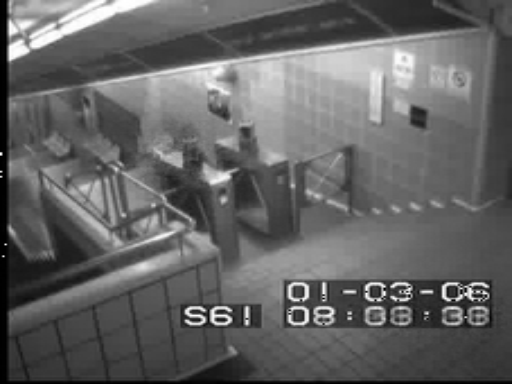}\\
		\multicolumn{4}{c}{{\footnotesize Video \# 2, Frame \# 5540}} \\
		\includegraphics[scale=0.22]{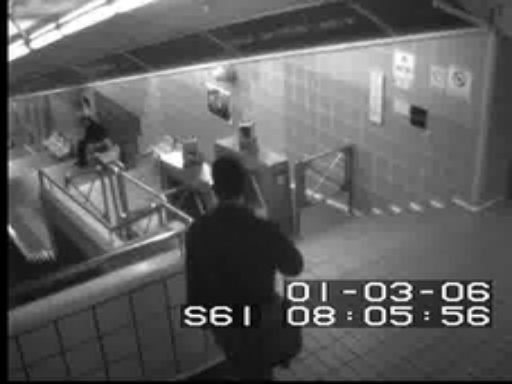}
		&\includegraphics[scale=0.22]{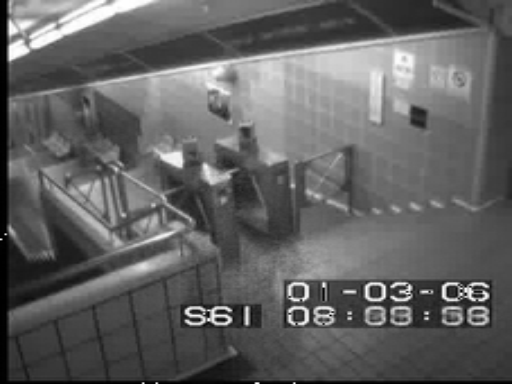}
		&\includegraphics[scale=0.22]{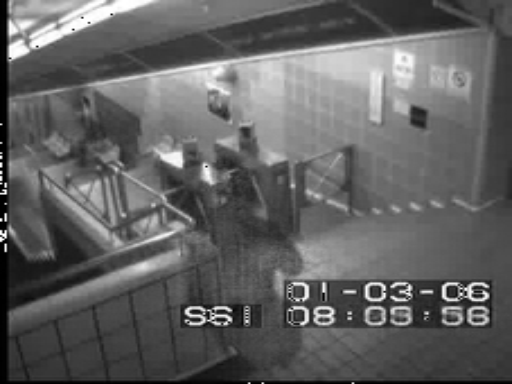}
		&\includegraphics[scale=0.22]{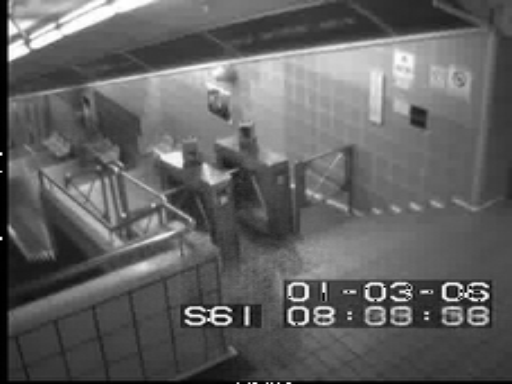}\\
		\multicolumn{4}{c}{{\footnotesize Video \# 2, Frame \# 12170}} \\
	\end{tabular}
	\caption{Same layout as discussed in Section \ref{sec:frame_pred}.}
	\label{fig:frame_pred_enter}
\end{figure}

\begin{center}
	\hyperlink{page.11}{Go to Table of Contents}
\end{center}

\clearpage

\subsection{Subway Exit}
\label{sec:frame_pred_exit}
We do not provide a video for this dataset but figure will explain it.

\begin{figure}[h]
	\centering
	\begin{tabular}{c|ccc}
		\includegraphics[scale=0.22]{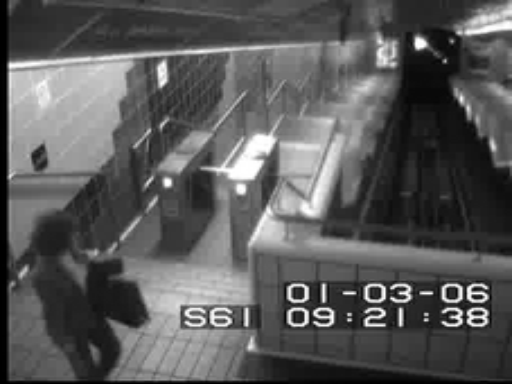}
		&\includegraphics[scale=0.22]{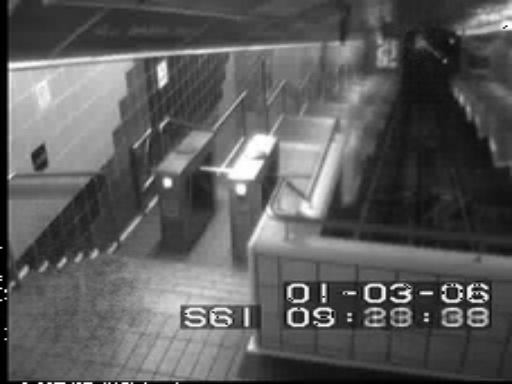}
		&\includegraphics[scale=0.22]{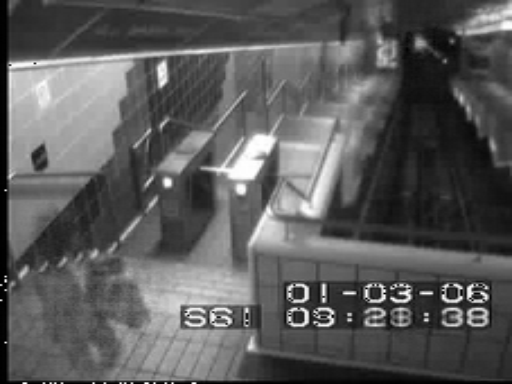}
		&\includegraphics[scale=0.22]{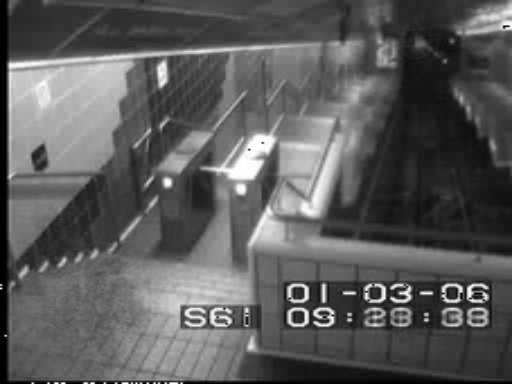}\\
		\multicolumn{4}{c}{{\footnotesize Video \# 1, Frame \# 12390}} \\
		\includegraphics[scale=0.22]{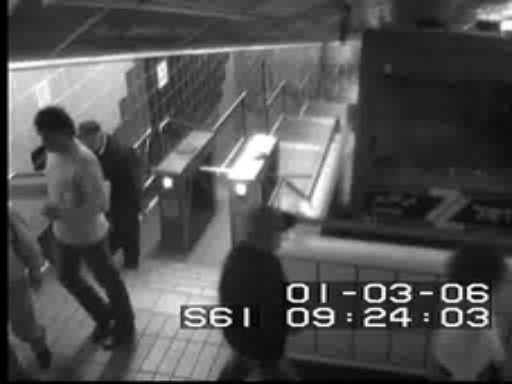}
		&\includegraphics[scale=0.22]{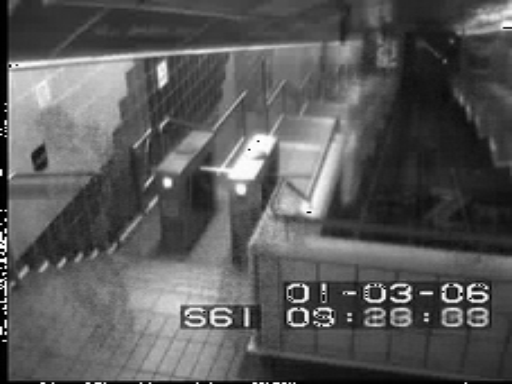}
		&\includegraphics[scale=0.22]{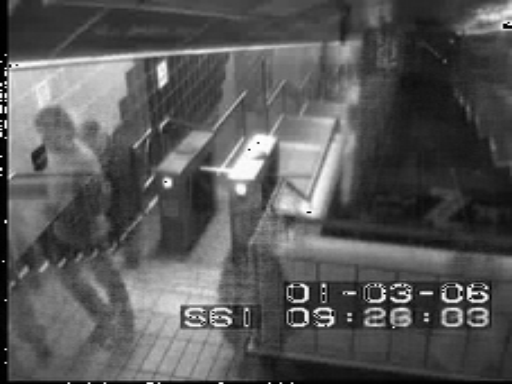}
		&\includegraphics[scale=0.22]{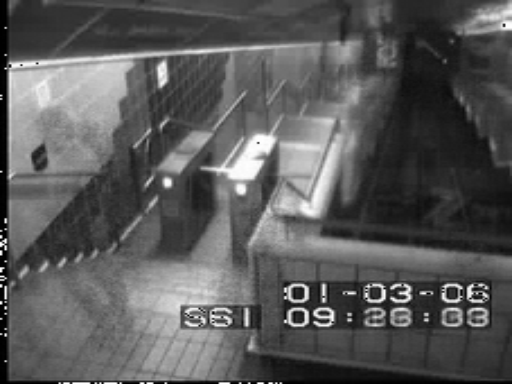}\\
		\multicolumn{4}{c}{{\footnotesize Video \# 1, Frame \# 1010}} \\
		\includegraphics[scale=0.22]{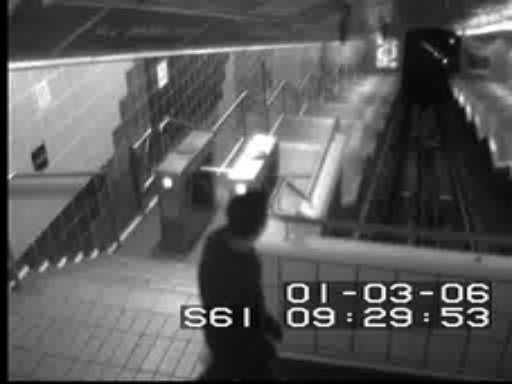}
		&\includegraphics[scale=0.22]{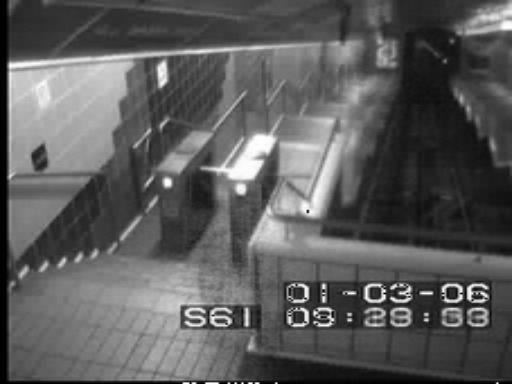}
		&\includegraphics[scale=0.22]{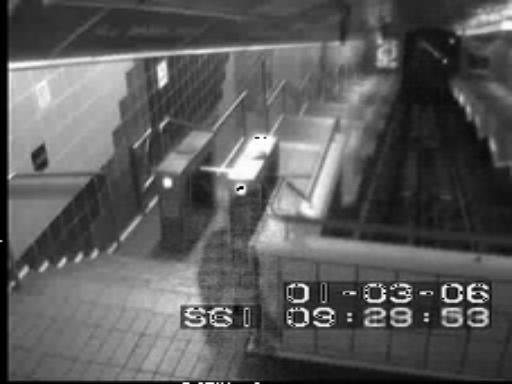}
		&\includegraphics[scale=0.22]{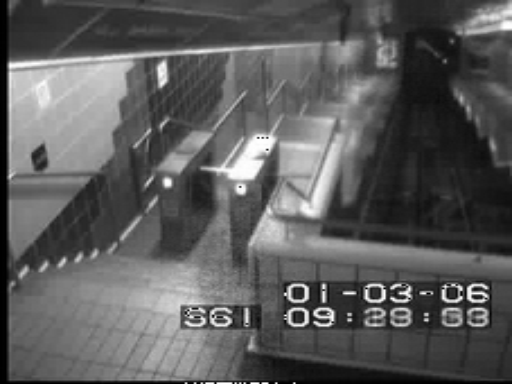}\\
		\multicolumn{4}{c}{{\footnotesize Video \# 2, Frame \# 9770}} \\
		\includegraphics[scale=0.22]{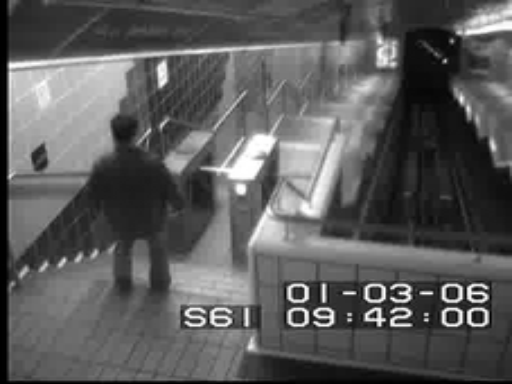}
		&\includegraphics[scale=0.22]{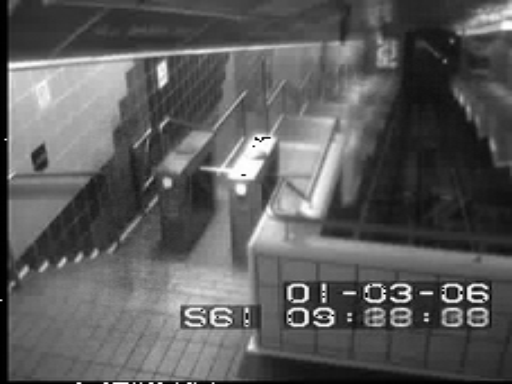}
		&\includegraphics[scale=0.22]{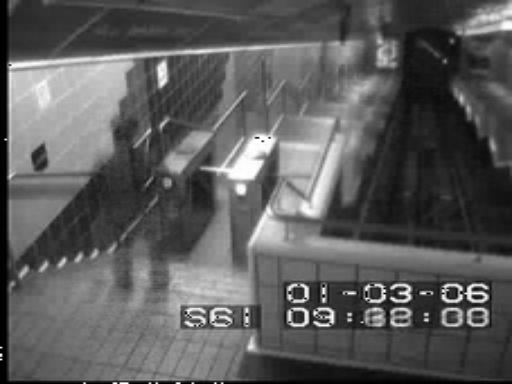}
		&\includegraphics[scale=0.22]{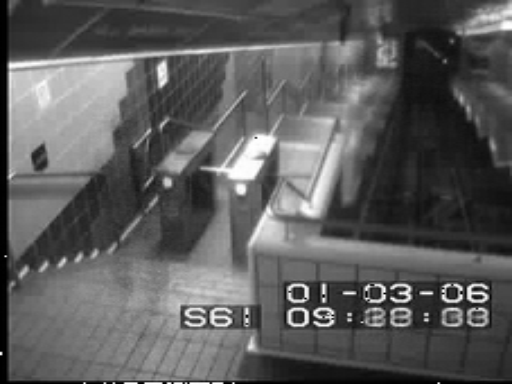}\\
		\multicolumn{4}{c}{{\footnotesize Video \# 3, Frame \# 12940}} \\
		\includegraphics[scale=0.22]{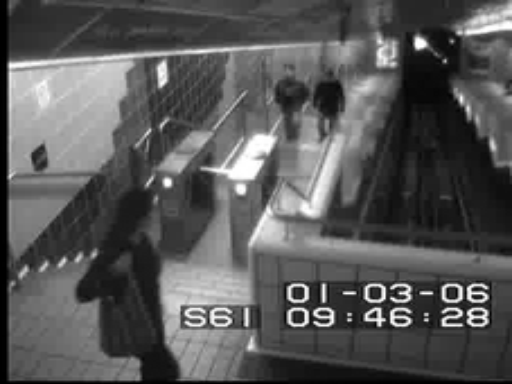}
		&\includegraphics[scale=0.22]{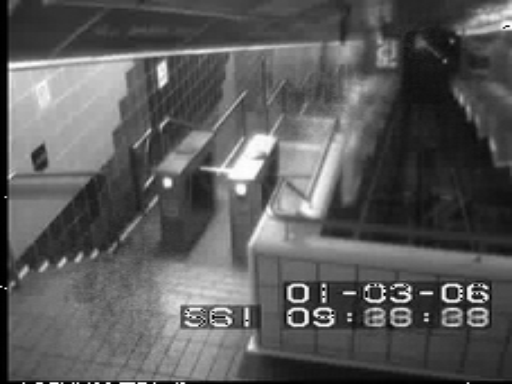}
		&\includegraphics[scale=0.22]{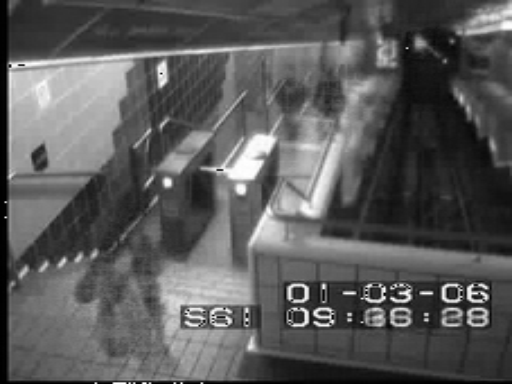}
		&\includegraphics[scale=0.22]{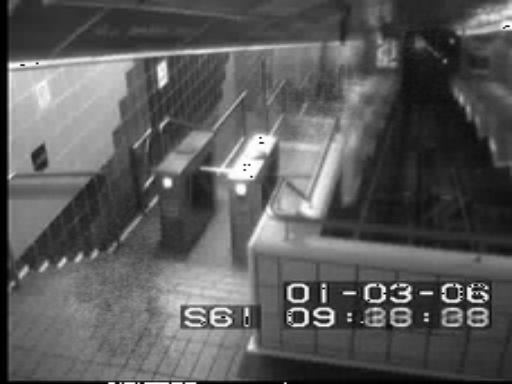}\\
		\multicolumn{4}{c}{{\footnotesize Video \# 4, Frame \# 4640}} \\
	\end{tabular}
		\caption{Same layout as discussed in Section \ref{sec:frame_pred}.}
		\label{fig:frame_pred_exit}
\end{figure}

\begin{center}
	\hyperlink{page.11}{Go to Table of Contents}
\end{center}

\clearpage

\section{Anomalous Event Detection and Generalization Analysis on Multiple Datasets}
\label{sec:anomaly}
We visualize the regularity score (defined in Eq.~(3) in the main paper) to detect anomalous events in video.
When the regularity score is low in a local temporal window, the video segment is determined containing anomalous events.
We additionally compare with the generalizability of the trained model using various training sets.
Blue (conventional) represents the score obtained by a model trained on the \emph{specific target} dataset.
Red (generalized) represents the score obtained by a model trained on \emph{all} datasets. (This is the model we use for all other experiments.)
Yellow (transfer) represents the score obtained by a model trained on \emph{all datasets except that specific target} datasets.

\subsection{CUHK Avenue Dataset}
\label{sec:anomaly_avenue}
The target dataset is CUHK Avenue.
Thus, the `conventional' represents the score obtained by a model trained only on the Avenue dataset.
The `generalized' represents the score obtained by a model trained on all datasets we used.
The `transfer' represents the score obtained by a model trained on all datasets except the Avenue dataset.
Surprisingly, the generalized model performs very well same as the target model (conventional).
And the transfer model also performs decently.

By comparing `conventional' and `generalized', we observe that the model is powerful enough not being harmed by other datasets.
At the same time, by comparing `transfer' and either `generalized' or `conventional', we observe that the model is not too much overfitting to the given dataset as it can generalized to \emph{unseen} videos.
Consequently, we believe that the proposed network structure is well balanced between overfitting and underfitting.
Red shaded region represents the ground truth anomalous temporal segments defined by each data curators.

\begin{figure}[h]
	\centering
	\begin{tabular}{cc}
		\includegraphics[scale=0.38]{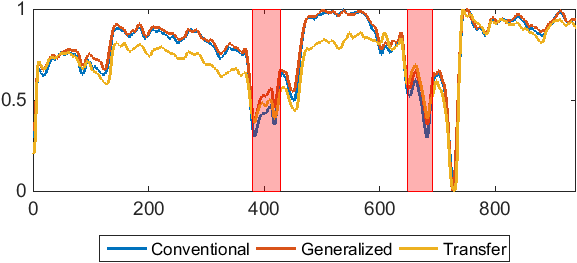}
		&\includegraphics[scale=0.38]{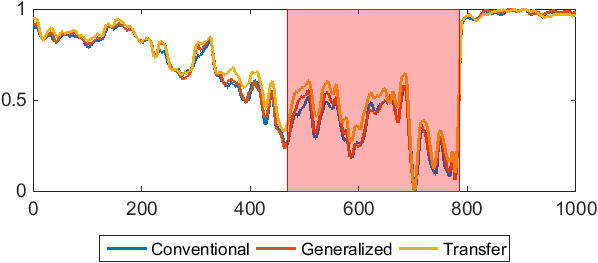}\\
		{\footnotesize Video \# 4} & {\footnotesize Video \# 5}\\
		\includegraphics[scale=0.38]{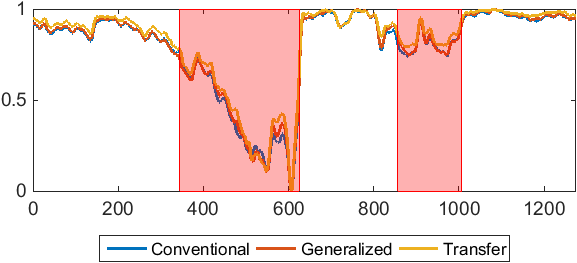}
		&\includegraphics[scale=0.38]{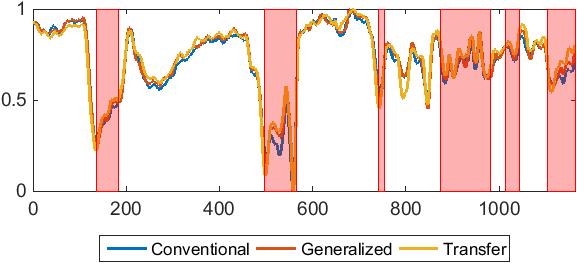}\\
		{\footnotesize Video \#6 } & {\footnotesize Video \# 9}\\
		\includegraphics[scale=0.38]{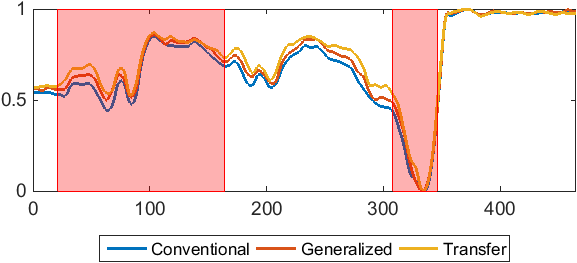}
		&\includegraphics[scale=0.38]{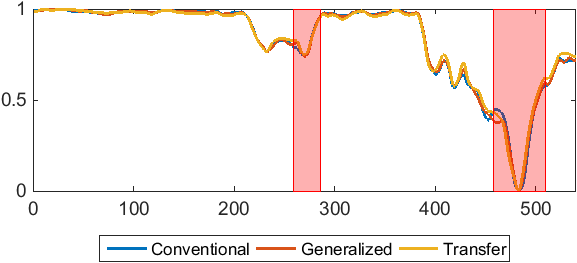}\\
		{\footnotesize Video \# 11} & {\footnotesize Video \# 13}\\
		\includegraphics[scale=0.38]{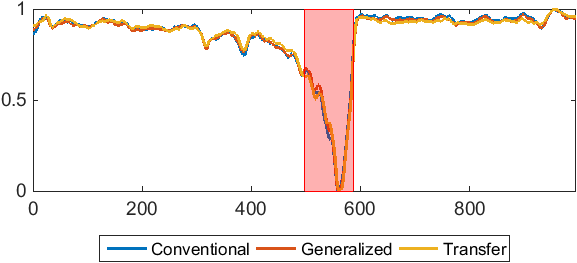}
		&\includegraphics[scale=0.38]{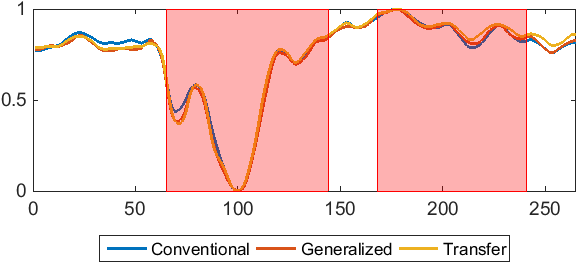}\\
		{\footnotesize Video \# 15} & {\footnotesize Video \# 20}\\
	\end{tabular}
	\caption{Our model captures anomalous regions as a form of local minima. In some of the ground truth anomalous region, however, the regularity score is not as much low as other regions. This is mainly due to the the anomalous action is happening in a small region or is well blended with the appearances of regular activity.}
\end{figure}
\vspace{-8mm}
\begin{center}
	\hyperlink{page.11}{Go to Table of Contents}
\end{center}

\clearpage

\subsection{UCSD Ped1}
\label{sec:anomaly_ped1}
Similar to Avenue dataset, the generalized model performs very well same as the target model (conventional) and the transfer model also performs very decently.

\begin{figure}[h]
	\centering
	\begin{tabular}{cc}
		\includegraphics[scale=0.58]{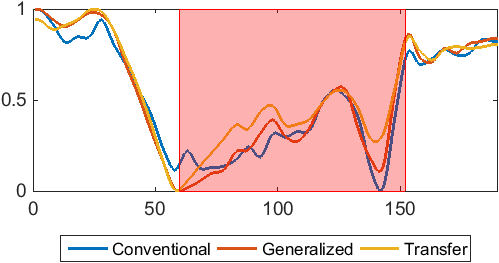}
		&\includegraphics[scale=0.58]{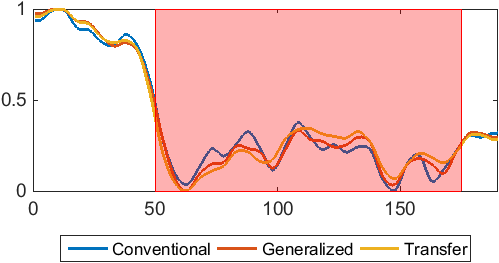}\\
		{\footnotesize Video \# 1} & {\footnotesize Video \# 2}\\
		\includegraphics[scale=0.58]{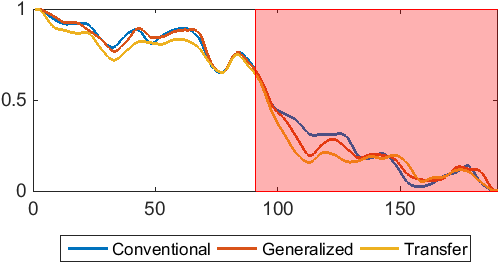}
		&\includegraphics[scale=0.58]{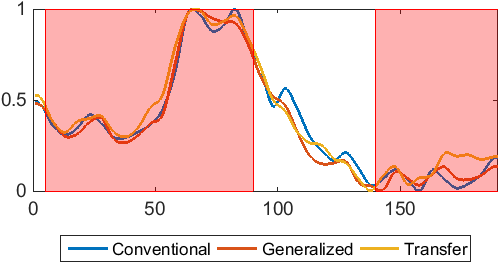}\\
		{\footnotesize Video \# 3} & {\footnotesize Video \# 5}\\
		\includegraphics[scale=0.58]{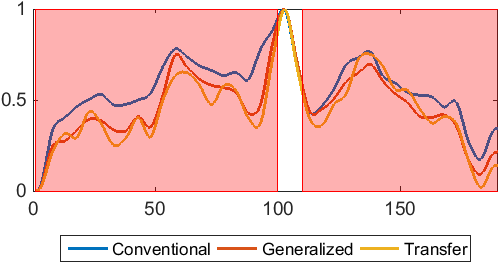}
		&\includegraphics[scale=0.58]{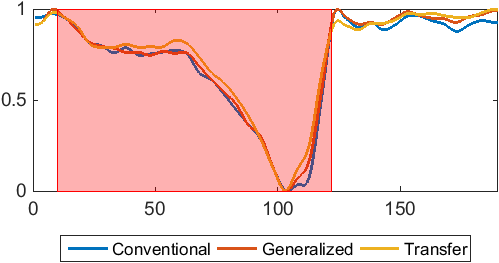}\\
		{\footnotesize Video \# 6} & {\footnotesize Video \# 27}\\
		\includegraphics[scale=0.58]{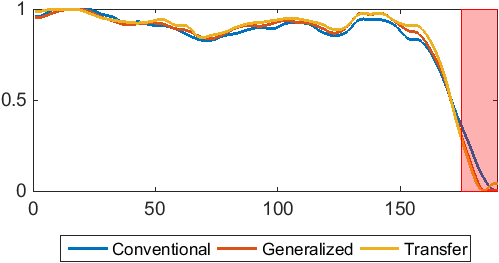}
		&\includegraphics[scale=0.58]{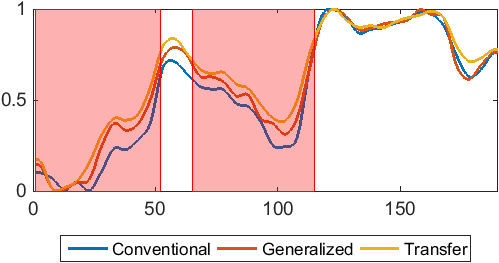}\\
		{\footnotesize Video \# 30} & {\footnotesize Video \# 32}\\
	\end{tabular}
	\caption{In video \#5, at the end of the first region we have high regularity score even though it is in anomalous regions. This is mainly because the definition of anomalous event is different from the definition of regularity; regularity means temporally ordinary motions whereas the anomalous event can be defined as necessary - thus regular motion can be defined as anomaly.}
\end{figure}

\begin{center}
	\hyperlink{page.11}{Go to Table of Contents}
\end{center}

\clearpage

\subsection{UCSD Ped2}
\label{sec:anomaly_ped2}

\begin{figure}[h]
	\centering
	\begin{tabular}{cc}
		\includegraphics[scale=0.6]{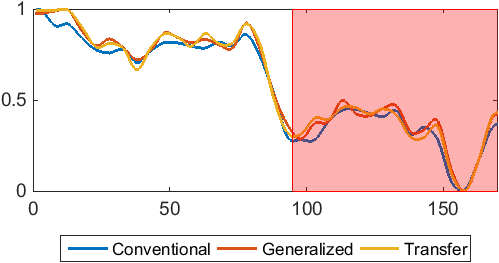}
		&\includegraphics[scale=0.6]{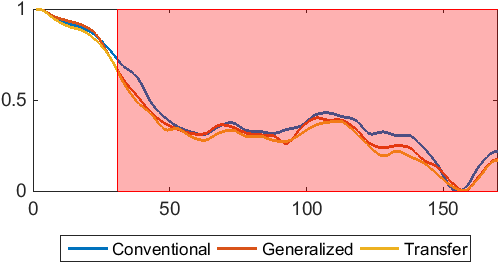}\\
		{\footnotesize Video \# 02} & {\footnotesize Video \# 04}\\
		\includegraphics[scale=0.6]{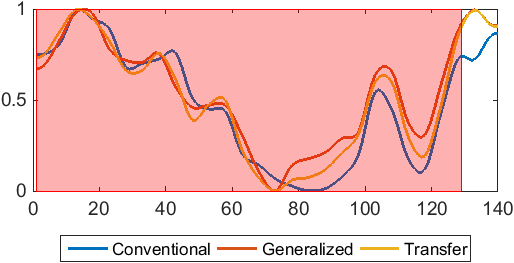}
		&\includegraphics[scale=0.6]{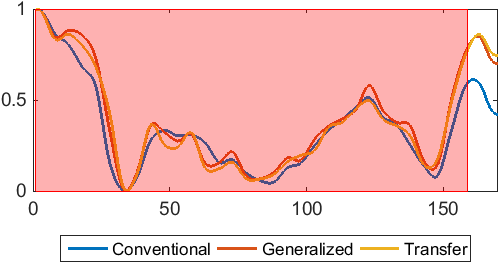}\\
		{\footnotesize Video \# 05} & {\footnotesize Video \# 06}\\
		\includegraphics[scale=0.6]{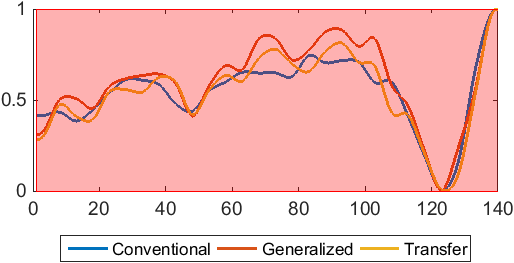}
		&\includegraphics[scale=0.6]{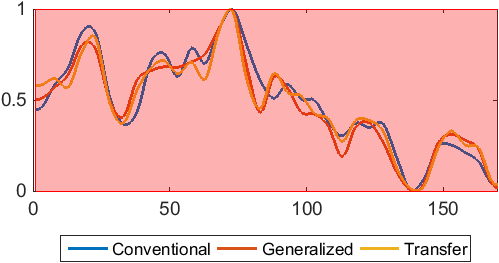}\\
		{\footnotesize Video \# 10} & {\footnotesize Video \# 11}\\
	\end{tabular}
	\caption{In video \# 10 and 11, entire sequence is defined as an anomalous event. Some frames, however, shows regular motions as discussed in the previous section.}
\end{figure}

\begin{center}
	\hyperlink{page.11}{Go to Table of Contents}
\end{center}

\clearpage

\subsection{Subway Enter}
\label{sec:anomaly_enter}
The anomalous events are well captured by the regularity score as the definition of anomalous events in this dataset is similar to our definition of regularity - no presence of any abruption motions.

\begin{figure}[h]
	\centering
	\includegraphics[scale=0.35]{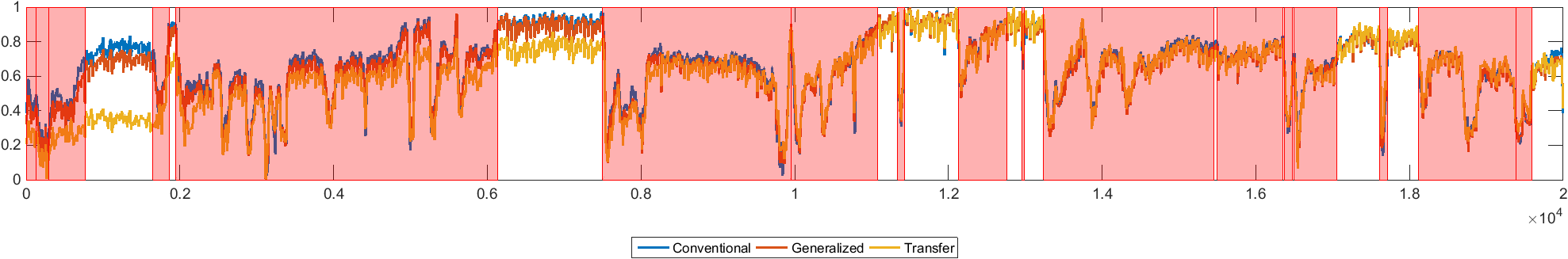}\\
	{\footnotesize Video \#1, Frame \# 20,000-40,000}\\
	\includegraphics[scale=0.35]{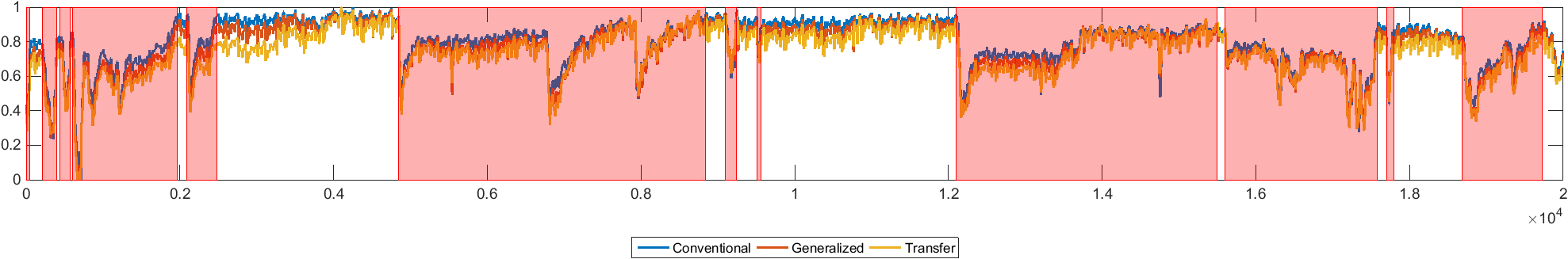}\\
	{\footnotesize Video \#1, Frame \# 40,000-60,000}\\
	\includegraphics[scale=0.35]{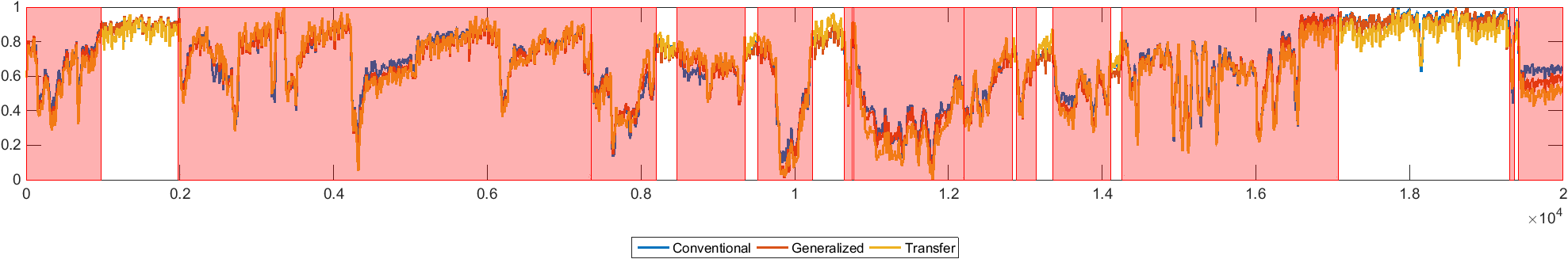}\\
	{\footnotesize Video \#1, Frame \# 60,000-80,000}\\
	\includegraphics[scale=0.35]{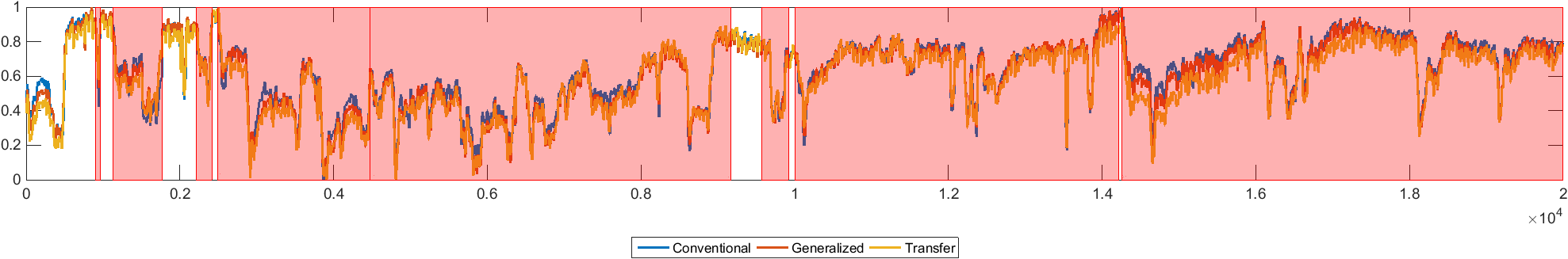}\\
	{\footnotesize Video \#1, Frame \# 80,000-100,000}\\
	\includegraphics[scale=0.35]{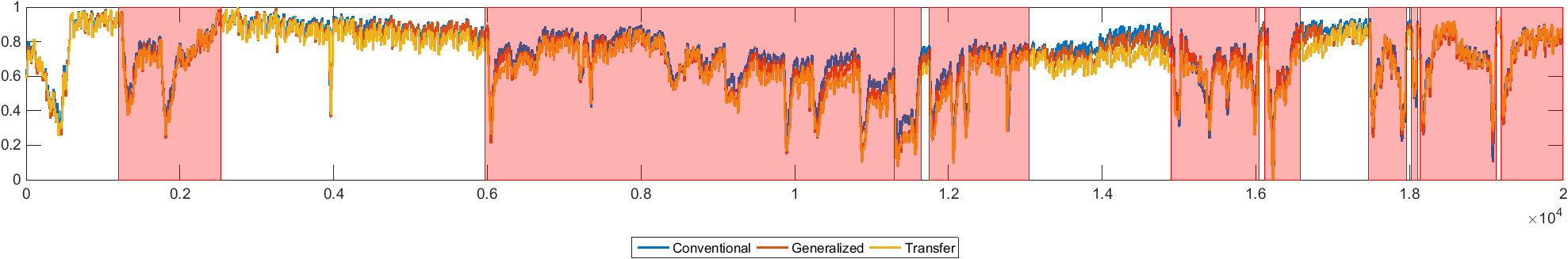}\\
	{\footnotesize Video \#1, Frame \# 100,000-120,000}\\
	\caption{Low score regions are well aligned with the temporal regions of anomalous events.}
\end{figure}

\begin{center}
	\hyperlink{page.11}{Go to Table of Contents}
\end{center}

\clearpage

\subsection{Subway Exit}
\label{sec:anomaly_exit}
Similar to Subway-Enter, the definition of anomalous events in this dataset is similar to our definition of regularity.

\begin{figure}[h]
	\centering
	\includegraphics[scale=0.43]{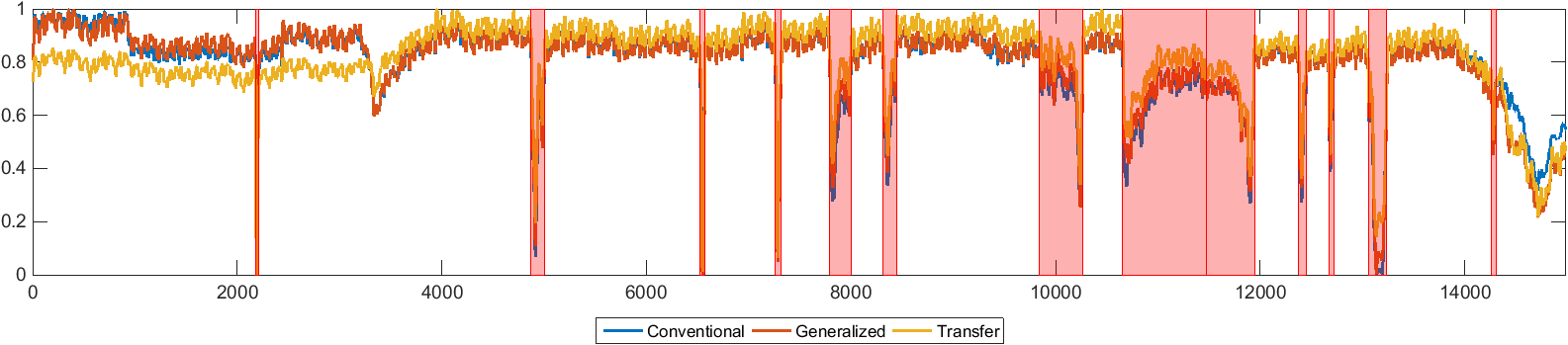}\\
	{\footnotesize Video \#1, Frame \# 7,500-22,500}\\
	\includegraphics[scale=0.43]{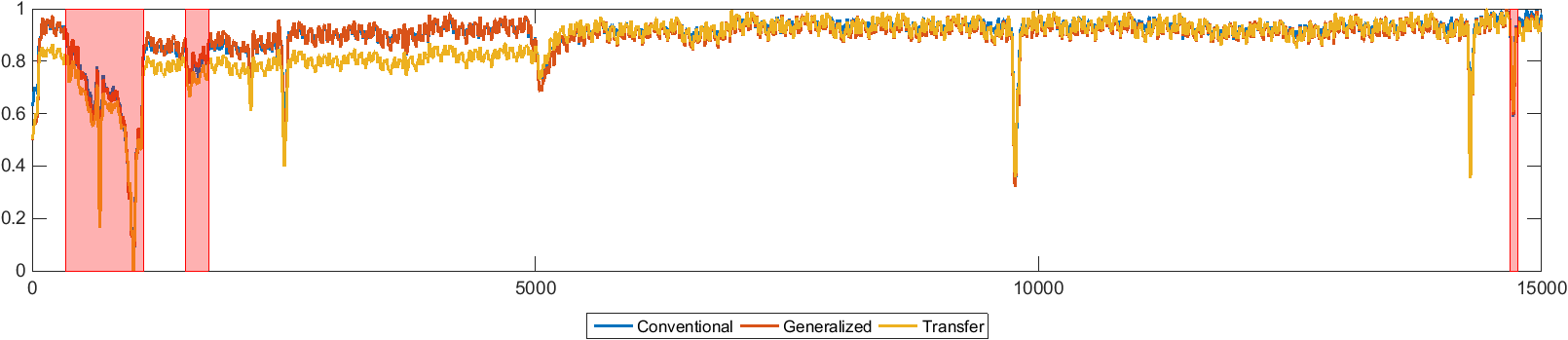}\\
	{\footnotesize Video \#1, Frame \# 22,500-37,500}\\
	\includegraphics[scale=0.43]{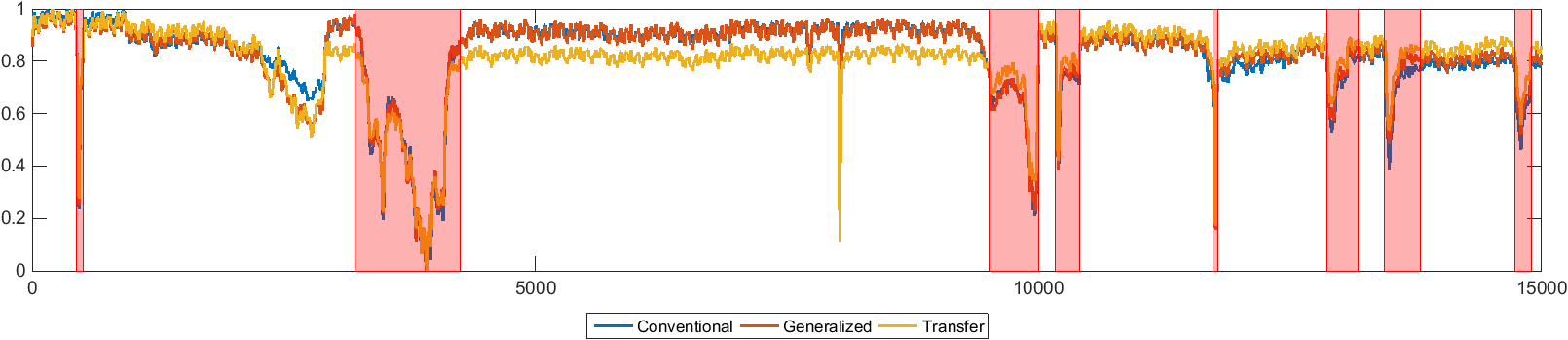}\\
	{\footnotesize Video \#1, Frame \# 37,500-52,500}\\
	\includegraphics[scale=0.43]{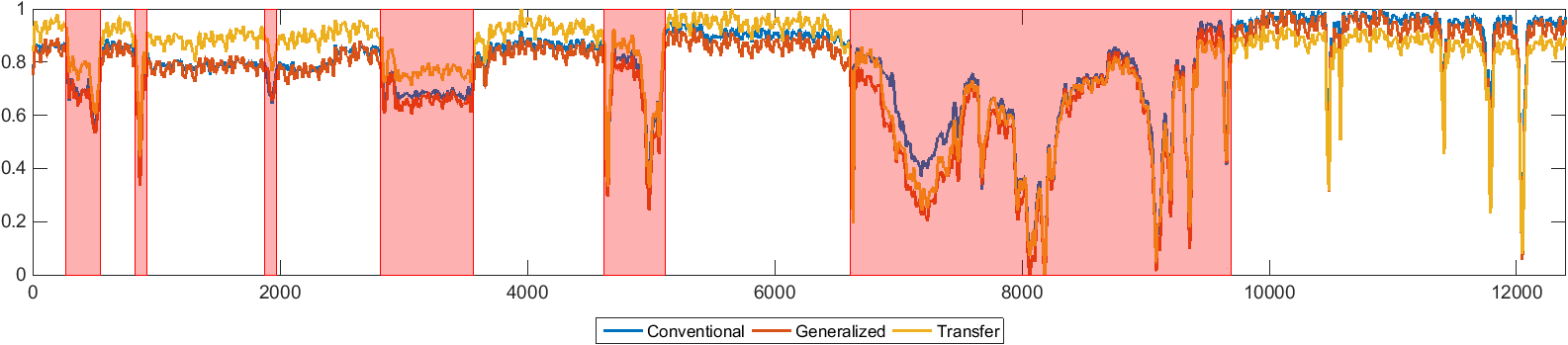}\\
	{\footnotesize Video \#1, Frame \# 52,500-64,000}\\
	\caption{Low score regions are well aligned with the temporal regions of anomalous events.}
\end{figure}

\begin{center}
	\hyperlink{page.11}{Go to Table of Contents}
\end{center}

\clearpage

\newcommand{\dl}{0.17}
\newcommand{\caaal}{0.14}
\newcommand{\cbl}{0.11}
\newcommand{\ccl}{0.08}

\newcommand{\dla}{0.15}
\newcommand{\caaala}{0.12}
\newcommand{\cbla}{0.09}
\newcommand{\ccla}{0.06}

\section{Filter Response Visualization}
\label{sec:filter_res_vis}
We visualize the responses of learned convolutional filters in every layer.
In the early convolutional layers, the filters capture various low level structural patches. 
Various learned filters capture complementary information as different filters shows very different responses on the same patch.
As the layer goes deeper in convolution, the filters capture higher level structure in scale.
The deconvolutional layers try to unpack the encoded (and noiseless) information in a hierarchical way in scale.
Note that, for ease of visualization, we only show two frames of input and output.

\subsection{CUHK Avenue Dataset}
\label{sec:filter_res_vis_avenue}

\begin{figure}[h]
	\centering
	\includegraphics[scale=\dla]{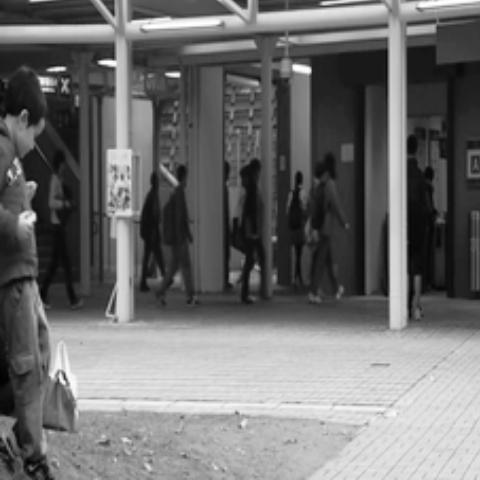}
	\includegraphics[scale=\dla]{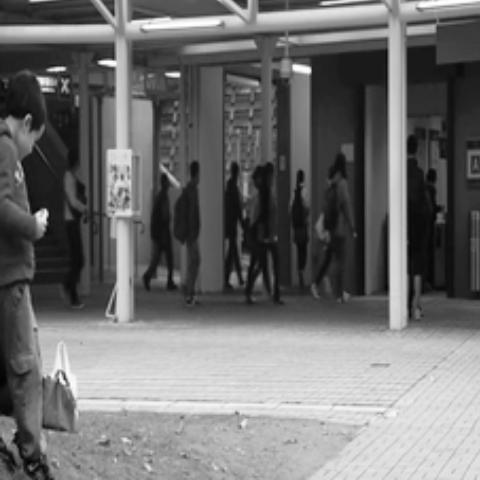}\\
	{\footnotesize Data layer}\\
	\includegraphics[scale=\caaala]{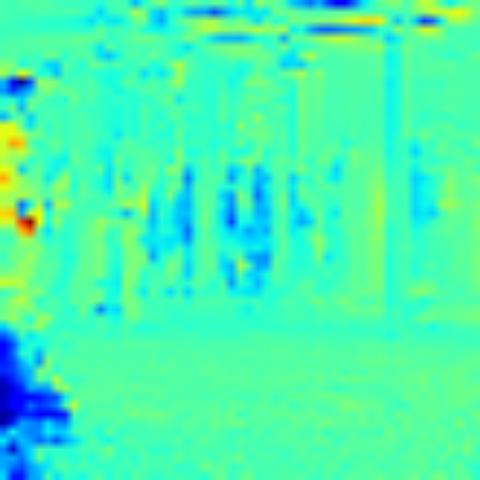}
	\includegraphics[scale=\caaala]{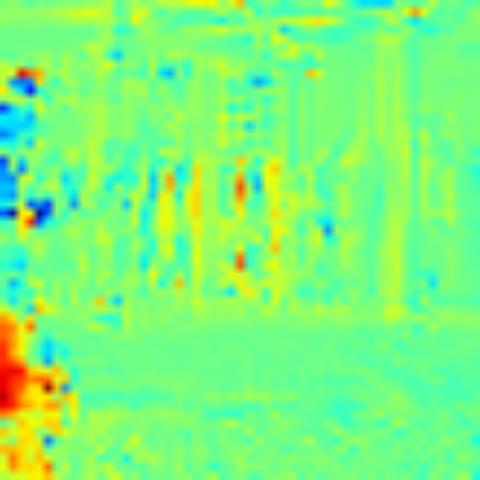}
	\includegraphics[scale=\caaala]{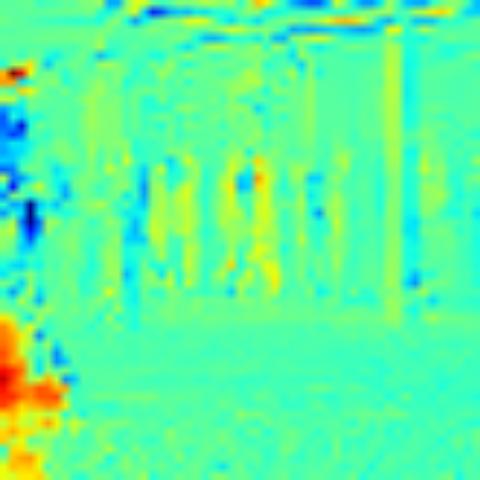}
	\includegraphics[scale=\caaala]{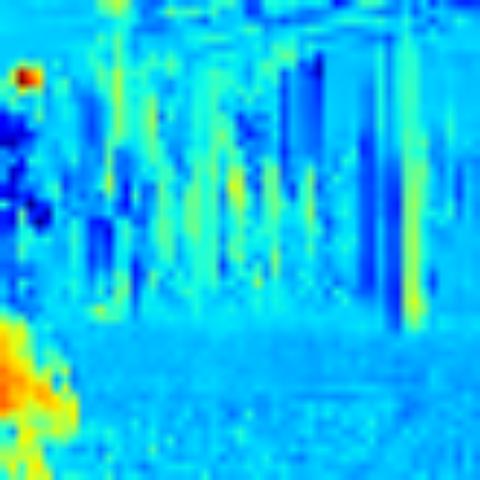}
	\includegraphics[scale=\caaala]{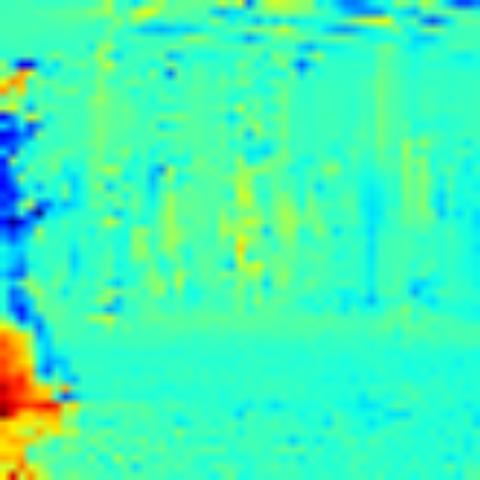}\\
	{\footnotesize First convolutional layer}\\
	\includegraphics[scale=\cbla]{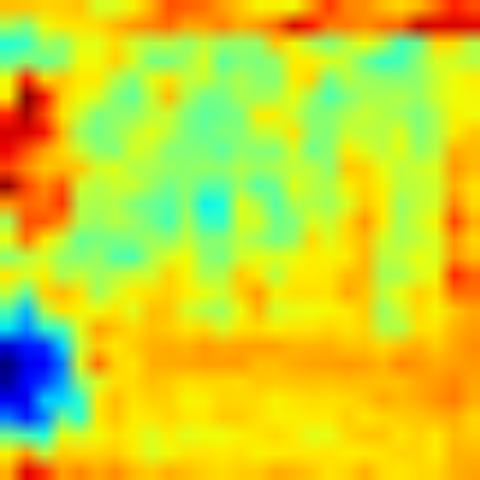}
	\includegraphics[scale=\cbla]{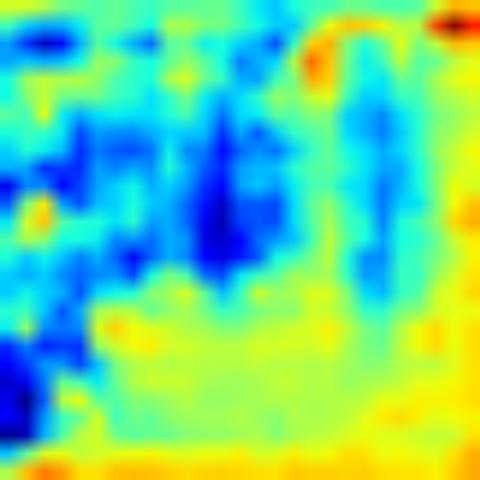}
	\includegraphics[scale=\cbla]{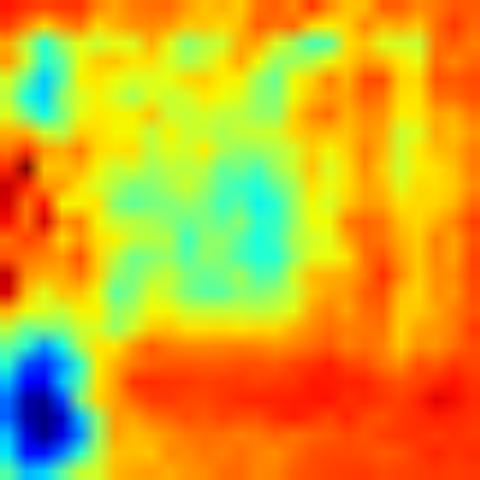}
	\includegraphics[scale=\cbla]{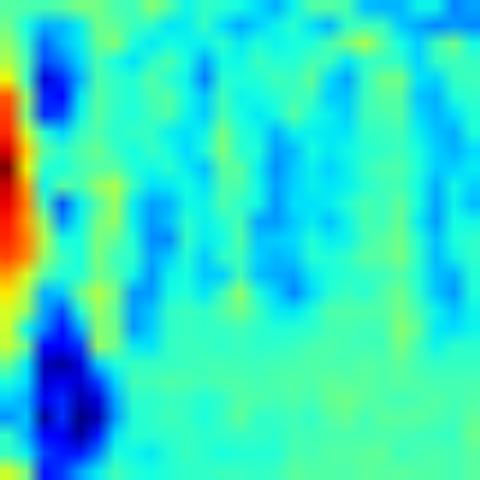}\\
	{\footnotesize Second convolutional layer}\\
	\includegraphics[scale=\ccla]{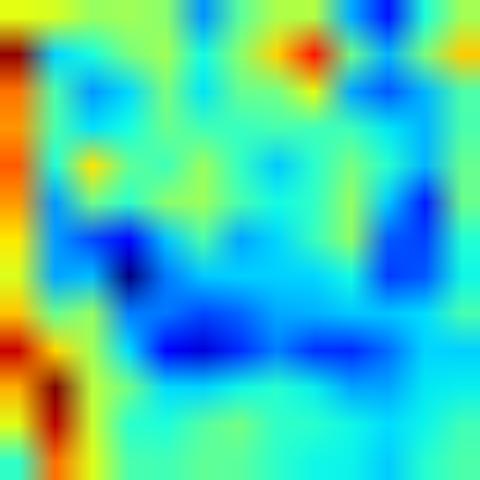}
	\includegraphics[scale=\ccla]{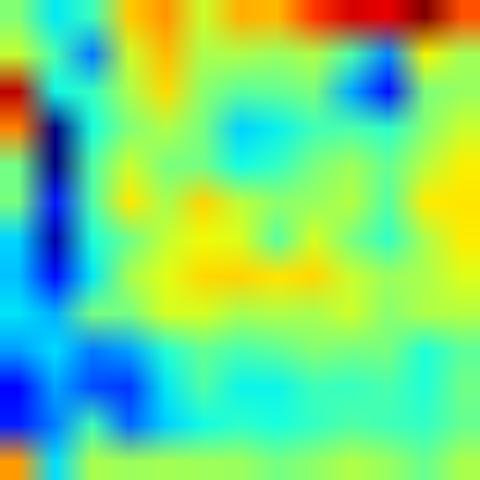}
	\includegraphics[scale=\ccla]{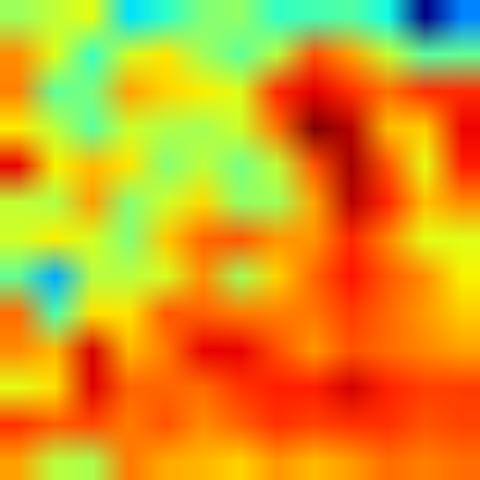}\\
	{\footnotesize Third convolutional layer}\\
	\includegraphics[scale=\cbla]{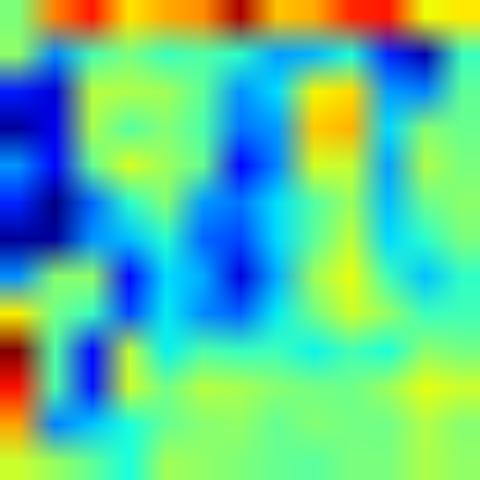}
	\includegraphics[scale=\cbla]{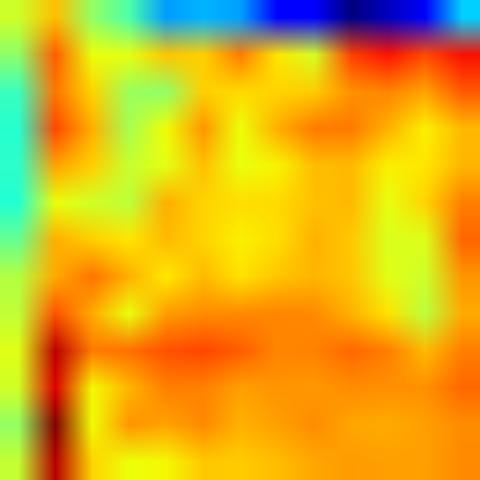}
	\includegraphics[scale=\cbla]{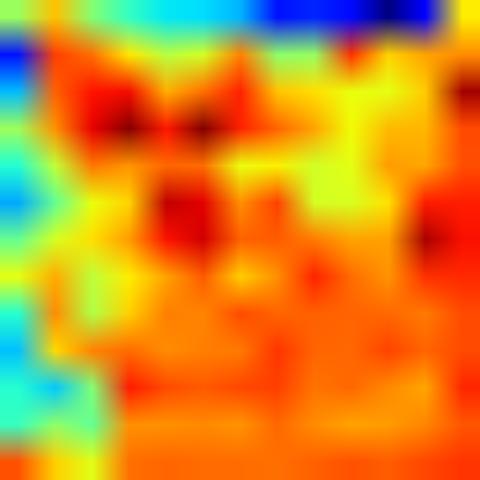}
	\includegraphics[scale=\cbla]{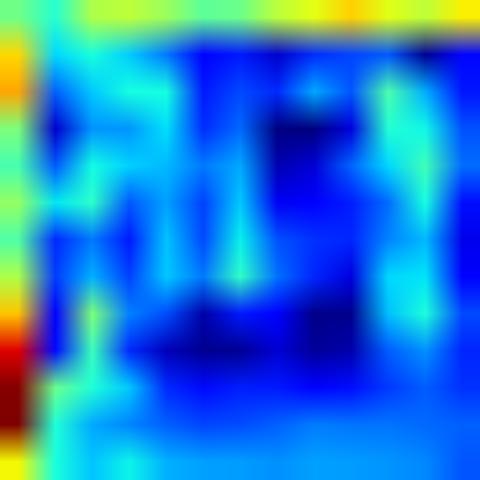}\\
	{\footnotesize First deconvolutional layer}\\
	\includegraphics[scale=\caaala]{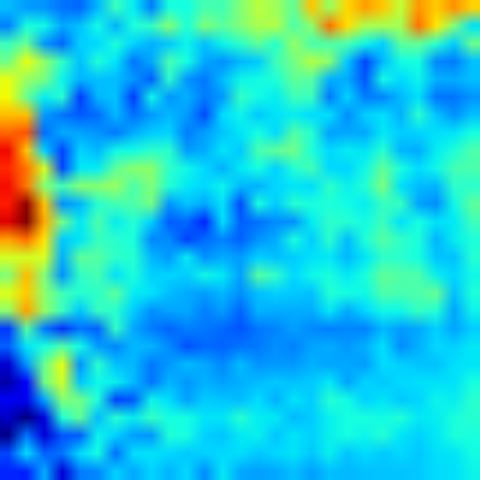}
	\includegraphics[scale=\caaala]{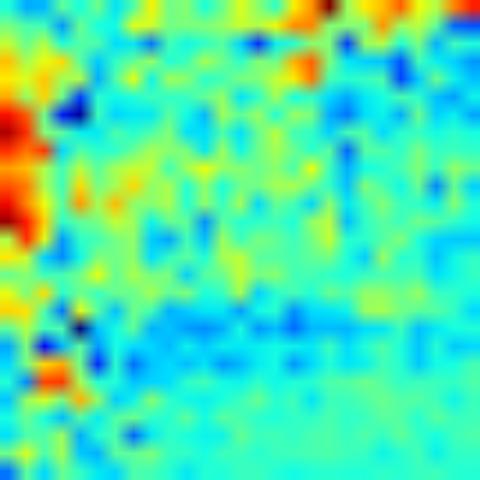}
	\includegraphics[scale=\caaala]{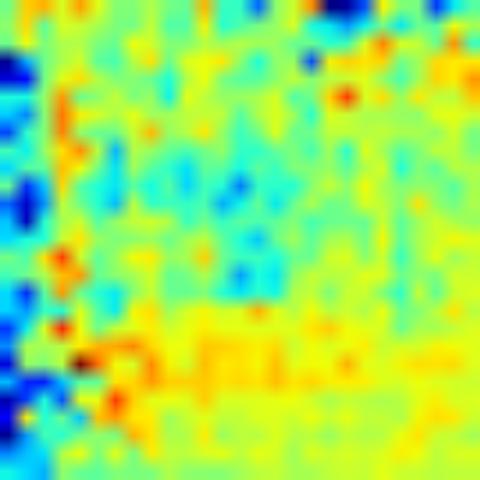}
	\includegraphics[scale=\caaala]{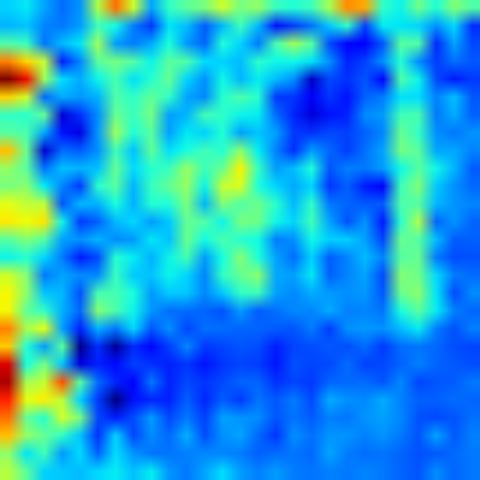}
	\includegraphics[scale=\caaala]{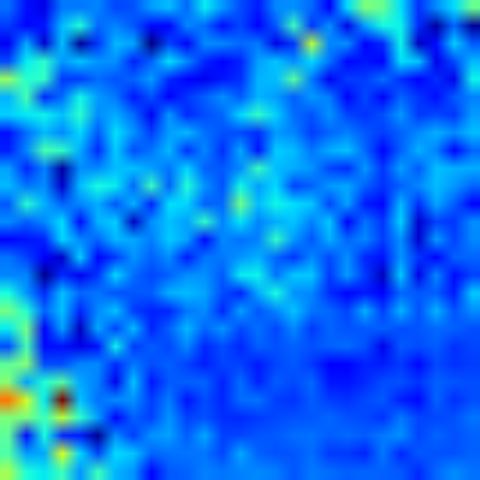}\\
	{\footnotesize Second deconvolutional layer}\\
	\includegraphics[scale=\dla]{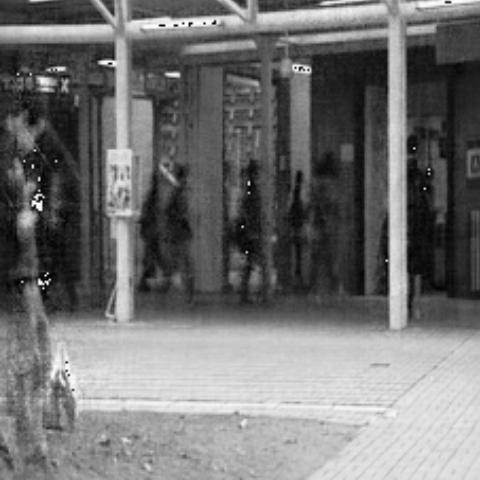}
	\includegraphics[scale=\dla]{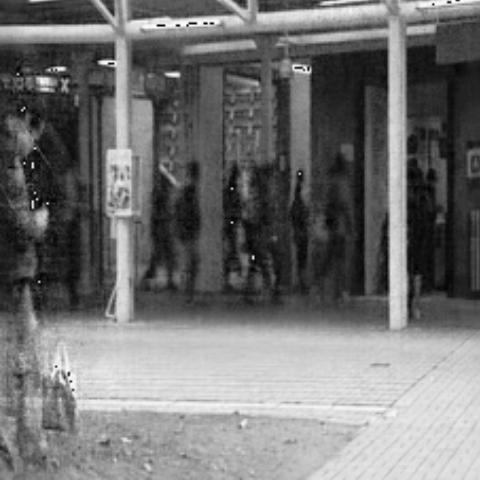}\\
	{\footnotesize Third deconvolutional layer}\\
	\caption{Responses of learned filters, evaluated on a video in CHUK Avenue dataset.}
\end{figure}
\vspace{-5mm}
\begin{center}
	\hyperlink{page.11}{Go to Table of Contents}
\end{center}

\clearpage

\subsection{UCSD Ped1}
\label{sec:filter_res_vis_ped1}

\begin{figure}[h]
	\centering
	\includegraphics[scale=\dl]{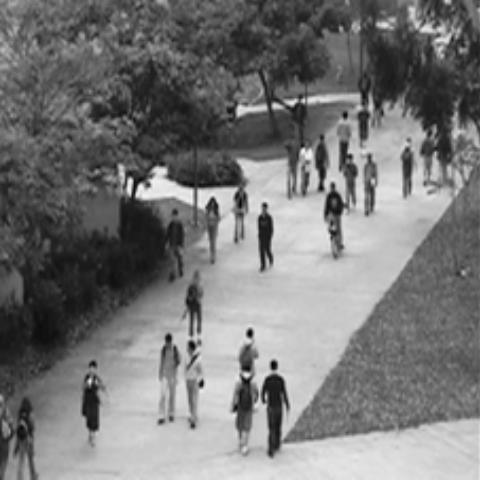}
	\includegraphics[scale=\dl]{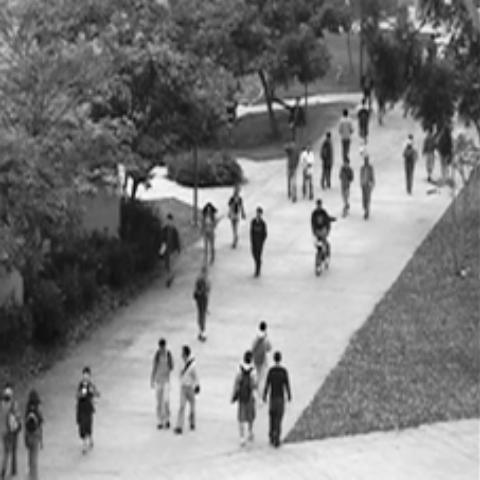}\\
	{\footnotesize Data layer}\\
	\includegraphics[scale=\caaal]{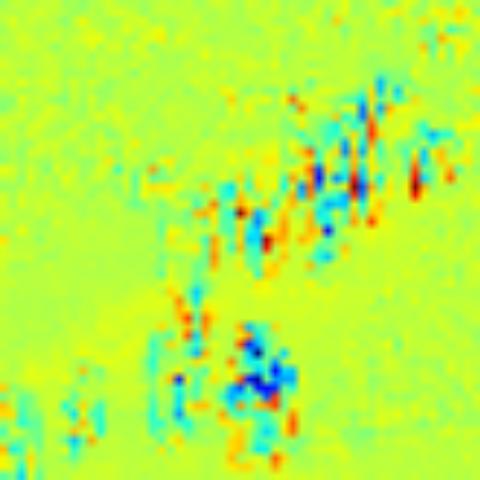}
	\includegraphics[scale=\caaal]{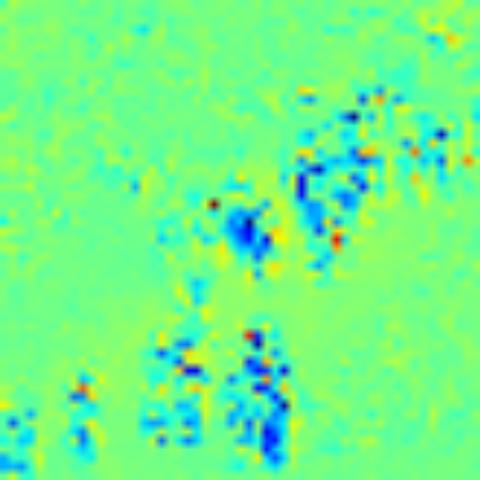}
	\includegraphics[scale=\caaal]{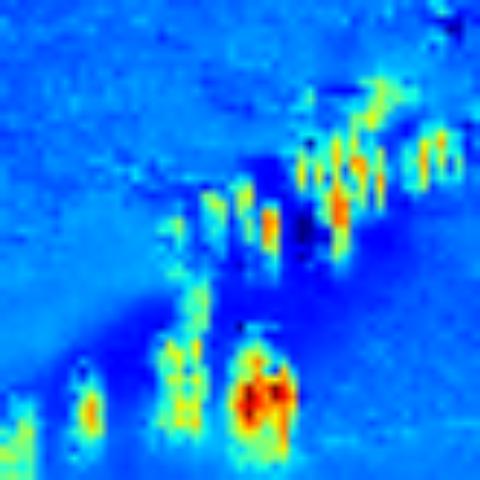}
	\includegraphics[scale=\caaal]{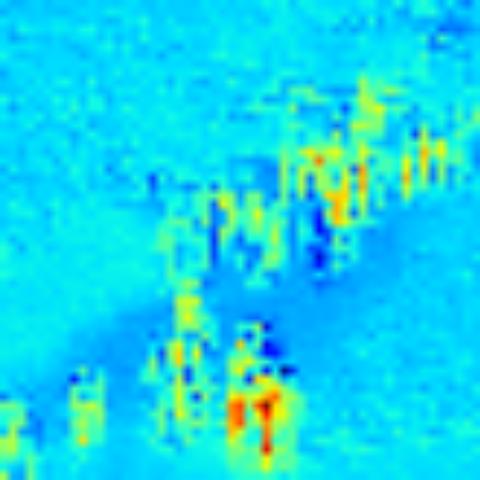}
	\includegraphics[scale=\caaal]{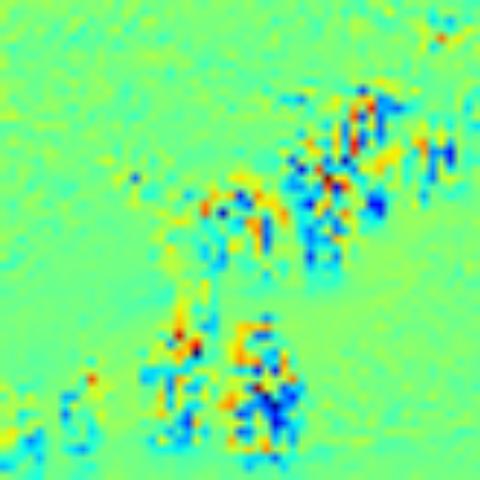}\\
	{\footnotesize First convolutional layer}\\
	\includegraphics[scale=\cbl]{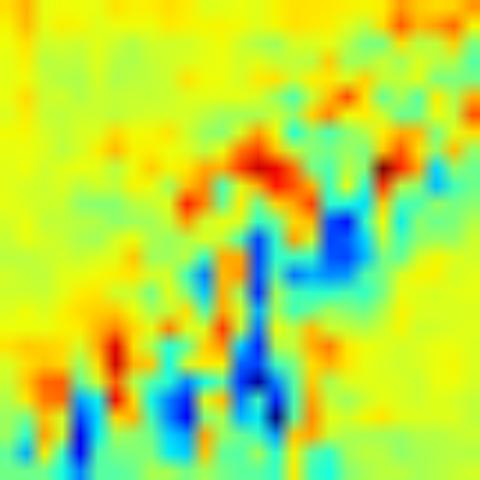}
	\includegraphics[scale=\cbl]{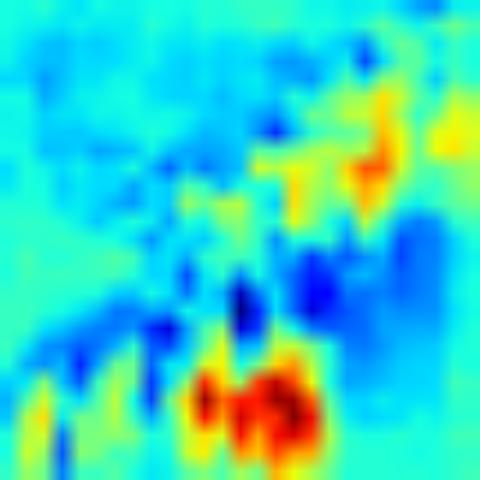}
	\includegraphics[scale=\cbl]{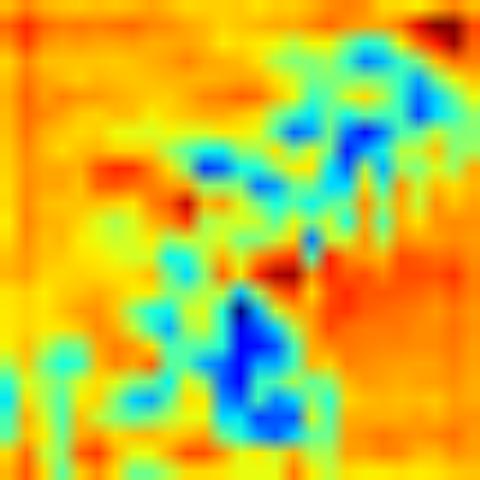}
	\includegraphics[scale=\cbl]{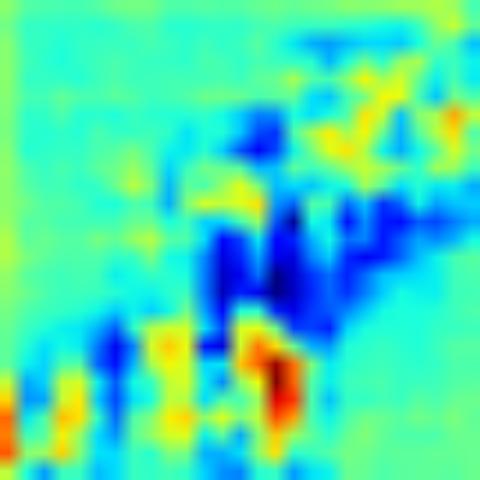}\\
	{\footnotesize Second convolutional layer}\\
	\includegraphics[scale=\ccl]{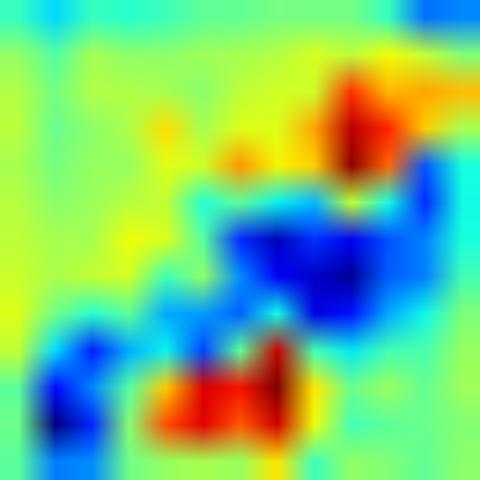}
	\includegraphics[scale=\ccl]{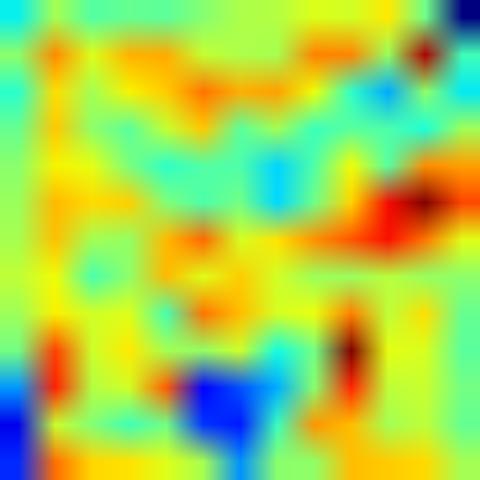}
	\includegraphics[scale=\ccl]{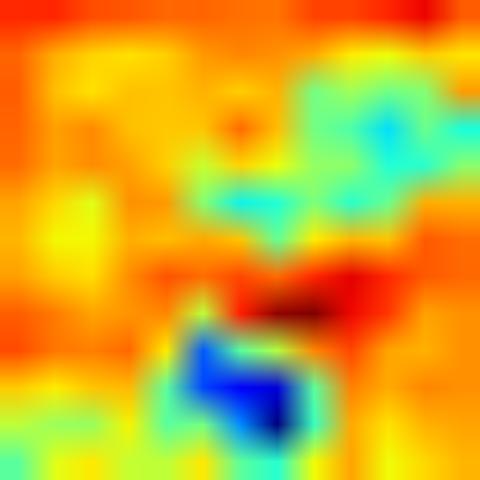}\\
	{\footnotesize Third convolutional layer}\\
	\includegraphics[scale=\cbl]{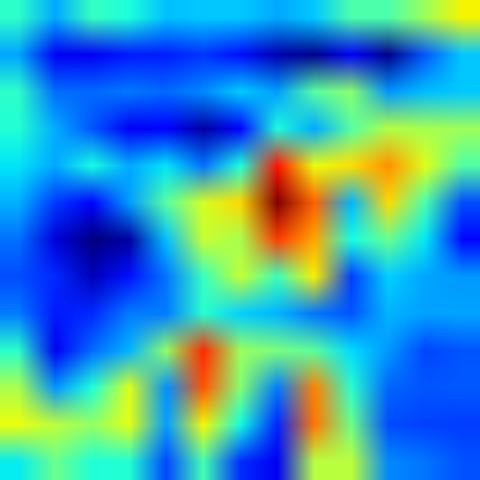}
	\includegraphics[scale=\cbl]{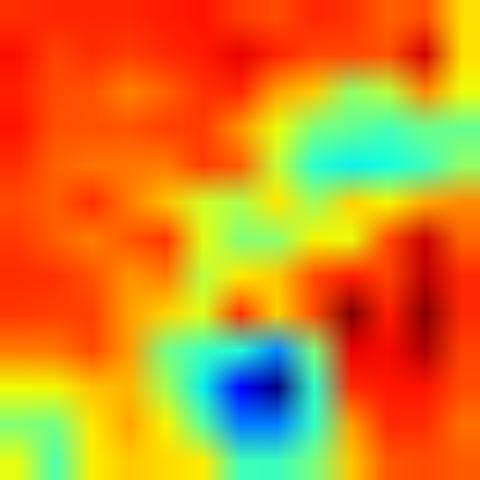}
	\includegraphics[scale=\cbl]{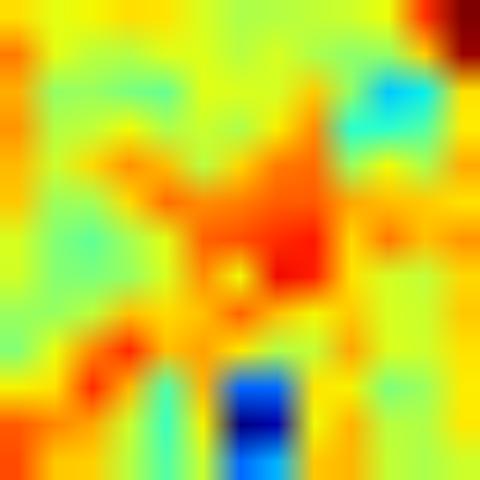}
	\includegraphics[scale=\cbl]{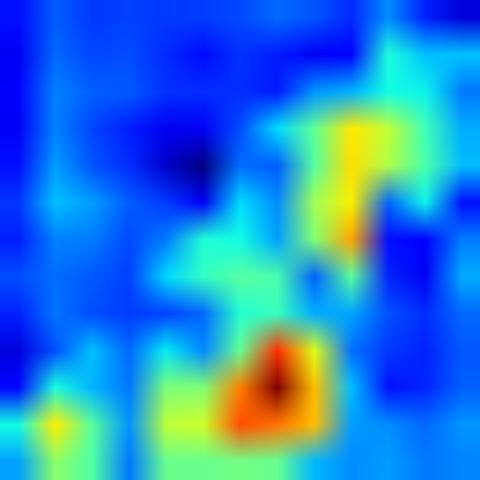}\\
	{\footnotesize First deconvolutional layer}\\
	\includegraphics[scale=\caaal]{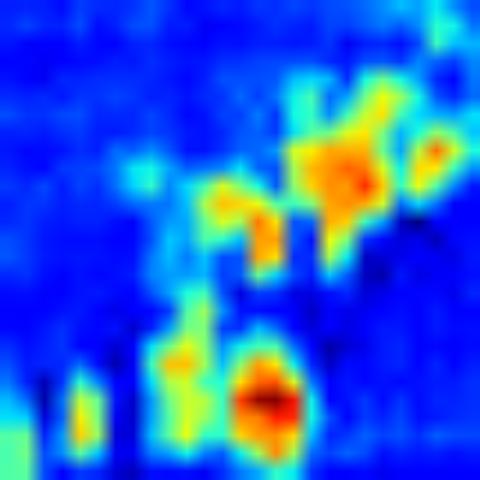}
	\includegraphics[scale=\caaal]{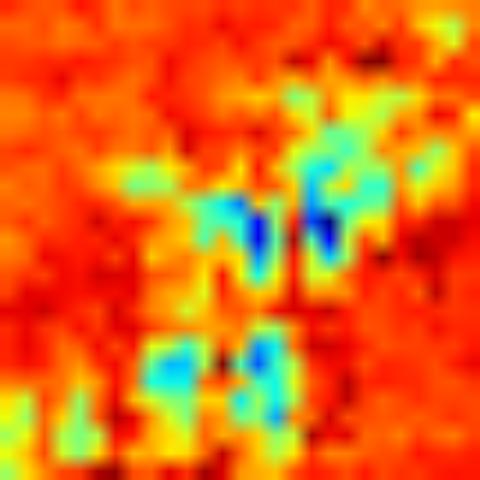}
	\includegraphics[scale=\caaal]{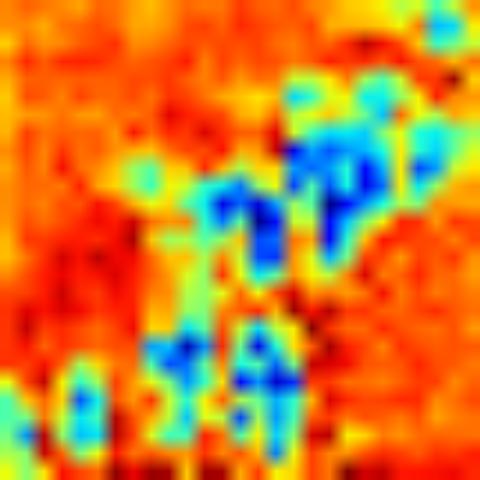}
	\includegraphics[scale=\caaal]{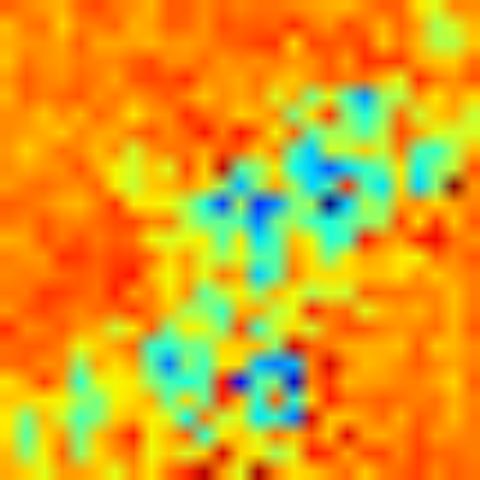}
	\includegraphics[scale=\caaal]{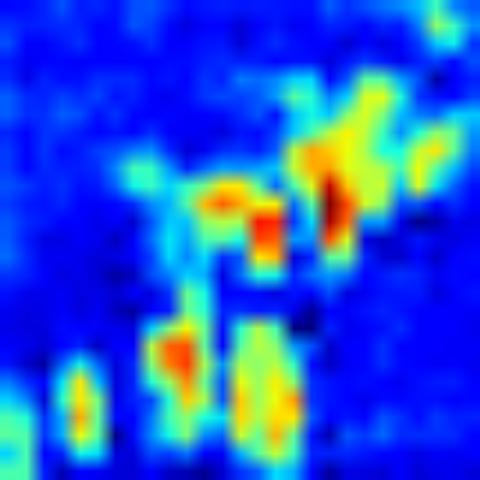}\\
	{\footnotesize Second deconvolutional layer}\\
	\includegraphics[scale=\dl]{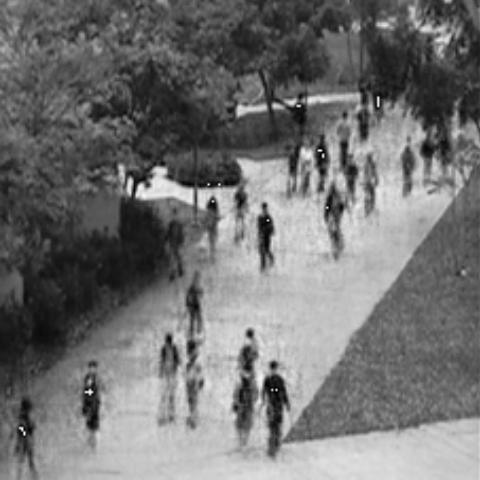}
	\includegraphics[scale=\dl]{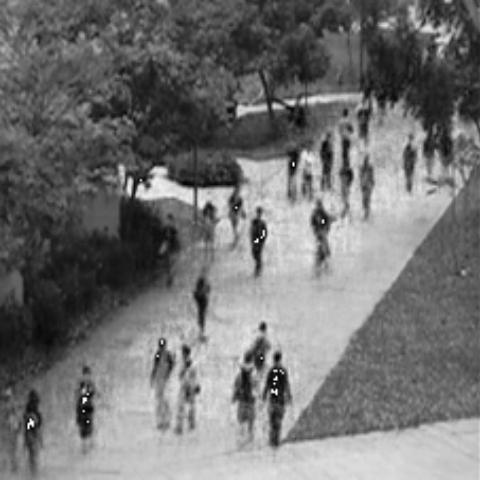}\\
	{\footnotesize Third deconvolutional layer}\\
	\caption{Responses of learned filters, evaluated on a video in UCSD-Ped1 dataset. Note that the filters captures various aspects of regularity as shown by various colored responses on the same region.}
\end{figure}

\begin{center}
	\hyperlink{page.11}{Go to Table of Contents}
\end{center}

\clearpage

\subsection{UCSD Ped2}
\label{sec:filter_res_vis_ped2}

\begin{figure}[h]
	\centering
	\includegraphics[scale=\dl]{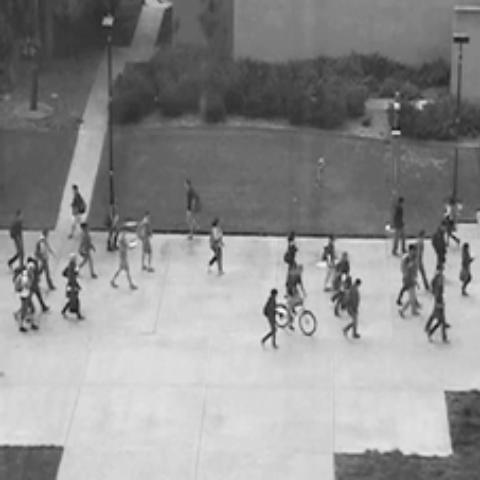}
	\includegraphics[scale=\dl]{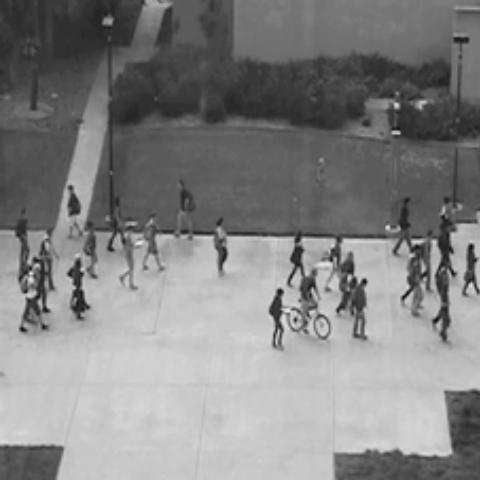}\\
	{\footnotesize Data layer}\\
	\includegraphics[scale=\caaal]{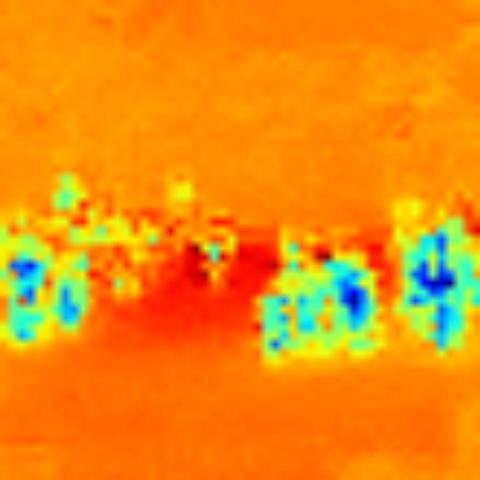}
	\includegraphics[scale=\caaal]{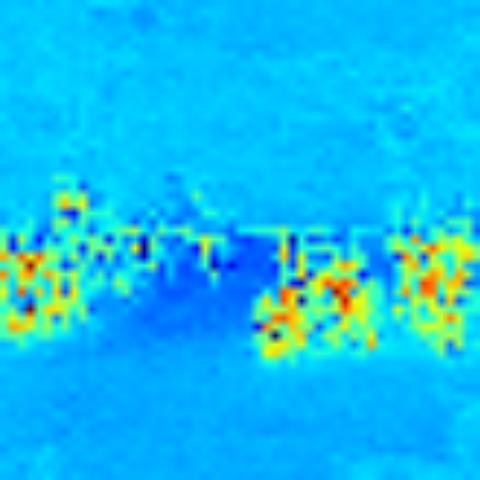}
	\includegraphics[scale=\caaal]{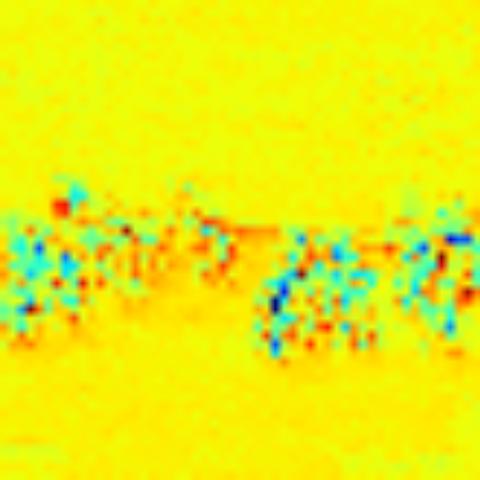}
	\includegraphics[scale=\caaal]{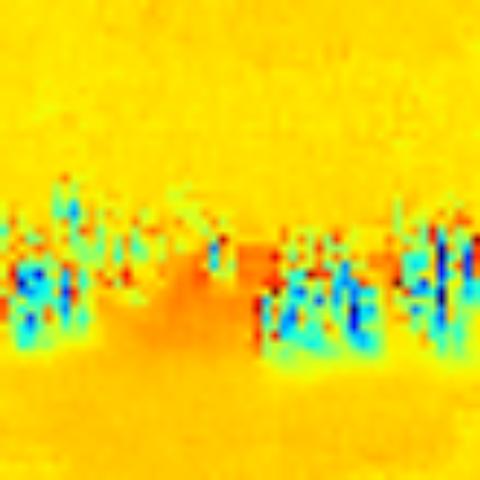}
	\includegraphics[scale=\caaal]{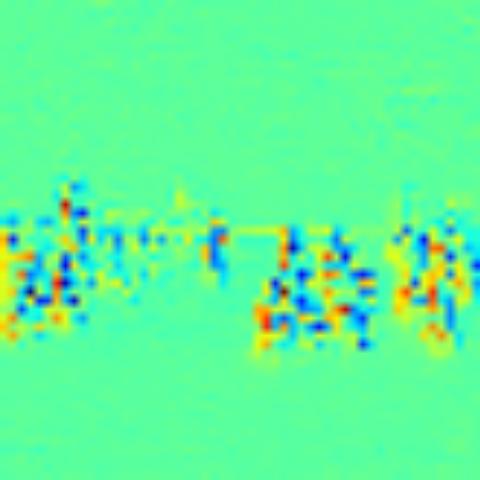}\\
	{\footnotesize First convolutional layer}\\
	\includegraphics[scale=\cbl]{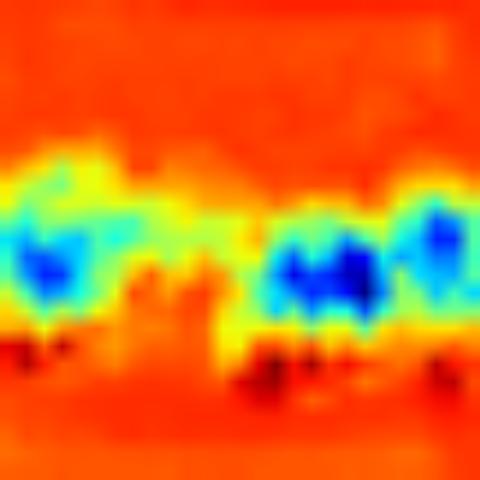}
	\includegraphics[scale=\cbl]{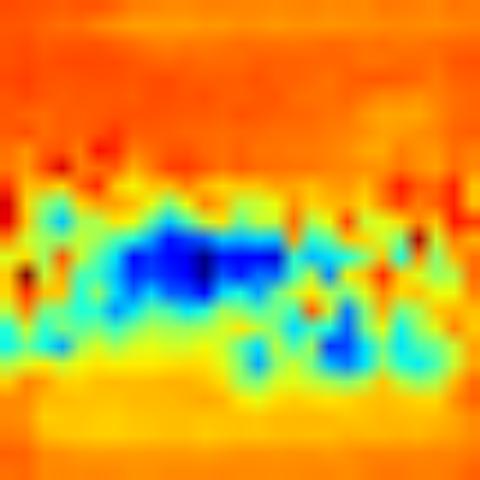}
	\includegraphics[scale=\cbl]{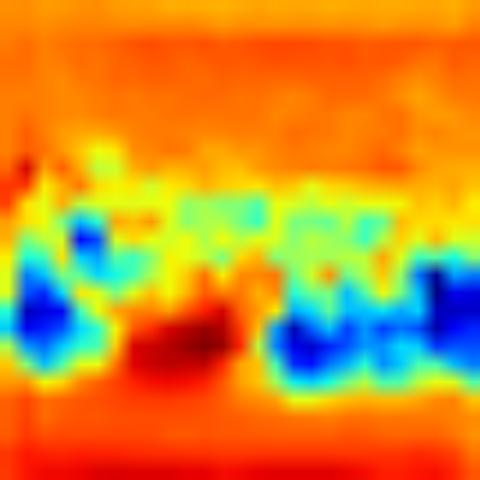}
	\includegraphics[scale=\cbl]{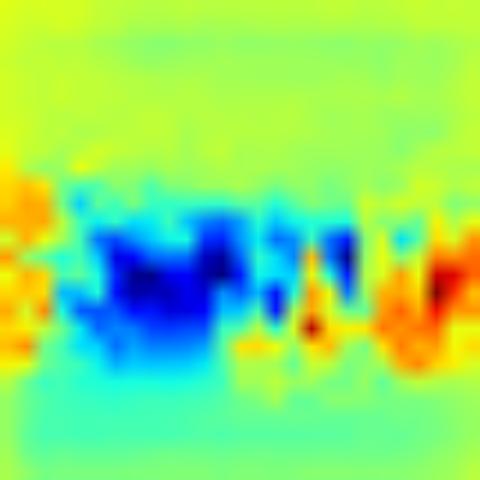}\\
	{\footnotesize Second convolutional layer}\\
	\includegraphics[scale=\ccl]{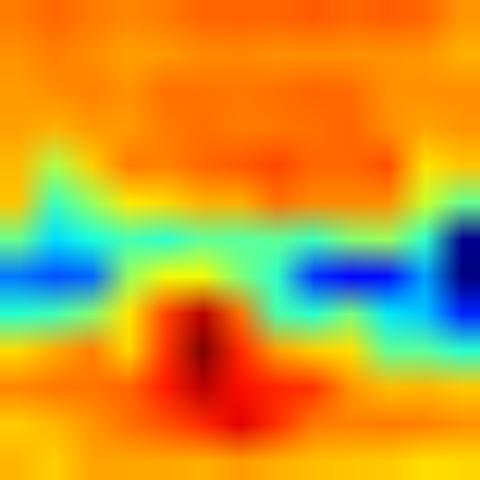}
	\includegraphics[scale=\ccl]{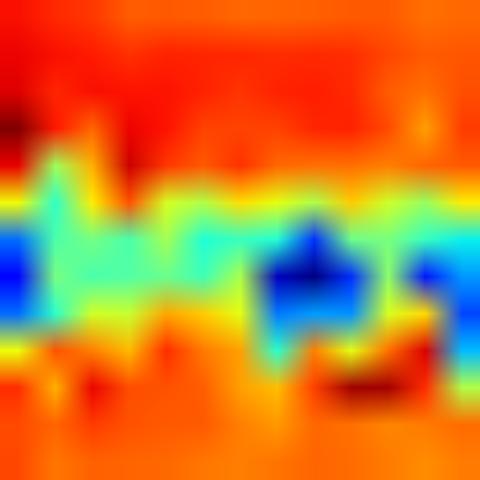}
	\includegraphics[scale=\ccl]{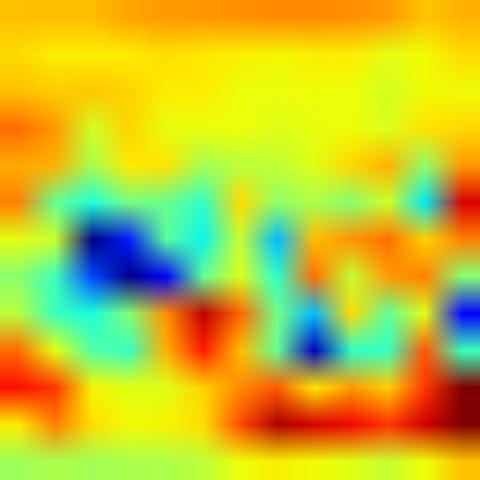}\\
	{\footnotesize Third convolutional layer}\\
	\includegraphics[scale=\cbl]{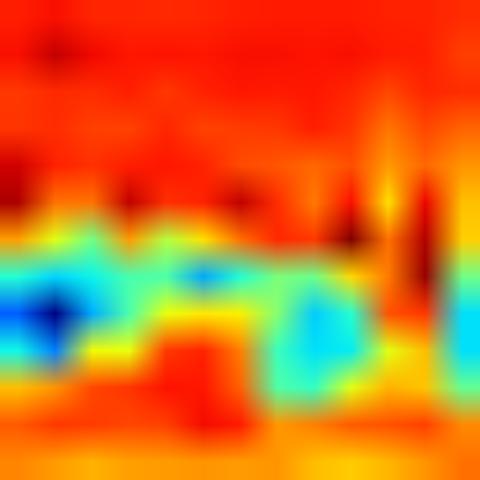}
	\includegraphics[scale=\cbl]{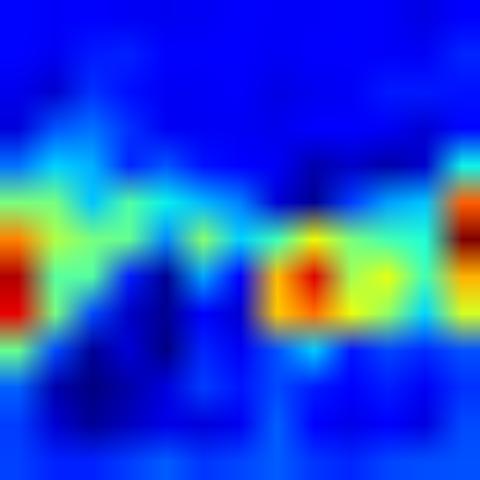}
	\includegraphics[scale=\cbl]{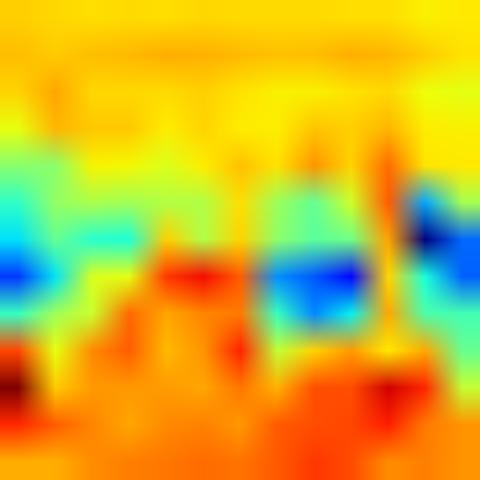}
	\includegraphics[scale=\cbl]{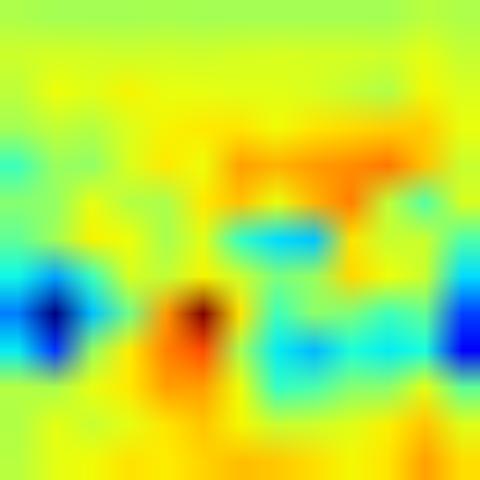}\\
	{\footnotesize First deconvolutional layer}\\
	\includegraphics[scale=\caaal]{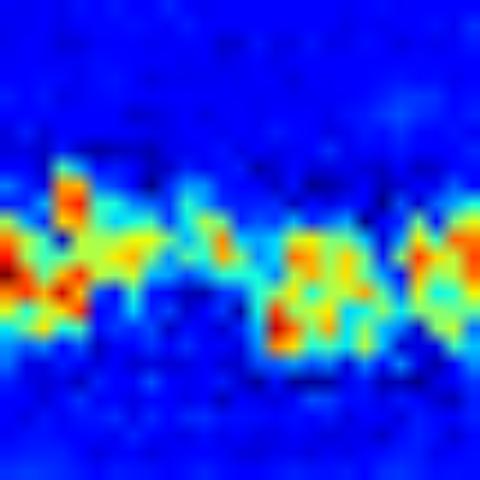}
	\includegraphics[scale=\caaal]{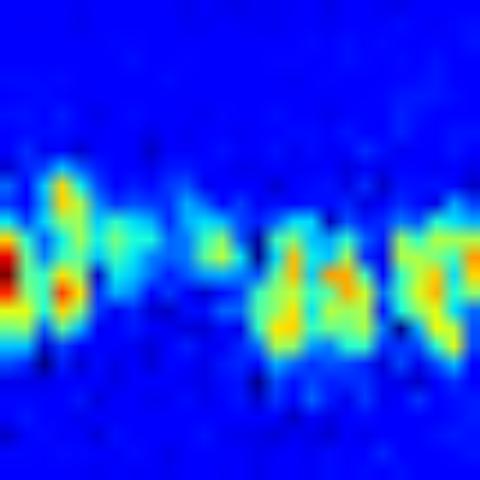}
	\includegraphics[scale=\caaal]{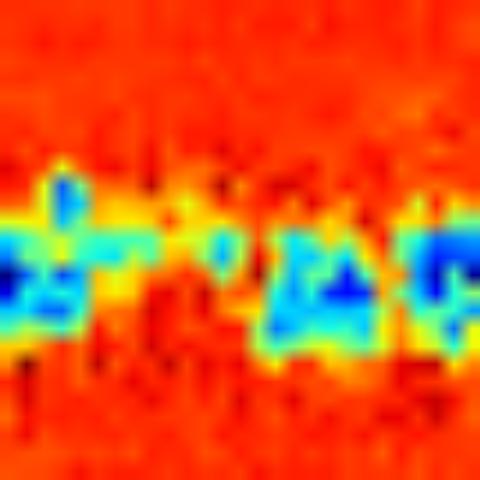}
	\includegraphics[scale=\caaal]{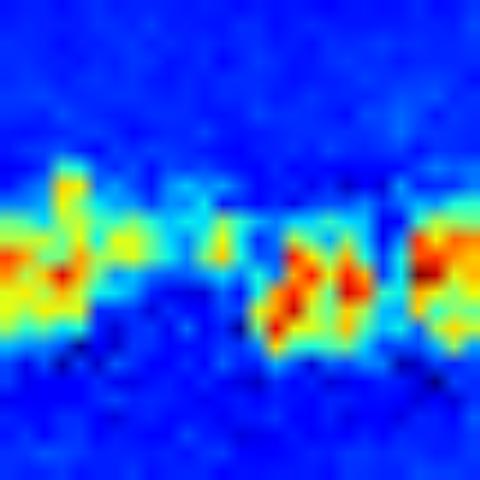}
	\includegraphics[scale=\caaal]{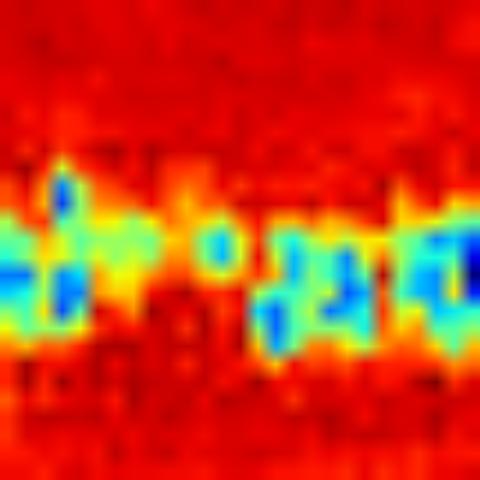}\\
	{\footnotesize Second deconvolutional layer}\\
	\includegraphics[scale=\dl]{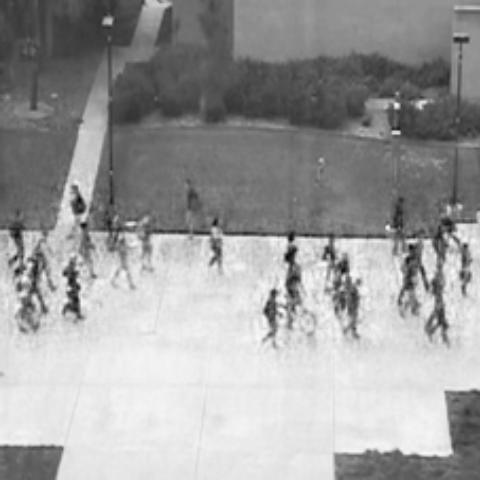}
	\includegraphics[scale=\dl]{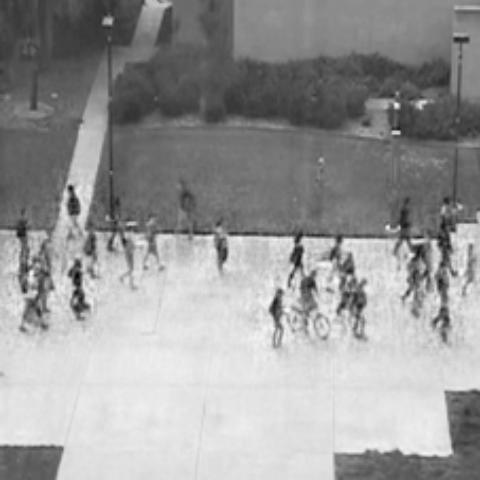}\\
	{\footnotesize Third deconvolutional layer}\\
	\caption{Responses of learned filters, evaluated on a video in UCSD-Ped2 dataset. Note that the filters captures various aspects of regularity as shown by various colored responses on the same region. Noticeably, the background of first convolutional layer outputs are in various colors.}
\end{figure}

\begin{center}
	\hyperlink{page.11}{Go to Table of Contents}
\end{center}

\clearpage

\subsection{Subway Enter}
\label{sec:filter_res_vis_enter}

\begin{figure}[h]
	\centering
	\includegraphics[scale=\caaal5]{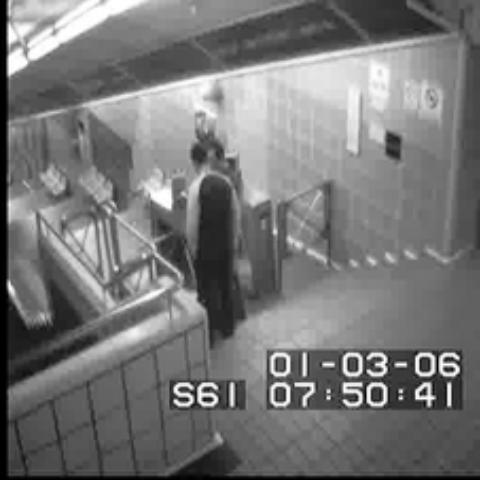}
	\includegraphics[scale=\caaal5]{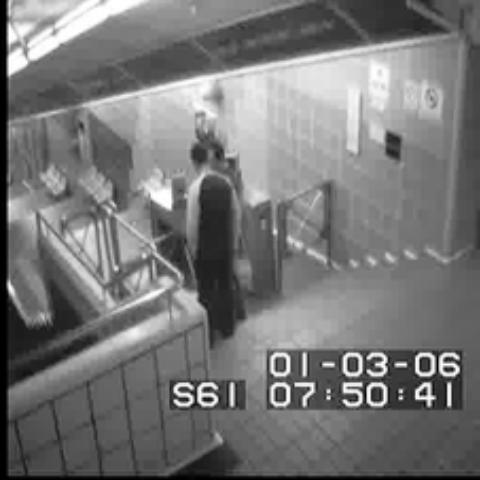}\\
	{\footnotesize Data layer}\\
	\includegraphics[scale=\caaal]{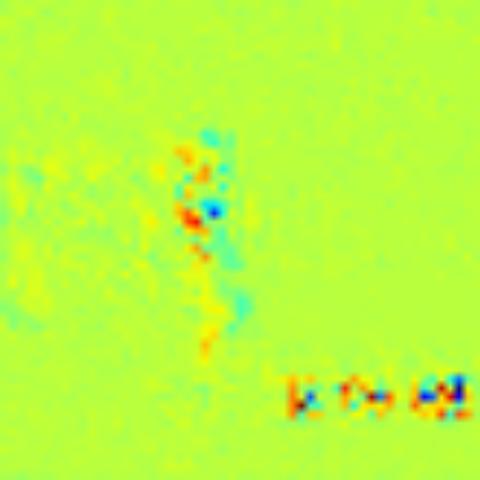}
	\includegraphics[scale=\caaal]{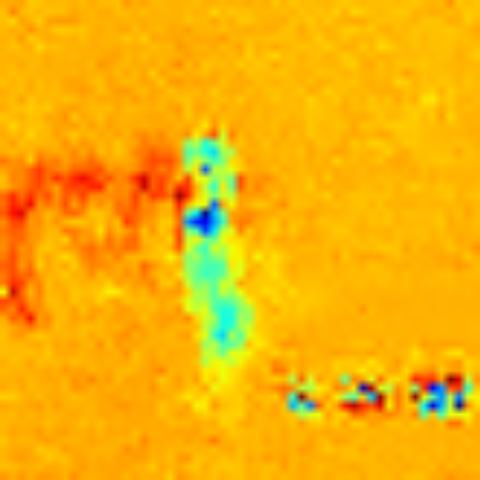}
	\includegraphics[scale=\caaal]{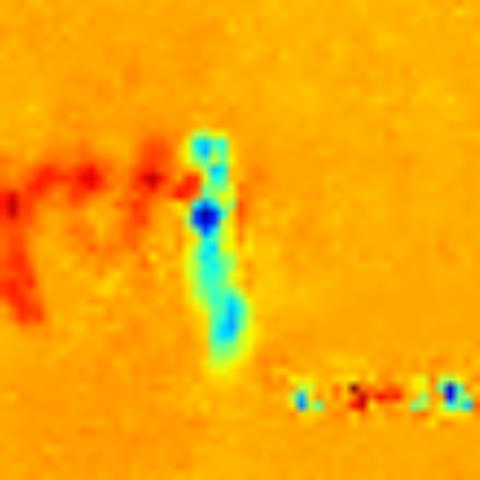}
	\includegraphics[scale=\caaal]{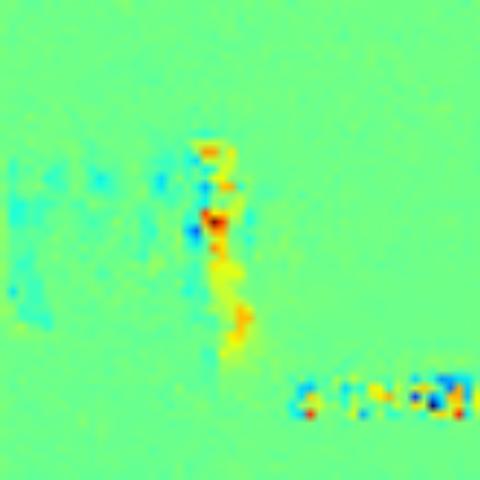}
	\includegraphics[scale=\caaal]{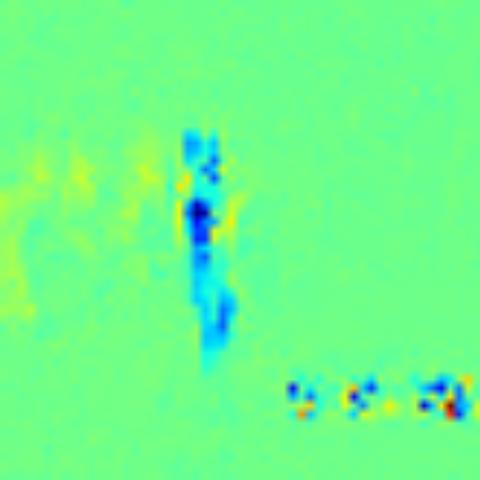}\\
	{\footnotesize First convolutional layer}\\
	\includegraphics[scale=\cbl]{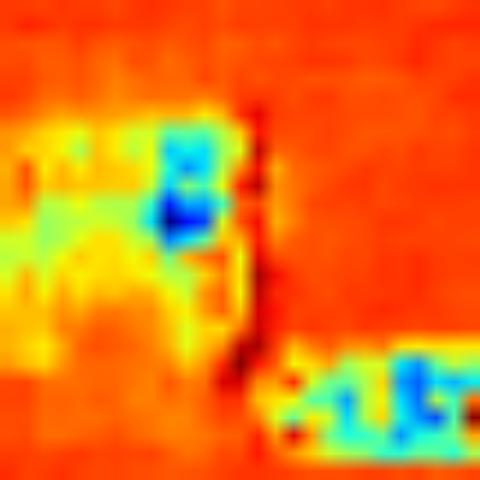}
	\includegraphics[scale=\cbl]{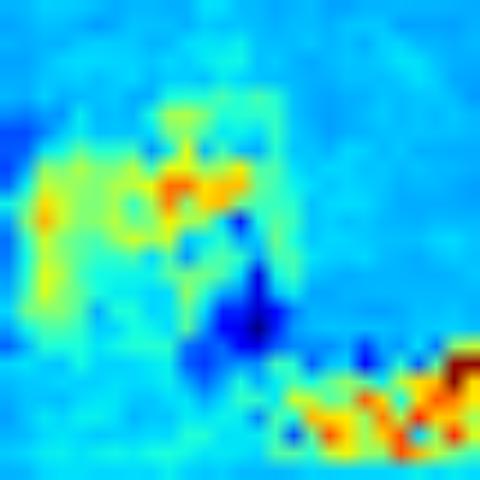}
	\includegraphics[scale=\cbl]{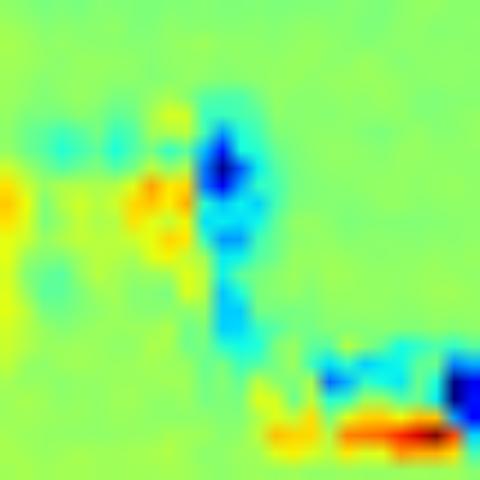}
	\includegraphics[scale=\cbl]{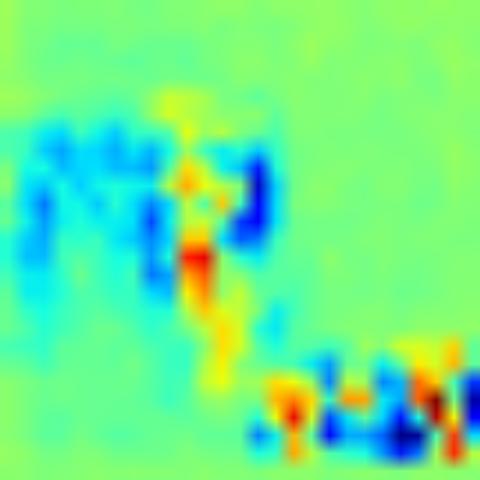}\\
	{\footnotesize Second convolutional layer}\\
	\includegraphics[scale=\ccl]{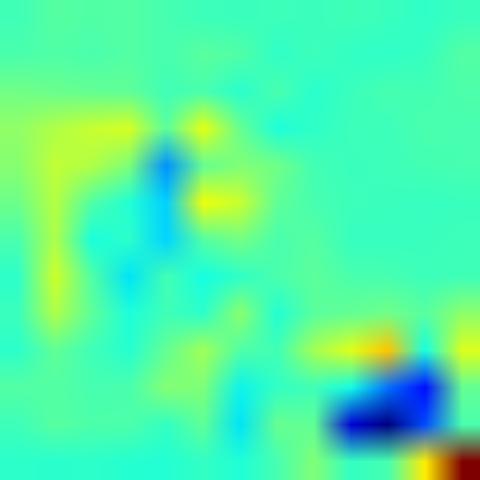}
	\includegraphics[scale=\ccl]{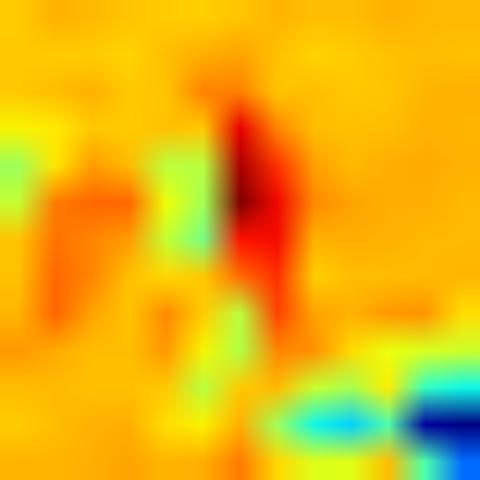}
	\includegraphics[scale=\ccl]{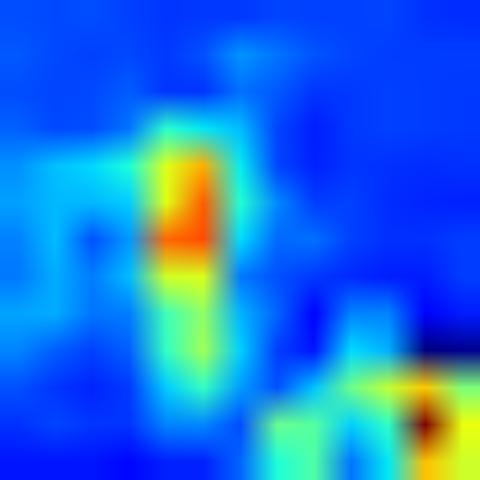}\\
	{\footnotesize Third convolutional layer}\\
	\includegraphics[scale=\cbl]{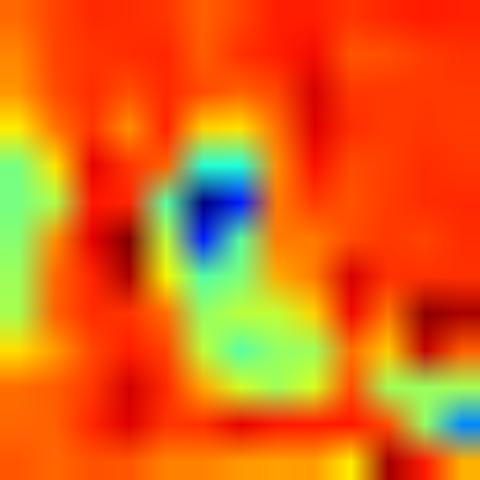}
	\includegraphics[scale=\cbl]{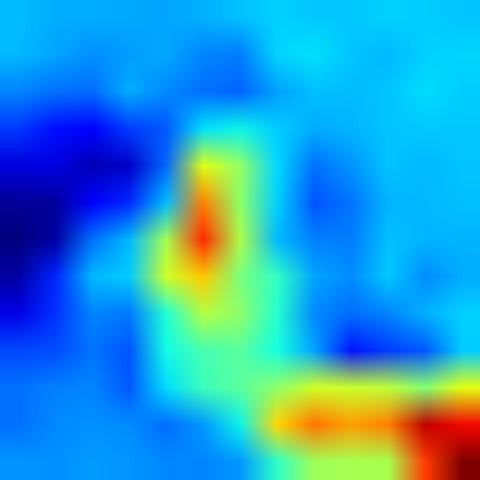}
	\includegraphics[scale=\cbl]{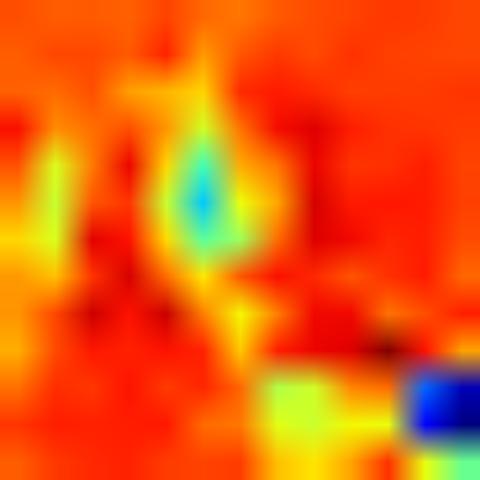}
	\includegraphics[scale=\cbl]{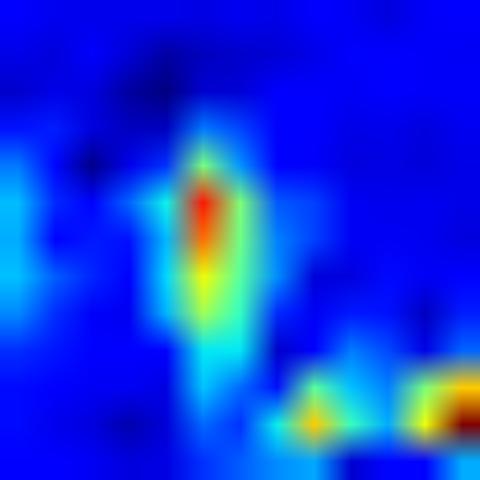}\\
	{\footnotesize First deconvolutional layer}\\
	\includegraphics[scale=\caaal]{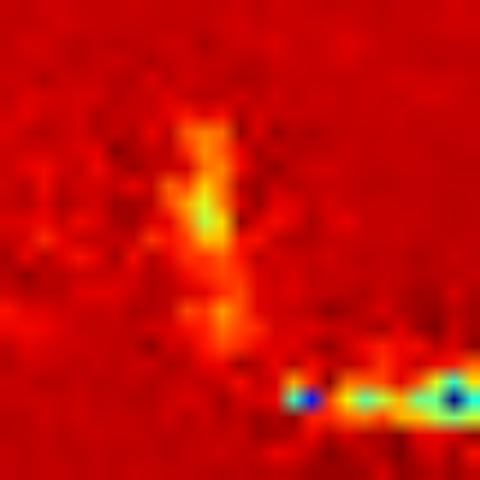}
	\includegraphics[scale=\caaal]{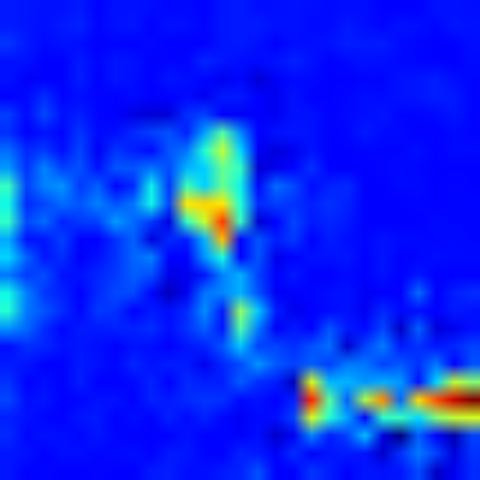}
	\includegraphics[scale=\caaal]{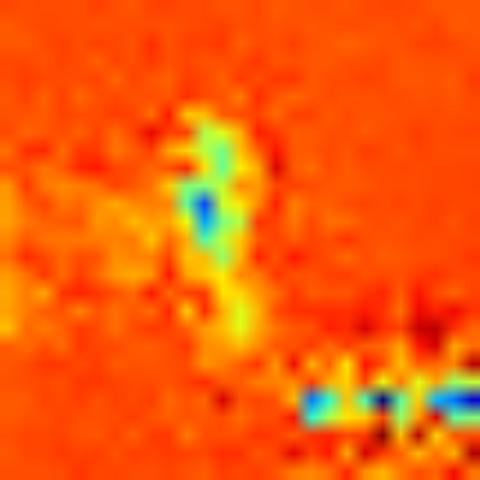}
	\includegraphics[scale=\caaal]{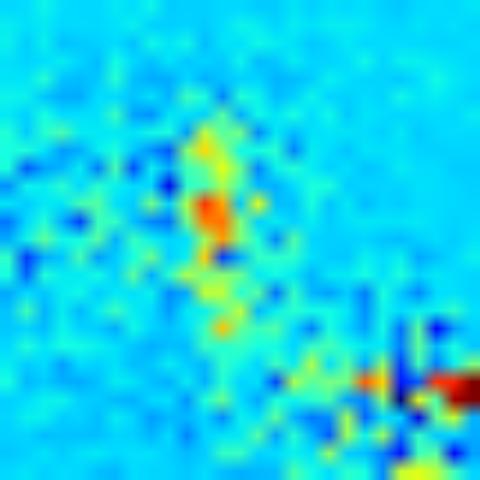}
	\includegraphics[scale=\caaal]{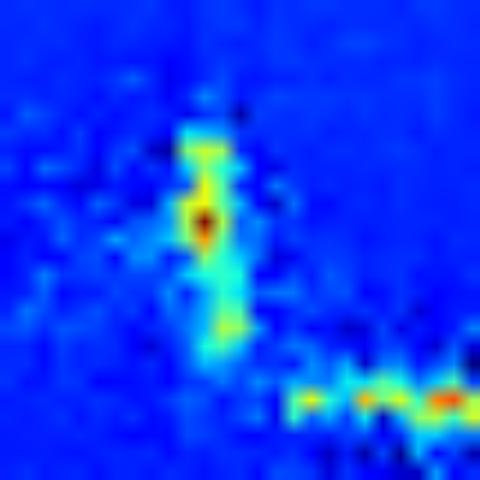}\\
	{\footnotesize Second deconvolutional layer}\\
	\includegraphics[scale=\dl]{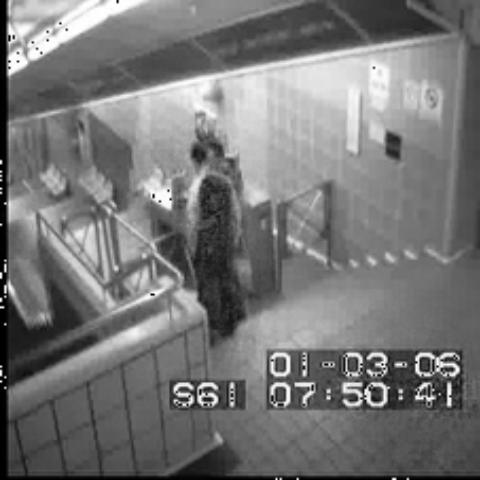}
	\includegraphics[scale=\dl]{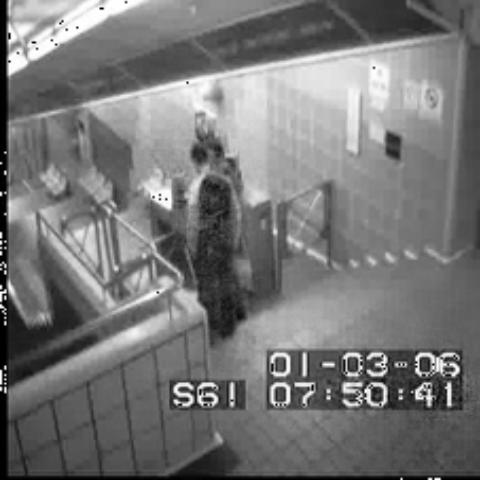}\\
	{\footnotesize Third deconvolutional layer}\\
	\caption{Responses of learned filters, evaluated on a video in Subway Enter dataset.}
\end{figure}

\begin{center}
	\hyperlink{page.11}{Go to Table of Contents}
\end{center}

\clearpage

\subsection{Subway Exit}
\label{sec:filter_res_vis_exit}

\begin{figure}[h]
	\centering
	\includegraphics[scale=\caaal5]{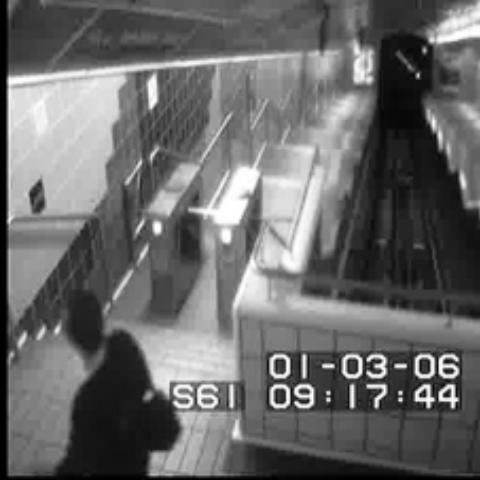}
	\includegraphics[scale=\caaal5]{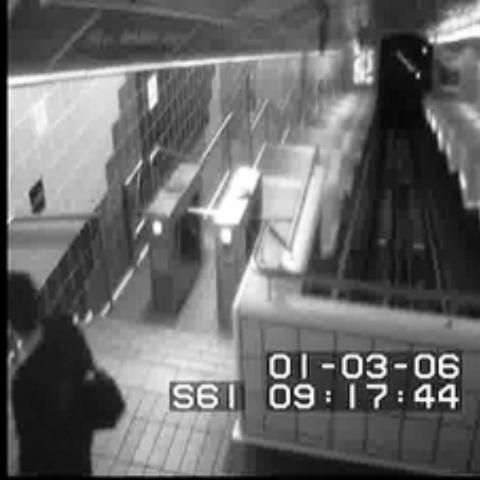}\\
	{\footnotesize Data layer}\\
	\includegraphics[scale=\caaal]{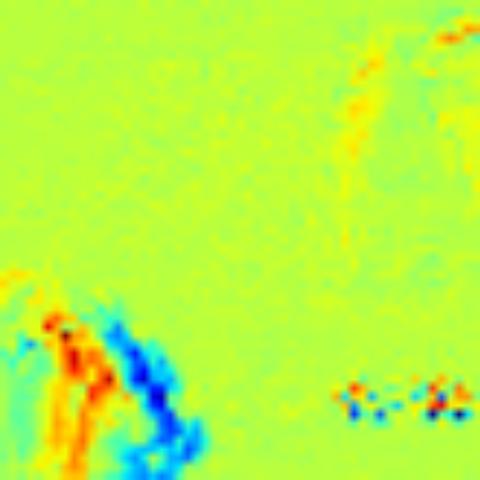}
	\includegraphics[scale=\caaal]{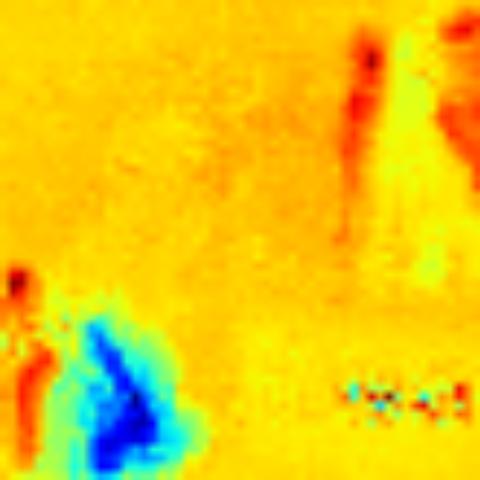}
	\includegraphics[scale=\caaal]{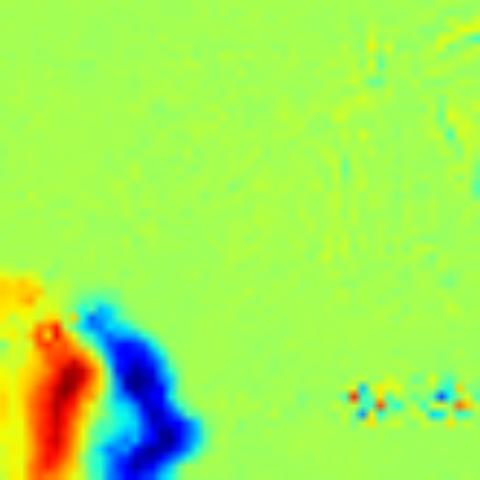}
	\includegraphics[scale=\caaal]{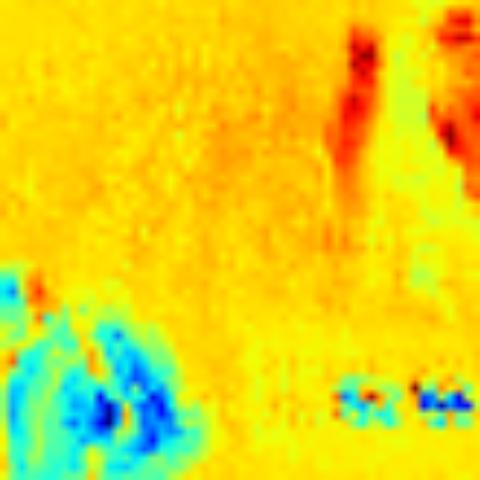}
	\includegraphics[scale=\caaal]{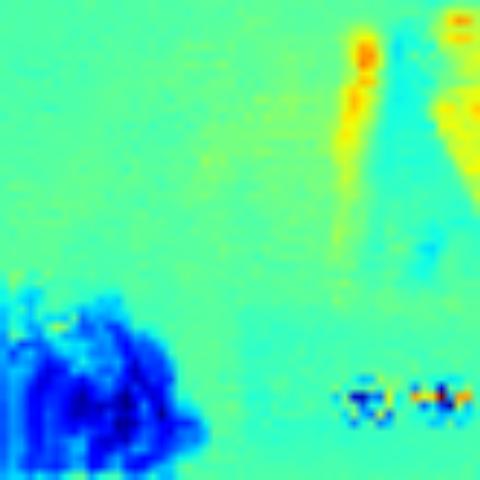}\\
	{\footnotesize First convolutional layer}\\
	\includegraphics[scale=\cbl]{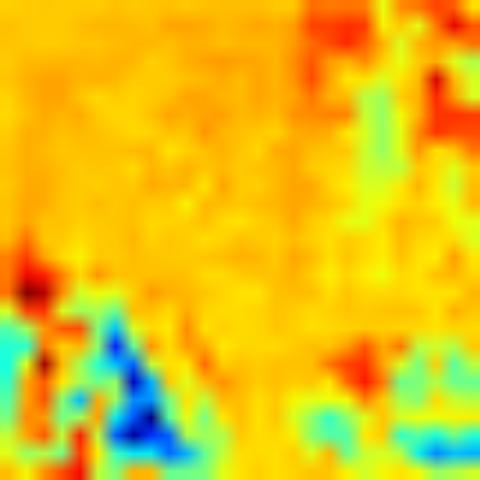}
	\includegraphics[scale=\cbl]{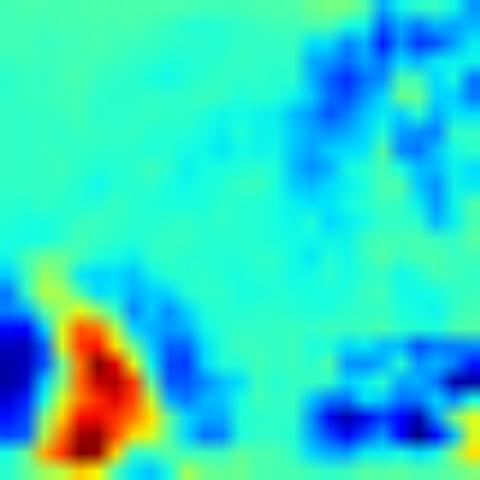}
	\includegraphics[scale=\cbl]{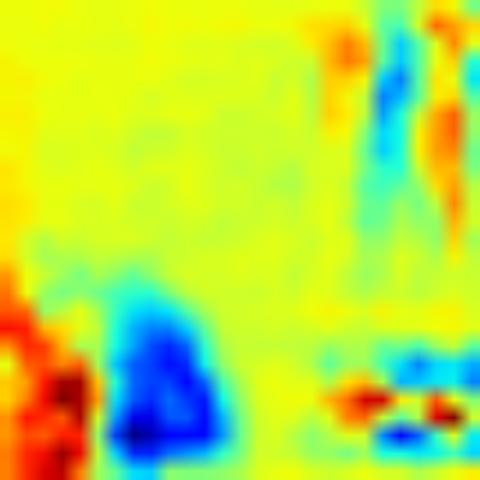}
	\includegraphics[scale=\cbl]{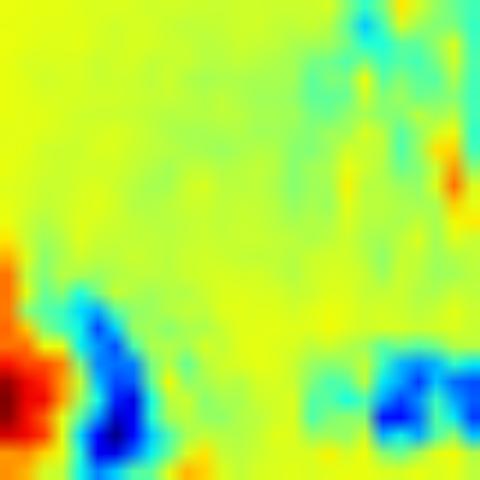}\\
	{\footnotesize Second convolutional layer}\\
	\includegraphics[scale=\ccl]{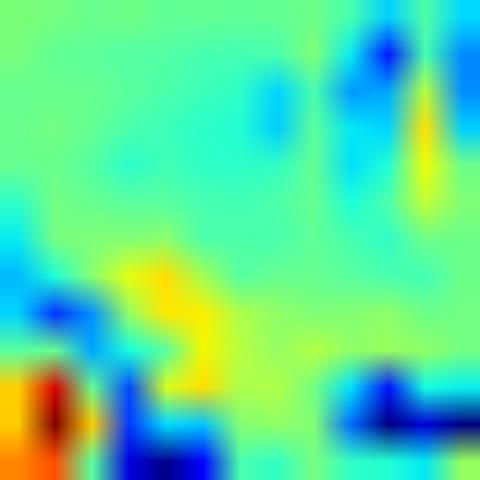}
	\includegraphics[scale=\ccl]{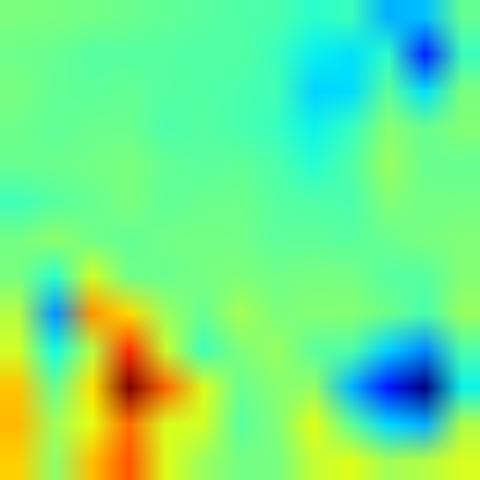}
	\includegraphics[scale=\ccl]{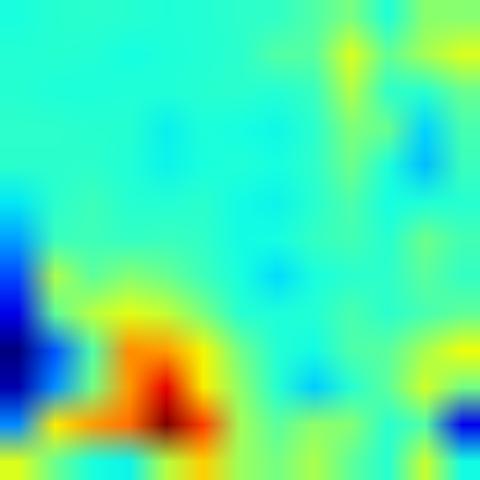}\\
	{\footnotesize Third convolutional layer}\\
	\includegraphics[scale=\cbl]{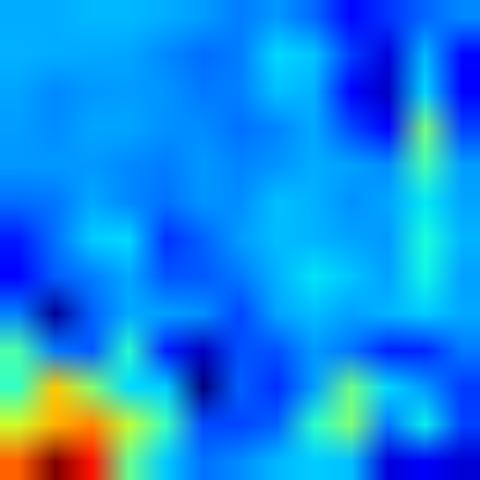}
	\includegraphics[scale=\cbl]{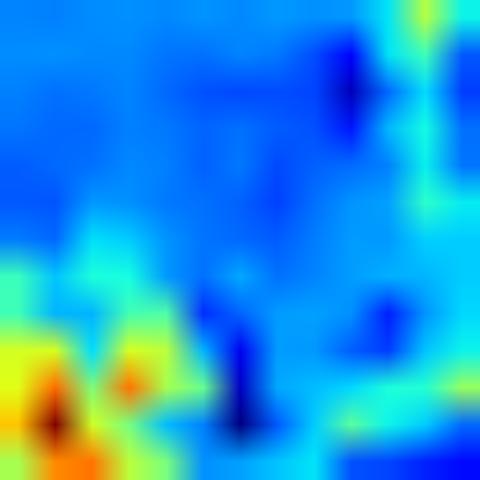}
	\includegraphics[scale=\cbl]{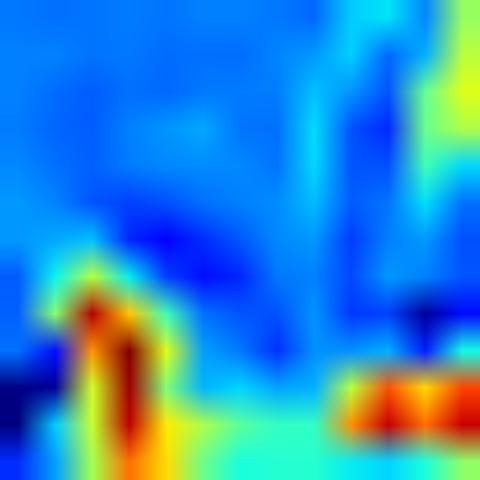}
	\includegraphics[scale=\cbl]{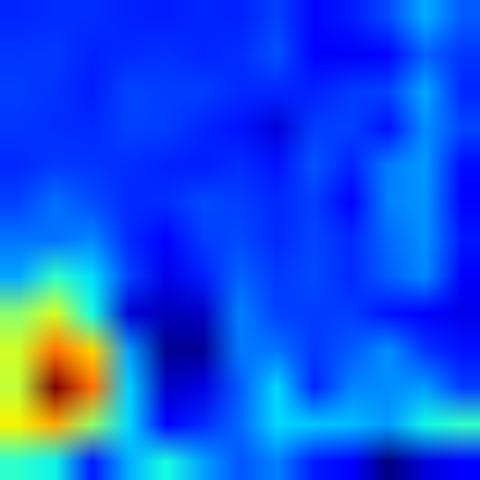}\\
	{\footnotesize First deconvolutional layer}\\
	\includegraphics[scale=\caaal]{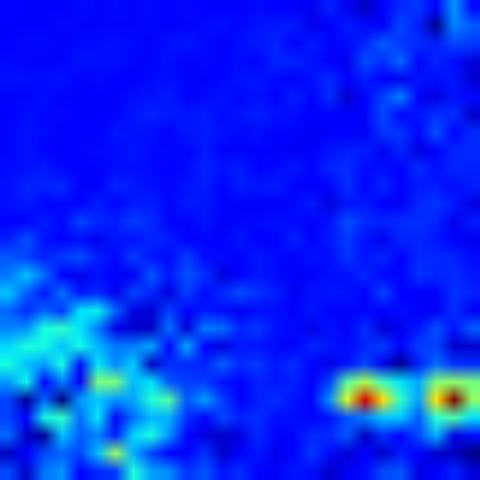}
	\includegraphics[scale=\caaal]{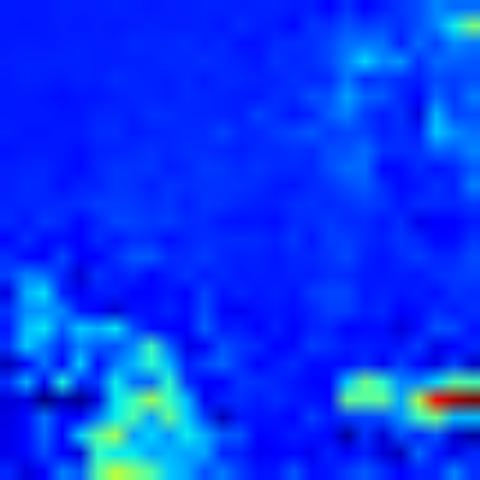}
	\includegraphics[scale=\caaal]{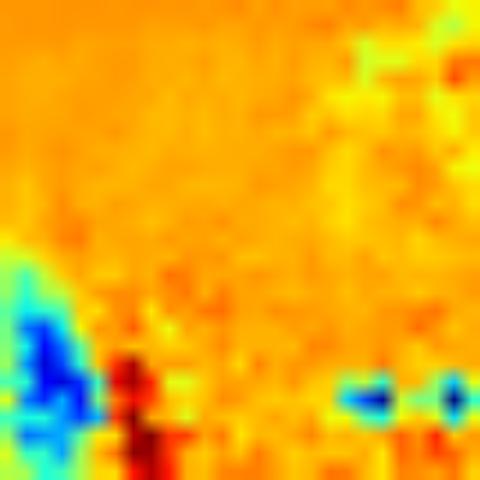}
	\includegraphics[scale=\caaal]{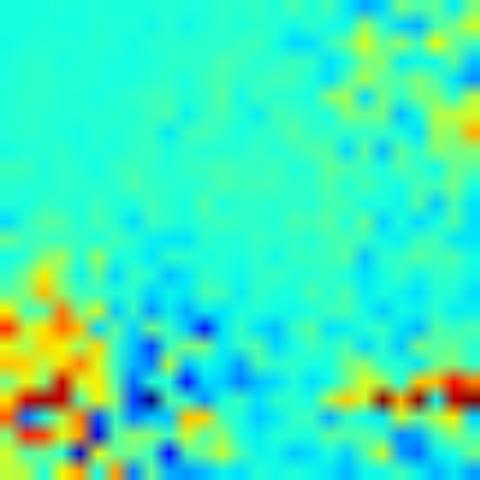}
	\includegraphics[scale=\caaal]{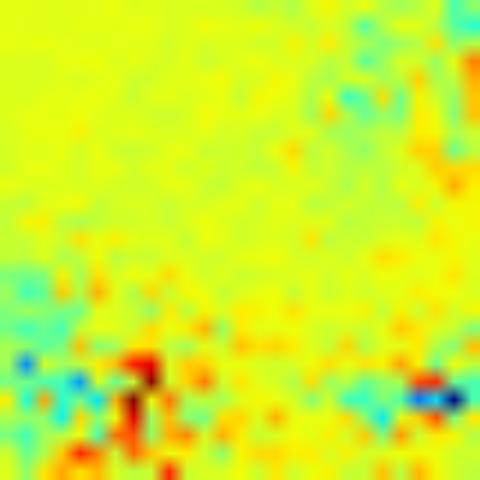}\\
	{\footnotesize Second deconvolutional layer}\\
	\includegraphics[scale=\dl]{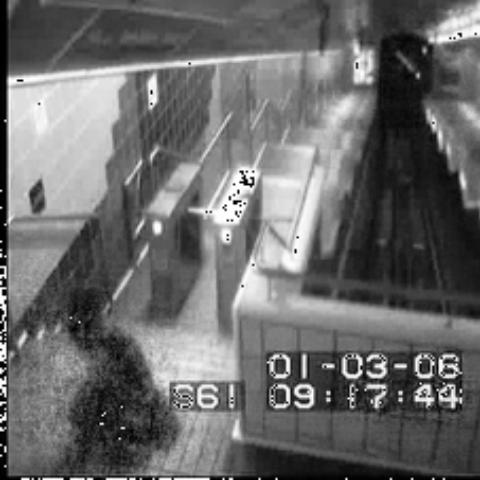}
	\includegraphics[scale=\dl]{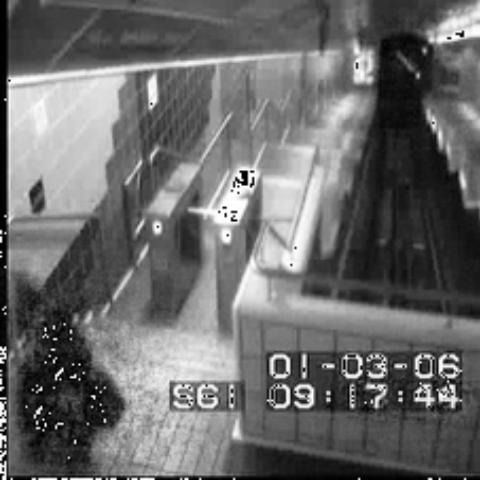}\\
	{\footnotesize Third deconvolutional layer}\\
	\caption{Responses of learned filters, evaluated on a video in Subway Exit dataset.}
\end{figure}

\begin{center}
	\hyperlink{page.11}{Go to Table of Contents}
\end{center}

\clearpage

\section{Filter Weights Visualization}
\label{sec:filter_w_vis}
In addition to visualizing filter responses, we visualize the learned filters themselves.
The learned filters are on a small spatial region and span in temporal dimensions up to 10 frames.
Since ten frame cube is hard to visualize, we select the first 3 frames to visualize the filters of temporal regularity.
Compared to the filters for object recognition that capture spatial structures~\cite{zeilerF13}, the temporal regular patterns do not have obvious spatial structure since they capture both spatial and temporal appearance thus look like random patterns.
But they exhibit some forms of horizontal and vertical motions in a form of implicit horizontal and vertical lines of same colored pixels (best viewed in zoom-in).

\begin{figure}[h]
	\centering
	\includegraphics[scale=0.70]{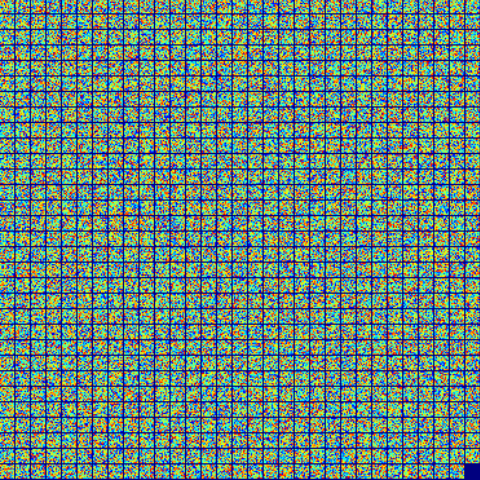}\\
	First convolutional layer \\
	~\\
	\includegraphics[scale=0.70]{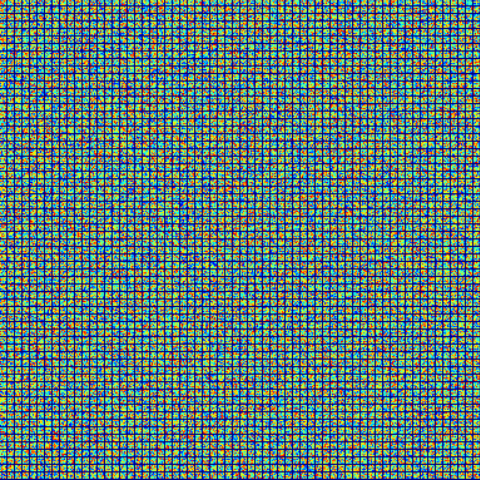}\\
	Second convolutional layer \\
	~\\
\end{figure}

\begin{figure}[h]
	\centering
	\includegraphics[scale=0.825]{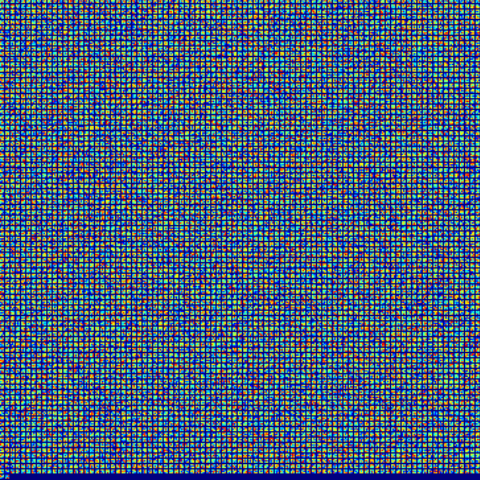}\\
	Third convolutional layer \\
	~\\
	\includegraphics[scale=0.825]{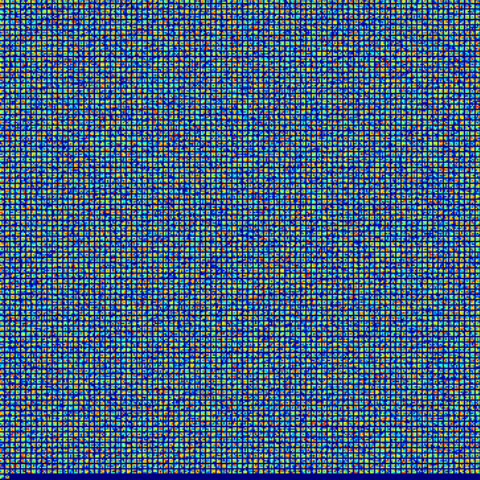}\\
	First deconvolutional layer \\
	~\\
\end{figure}

\begin{figure}[h]
	\centering
	\includegraphics[scale=0.8]{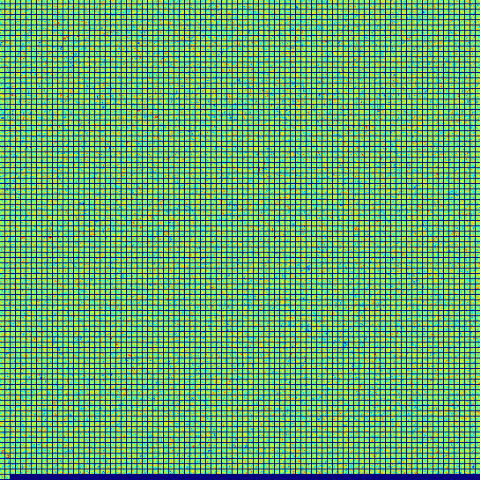}\\
	Second deconvolutional layer \\
	~\\
	\includegraphics[scale=0.8]{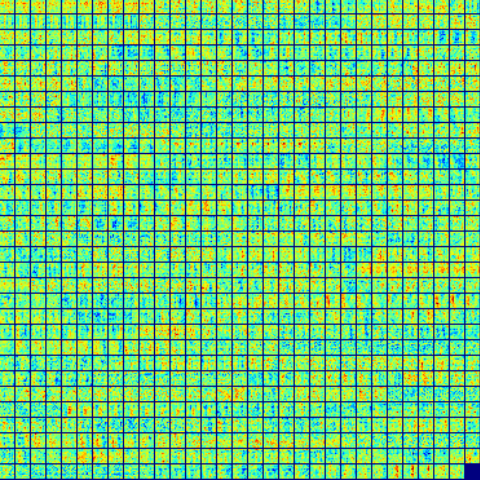}\\
	Third dconvolutional layer \\
	~\\
\end{figure}

\begin{center}
	\hyperlink{page.11}{Go to Table of Contents}
\end{center}

\clearpage

\end{document}